\documentclass{article}

\usepackage{PRIMEarxiv}

\usepackage[utf8]{inputenc} 
\usepackage[T1]{fontenc}    
\usepackage{textcomp}
\usepackage{hyperref}       
\usepackage{url}            
\usepackage{booktabs}       
\usepackage{amsmath,amssymb,amsfonts}
\usepackage{bm}
\usepackage{nicefrac}       
\usepackage{microtype}      
\usepackage{fancyhdr}       
\usepackage{graphicx}       
\usepackage{array}
\usepackage{float}
\usepackage{subfig}
\usepackage{newunicodechar}
\newunicodechar{→}{\ensuremath{\rightarrow}}
\newunicodechar{≥}{\ensuremath{\geq}}
\newunicodechar{≤}{\ensuremath{\leq}}
\newunicodechar{±}{\ensuremath{\pm}}
\newunicodechar{×}{\ensuremath{\times}}
\newunicodechar{°}{\ensuremath{^\circ}}
\newunicodechar{ε}{\ensuremath{\varepsilon}}
\newunicodechar{σ}{\ensuremath{\sigma}}
\newunicodechar{κ}{\ensuremath{\kappa}}
\graphicspath{{media/}}     

\title{Interpretable Deep Learning Models for Breast Cancer Diagnosis Using Medical Imaging}

\author{%
\normalfont\small
\begin{tabular}{@{}>{\centering\arraybackslash}p{0.46\textwidth}@{\hspace{0.04\textwidth}}>{\centering\arraybackslash}p{0.46\textwidth}@{}}
  \begin{tabular}[t]{@{}c@{}}\textbf{Sami Alnawali\thanks{Corresponding author.}} \\
    Information Systems Department \\
    College of Computer Science and Information Technology \\
    King Faisal University, Alahsaa, Saudi Arabia \\
    \texttt{salnawali@kfu.edu.sa}\end{tabular}
  &
  \begin{tabular}[t]{@{}c@{}}\textbf{Muhammad Atif Saeed\thanks{These authors contributed equally to this work.}} \\
    Department of Artificial Intelligence \& Data Science \\
    National University of Computer and Emerging Sciences \\
    Islamabad, Pakistan \\
    \texttt{atif.saeed@isb.nu.edu.pk}\end{tabular} \\ \addlinespace[1.2em]
  \begin{tabular}[t]{@{}c@{}}\textbf{Abdul Hadi\footnotemark[2]} \\
    Department of Robotics \& Artificial Intelligence \\
    Shaheed Zulfikar Ali Bhutto Institute\\ of Science and Technology \\
    Islamabad, Pakistan \\
    \texttt{abdulhaadii146@gmail.com}\end{tabular}
  &
  \begin{tabular}[t]{@{}c@{}}\textbf{Azka Atiq\footnotemark[2]} \\
    Department of Artificial Intelligence \& Data Science \\
    National University of Computer and Emerging Sciences \\
    Islamabad, Pakistan \\
    \texttt{azka.atiq@isb.nu.edu.pk}\end{tabular} \\ \addlinespace[1.2em]
  \begin{tabular}[t]{@{}c@{}}\textbf{Akhtar Jamil} \\
    Department of Computer Science \\
    National University of Computer and Emerging Sciences \\
    Islamabad, Pakistan \\
    \texttt{akhtar.jamil@nu.edu.pk}\end{tabular}
  &
  \begin{tabular}[t]{@{}c@{}}\textbf{Ahmad Din} \\
    Department of Artificial Intelligence \& Data Science \\
    National University of Computer and Emerging Sciences \\
    Islamabad, Pakistan \\
    \texttt{ahmad.din@nu.edu.pk}\end{tabular}
\end{tabular}%
}

\begin{document}
\maketitle

\begin{abstract}
Breast cancer continues to be a leading cause of mortality among women globally, highlighting the necessity for precise and comprehensible diagnostic tools. Although deep learning demonstrates significant potential, several current methodologies emphasize accuracy at the expense of interpretability, hence limiting their clinical applicability. To tackle this issue, we established a thorough framework that assesses ten advanced transfer learning models ResNet50, VGG19, InceptionV3, DenseNet121, Xception, MobileNetV2, EfficientNetV2S, EfficientNet-B7, Vision Transformer, and Swin Transformer, using the BreakHis dataset for automated breast cancer classification. All models underwent pretraining on ImageNet and were subsequently fine-tuned using bespoke classification heads, strengthened by comprehensive preprocessing and data augmentation methods including rotation, flipping, zooming, normalization, and contrast enhancement to increase generalization. To improve transparency, explainable AI (XAI) techniques such as Grad-CAM, LIME, SHAP, and LRP were utilized to display distinguishing histological features, emphasizing nuclei-rich and glandular areas aligned with clinical significance. Experimental results indicated that DenseNet121 attained superior performance, achieving a test accuracy of 99.4\% and an AUC-ROC of 0.9997 across various magnifications, whilst EfficientNet also exhibited robust performance with an accuracy of 98.5\% and continuously low false positive rates. Transformer-based models as Swin Transformer and ViT demonstrated strong generalization in intricate scenarios, while the lightweight MobileNetV2 provided competitive performance for implementation in resource-constrained settings. This study demonstrates the efficacy of integrating advanced transfer learning with explainable artificial intelligence to provide clinically relevant and interpretable frameworks for breast cancer detection. Future work will extend this framework to include multi-modal datasets, detailed tumor subtyping, and more advanced explainable artificial intelligence techniques to augment diagnostic reliability and facilitate clinical adoption.
\end{abstract}


\section{Introduction}\label{sec:intro}

Breast cancer is among the most common and lethal malignancies impacting women globally. The World Health Organization reports over 2.3 million new cases and over 680,000 deaths per year, establishing it as a primary cause of cancer-related mortality in women\cite{emam2024breast}. Timely and precise identification of breast cancer is crucial for enhancing treatment efficacy, survival probabilities, and the overall quality of life for patients\cite{wang2024breast}. Traditional diagnostic methods, including mammography, histological examination, and ultrasound imaging, while useful, are prone to inter-observer variability, necessitate professional interpretation, and may overlook small abnormalities in the early stages\cite{potsangbam2024classification}. These constraints have catalyzed the integration of artificial intelligence (AI) and deep learning (DL) methodologies in medical imaging to aid radiologists in achieving precise and prompt diagnosis\cite{lanjewar2024fusion}, \cite{arravalli2025detection}.

Deep learning, especially convolutional neural networks (CNNs), has demonstrated exceptional efficacy in visual recognition tasks, encompassing medical image classification, segmentation, and detection\cite{alom2025explainable}. Numerous research have utilized CNN architectures, including VGG, ResNet, Inception, and DenseNet, for breast cancer classification, with remarkable accuracy\cite{emam2024breast}. Nonetheless, these models frequently necessitate substantial quantities of annotated training data and considerable computational resources, posing challenges in the medical field where labeled data is limited and costly to obtain\cite{rani2024multi}. Furthermore, deep learning models are often criticized for their opacity and lack of interpretability, presenting a substantial obstacle to clinical implementation\cite{sannasi2024deep}. Comprehending the rationale behind a model's decision-making process is essential in healthcare applications, as erroneous predictions can lead to severe repercussions\cite{maurya2024fccs}.

To mitigate data scarcity and enhance model generalisability, transfer learning has emerged as an effective technique that facilitates the utilization of pretrained models on extensive datasets (e.g., ImageNet) for domain-specific applications such as breast cancer classification\cite{souza2024earlynet}, \cite{mirimoghaddam2024her2gan}. Transfer learning decreases computational demands and training duration while improving performance on limited, specialized medical datasets. The use of Explainable Artificial Intelligence (XAI) techniques, including Grad-CAM, LIME, SHAP, and XRAI, has created new opportunities for visualizing model behavior, enhancing clinician trust, and streamlining regulatory approval\cite{rani2024multi}, \cite{bouallegue2025enhanced}.

This paper presents a comprehensive and interpretable deep learning framework that integrates various transfer learning models with explainable artificial intelligence methodologies for robust breast cancer classification\cite{sannasi2024deep}. We assess ten cutting-edge models, ResNet50, VGG19, InceptionV3, DenseNet121, Xception, MobileNetV2, EfficientNetV2S, EfficientNet-B7, Vision Transformer (ViT), and Swin Transformer, across four prevalent breast cancer datasets\cite{jaganathan2024revolutionizing}. To improve model performance and generalisation, we employ customised data augmentation methods, normalisation, and patch-based image preprocessing\cite{lanjewar2024fusion}. Additionally, we incorporate various XAI techniques to deliver both local and global interpretability, assisting physicians in comprehending the reasoning behind the model's predictions\cite{potsangbam2024classification}.

The subsequent sections of this work are structured as follows: Section 1 delineates the pertinent literature and recent progress in deep learning for breast cancer diagnosis. Section 2 delineates the datasets employed, encompassing their structure, clinical context, and preparation procedures. Section 3 delineates the methodology, encompassing model architectures, transfer learning procedures, data augmentation techniques, and explainability methods. Section 4 delineates the experimental configuration and assessment criteria. Section 5 examines the outcomes derived from various models and datasets. Section 6 ultimately summarises the paper by presenting insights, limitations, and prospective future directions.

\section{Literature Review}\label{lr}

Breast cancer remains a major global health issue and one of the leading causes of cancer-related deaths among women. Early and precise detection is still crucial for improving prognosis, providing timely treatment, and enhancing survival rates\cite{alom2025explainable}. In recent years, deep learning (DL) algorithms have demonstrated exceptional potential for automating breast cancer diagnosis by understanding complicated patterns from medical pictures. However, a number of difficulties remain that prevent widespread clinical implementation\cite{jaganathan2024revolutionizing}. Key drawbacks include a lack of model generalisability across varied patient groups, biases caused by uneven or insufficient datasets, and a lack of transparency in decision-making. Many contemporary DL models behave like "black boxes," making it difficult for clinicians to trust and interpret their predictions\cite{alom2025explainable}. To fill these gaps, we present a complete system that combines a variety of pre-trained transfer learning modelsj, including EfficientNetV2S, InceptionResNetV2, and XceptionNet, among others, with explainable AI (XAI) techniques such as LIME and Faster Score-CAM\cite{arravalli2025detection}. This integration intends to improve classification performance while also providing interpretable insights into the model's predictions, allowing for better informed and reliable clinical decision-making.

Silva-Aravena et al. [2023]\cite{silva2023hybrid} introduced a hybrid machine learning and explainable artificial intelligence system for the early prevention of breast cancer, utilizing risk factor analysis on an Indonesian dataset of 400 persons (200 cases and 200 controls). The system utilized an XGBoost ensemble, compared with logistic regression, SVMs, and random forests, following the CRISP-ML(Q) framework to guarantee transparency and quality\cite{rani2024multi}. Preprocessing encompassed feature encoding and stratified division of lifestyle and reproductive variables, resulting in 85\% accuracy. SHAP explanations identified high-fat diet, absence of breastfeeding, and familial predisposition as significant risk factors; but, the study's limited, self-reported dataset constrains generalisability.

Nasir et al. [2025]\cite{nasir2025breast} performed a comparative deep learning analysis on the Breast Cancer Wisconsin Diagnostic Dataset (569 cases, 33 tumor characteristics), assessing CNNs, RNNs, MLPs, and hybrid models (CNN+LSTM, CNN+MLP), in addition to pretrained VGG16, ResNet, and EfficientNet\cite{lanjewar2024fusion}. Preprocessing encompassed normalization, feature correlation analysis, and stratified partitioning, while data insufficiency was mitigated through augmentation (rotation, flipping, contrast adjustment), SMOTE, and GAN-based synthesis. Grad-CAM and SHAP were utilized for interpretability, with CNN+LSTM attaining the highest accuracy (98.2\%), illustrating the efficacy of integrating spatial and sequential feature learning. Nonetheless, the study was constrained by the size of the dataset, the lack of image inputs, and diminished generalizability to clinical imaging\cite{rani2024multi}.

Patho-Net \cite{manojee2025patho} introduced a GRU-augmented U-Net segmentation framework for breast cancer histopathology, utilizing the BreakHis dataset of 2,048 pictures at 100× magnification. The pipeline utilized Reinhard color normalization with automated reference selection, EANLM noise reduction, and I-GLCM-based feature extraction, culminating in GRU-based classification\cite{potsangbam2024classification}. It attained an accuracy of 98.9\%, precision and recall of 98.6\%, and specificity of 99\%, with Grad-CAM and XRAI augmenting interpretability. Notwithstanding its robust performance, the model was confined to binary classification and depended on a resource-intensive preprocessing procedure, hence limiting cross-domain generalization.

Kabir et al. [2025] \cite{kabir2025breast} introduced a deep learning framework for the classification of breast cancer utilizing the BUSI ultrasound dataset, which comprises 780 pictures categorized as benign, malignant, and normal. Four pretrained CNNs—EfficientNetV2S, EfficientNetV2M, InceptionResNetV2, and XceptionNet\cite{emam2024breast} were fine-tuned utilizing scaling, flipping, rotation, zooming, and stratified splitting techniques. To enhance interpretability, Faster Score-CAM and LIME were utilized, emphasizing critical diagnostic areas. EfficientNetV2S attained superior performance (91.02\% accuracy, precision 0.92, recall 0.91, F1-score 0.91); however, dataset size, class imbalance, and absence of external validation limited generalizability\cite{gupta2024enhancing}.

Talukder [2025] \cite{talukder2025improved} proposed an improved DenseNet169 model for breast cancer histopathology, validated on the BreakHis (40×–400×), BACH, and BHI datasets. Preprocessing encompassed resizing (256×256), sharpening, normalization, and label encoding, along with augmentation by ±20° rotation, shifting, and flipping. Class Activation Mapping (CAM) and saliency maps were utilized for interpretability. The model attained 99.5\% accuracy on BreakHis 40×, 98.8\% on 100×, 97.3\% on 200×, 96.9\% on 400×, 94.8\% on BACH, and 88\% on BHI; nonetheless, difficulties remained in categorizing benign cases at elevated magnifications, highlighting the necessity for clinical validation.

Sharmi et al. [2025] \cite{ridha2025explainable} proposed an explainable framework for breast cancer diagnosis utilizing thermal images from the DMR dataset, comprising 2,010 samples. The pipeline integrated Attention U-Net for segmentation, K-Means clustering for region extraction, and EfficientNet-B7 for classification, incorporating preprocessing processes such as resizing (224×224), masking, and array conversion\cite{arravalli2025detection}. LIME was utilized for interpretability based on superpixels. The model attained a training accuracy of 96.5\%, a validation accuracy of 91.7\%, a recall of 91.9\%, a specificity of 91.4\%, a precision of 89.9\%, and an F1-score of 90.9\%. However, its generalizability was constrained by class imbalance, demographic bias, and insufficient cross-validation\cite{bouallegue2025enhanced}.

An ensemble-based transfer learning method\cite{rao2024intelligent} for breast cancer diagnosis utilizing ultrasound imagery was presented in IEEE Access [2024]. The research employed a dataset including 780 photos classified into benign, malignant, and normal categories, obtained from UCI and Kaggle. The authors employed an ensemble stacking technique to improve classification, integrating pretrained CNNs (InceptionV3, VGG16, VGG19) with machine learning classifiers, including MLPs and SVMs\cite{sannasi2024deep}. Preprocessing encompassed resizing, cropping, normalization, and data supplementation to enhance image quality and model resilience. The optimal ensemble model, utilizing InceptionV3 with stacking, attained an accuracy of 85.77\% and an AUC of 0.947, underscoring its diagnostic efficacy\cite{munshi2024novel}. Nonetheless, the approach failed to incorporate explainable AI (XAI) methodologies such as Grad-CAM or SHAP, which are essential for clinical transparency, and its scalability to extensive, heterogeneous ultrasound datasets has yet to be investigated\cite{maurya2024fccs}.

\section{Research Gap}
Although substantial advancements have been made in deep learning for breast cancer classification, current research frequently prioritizes accuracy over interpretability, so limiting medical applicability. The integration of XAI approaches (e.g., Grad-CAM, LIME, SHAP) is superficial and inconsistent among architectures, while dependence on a singular dataset limits generalizability. Histopathology datasets, like BreakHis and BACH, exhibit class imbalance, while mammography datasets such as CBIS-DDSM and INbreast are inadequately employed for cross-model comparisons. Moreover, there is a paucity of benchmarks involving both CNNs and transformer-based models (ResNet, EfficientNet, MobileNet, Xception, ViT, Swin Transformer), and the synergistic impact of augmentation and preprocessing on interpretability remains little examined. This study fills these gaps by doing a thorough assessment of transfer learning models utilizing several XAI methodologies to establish a transparent, precise, and clinically dependable diagnostic framework.

\section{Proposed Methodology}\label{methodology}

\subsection{Dataset}
In this paper, we will use the BreakHis (Breast Cancer Histopathological Image Classification) dataset which is one of the most popular benchmarks of 7,909 images of 82 patients with 40, 100, 200, and 400 magnifications of both benign and malignant tumor cells. In order to guarantee methodological rigor and prevent data leakage, a patient-level strict splitting strategy was utilized, and pictures of a patient were only used in training (70\%), validation (20\%), or testing (10\%) sets, which ensured that no overlap occurred between splits and simulated clinical conditions in real-life. The data has an inherent imbalance of classes and this was overcome by using weighted sampling methodology in training. All the images were subjected to a standardized processing procedure that included resizing (224x224 or 299x299 based on the needs of the model), ImageNet normalization, and augmentation as rotation, flipping, zooming, and contrast adjustment to enhance generalization. Also, evaluations of the magnification specifics were performed to measure the robustness of various magnifications of images. In spite of these measures, there can still be intra-patient and inter-magnification similarities in the dataset, thus, can bring subtle correlations, this is a limitation that is noted in order to make the results transparently reported and carefully interpreted.

\subsection{Data Augmentation}

This study utilizes customized data augmentation techniques to enhance generalization and reduce overfitting in deep learning models for breast cancer classification, addressing the lack of labeled medical images and class imbalance\cite{fayaz2024advancements}. Augmentations have been proposed to maintain diagnostic integrity while improving robustness: rotation (±20°) to establish rotational invariance, horizontal/vertical flipping to enhance spatial diversity\cite{potsangbam2024classification}, shearing to replicate distortions from scanning or tissue preparation, zooming (±20\%) to represent scale variations across magnifications\cite{fayaz2024advancements}., and contrast adjustment to accommodate discrepancies in staining and lighting conditions\cite{sumaiyya2024breast}. Bilinear interpolation was employed during scaling and translating to maintain smooth transitions and reduce aliasing problems\cite{bouallegue2025enhanced}. These augmentations collectively mimic real-world variability, allowing models to learn invariant, clinically significant characteristics that enhance classification performance across various imaging settings.

\subsection{Transfer Learning}
Transfer learning has become a potent approach in medical image analysis, especially in contexts with scarce labelled data. Transfer learning utilises deep neural networks pretrained on extensive datasets like ImageNet to generate robust, generalisable characteristics that may be refined for specific applications, such as breast cancer prediction. This work utilised many advanced pretrained models, including ResNet, VGG, DenseNet, EfficientNet, Xception, MobileNet, and vision transformer architectures, to evaluate their efficacy in breast cancer detection and classification. The models were modified by substituting their final classification layers and fine-tuning specific layers on target datasets to enhance performance for the particular diagnostic task.

\subsubsection{DenseNet121}

This study utilized DenseNet-121, a densely connected convolutional neural network pre-trained on ImageNet, as the foundation for classifying breast cancer histopathology images. The architecture utilizes extensive connectivity, allowing each layer to receive inputs from all prior levels, thus enhancing gradient flow, optimizing parameter efficiency, and alleviating vanishing gradient problems. The backbone consists of four dense blocks (6, 12, 24, and 16 layers, growth rate = 32) interspersed with transition layers utilizing 1×1 convolutions and 2×2 average pooling, culminating in global average pooling\cite{hiremath2023hybrid}, \cite{pattanaik2022breast}. A bespoke classifier head was incorporated, comprising a fully connected layer (1024→512) with ReLU activation, dropout (p=0.5) for regularization, and a concluding dense layer (512→2) generating logits for binary classification (benign versus malignant)\cite{wang2024breast}. Input photos were downsized to 224×224, normalized, then augmented by random rotations, shifts, scaling, and flips to improve generality. The model underwent fine-tuning utilizing Label Smoothing Cross-Entropy loss ($\varepsilon$=0.1), optimized via AdamW (weight decay = 0.01) with differential learning rates, and trained employing Cosine Annealing Warm Restarts\cite{hiremath2023hybrid}, \cite{pattanaik2022breast}. Supplementary regularization encompassed RandAugment and gradual layer unfreezing. Training was conducted with a batch size of 32 over 50 epochs, incorporating early stopping (patience = 15, monitored on validation F1-score).

\subsubsection{EfficientNet}
This study utilized EfficientNet, a scalable convolutional neural network created by Google AI, owing to its exceptional balance between accuracy and computational efficiency in medical image classification\cite{souza2024earlynet}, \cite{folorunso2024efficientnets}. The model uses the compound scaling method, which uniformly adjusts network depth, width, and input resolution, and is constructed using MBConv (Mobile Inverted Residual Bottleneck) blocks integrated with squeeze-and-excitation (SE) modules for channel-wise recalibration. We specifically fine-tuned EfficientNet-B0 and its larger variants (EfficientNetV2S and EfficientNet-B7), which were pretrained on ImageNet, by substituting the classifier head with a bespoke binary classification architecture\cite{behzadpour2024breast}. The backbone generated 1280 feature maps, which underwent global average pooling, succeeded by dropout (p = 0.5), a fully connected layer (1280 → 512, ReLU), an additional dropout, and a concluding linear layer (512 → 2) to yield logits for benign and malignant classifications. The training utilized Cross Entropy Loss with label smoothing and softmax activation during inference. To mitigate overfitting and enhance resilience, input photos were downsized (224×224 or 380×380), normalized, and subjected to augmentation by random flipping, zooming, shifting, and rotation with Keras’ ImageDataGenerator\cite{behzadpour2024breast}. The model underwent training with a progressive unfreezing of the pretrained backbone and varying learning rates, resulting in roughly 4 million trainable parameters post fine-tuning. 

\subsubsection{ConvNeXt}
Using the BreakHis dataset, we optimized ConvNeXt, a state-of-the-art CNN architecture that draws inspiration from ResNet and Vision Transformers, for the purpose of breast cancer histological picture categorization. ConvNeXt maintains CNN inductive biases, such as locality and translation equivariance, while integrating transformer-inspired design elements, including extensive kernel depthwise convolutions, inverted bottlenecks, Layer Normalization, GELU activation, stochastic depth, AdamW optimization, and cosine learning rate scheduling\cite{peter2025transformer}, \cite{khan2025ensemble}. We employed the ConvNeXt-Tiny/Base variation pretrained on ImageNet-21k and substituted the classification head with a bespoke architecture using LayerNorm, GELU activation, dropout (p=0.5), and fully connected layers (768 → 512 → 2) for the binary classification of benign and malignant instances. Input photos were reduced to 224×224 pixels, normalized with ImageNet statistics, and augmented by RandAugment (n=2, m=10), random cropping, color jittering, flipping, and rotation, while test-time augmentation (TTA) was utilized for enhanced inference robustness\cite{khan2025ensemble}. Training utilized label-smoothing cross-entropy (smoothing=0.1), the AdamW optimizer (weight decay=0.01), cosine annealing for learning rate scheduling, and mixed-precision training, alongside a gradual unfreezing of backbone layers to modify pretrained features\cite{ariyametkul2024explainable}. The model comprised around 28.6 million trainable parameters.

\subsubsection{ResNet}
This study utilized the fine-tuning of a ResNet-50 backbone, pretrained on ImageNet, for the classification of breast cancer utilizing the BreaKHis dataset. The input photos were downsized to 224×224 RGB, normalized using ImageNet statistics, and augmented by flipping, rotation, zooming, and RandAugment\cite{reenadevi2021breast}. The backbone (about 23.5 million parameters) was initially fixed and subsequently unfixed with varying learning rates, while the custom classifier head (approximately 1.05 million parameters) comprised fully connected layers of 2048→512→2 with ReLU activation and a dropout rate of 0.5\cite{khalid2021histological}. The training employed CrossEntropyLoss with label smoothing, the Adam optimizer, and cosine annealing with warm restarts, while the assessment metrics comprised accuracy, precision, recall, specificity, F1-score, and AUC-ROC\cite{reenadevi2021breast}. To improve interpretability, Grad-CAM, LIME, SHAP, and LRP were combined for visualizing decision boundaries.

\subsubsection{InceptionV3}
The study used InceptionV3, a deep convolutional neural network architecture designed for effective multi-scale feature extraction, as a transfer learning model for breast cancer classification\cite{admass2024integration}. InceptionV3 enhances the original GoogLeNet by integrating factorized convolutions, auxiliary classifiers, asymmetric filters (1×7 and 7×1), and batch normalization, hence facilitating computational efficiency and superior generalization. The model employs Inception modules that concurrently process various convolution filters (1×1, 3×3, 5×5), effectively capturing both local and global representations\cite{emam2024breast}. In our approach, the pretrained backbone was initialized using ImageNet weights, and the final classification head was substituted with a dense block (2048→512→2) with ReLU activation and dropout (p=0.5), in addition to a modified auxiliary classifier. Histopathology photos were downsized to 299×299, normalized using ImageNet statistics, and augmented with RandAugment, including cropping, flipping, rotating, zooming, and color jittering\cite{admass2024integration}. Training utilized the AdamW optimizer, label smoothing cross-entropy, cosine annealing warm restarts, mixed precision, and a gradual unfreezing of backbone layers for stable fine-tuning\cite{emam2024breast}.

\subsubsection{VGG19}
This research used VGG-19, pretrained on ImageNet, as a transfer learning model for the classification of breast cancer. The backbone, consisting of 16 convolutional and 5 pooling layers (about 20 million parameters), was initially frozen and subsequently fine-tuned using differential learning rates and a gradual unfreezing approach\cite{kanimozhi2024breast}. The initial head was substituted with a bespoke classifier (~102.8M parameters) comprising Linear (25088→4096→4096→512→2) layers, ReLU activations, and several dropout layers (p=0.5) for regularization purposes. The input photos were downsized to 224×224, normalized, and augmented using RandAugment, along with cropping, flipping, rotation, and color jittering\cite{souza2024earlynet}. The training employed Label Smoothing Cross-Entropy (smoothing=0.1), the AdamW optimizer, cosine annealing warm restarts, AMP mixed-precision, and early halting, resulting in about 122.8 million trainable parameters optimized for histopathology picture classification\cite{kanimozhi2024breast}.

\subsubsection{MobileNetV2}
This research utilized MobileNetV2, a lightweight convolutional neural network optimized for processing efficiency by depthwise separable convolutions and inverted residual bottlenecks with linear projections\cite{ariyametkul2024explainable}. The model accepts 224×224 RGB inputs, commencing with a 3×3 convolution (32 filters, stride 2), succeeded by a series of 17 bottleneck layers incorporating expansion through 1×1 convolutions, depthwise filtering, and projection layers. Each block integrates ReLU6 activations and residual shortcuts when the input and output dimensions are congruent\cite{emam2024breast}. The backbone generates 1280 feature maps, which are subsequently processed by a global average pooling layer and a bespoke classifier head (Linear 1280→512, ReLU, Dropout 0.5, Linear 512→2) for binary classification. MobileNetV2, with around 3.5 million parameters, attains a balance between precision and low complexity, rendering it appropriate for medical imaging applications where efficiency and scalability are paramount\cite{mehedi2024lightweight}, \cite{sumaiyya2024breast}.

\subsubsection{Vision Transformer (ViT)}
This study utilized a Vision Transformer (ViT-B/16) pretrained on ImageNet-21k and subsequently fine-tuned for breast cancer categorization\cite{naas2025explainable}. Input images (3×224×224) were segmented into 16×16 non-overlapping patches, resulting in 196 tokens embedded in 768-dimensional vectors, together with an additional learnable class token and positional embeddings\cite{al2025hybrid}. The sequence underwent processing through 12 transformer encoder blocks, each consisting of multi-head self-attention (12 heads), MLP layers (hidden=3072, GELU), LayerNorm, residual connections, and dropout (p=0.1). The ultimate representation underwent LayerNorm, followed by a bespoke classifier head incorporating LayerNorm, Linear (768→512), GELU, Dropout, and Linear (512→2) for binary classification\cite{balaha2025harnessing}. The architecture comprised over 86 million parameters, with around 400,000 derived from the custom head, facilitating efficient fine-tuning on histopathological pictures while maintaining the ViT's robust feature extraction and contextual modeling abilities\cite{mehedi2024lightweight}, \cite{ayana2024vision}.

\subsubsection{Swin Transformer (Shifted Window Transformer)}
This research used the Swin Transformer Tiny, a hierarchical vision transformer pretrained on ImageNet-1K, for the classification of breast cancer images\cite{ahmed2025hierarchical}, \cite{ayana2024vision}. Input images (3×224×224) were divided into 4×4 patches and transformed into 768-dimensional tokens, which underwent four steps of shifted window-based multi-head self-attention (W-MSA, SW-MSA), residual connections, LayerNorm, and MLPs with GELU activations. Patch merging systematically diminished resolution while augmenting channel depth (768→1536→3072)\cite{ahmed2025hierarchical}. The backbone output, a 3072-dimensional feature vector, was consolidated using global average pooling and forwarded to a bespoke classifier head consisting of Linear (768→512), ReLU, Dropout (0.5), Linear (512→256), ReLU, Dropout (0.5), and Linear (256→2) for the prediction of benign versus malignant outcomes\cite{peter2025transformer}. The architecture had approximately 28.3 million parameters, with around 0.5 million being trainable during the initial fine-tuning phase, integrating transformer-based global context modeling with computational efficiency appropriate for high-resolution medical imaging\cite{peter2025transformer}.

\subsubsection{Xception}
The study utilized Xception (Extreme Inception), a deep convolutional neural network pretrained on ImageNet, modified for binary breast cancer classification\cite{manojee2025patho}. Xception substitutes conventional convolutions with depthwise separable convolutions integrated with residual connections, facilitating efficient feature extraction. The network processes 299×299 RGB inputs via an entering flow of convolutional and separable convolutional layers (32–728 channels), succeeded by a middle flow including 8 repeating blocks (728 channels), and concluding with an exit flow that expands to 2048 channels\cite{farabi2023explainable}. The backbone output undergoes global average pooling, dropout (p=0.5), and a concluding linear layer (2048→2) for the classification of benign and malignant cases. The training utilized the AdamW optimizer, cosine annealing warm restarts, and label-smoothed cross-entropy loss, rendering the model both computationally efficient and highly expressive for histopathology picture processing\cite{ariyametkul2024explainable}.

\subsection{Explainable Artificial Intelligence (XAI)}
The black-box aspect of deep learning models presents major obstacles to their clinical adoption, where trust and transparency are of the utmost importance, notwithstanding their impressive accuracy in breast cancer detection. Explainable Artificial Intelligence (XAI) methodologies seek to mitigate this constraint by offering human-comprehensible insights into model predictions\cite{murugan2025efficient}. This study utilised XAI technologies, including Grad-CAM, LIME, SHAP, and XRAI, to visualise and elucidate the decision-making processes of diverse deep learning models. These tools not only improve model transparency but also aid doctors in verifying the regions or features most significant in diagnosis results.

\subsubsection{Grad-CAM (Gradient-weighted Class Activation Mapping)}
Grad-CAM (Gradient-weighted Class Activation Mapping) was utilized to improve the interpretability of deep learning models by producing class-discriminative heatmaps that emphasize the picture regions most significant for classification\cite{naas2025explainable, raghavan2024attention}. It calculates the gradients of the projected class score relative to the feature maps of the final convolutional layer, use global average pooling to obtain channel weights, and generates a ReLU-activated weighted sum to create the final heatmap. The heatmap is subsequently overlaid on the original image, clearly highlighting areas essential for decision-making\cite{behzadpour2024breast, ariyametkul2024explainable}. This study employed Grad-CAM across various architectures (ResNet50, DenseNet121, Xception, EfficientNetV2, Swin Transformer) to assess the models' attention to aberrant tissue patterns. Its post hoc nature, necessitating no architectural alterations, renders it especially appropriate for medical imaging applications where trust and transparency are important\cite{peter2025transformer}.

\subsubsection{LIME (Local Interpretable Model-Agnostic Explanations)}
The study used LIME (Local Interpretable Model-Agnostic Explanations) to produce localized, case-specific elucidations of breast cancer classification models. In contrast to gradient-based approaches, LIME is model-agnostic, approximating intricate predictions with a straightforward interpretable surrogate by perturbing inputs into superpixels and analyzing output fluctuations\cite{maurya2024fccs}. Each image underwent masking or modification, and a local linear model was applied to assess the significance of various image segments. LIME was utilized on models such as ResNet50, EfficientNet, DenseNet121, and Swin Transformer, emphasizing the areas most critical for benign or malignant predictions. This offered clear, instance-specific interpretability, augmenting Grad-CAM by elucidating local decision limits that global saliency maps would neglect, thus enhancing clinical confidence in AI-assisted breast cancer diagnosis\cite{sahu2024efficient}.

\subsubsection{SHAP (SHapley Additive Explanations)}
SHAP was used to measure feature significance through principles of cooperative game theory, assigning the contribution of each input region to the model's output. By estimating Shapley values, SHAP equitably allocates prediction influence among picture superpixels or patches\cite{alhussen2025xai}. This work employed SHAP on models such as ResNet50, EfficientNet, DenseNet121, and Swin Transformer, generating heatmaps that emphasized areas influencing malignant or benign predictions. In contrast to gradient-based techniques, SHAP delivers globally consistent and theoretically robust explanations, furnishing doctors with insights into the influence of each input feature on decision outcomes\cite{sumaiyya2024breast}.

\subsubsection{Integrated Gradients (IG)}
Integrated Gradients were used to attribute model predictions by calculating the path integral of gradients between a baseline (e.g., a black images) and the input image. This method offers a systematic methodology for allocating contribution scores to each pixel by averaging gradients along the interpolation trajectory\cite{alhussen2025xai}. This study employed IG on CNN and transformer-based architectures to elucidate subtle tissue patterns affecting benign-malignant classification. Its efficacy is rooted in generating smoother and more accurate attributions than raw gradients, rendering it particularly suitable for medical imaging, where detailed interpretability is essential\cite{murugan2025efficient}.

\subsubsection{Layer-wise Relevance Propagation (LRP)}
LRP was utilized as a backpropagation-based explanation technique, redistributing the prediction score backwards through the network to identify the input areas that were most responsible for the decision\cite{tanim2024breast}. Using relevance conservation concepts, LRP assigns pixel-level contributions that aggregate to the model's output, producing comprehensive and intuitive visual representations. LRP was incorporated into CNNs (ResNet, DenseNet, Xception) for the analysis of breast cancer images, emphasizing significant histological features. LRP's ability to deliver detailed pixel attributions improves clinical confidence by guaranteeing that diagnostic predictions correspond with medically relevant structures.

\section{Experimental Details}

\section{Results and Discussion}
\subsection{Datasets}
The BreakHis dataset is a collection of histopathological pictures created by the P\&D Lab in Brazil, comprising 7,909 microscopic images of breast tissue samples from 82 patients, classified as benign or malignant. Images are obtained at four magnification levels: 40X, 100X, 200X, and 400X, utilising an Olympus BX-50 microscope. The dataset has eight subtypes: four benign (adenosis, fibroadenoma, phyllodes tumour, tubular adenoma) and four malignant (ductal carcinoma, lobular carcinoma, mucinous carcinoma, papillary carcinoma). The dataset exhibits imbalance and significant intra-class variability, rendering it appropriate for training and assessing deep learning models in fine-grained classification tasks. Each image is in RGB format and possesses a resolution of 700×460 pixels.

\subsection{Experimental Results}

\begin{table}[ht]
\caption{Performance of DenseNet-121 on the BreakHis dataset}
\label{tab:densenet121}
\centering
\begin{tabular}{@{}lcccccc@{}}
\toprule
\textbf{Dataset} 
& \textbf{Accuracy (\%)}
& \textbf{Precision (\%)}
& \textbf{Recall (\%)}
& \textbf{F1-score (\%)}
& \textbf{Specificity (\%)}
& \textbf{AUC} \\
\midrule
Validation (All Mag.) 
& 89.2 
& 90.7 
& 95.5 
& 93.0 
& 70.3 
& 0.936 \\

Test (All Mag.)    
& \textbf{99.4} 
& \textbf{99.9} 
& \textbf{99.5} 
& \textbf{99.6} 
& \textbf{99.2} 
& \textbf{0.999} \\

Test (40$\times$)      
& 100.0 
& 100.0 
& 100.0 
& 100.0 
& 100.0 
& 1.000 \\

Test (100$\times$)      
& 100.0 
& 100.0 
& 100.0 
& 100.0 
& 100.0 
& 1.000 \\

Test (200$\times$)      
& 100.0 
& 100.0 
& 100.0 
& 100.0 
& 100.0 
& 1.000 \\

Test (400$\times$)      
& 97.5 
& 99.4 
& 97.7 
& 98.5 
& 96.4 
& 0.998 \\
\bottomrule
\end{tabular}
\end{table}

DenseNet-121 performed well on the BreakHis dataset, showing the ability to generalize across magnifications. On the validation set, it achieved 89.24\% accuracy, 90.72\% precision, 95.46\% recall, 70.31\% specificity, an F1-score of 93.03\%, and an AUC-ROC of 0.9362, indicating high sensitivity but some misclassification of benign cases(Table 1 \& Figure 1). On the test set (at all magnifications), the model achieved near-perfect results with 99.41\% accuracy, 99.86\% precision, 99.45\% recall, 99.17\% specificity, 99.65\% F1-score, and 0.9997 AUC-ROC, indicating strong generalization and balanced performance across both malignant and benign classes (Figure 2 \& 3). The model reliably extracted discriminative histopathological characteristics at individual magnification settings (40X, 100X, and 200X) with 100\% accuracy across all measures (Figure 4). These findings show the effectiveness of DenseNet-121's densely connected architecture in reducing vanishing gradients, encouraging feature reuse, and providing highly reliable breast cancer classification performance on histopathology pictures.

\begin{figure}[H]
 \centering
 \includegraphics[width=0.75\linewidth]{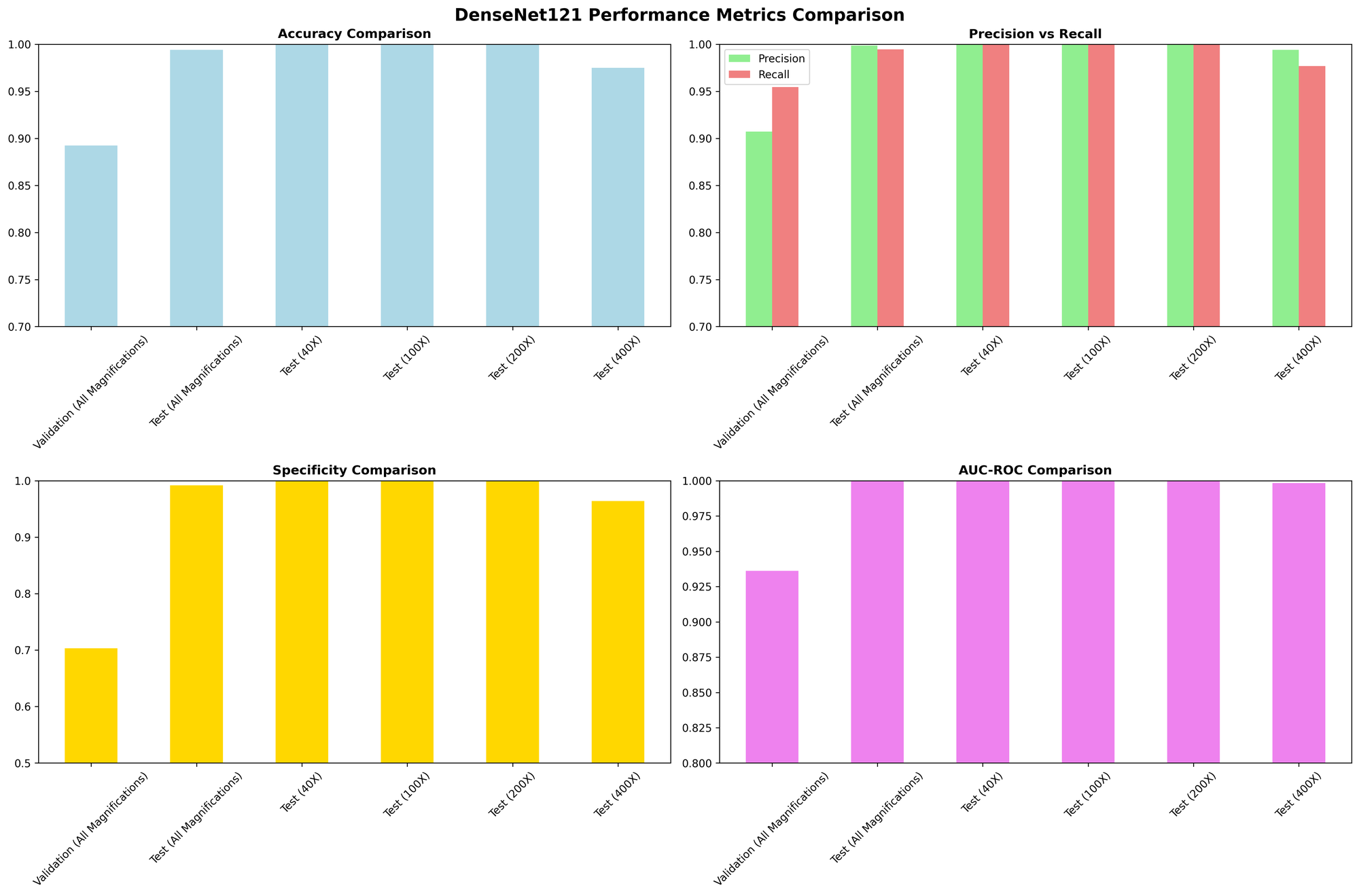}
 \caption{Performance metrics comparison of DenseNet121 on the BreakHis dataset across validation and test sets at different magnifications (40X, 100X, 200X, 400X).}
 \label{fig:densenet_metrics}
\end{figure}

\begin{figure}[H]
\centering

\subfloat[40X]{%
 \includegraphics[width=0.23\textwidth]{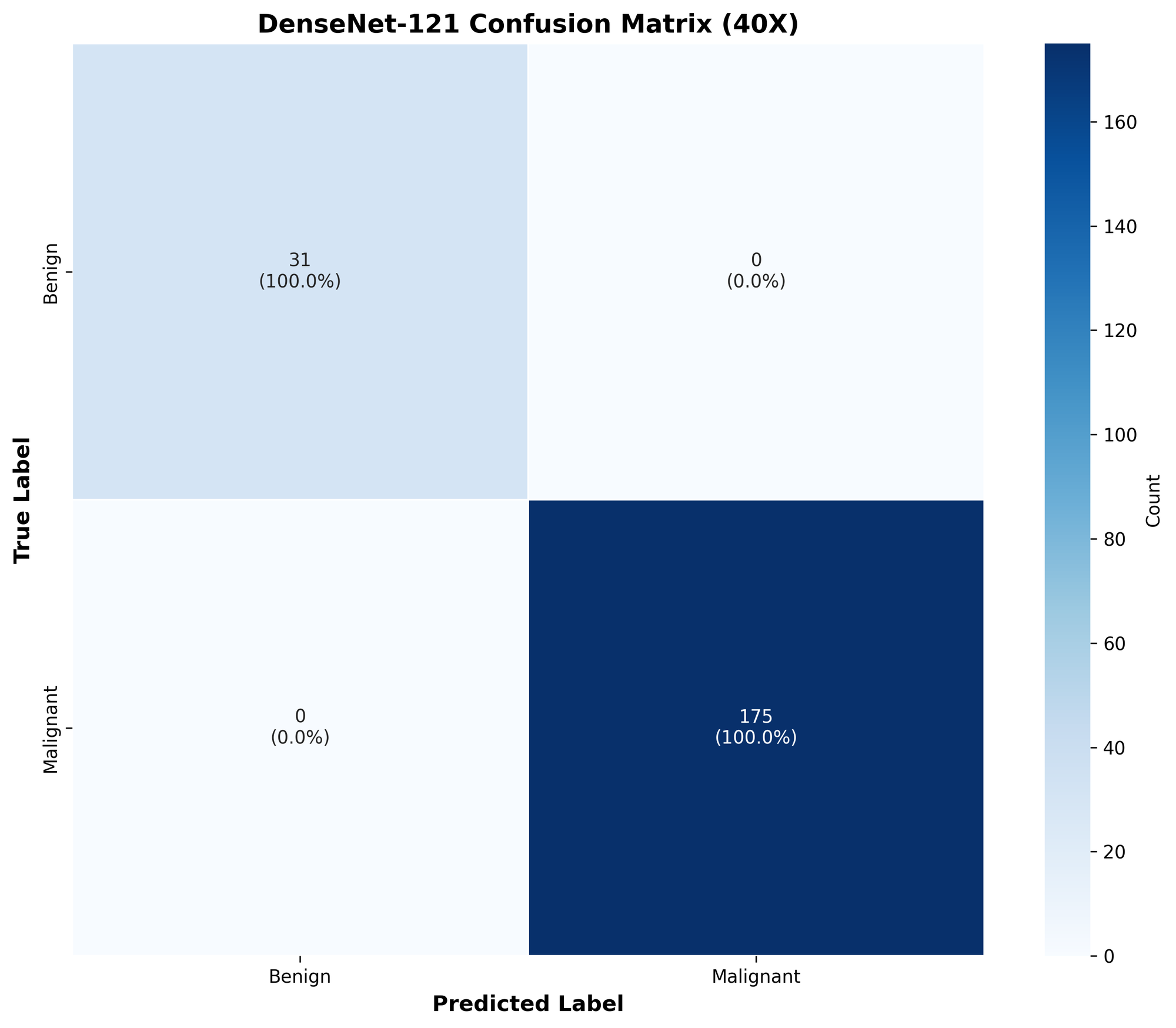}
}\hfill
\subfloat[100X]{%
 \includegraphics[width=0.23\textwidth]{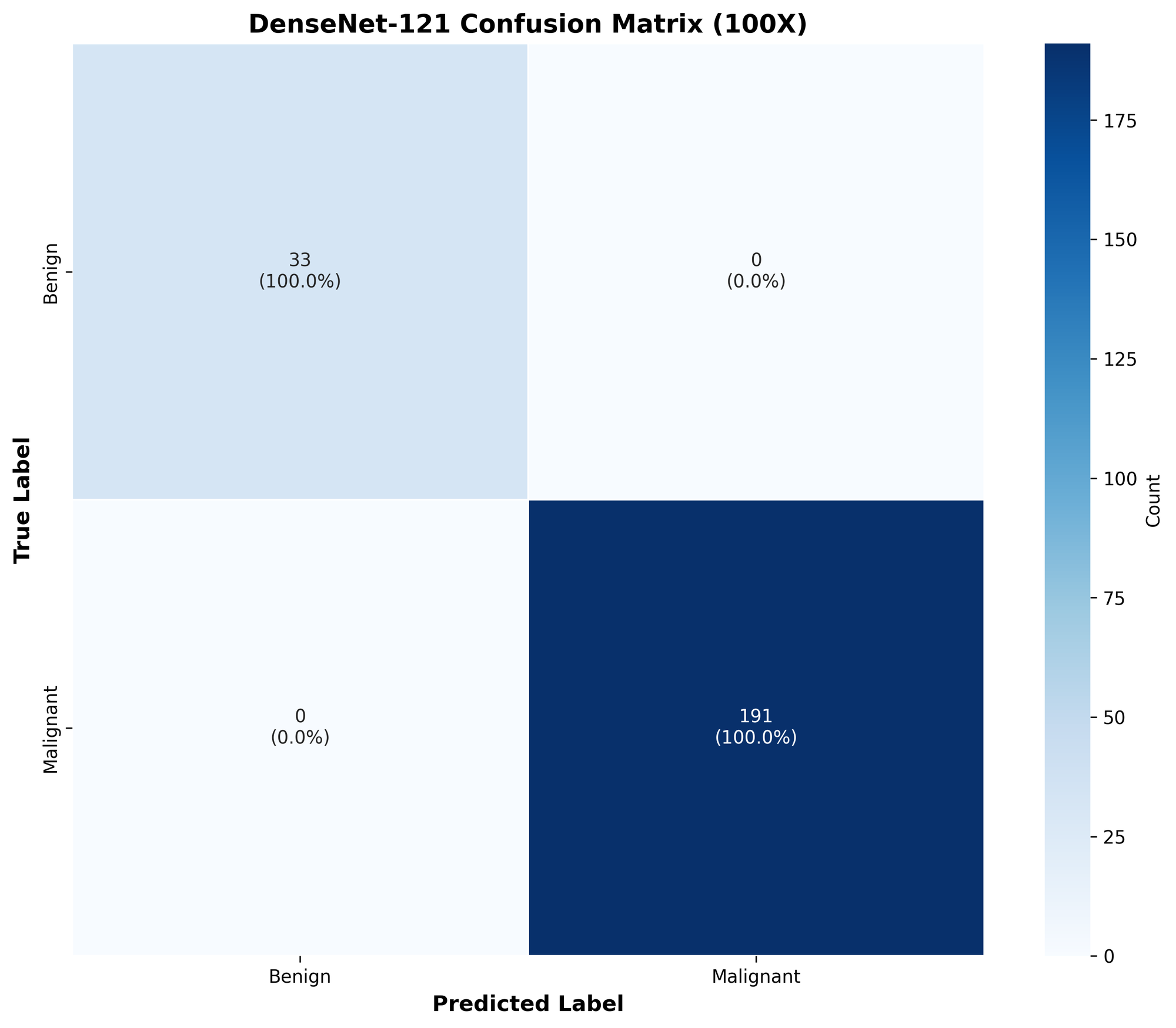}
}\hfill
\subfloat[200X]{%
 \includegraphics[width=0.23\textwidth]{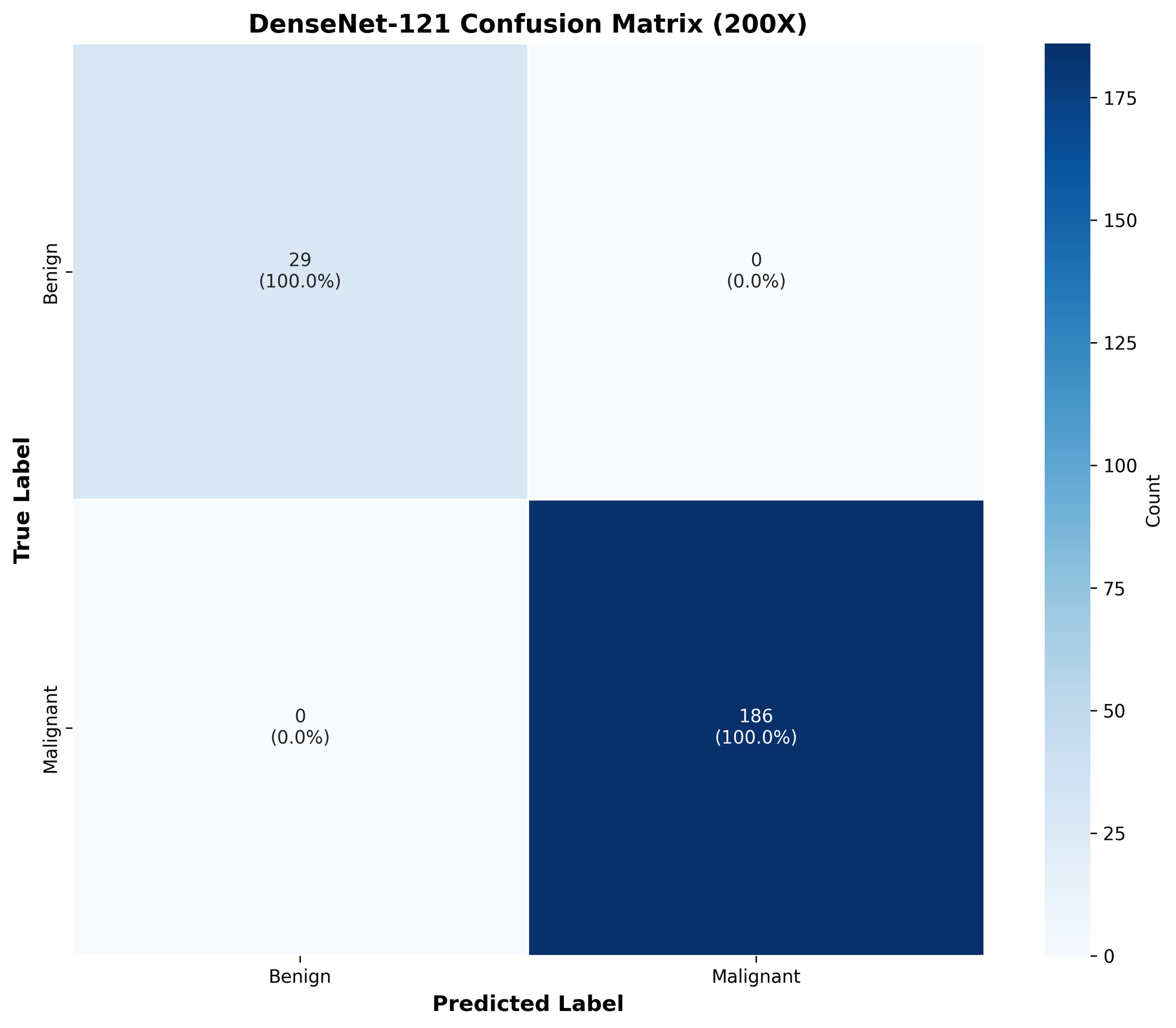}
}\hfill
\subfloat[400X]{%
 \includegraphics[width=0.23\textwidth]{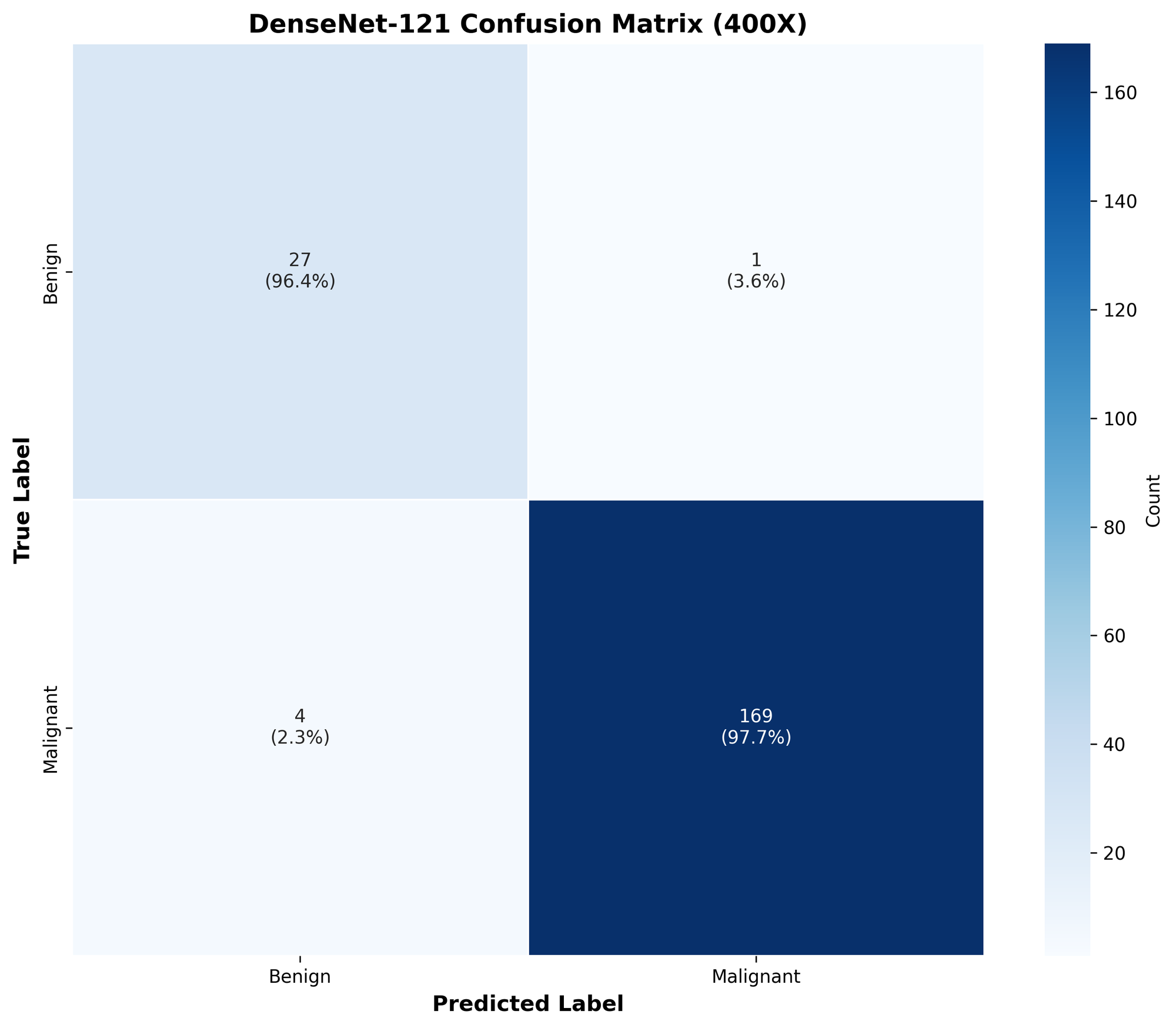}
}

\caption{Confusion matrices of the DenseNet-121 model across 40X, 100X, 200X, and 400X magnifications on the BreakHis dataset.}
\label{fig:densenet121_confusion_matrices}
\end{figure}

\begin{figure}[H]
\centering

\subfloat[Validation ROC]{%
 \includegraphics[width=0.23\textwidth]{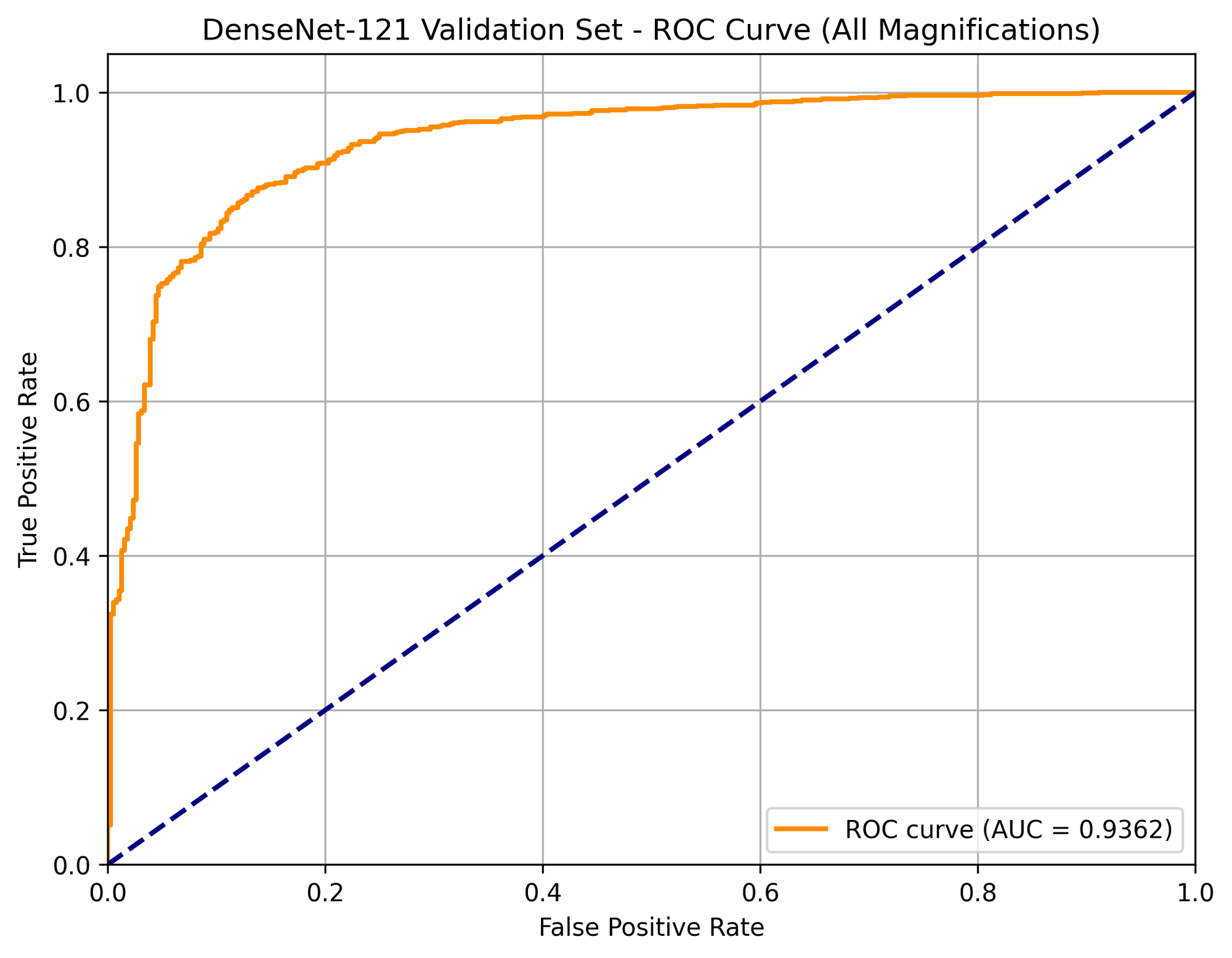}
}\hfill
\subfloat[Validation Confusion Matrix]{%
 \includegraphics[width=0.23\textwidth]{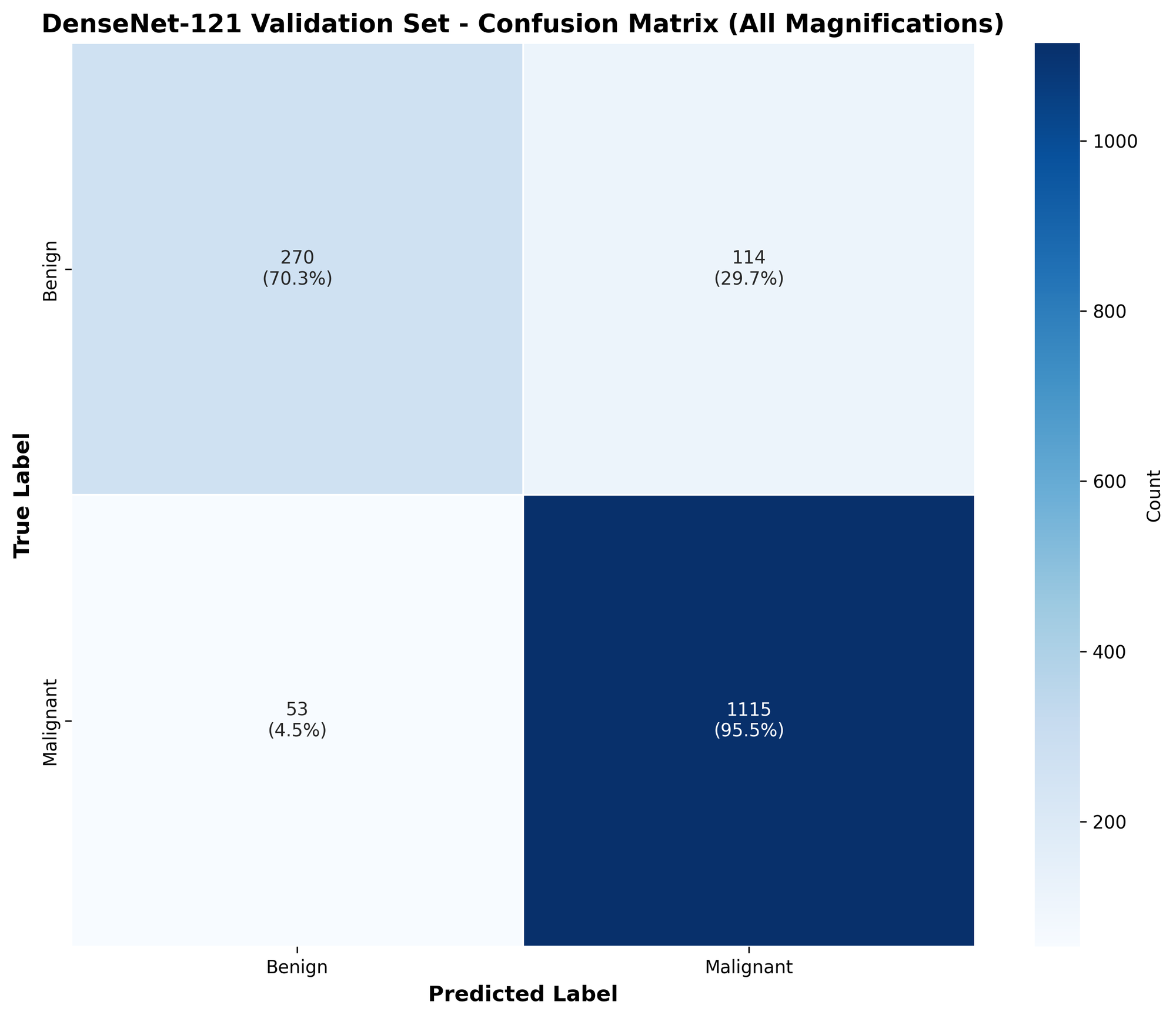}
}\hfill
\subfloat[Test ROC]{%
 \includegraphics[width=0.23\textwidth]{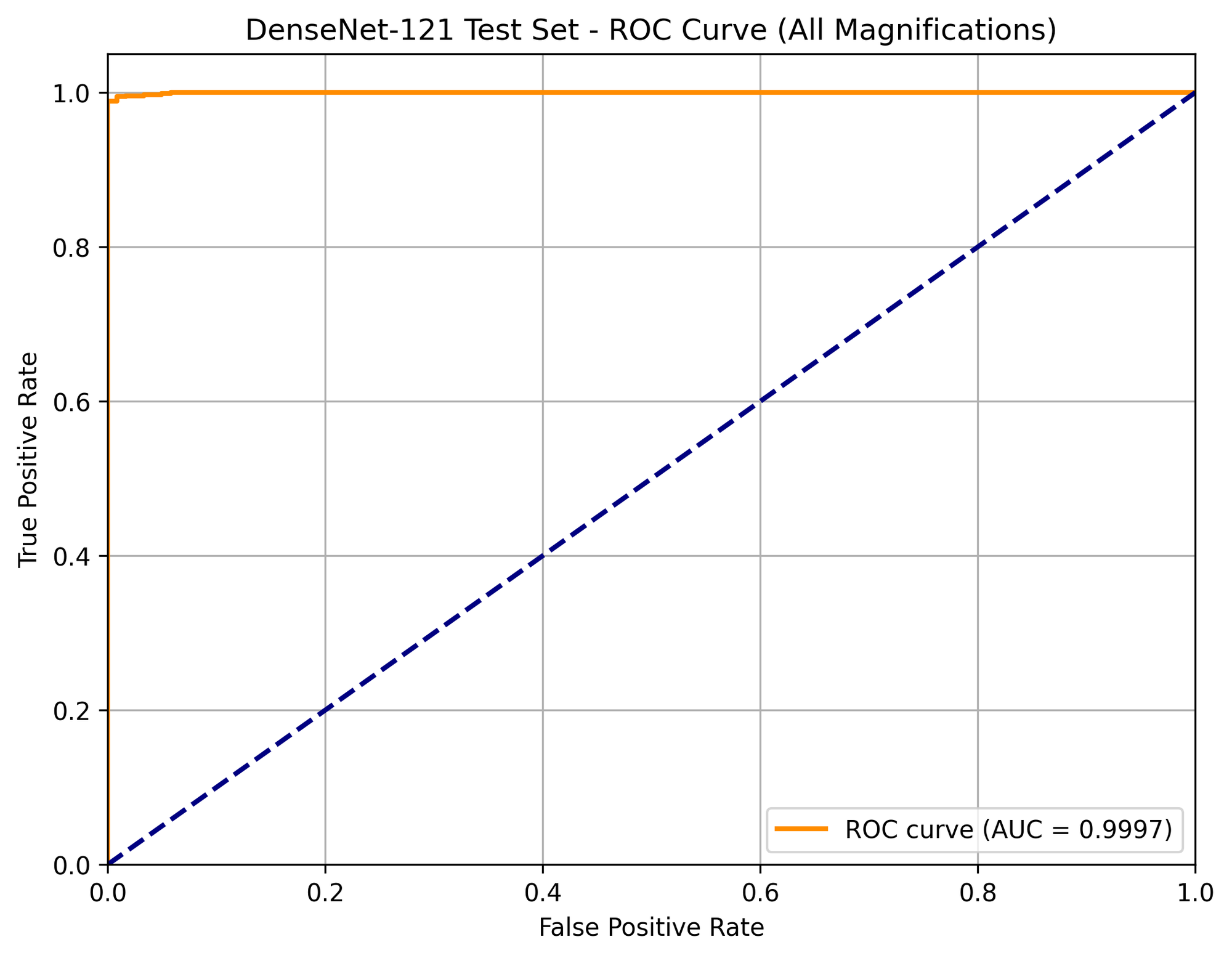}
}\hfill
\subfloat[Test Confusion Matrix]{%
 \includegraphics[width=0.23\textwidth]{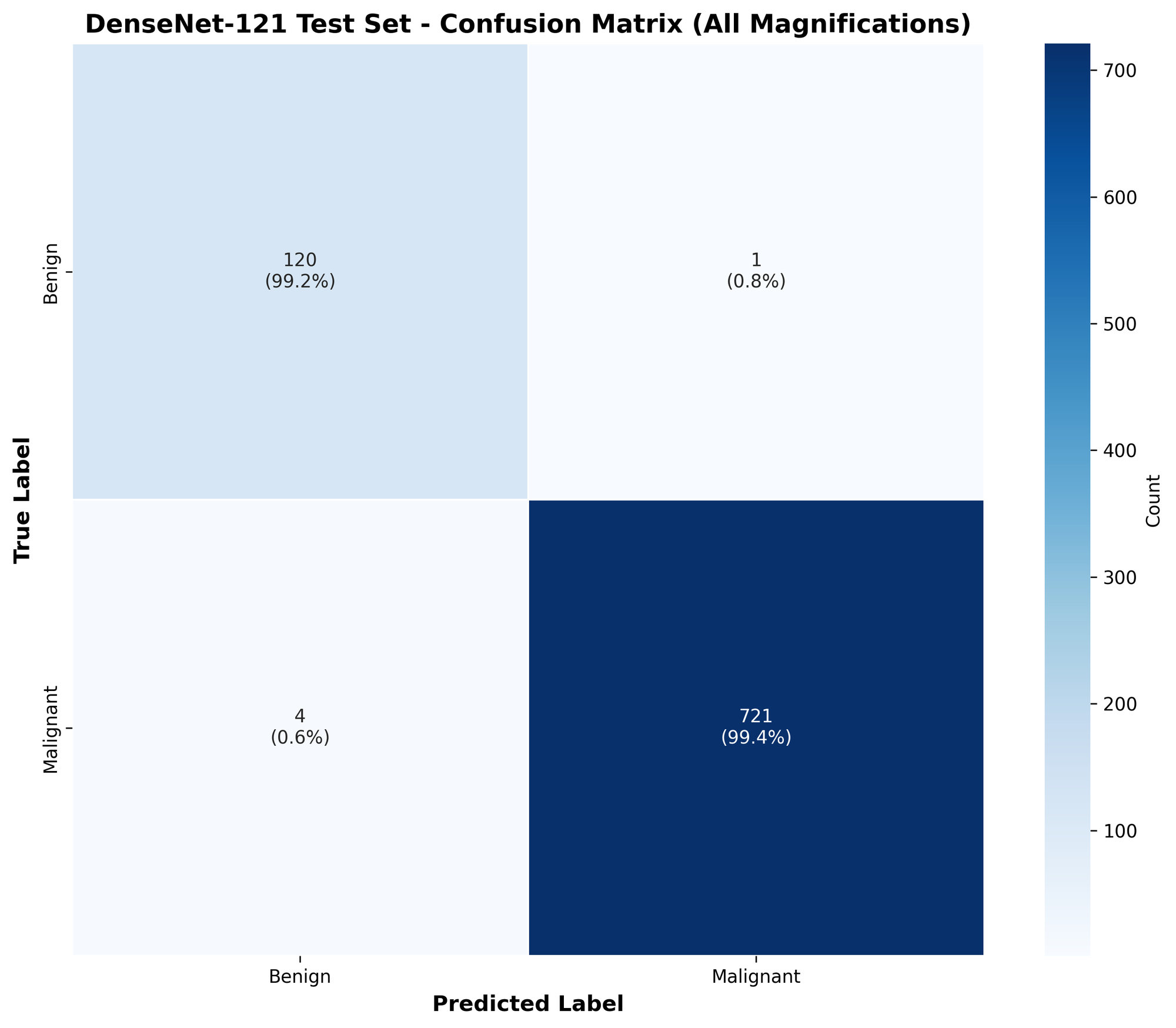}
}

\caption{Validation and test performance of the DenseNet-121 model on the BreakHis dataset across all magnifications.}
\label{fig:densenet121_val_test}
\end{figure}

\begin{figure}[H]
\centering

\subfloat[ROC 40X]{%
 \includegraphics[width=0.23\textwidth]{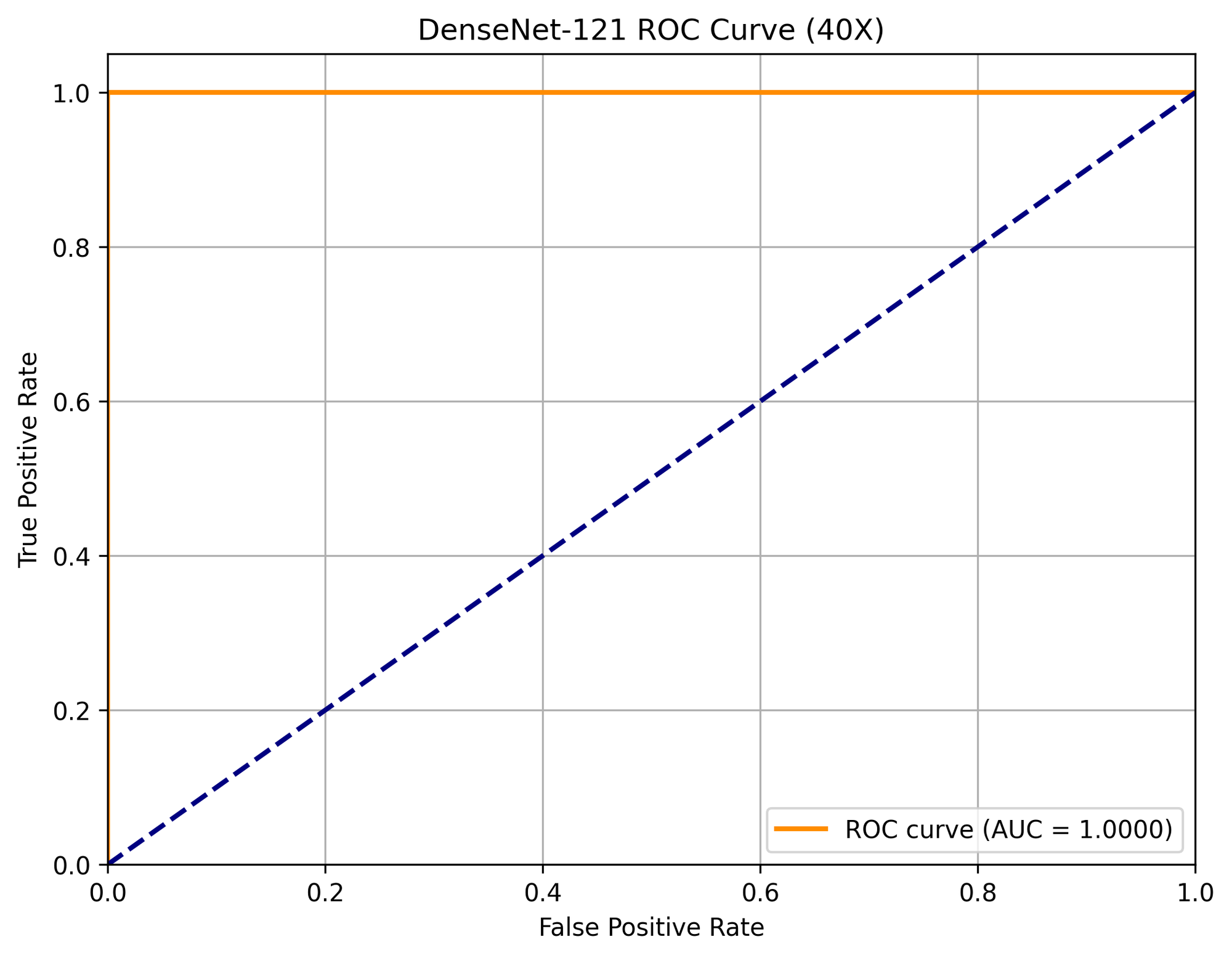}
}\hfill
\subfloat[ROC 100X]{%
 \includegraphics[width=0.23\textwidth]{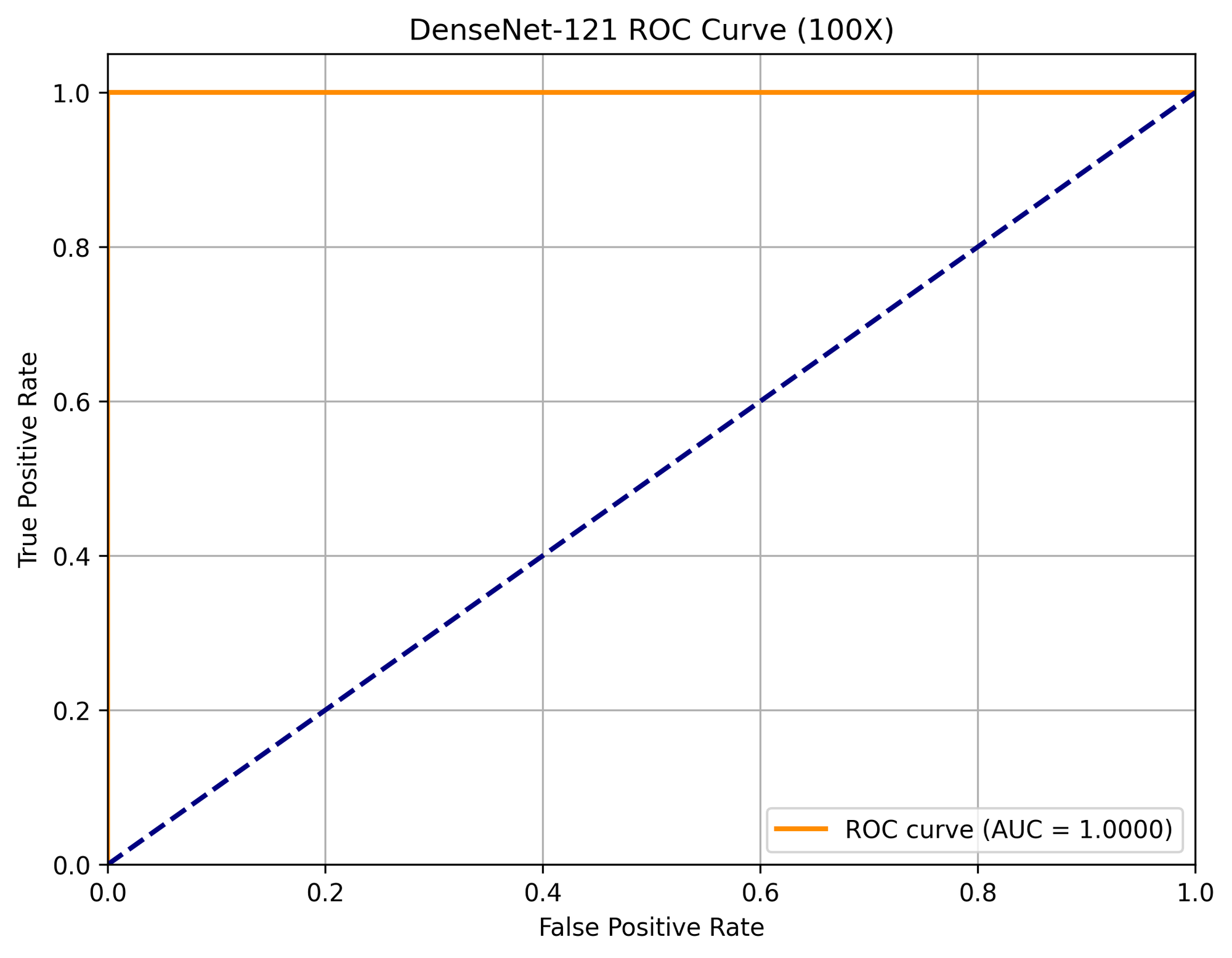}
}\hfill
\subfloat[ROC 200X]{%
 \includegraphics[width=0.23\textwidth]{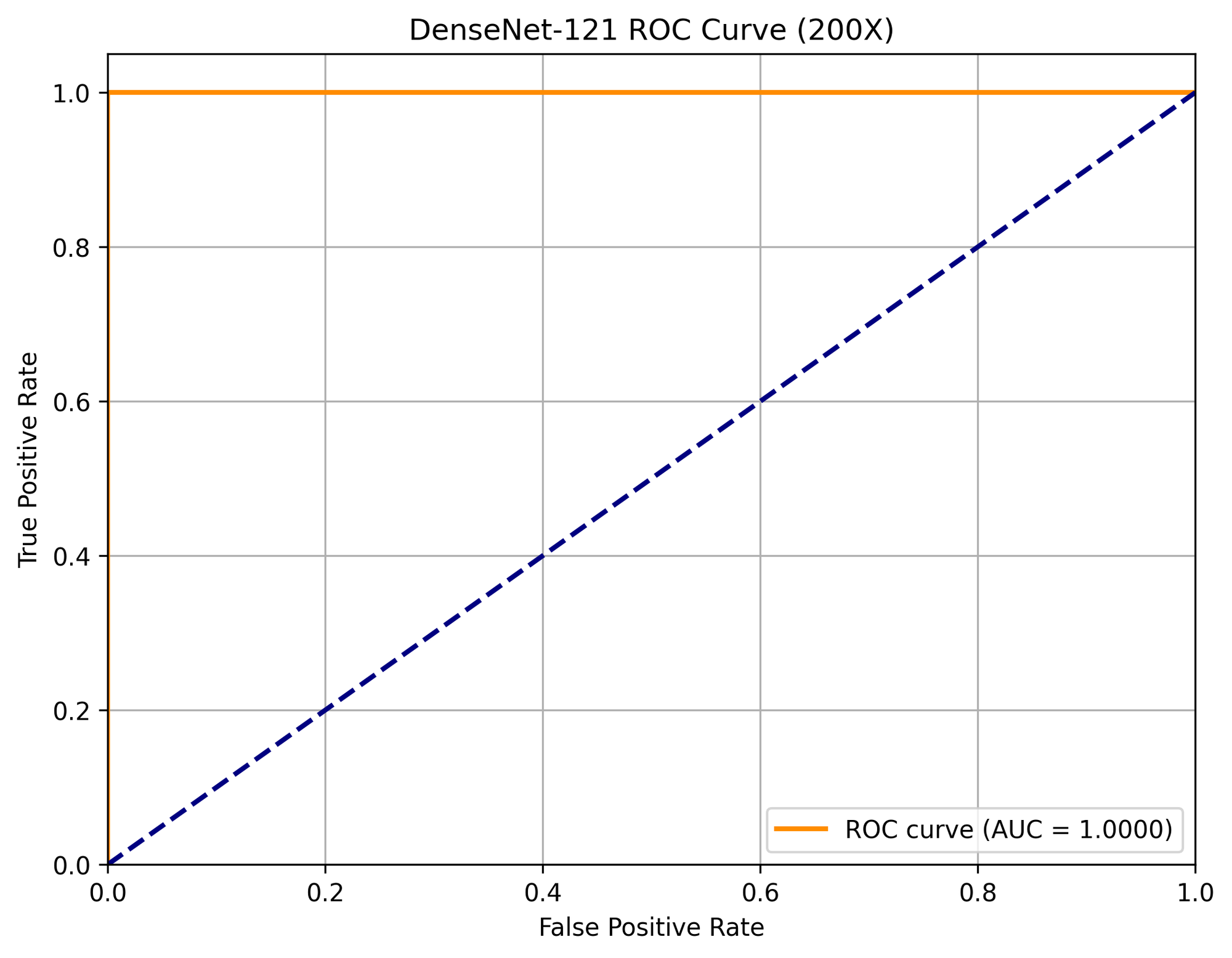}
}\hfill
\subfloat[ROC 400X]{%
 \includegraphics[width=0.23\textwidth]{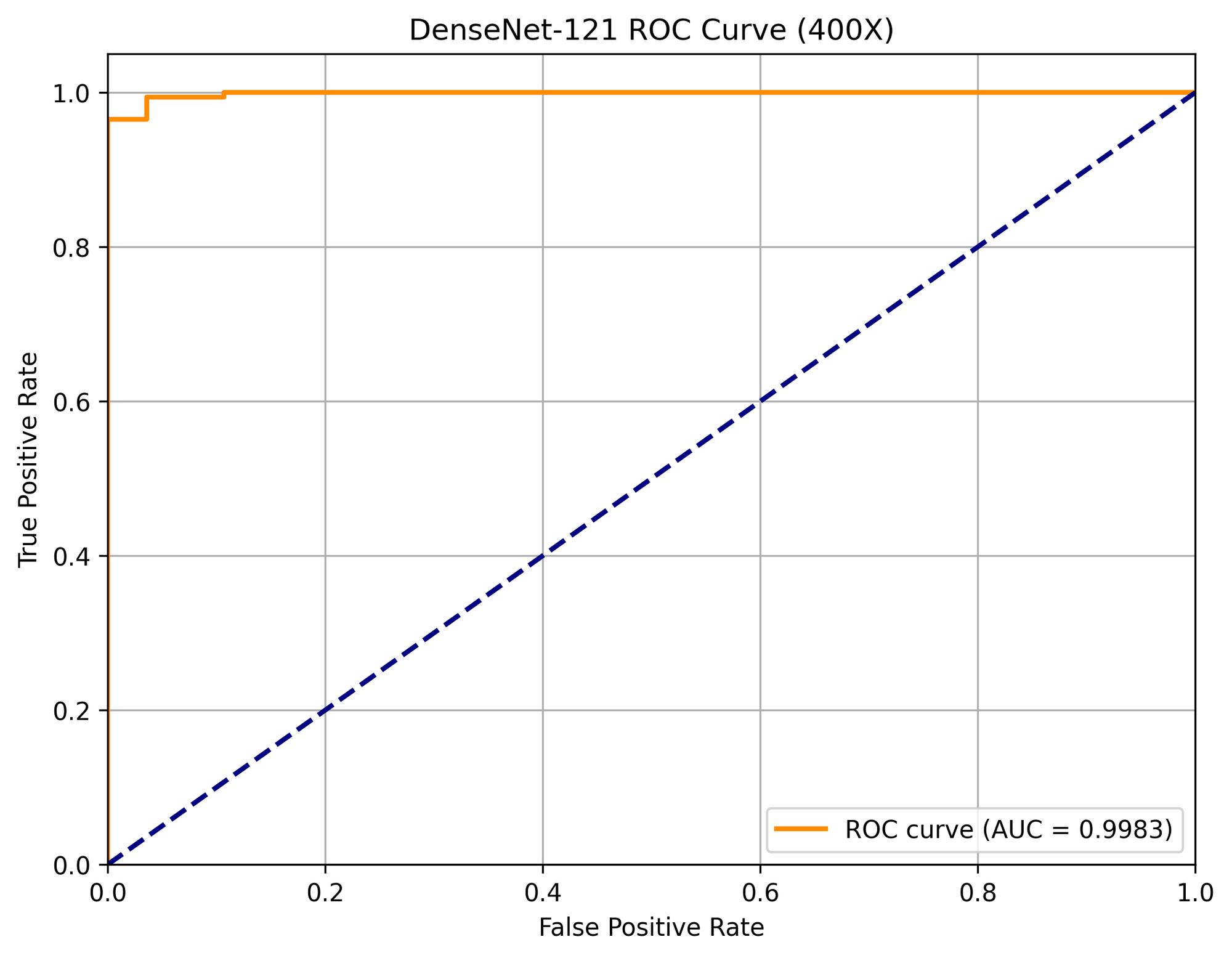}
}

\vspace{0.6ex}

\subfloat[PR 40X]{%
 \includegraphics[width=0.23\textwidth]{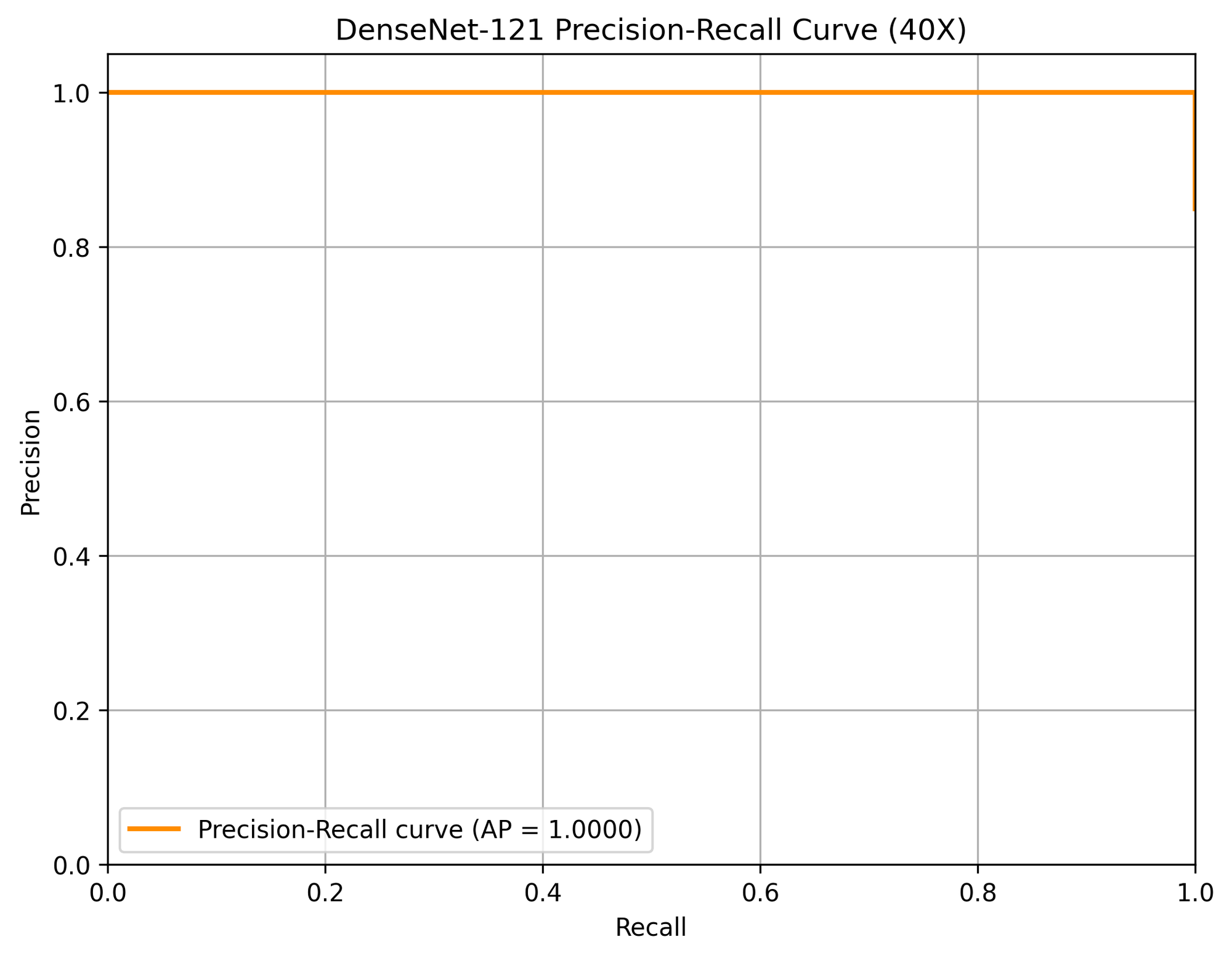}
}\hfill
\subfloat[PR 100X]{%
 \includegraphics[width=0.23\textwidth]{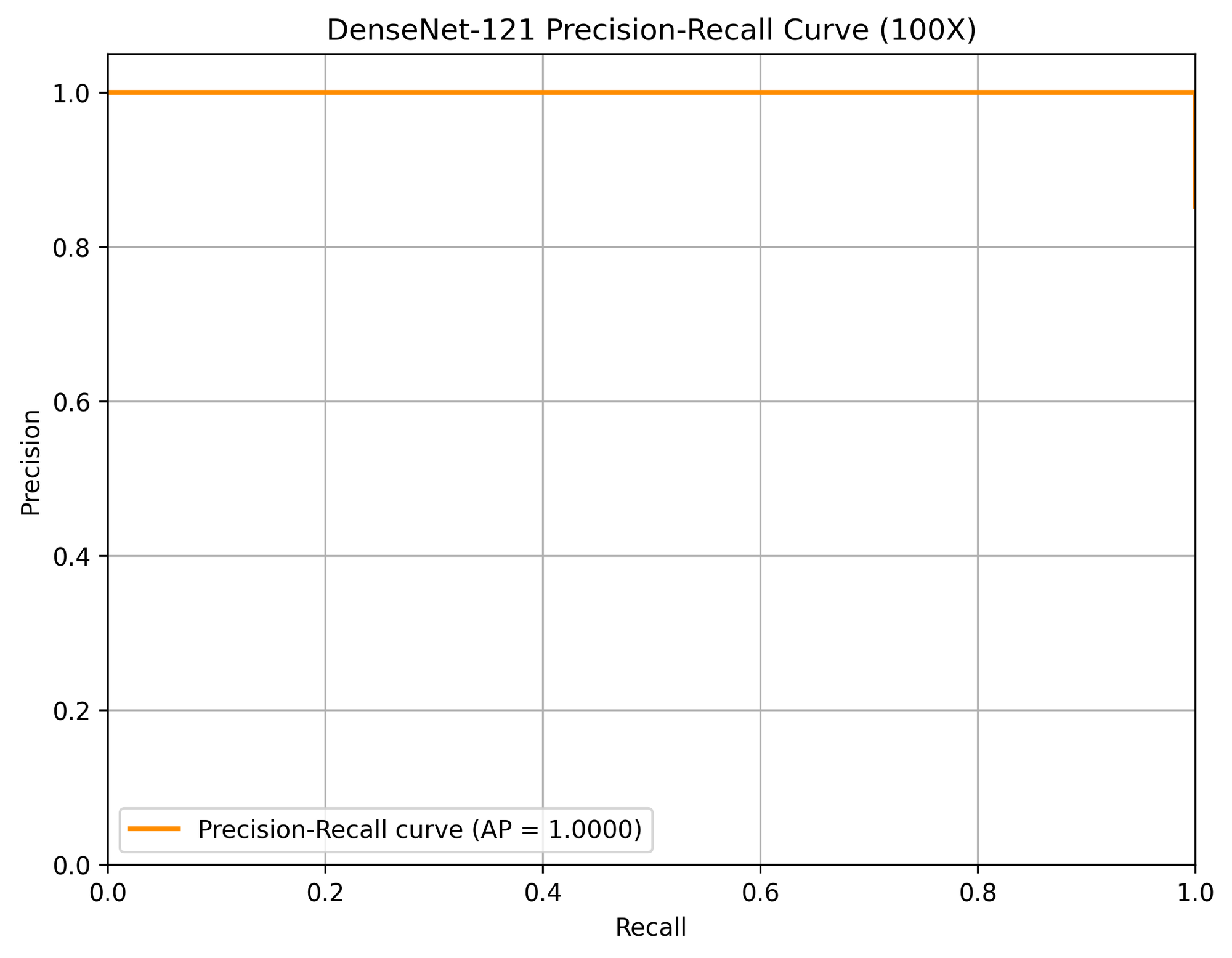}
}\hfill
\subfloat[PR 200X]{%
 \includegraphics[width=0.23\textwidth]{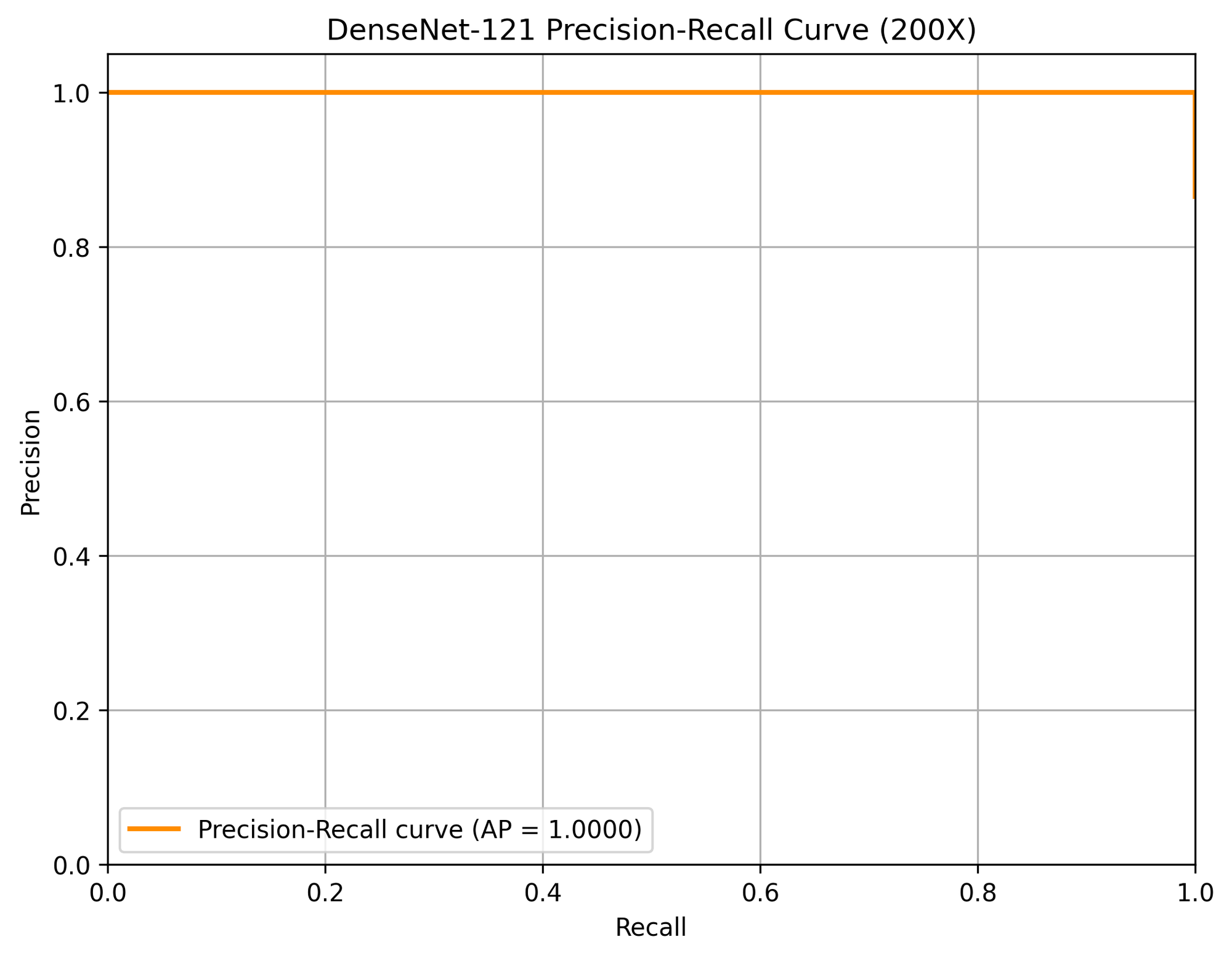}
}\hfill
\subfloat[PR 400X]{%
 \includegraphics[width=0.23\textwidth]{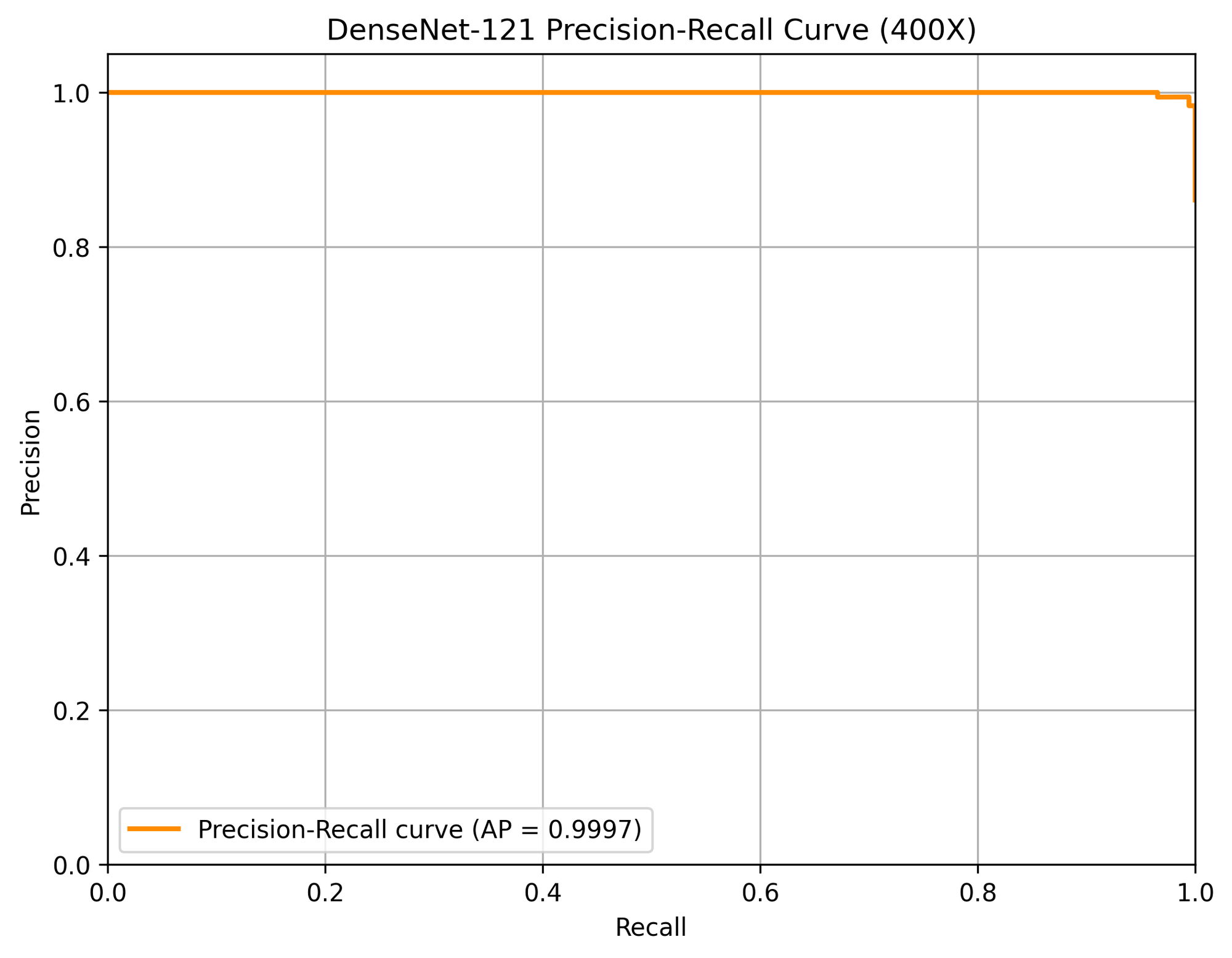}
}

\caption{Receiver Operating Characteristic (ROC) curves (a–d) and Precision–Recall (PR) curves (e–h) for the DenseNet-121 model evaluated on the BreakHis dataset across 40X, 100X, 200X, and 400X magnifications.}
\label{fig:densenet121_roc_pr}
\end{figure}

\begin{table}[ht]
\caption{Performance of EfficientNet on the BreakHis dataset}
\label{tab:efficientnet}
\centering
\begin{tabular}{@{}lcccccc@{}}
\toprule
\textbf{Dataset} 
& \textbf{Accuracy (\%)} 
& \textbf{Precision (\%)} 
& \textbf{Recall (\%)} 
& \textbf{F1-score (\%)} 
& \textbf{Specificity (\%)} 
& \textbf{AUC} \\
\midrule
Validation (All Mag.) 
& 93.1 
& 96.1 
& 94.3 
& 95.2 
& 89.7 
& 0.979 \\

Test (All Mag.)    
& 86.2 
& 83.6 
& 86.0 
& 84.8 
& 86.3 
& 0.940 \\

Test (40$\times$)      
& 85.9 
& 87.8 
& 80.4 
& 83.9 
& 90.6 
& 0.946 \\

Test (100$\times$)      
& 86.3 
& 86.4 
& 84.3 
& 85.4 
& 88.1 
& 0.941 \\

Test (200$\times$)      
& 85.5 
& 83.2 
& 83.2 
& 83.2 
& 87.3 
& 0.959 \\

Test (400$\times$)      
& 84.3 
& 78.7 
& 86.7 
& 82.5 
& 82.4 
& 0.915 \\
\bottomrule
\end{tabular}
\end{table}

On the BreakHis dataset, EfficientNet performed well though with a little bit of variation. With a recall of 94.31\%, a specificity of 89.74\%, an F1-score of 95.18\%, and an area under the curve (AUC-ROC) of 0.979 on the validation set (at all magnifications), it demonstrated strong generalizability and a balanced sensitivity-specificity trade-off (Table 2). Nevertheless, a decline in performance to 86.16\% accuracy, 83.6\% precision, 85.99\% recall, 86.29\% specificity, 84.78\% F1-score, and 0.9397 AUC-ROC was seen on the test set (at all magnifications), indicating difficulties with generalization (Figure 5). 40X (85.9\% accuracy, 94.64\% AUC), 100X (86.27\% accuracy, 94.13\% AUC), 200X (85.52\% accuracy, 95.87\% AUC), and 400X (84.28\% accuracy, 91.5\% AUC) were all results that were consistent but below validation (Figure 6 \& 7). These results demonstrate that EfficientNet is efficient and has good feature extraction capabilities, but they also show that it is sensitive to variance in magnification and that test performance is slightly lower than validation (Figure 8). This suggests that there may be overfitting in this dataset and that more optimization is needed.

\begin{figure}[H]
 \centering
 \includegraphics[width=0.75\linewidth]{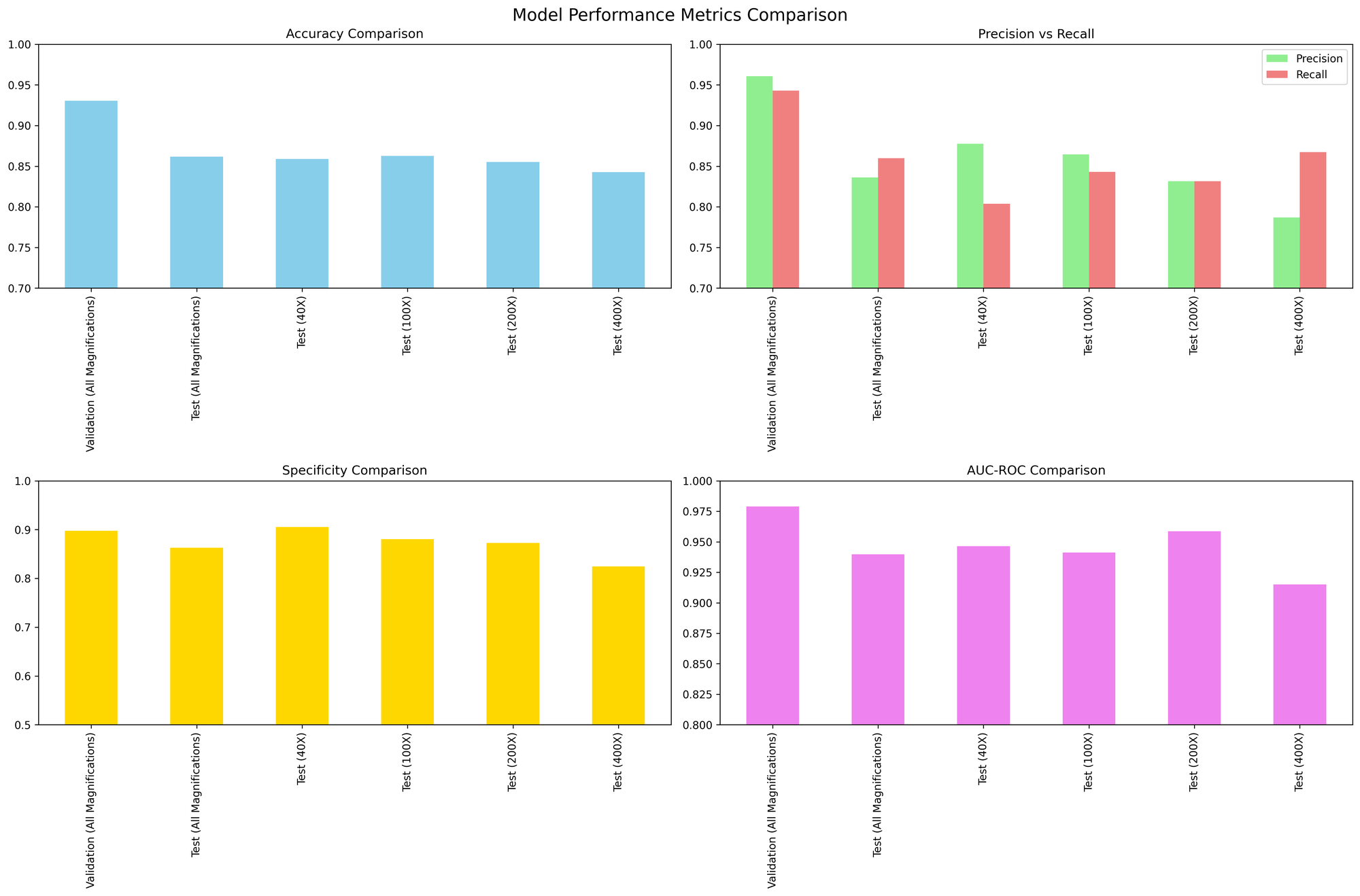}
 \caption{Model performance metrics on the BreakHis dataset across validation and test sets at different magnifications.}
 \label{fig:single1}
\end{figure}

\begin{figure}[H]
\centering

\subfloat[40X]{%
 \includegraphics[width=0.23\textwidth]{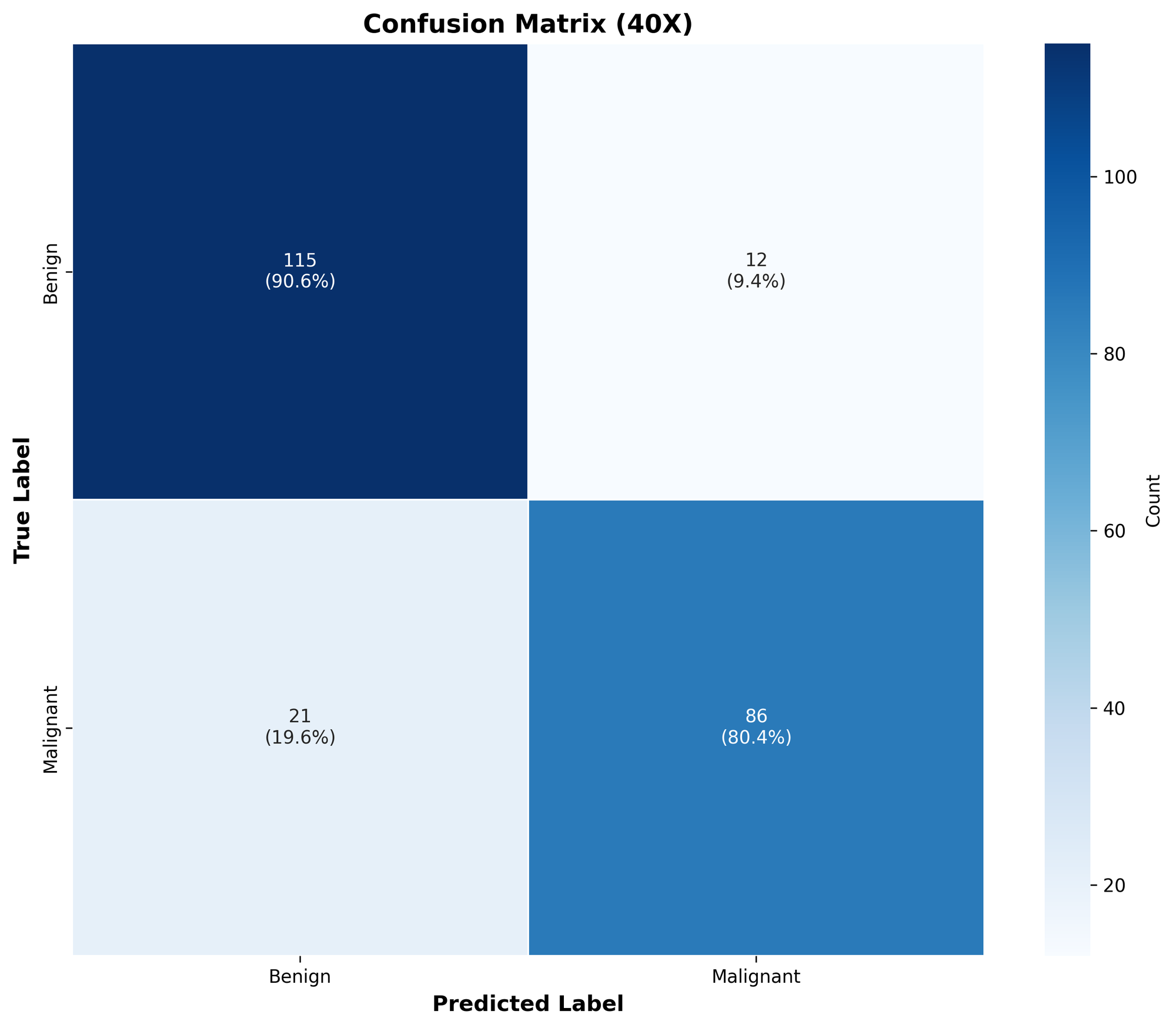}
}\hfill
\subfloat[100X]{%
 \includegraphics[width=0.23\textwidth]{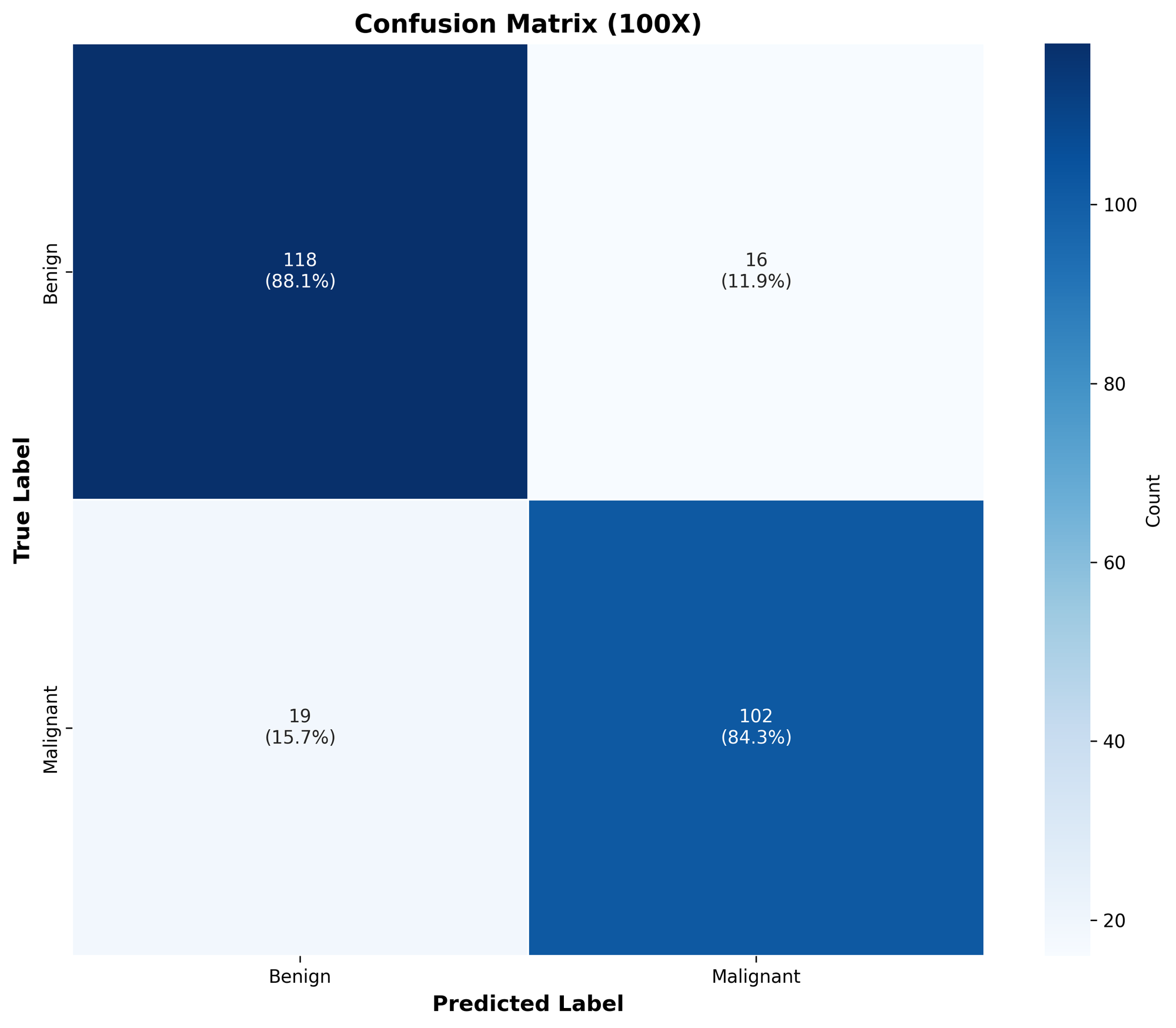}
}\hfill
\subfloat[200X]{%
 \includegraphics[width=0.23\textwidth]{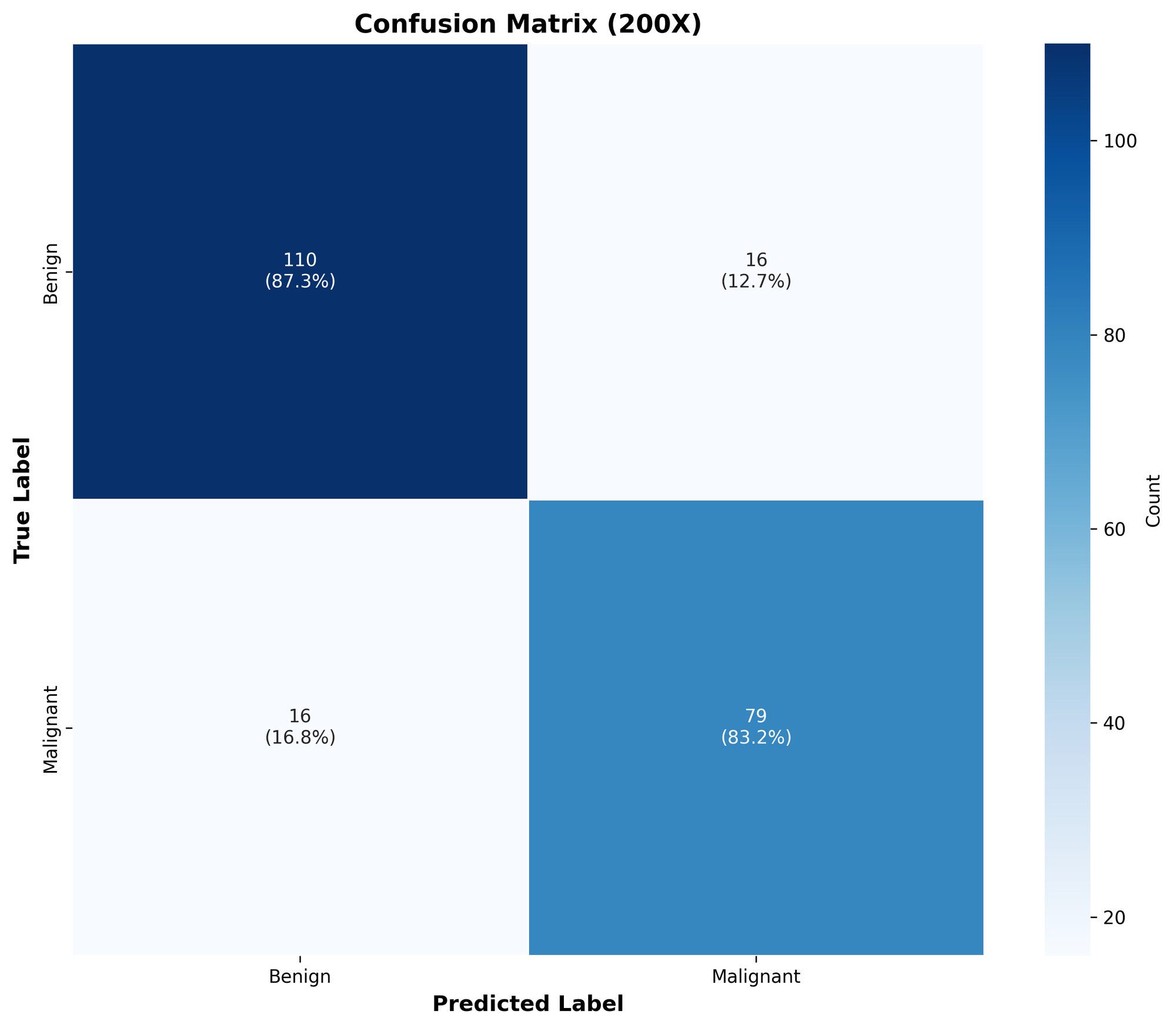}
}\hfill
\subfloat[400X]{%
 \includegraphics[width=0.23\textwidth]{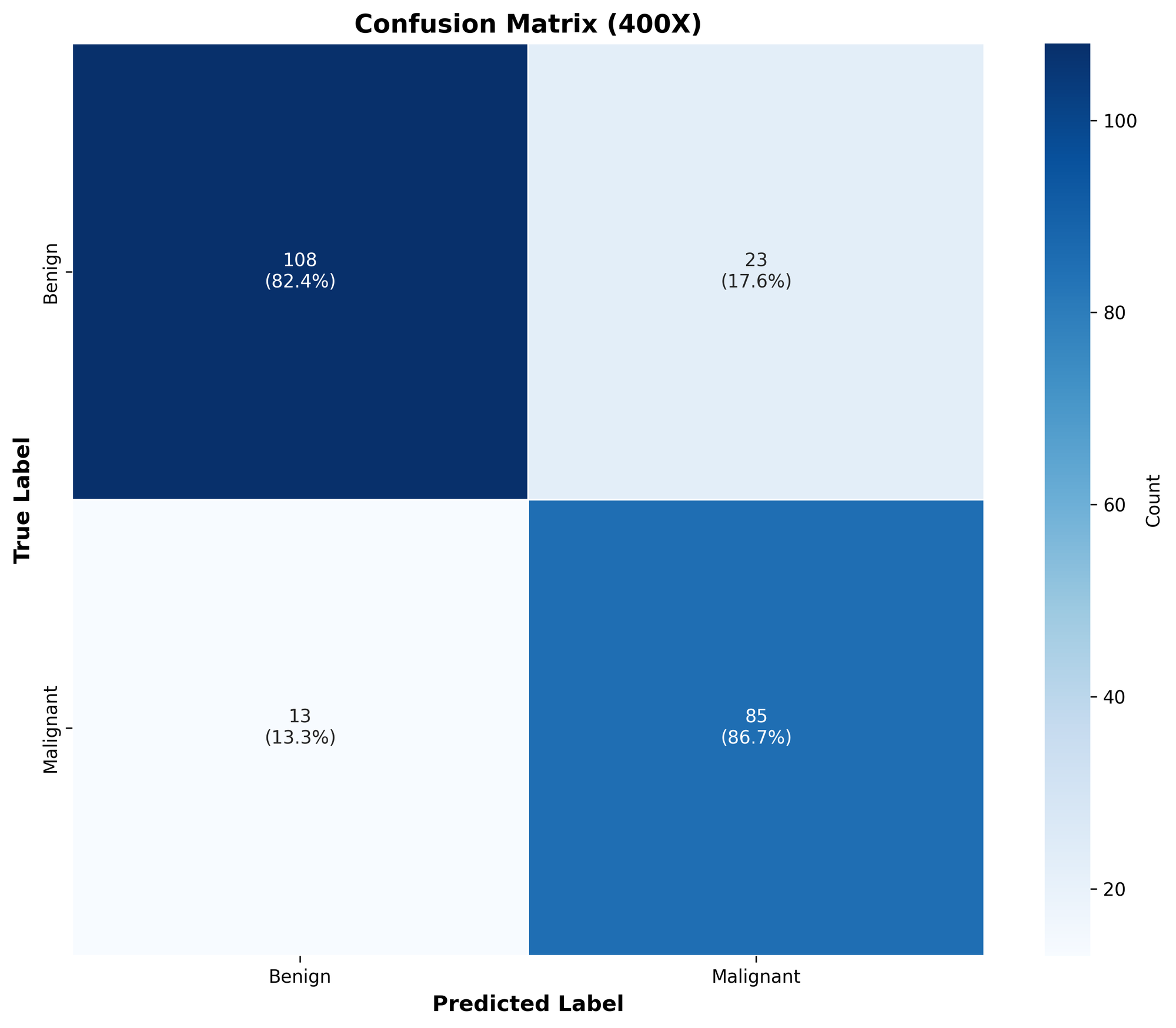}
}

\caption{Confusion matrices of the EfficientNet model across 40X, 100X, 200X, and 400X magnifications on the BreakHis dataset.}
\label{fig:efficientnet_confusion_matrices}
\end{figure}

\begin{figure}[H]
\centering

\subfloat[Validation ROC]{%
 \includegraphics[width=0.23\textwidth]{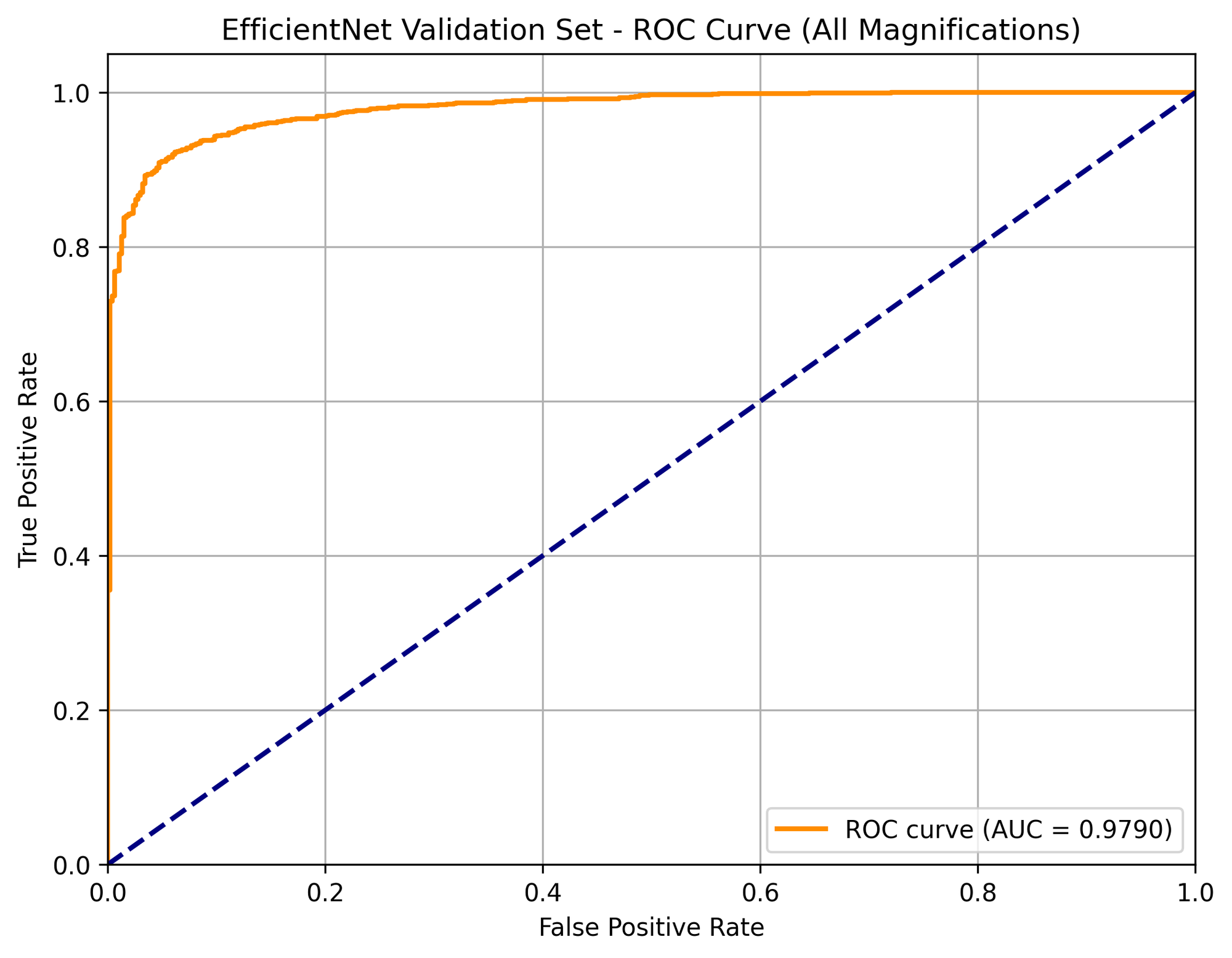}
}\hfill
\subfloat[Validation Confusion Matrix]{%
 \includegraphics[width=0.23\textwidth]{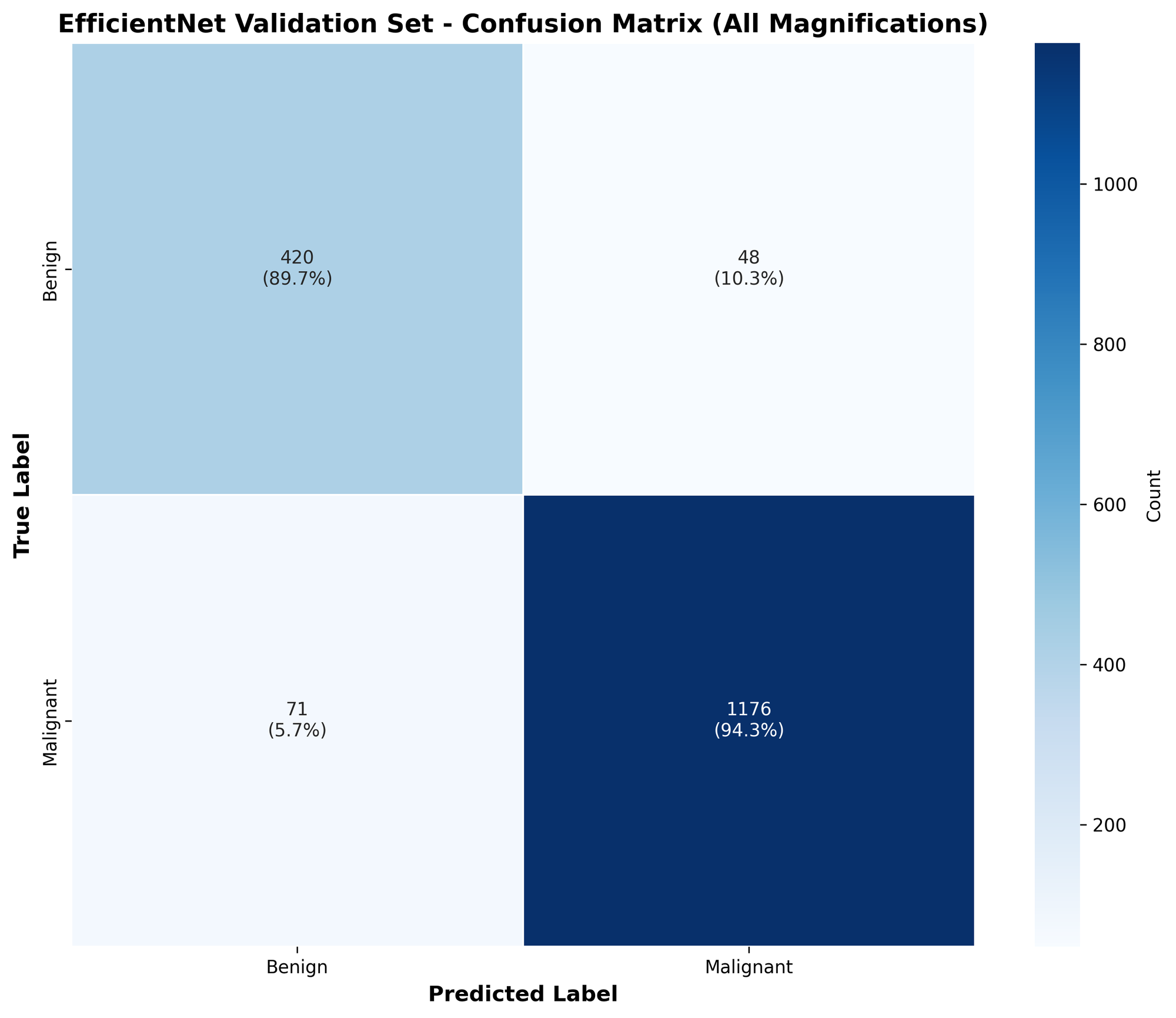}
}\hfill
\subfloat[Test ROC]{%
 \includegraphics[width=0.23\textwidth]{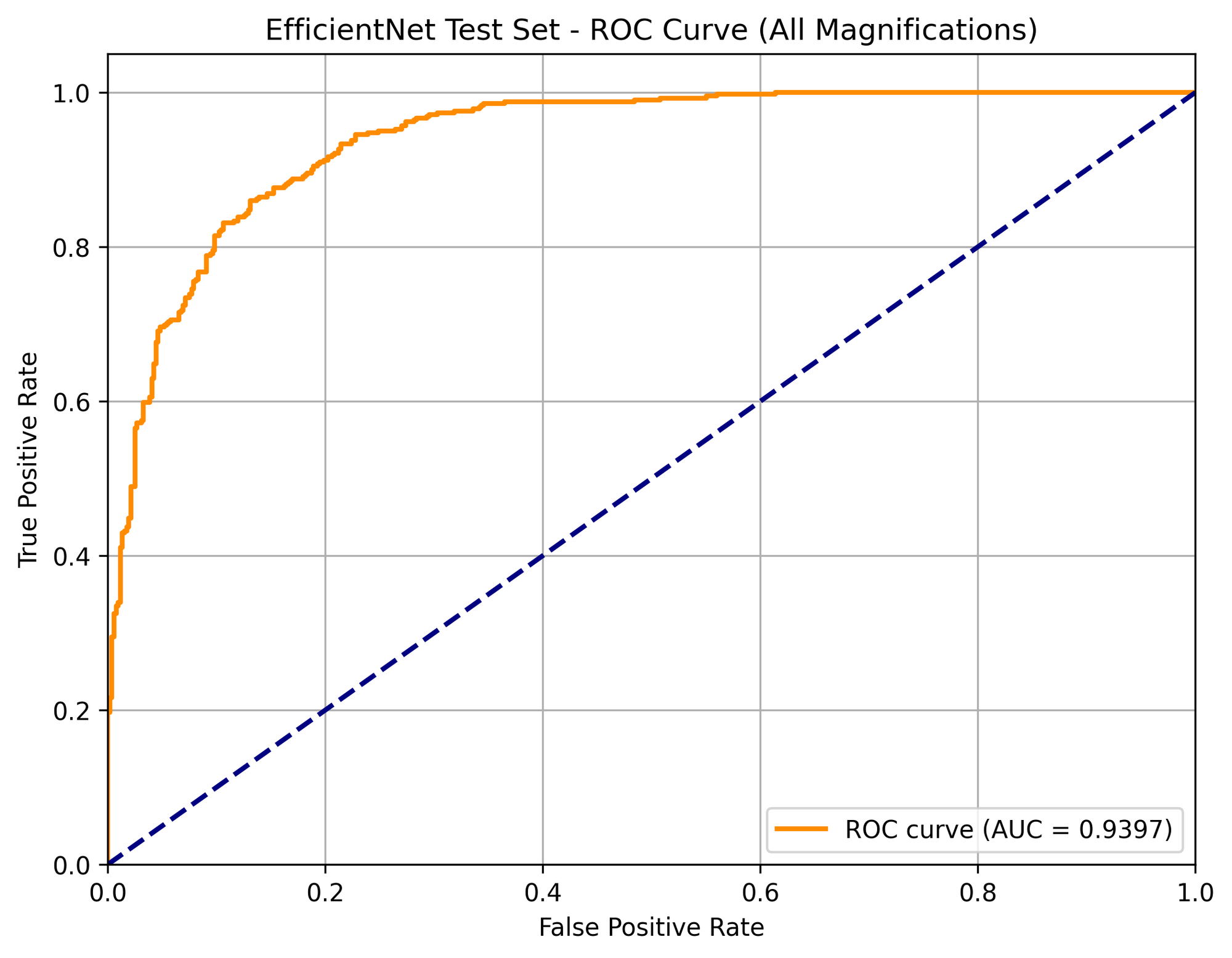}
}\hfill
\subfloat[Test Confusion Matrix]{%
 \includegraphics[width=0.23\textwidth]{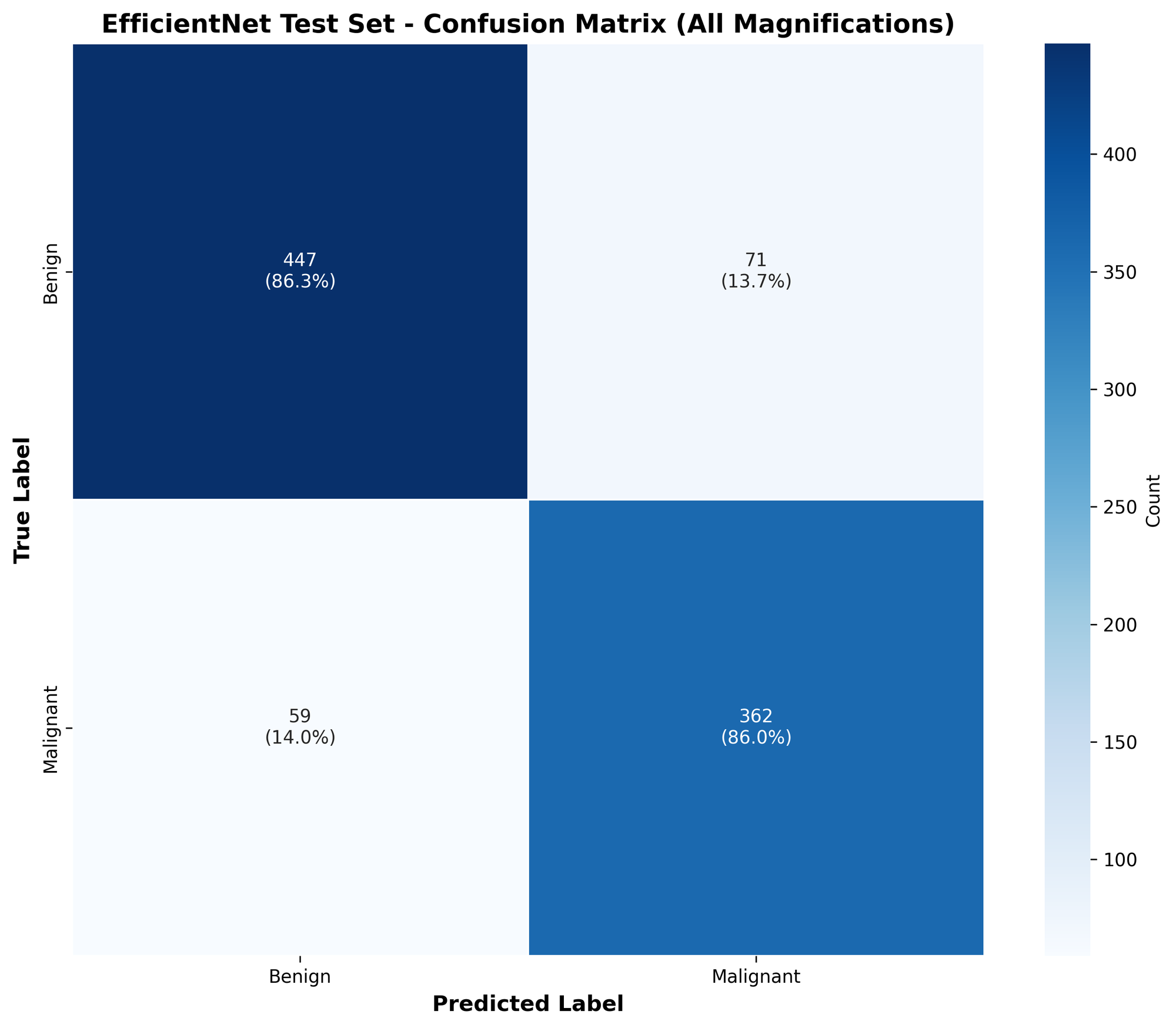}
}

\caption{Validation and test performance of the EfficientNet model on the BreakHis dataset across all magnifications.}
\label{fig:efficientnet_val_test}
\end{figure}

\begin{figure}[H]
\centering

\subfloat[ROC 40X]{%
 \includegraphics[width=0.23\textwidth]{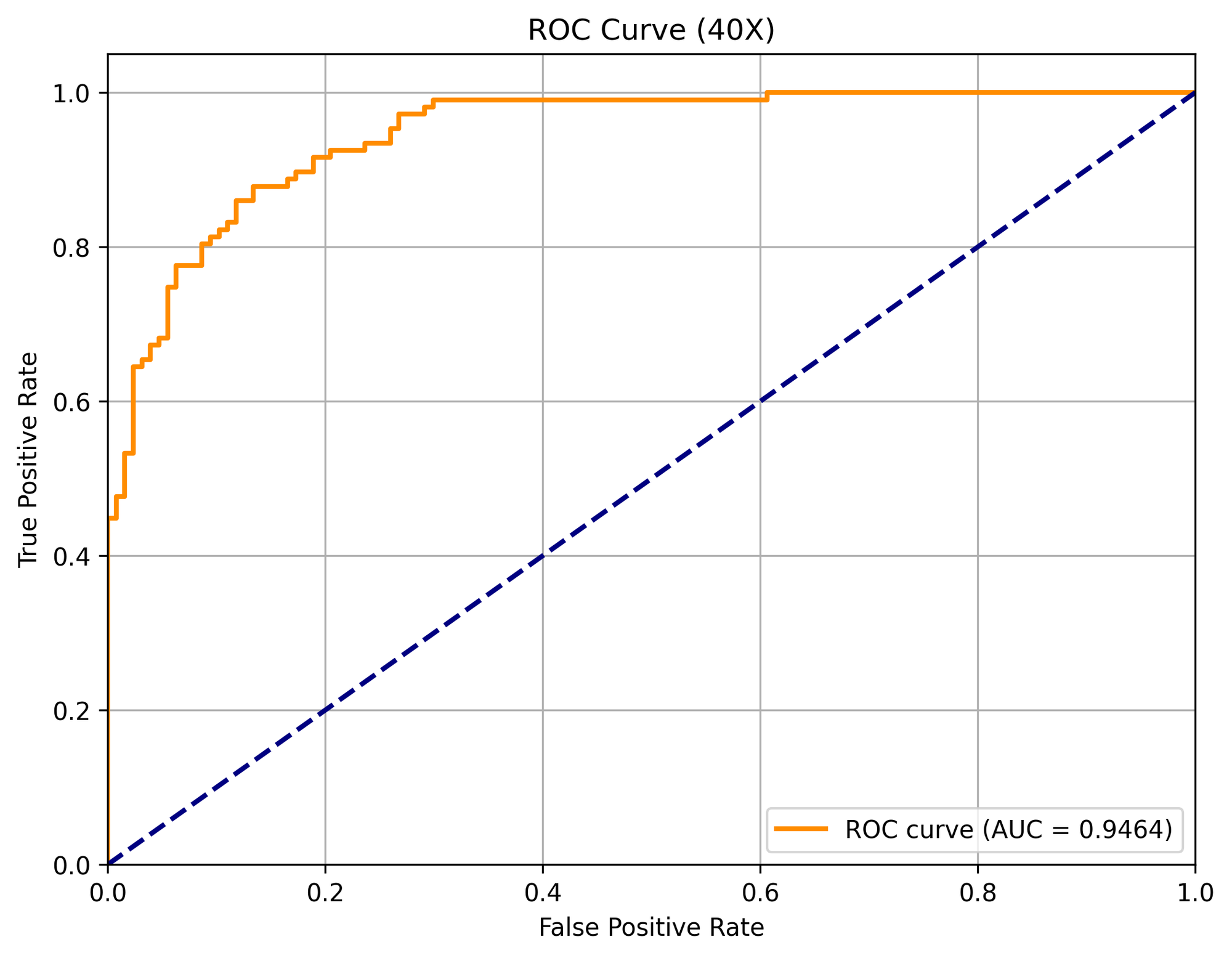}
}\hfill
\subfloat[ROC 100X]{%
 \includegraphics[width=0.23\textwidth]{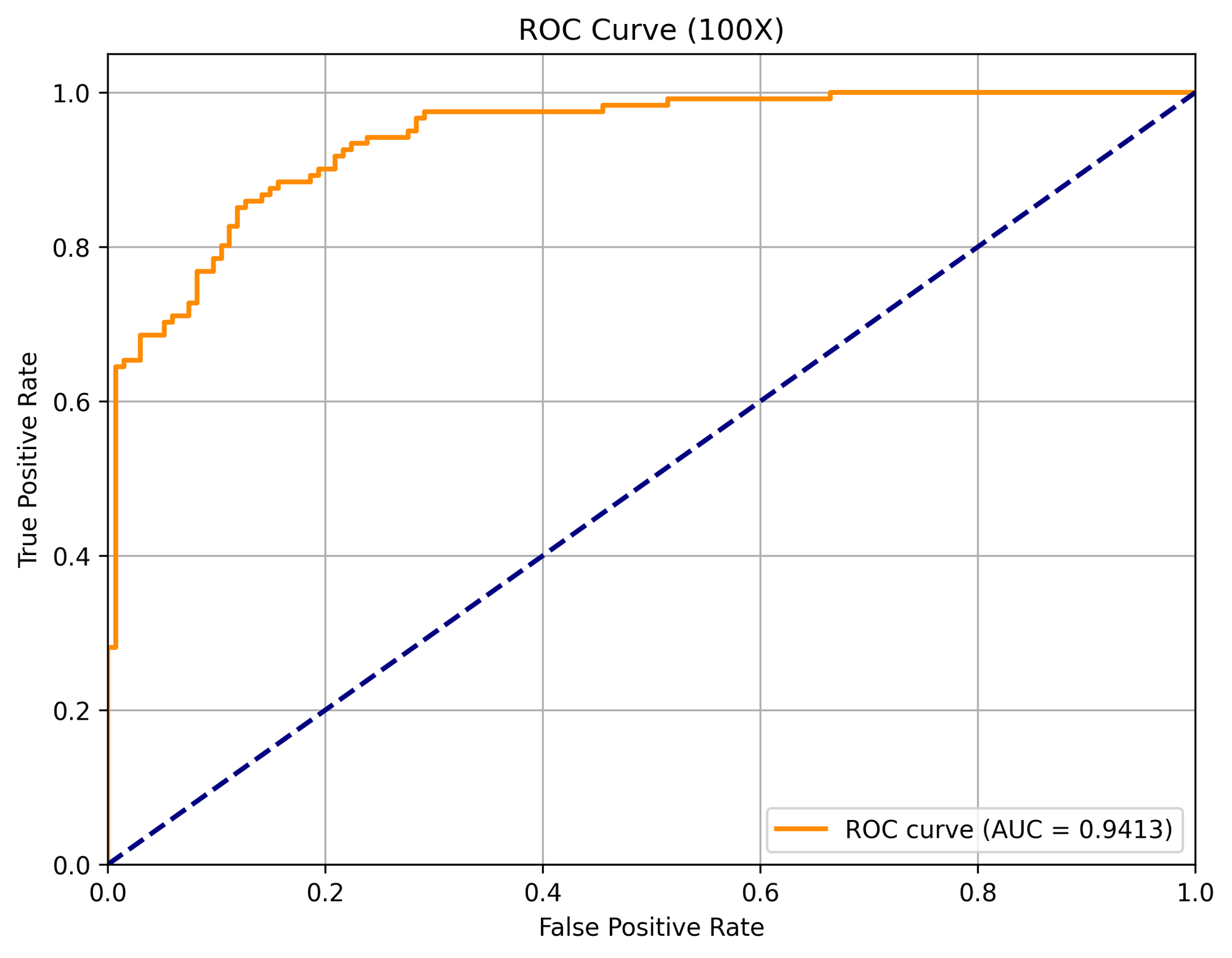}
}\hfill
\subfloat[ROC 200X]{%
 \includegraphics[width=0.23\textwidth]{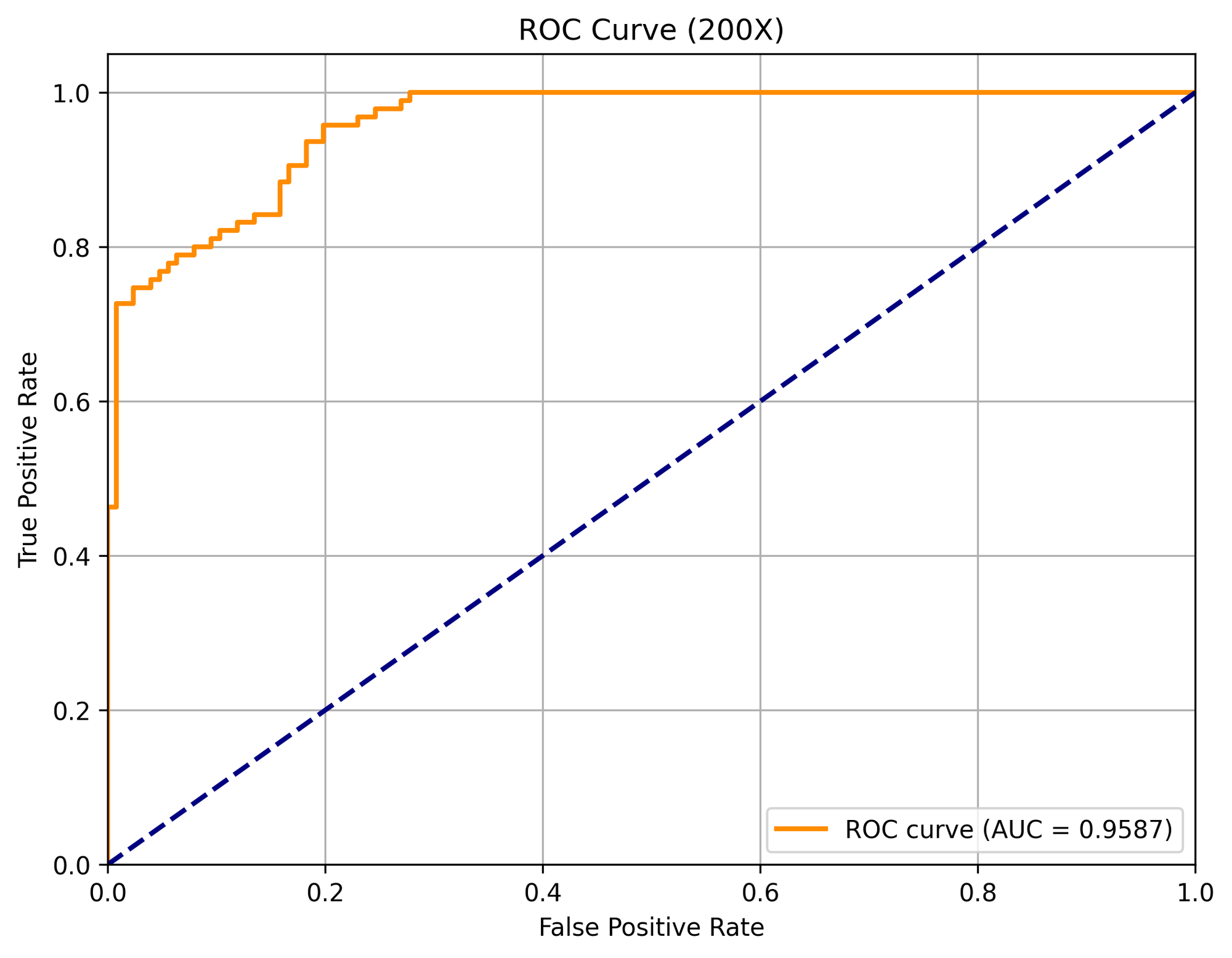}
}\hfill
\subfloat[ROC 400X]{%
 \includegraphics[width=0.23\textwidth]{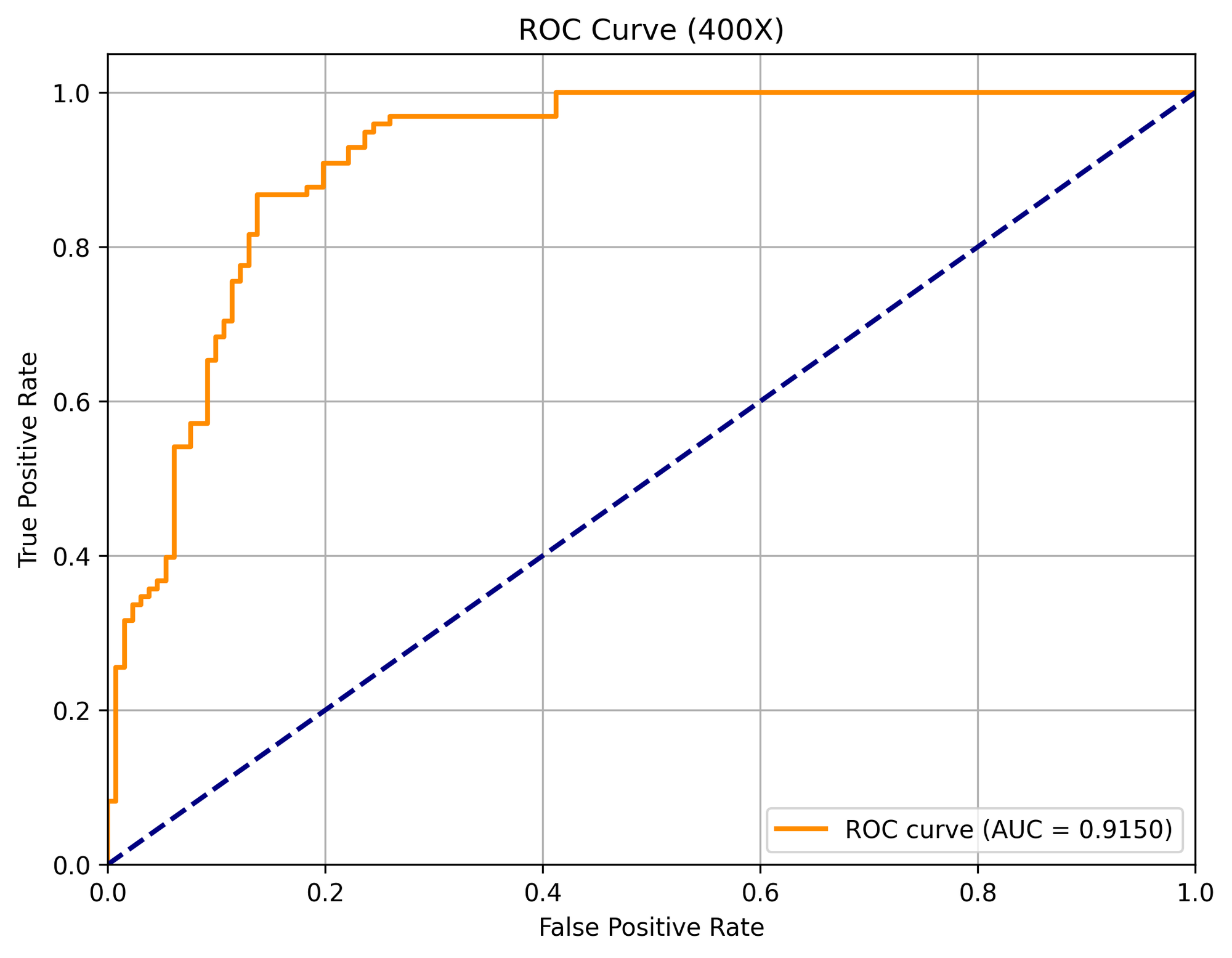}
}

\vspace{0.6ex}

\subfloat[PR 40X]{%
 \includegraphics[width=0.23\textwidth]{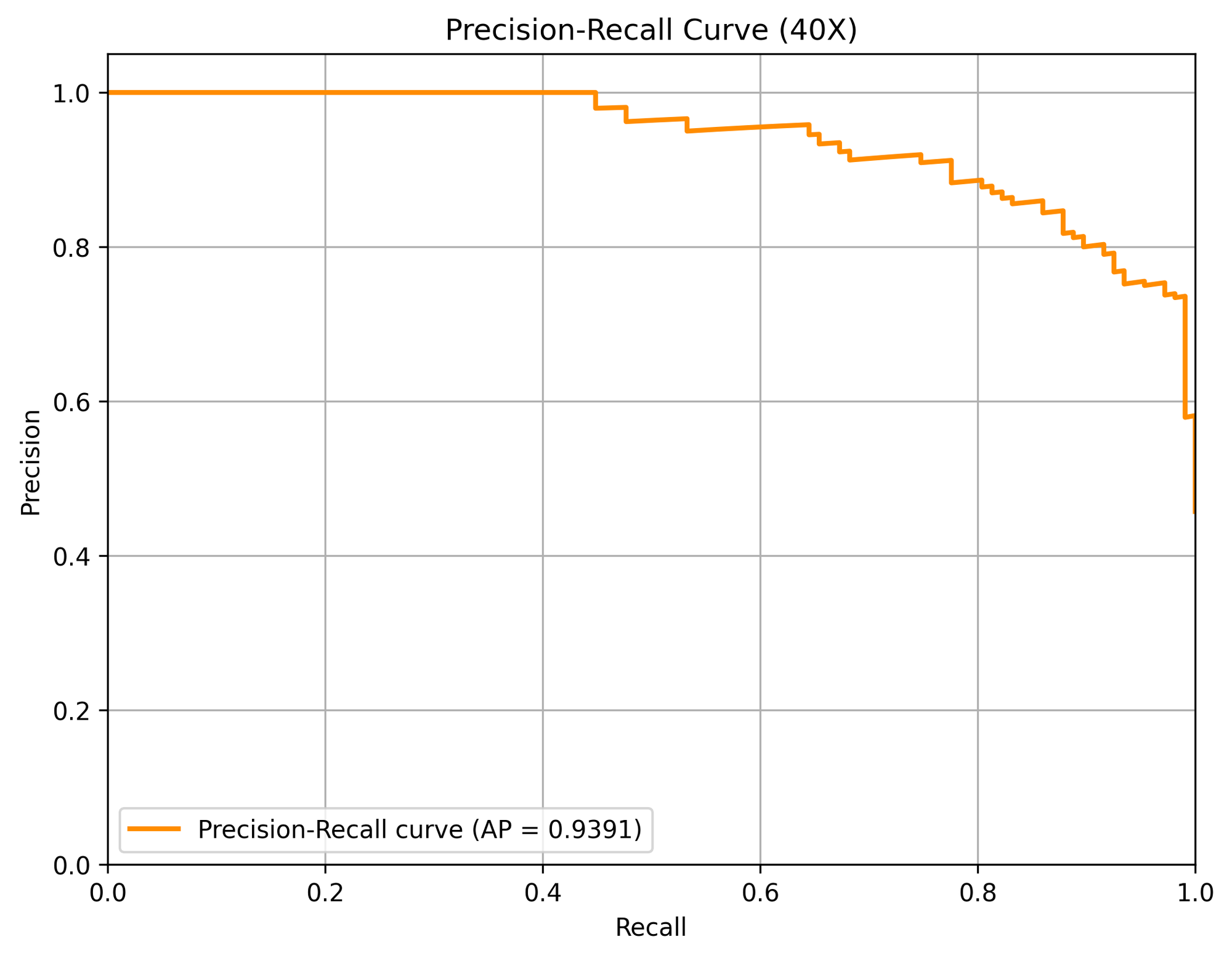}
}\hfill
\subfloat[PR 100X]{%
 \includegraphics[width=0.23\textwidth]{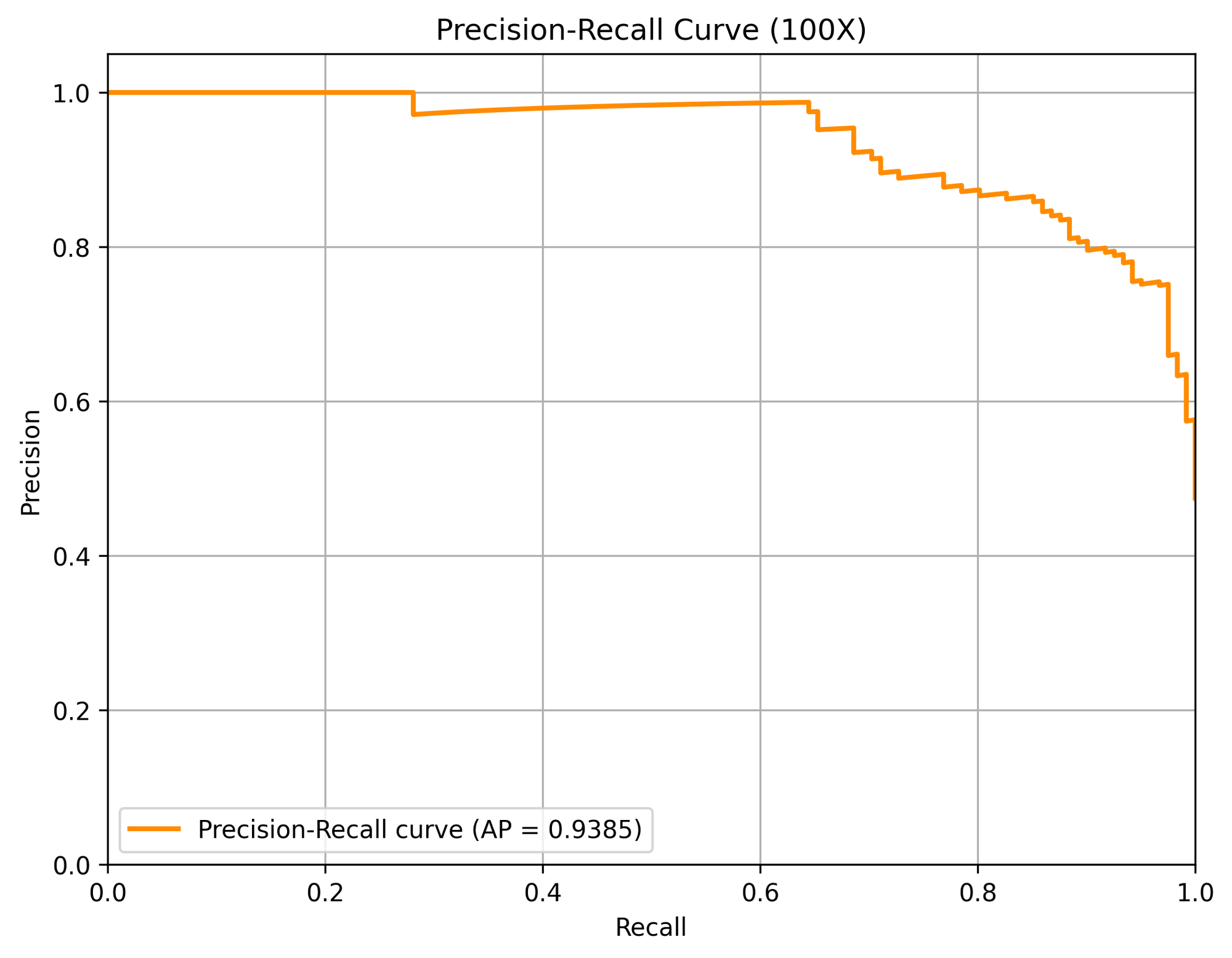}
}\hfill
\subfloat[PR 200X]{%
 \includegraphics[width=0.23\textwidth]{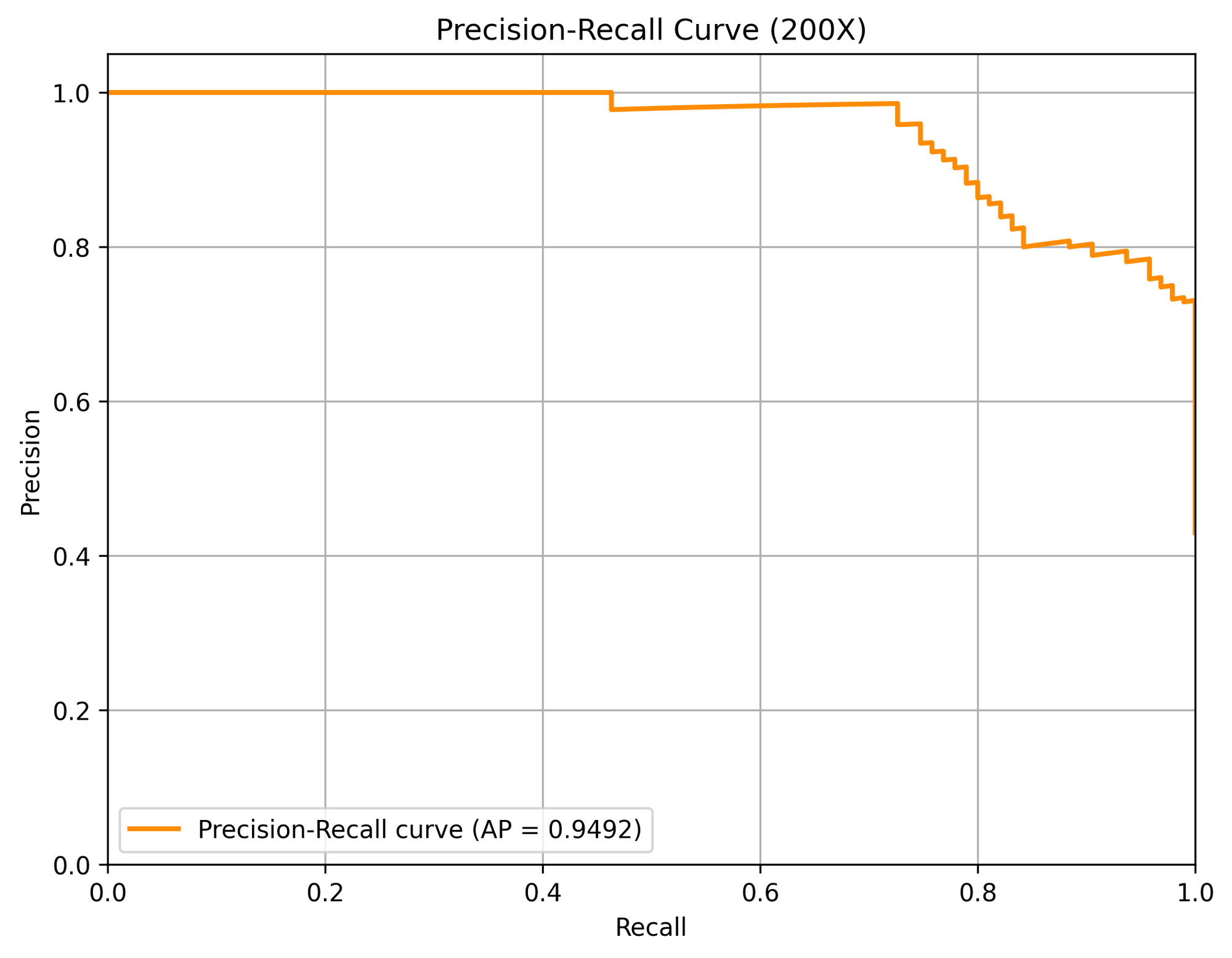}
}\hfill
\subfloat[PR 400X]{%
 \includegraphics[width=0.23\textwidth]{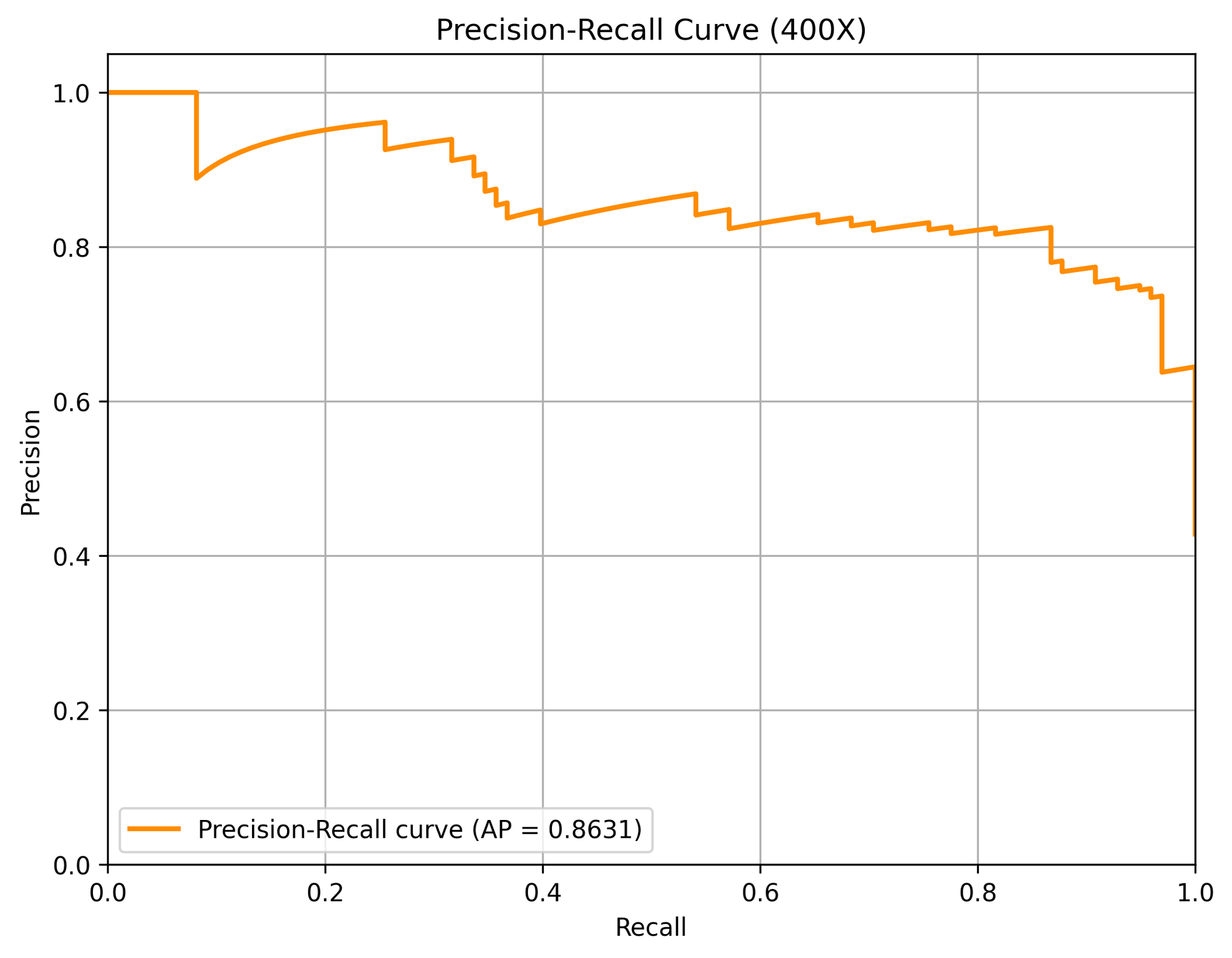}
}

\caption{Receiver Operating Characteristic (ROC) curves (a–d) and Precision–Recall (PR) curves (e–h) for the EfficientNet model evaluated on the BreakHis dataset across 40X, 100X, 200X, and 400X magnifications.}
\label{fig:efficientnet_roc_pr}
\end{figure}

\begin{table}[ht]
\caption{Performance of ConvNeXt on the BreakHis dataset}
\label{tab:convnext}
\centering
\begin{tabular}{@{}lcccccc@{}}
\toprule
\textbf{Dataset} 
& \textbf{Accuracy (\%)} 
& \textbf{Precision (\%)} 
& \textbf{Recall (\%)} 
& \textbf{F1-score (\%)} 
& \textbf{Specificity (\%)} 
& \textbf{AUC} \\
\midrule
Validation (All Mag.) 
& 80.9 
& 80.6 
& 91.1 
& 85.6 
& 64.2 
& 0.881 \\

Test (All Mag.)    
& 77.1 
& 85.1 
& 89.1 
& 87.0 
& 2.4 
& 0.653 \\

Test (40$\times$)      
& 76.1 
& 83.6 
& 89.5 
& 86.4 
& 0.0 
& 0.450 \\

Test (100$\times$)      
& 79.1 
& 84.7 
& 92.1 
& 88.3 
& 2.7 
& 0.708 \\

Test (200$\times$)      
& 79.7 
& 88.3 
& 88.8 
& 88.6 
& 8.7 
& 0.823 \\

Test (400$\times$)      
& 76.4 
& 85.2 
& 88.0 
& 86.6 
& 3.5 
& 0.780 \\
\bottomrule
\end{tabular}
\end{table}

ConvNeXt showed moderate efficacy on the BreakHis dataset, with the validation set achieving 80.92\% accuracy, 80.62\% precision, 91.14\% recall, 64.23\% specificity, 85.56\% F1-score, and 0.8806 AUC-ROC, indicating strong sensitivity but a diminished capacity to accurately detect benign instances (Table 3). In the test set (across all magnifications), the results exhibited a fall to 77.14\% accuracy, 85.12\% precision, 89.06\% recall, 2.42\% specificity, 87.04\% F1-score, and 0.6533 AUC-ROC, indicating a pronounced bias towards malignant categorization owing to the exceedingly low specificity (Figure 9, 10 \& 11)). ConvNeXt showed consistently high recall ($\geq$.88\%) across various magnifications but had inadequate specificity, especially at 40X (0\%) and only slight enhancements at 100X (2.7\%), 200X (8.7\%), and 400X (3.45\%), despite F1-scores ranging from 86\% to 88\% (Figure 12). The results indicate that although ConvNeXt effectively detects malignant samples owing to its hierarchical convolutional-transformer hybrid architecture, its failure to balance false positives undermines its diagnostic reliability in breast cancer classification.

\begin{figure}[H]
 \centering
 \includegraphics[width=0.75\linewidth]{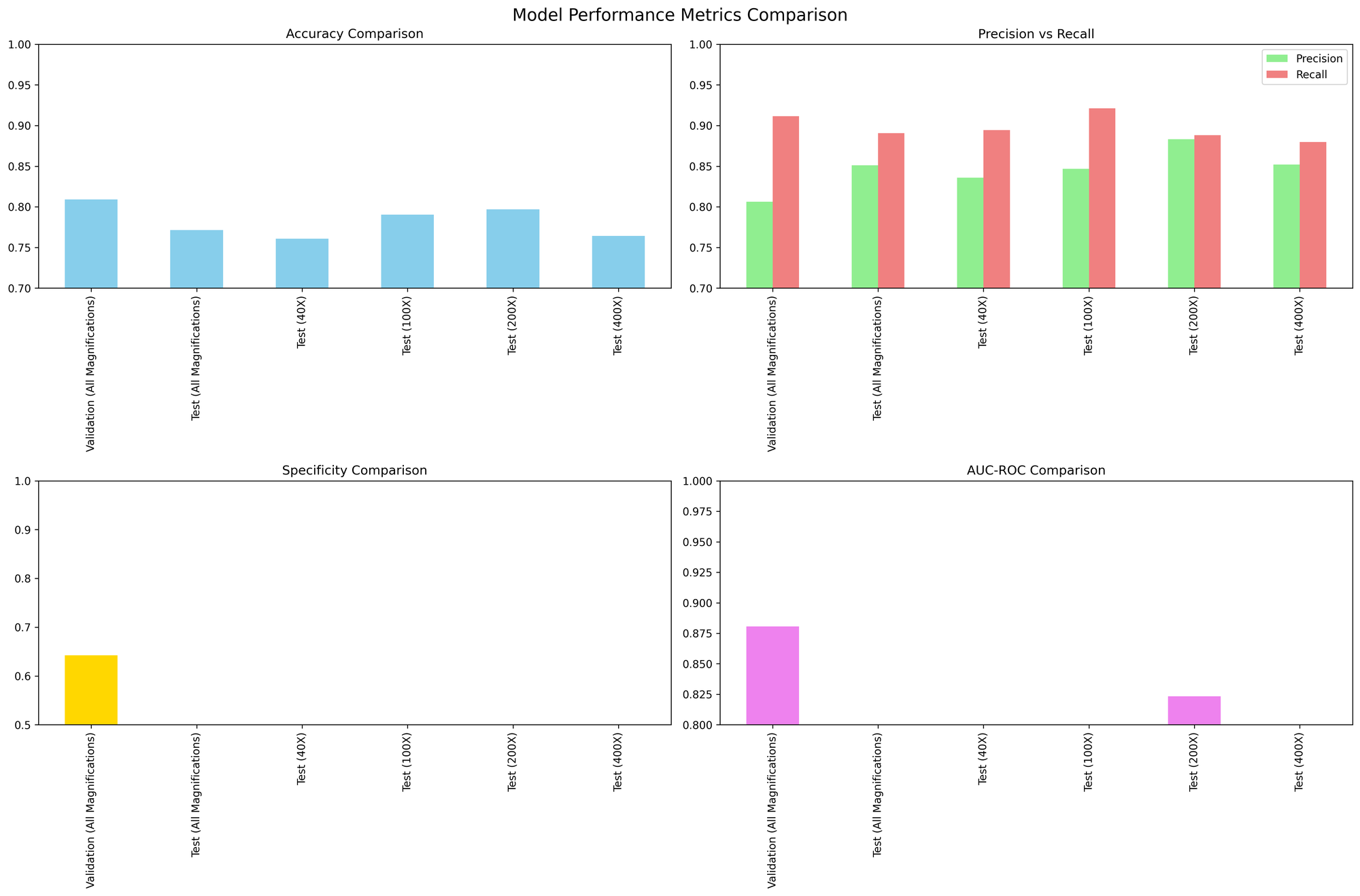}
 \caption{Performance metrics of the ConvNeXt model across validation and test sets at multiple magnifications.}
 \label{fig:single2}
\end{figure}

\begin{figure}[H]
\centering

\subfloat[40X]{%
 \includegraphics[width=0.23\textwidth]{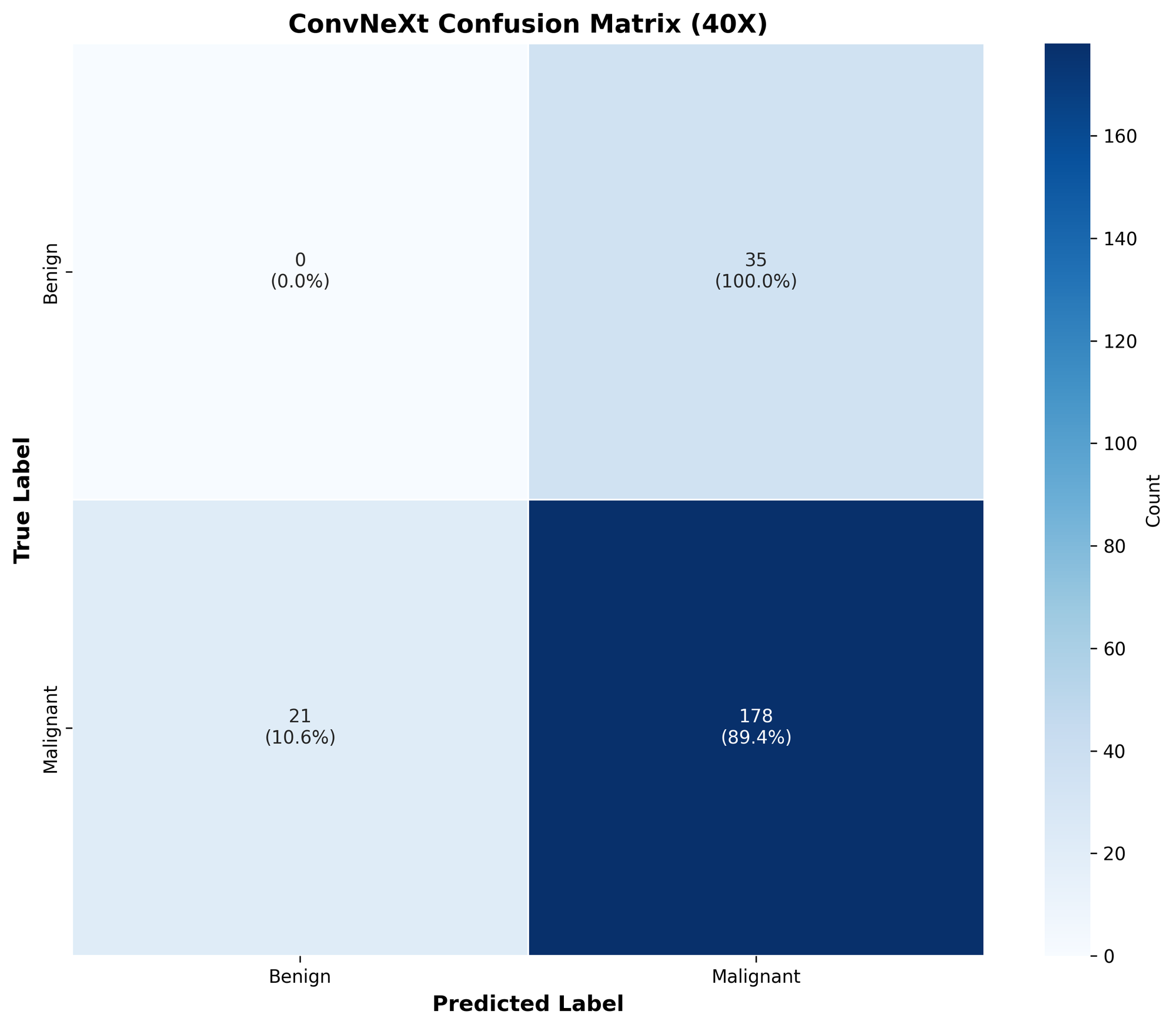}
}\hfill
\subfloat[(b) 100X]{%
 \includegraphics[width=0.23\textwidth]{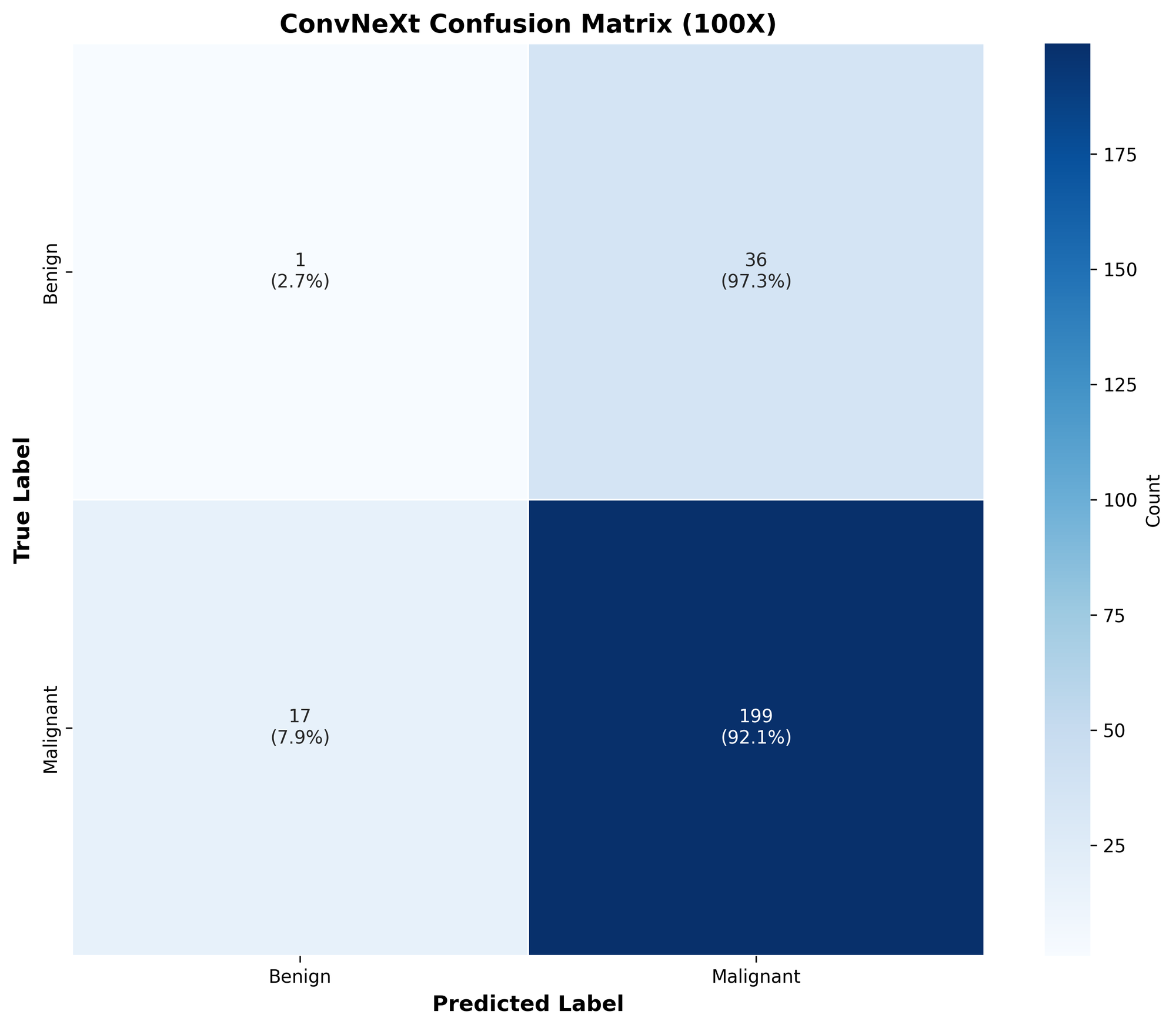}
}\hfill
\subfloat[(c) 200X]{%
 \includegraphics[width=0.23\textwidth]{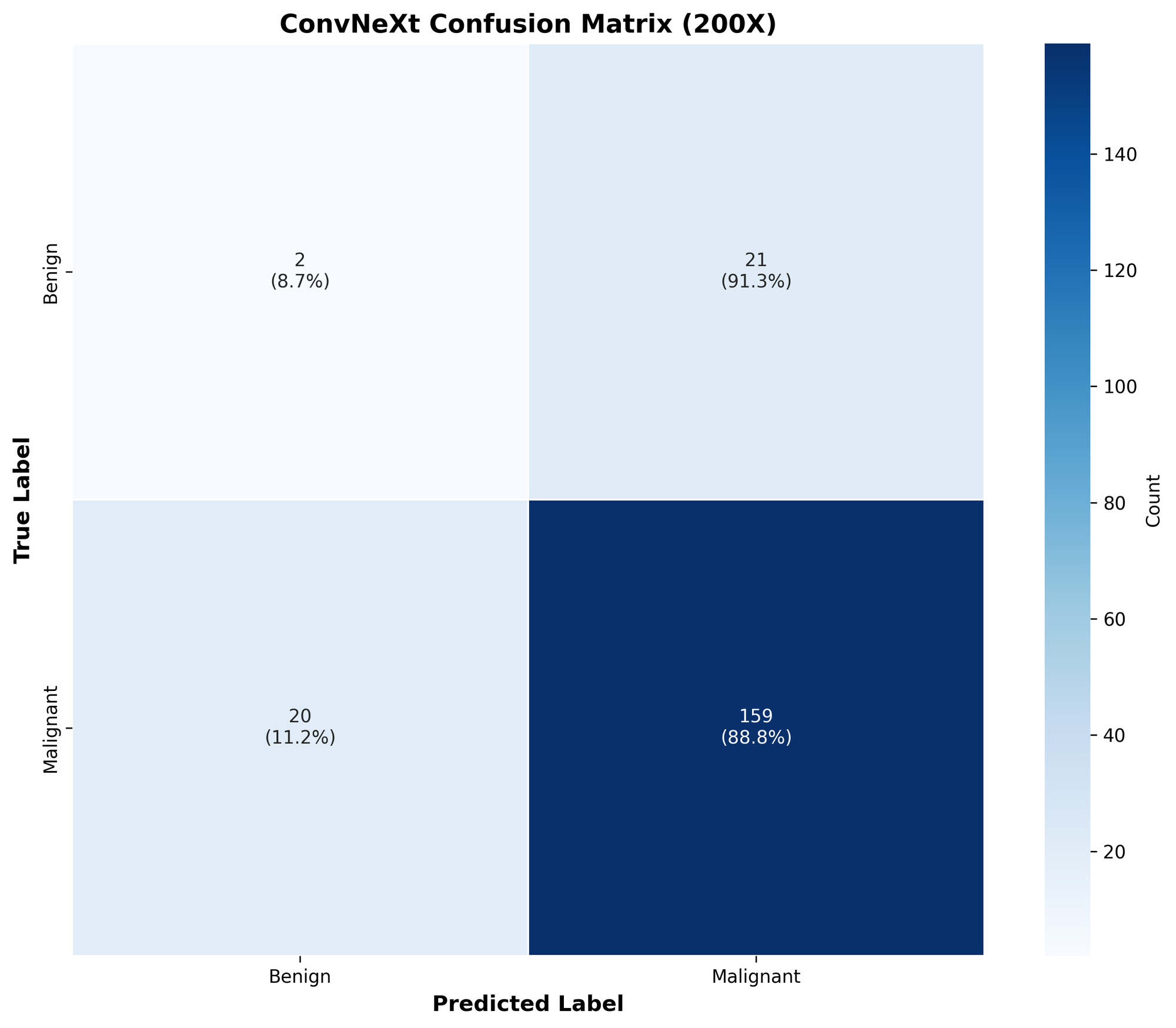}
}\hfill
\subfloat[(d) 400X]{%
 \includegraphics[width=0.23\textwidth]{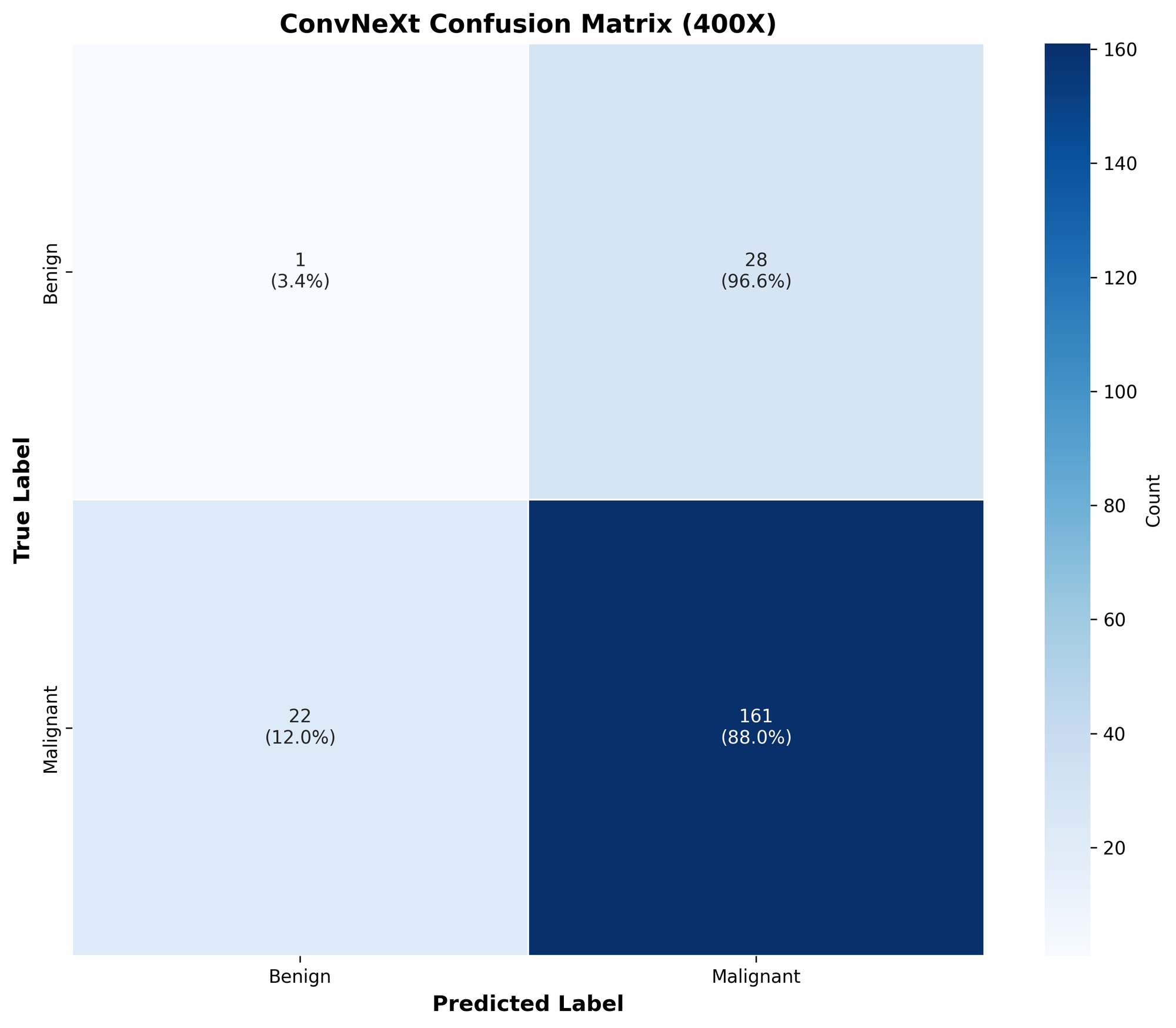}
}

\caption{Confusion matrices of the ConvNeXt model across 40X, 100X, 200X, and 400X magnifications on the BreakHis dataset.}
\label{fig:convnext_confusion_matrices}
\end{figure}

\begin{figure}[H]
\centering

\subfloat[Validation ROC]{%
 \includegraphics[width=0.23\textwidth]{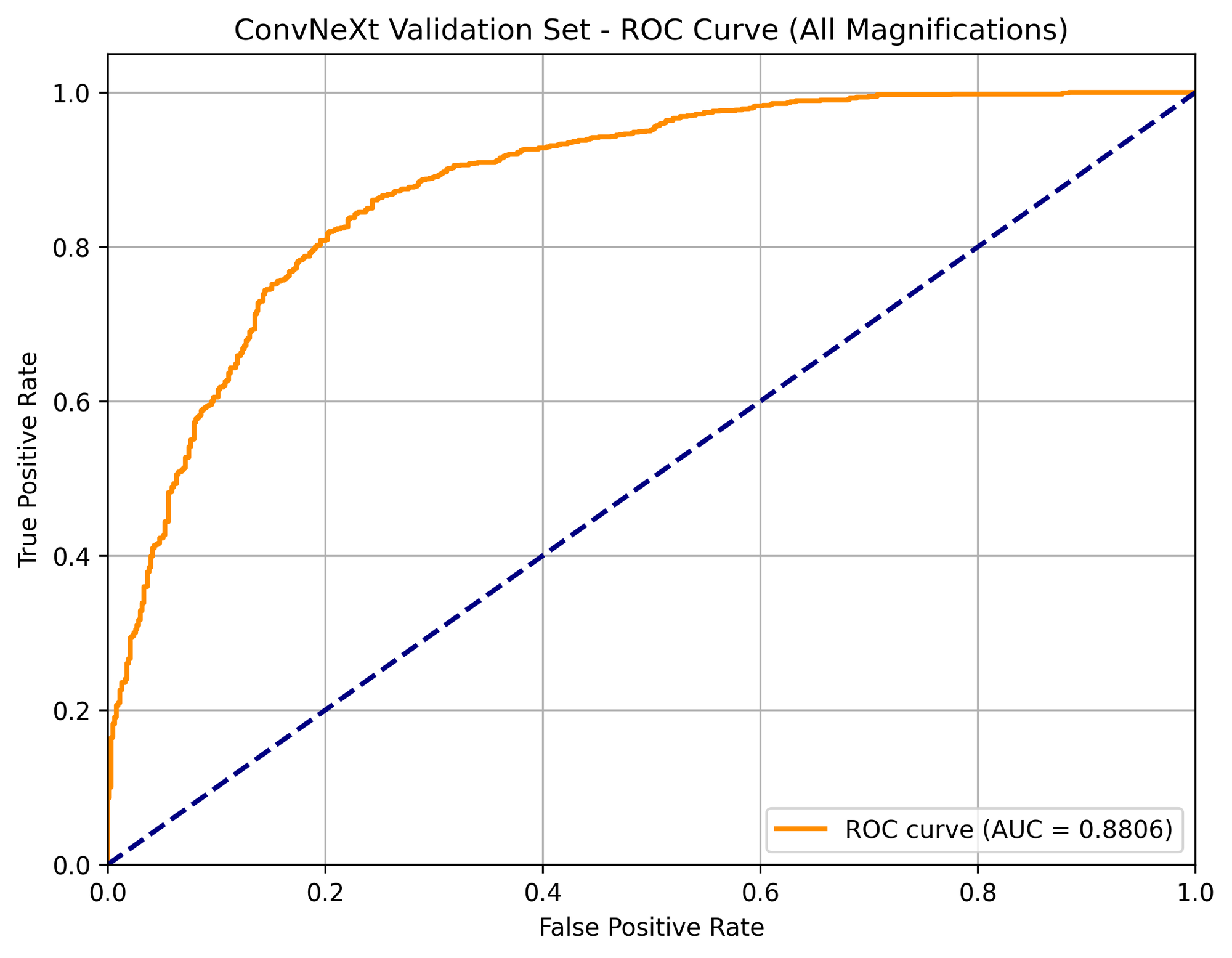}
}\hfill
\subfloat[Validation Confusion Matrix]{%
 \includegraphics[width=0.23\textwidth]{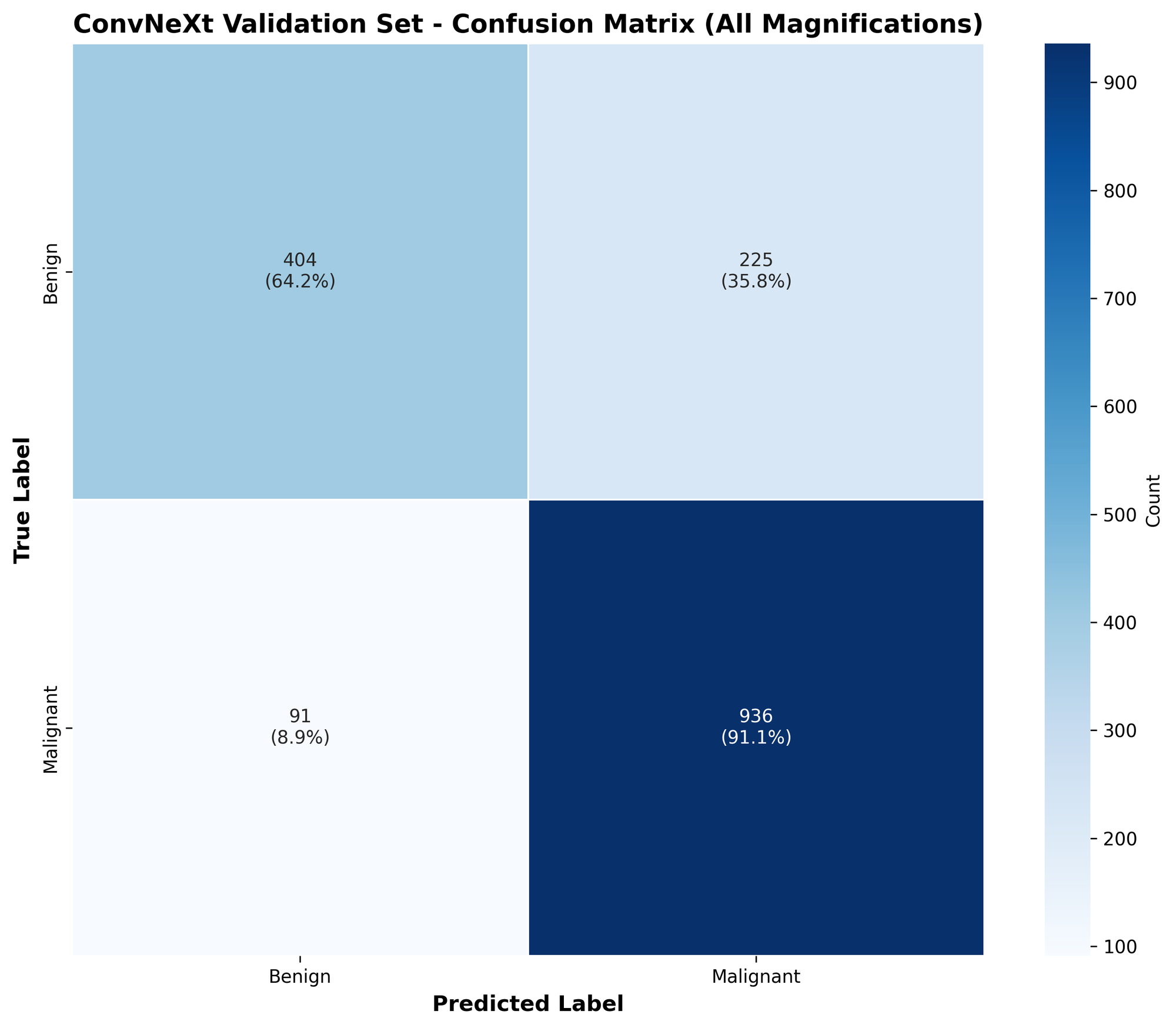}
}\hfill
\subfloat[Test ROC]{%
 \includegraphics[width=0.23\textwidth]{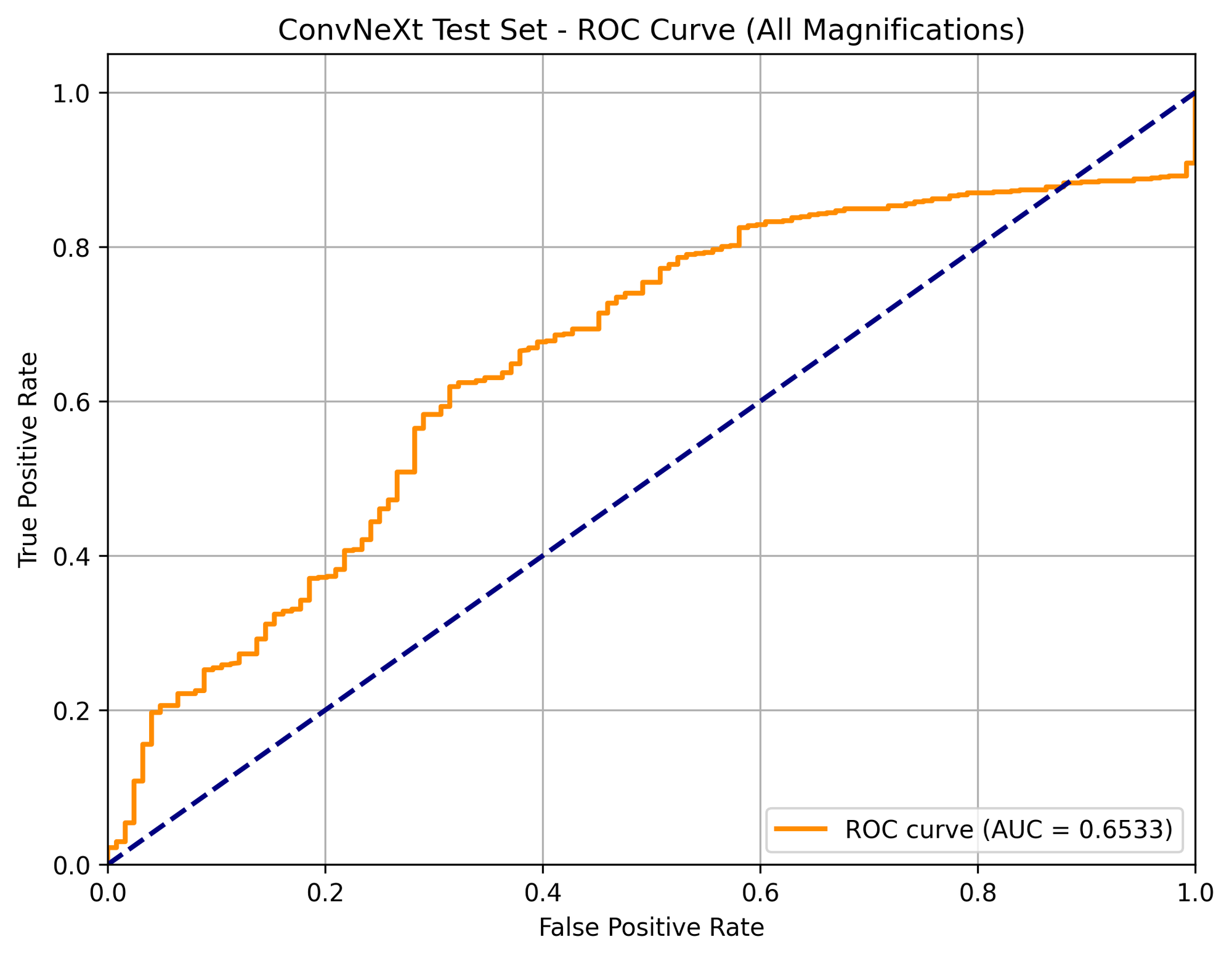}
}\hfill
\subfloat[Test Confusion Matrix]{%
 \includegraphics[width=0.23\textwidth]{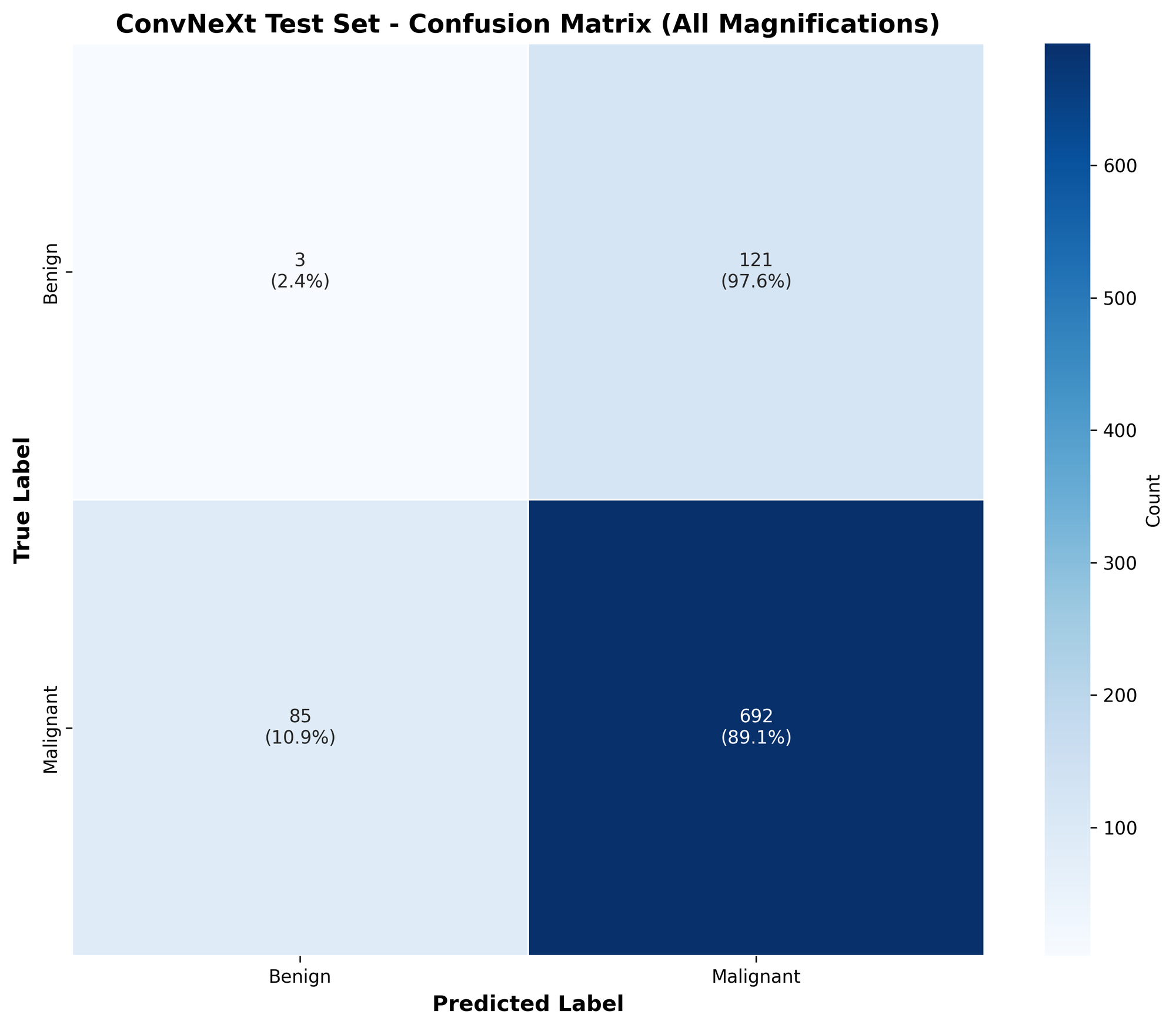}
}

\caption{Validation and test performance of the ConvNeXt model on the BreakHis dataset across all magnifications.}
\label{fig:convnext_val_test}
\end{figure}

\begin{figure}[H]
\centering

\subfloat[ROC 40X]{%
 \includegraphics[width=0.23\textwidth]{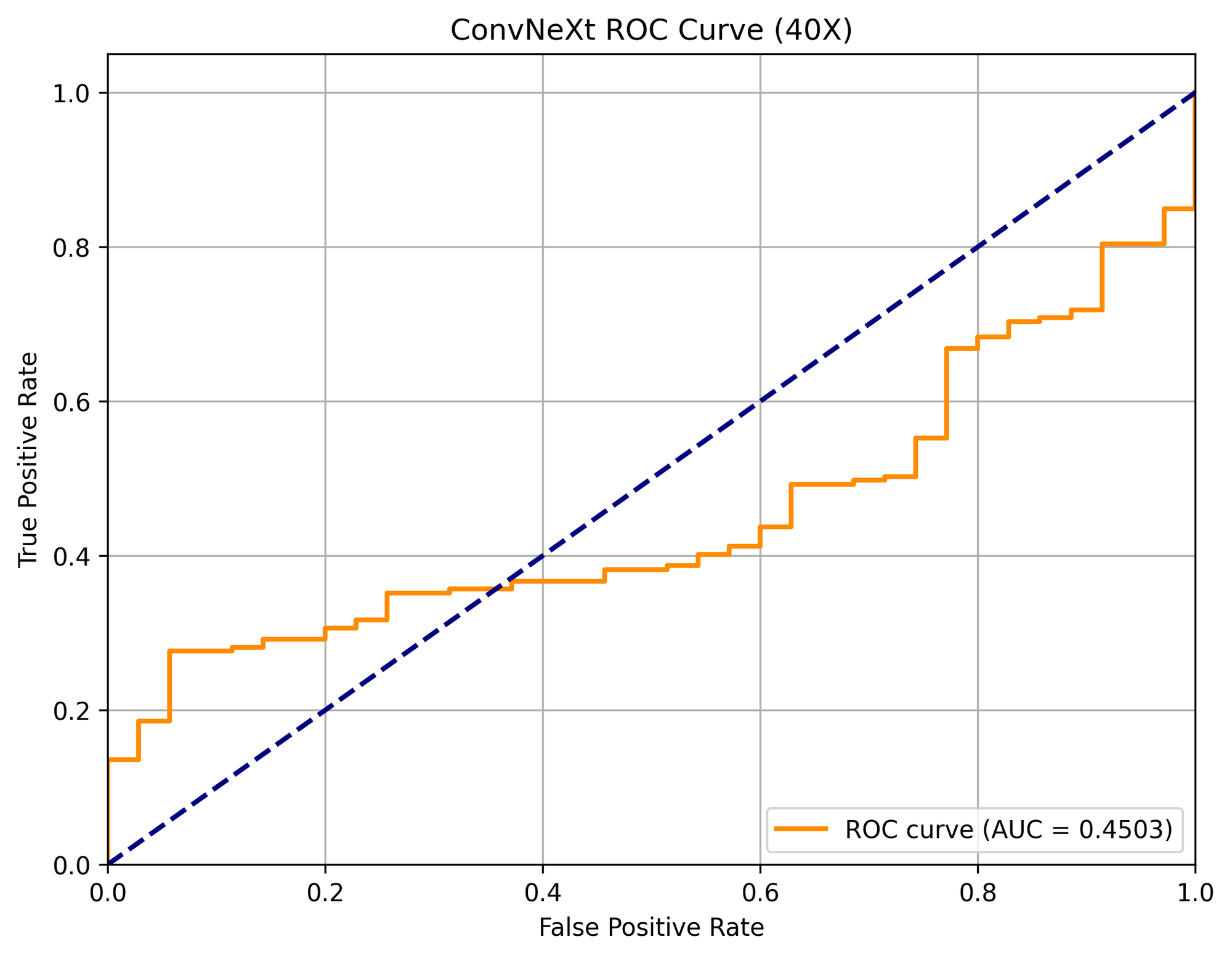}
}\hfill
\subfloat[ROC 100X]{%
 \includegraphics[width=0.23\textwidth]{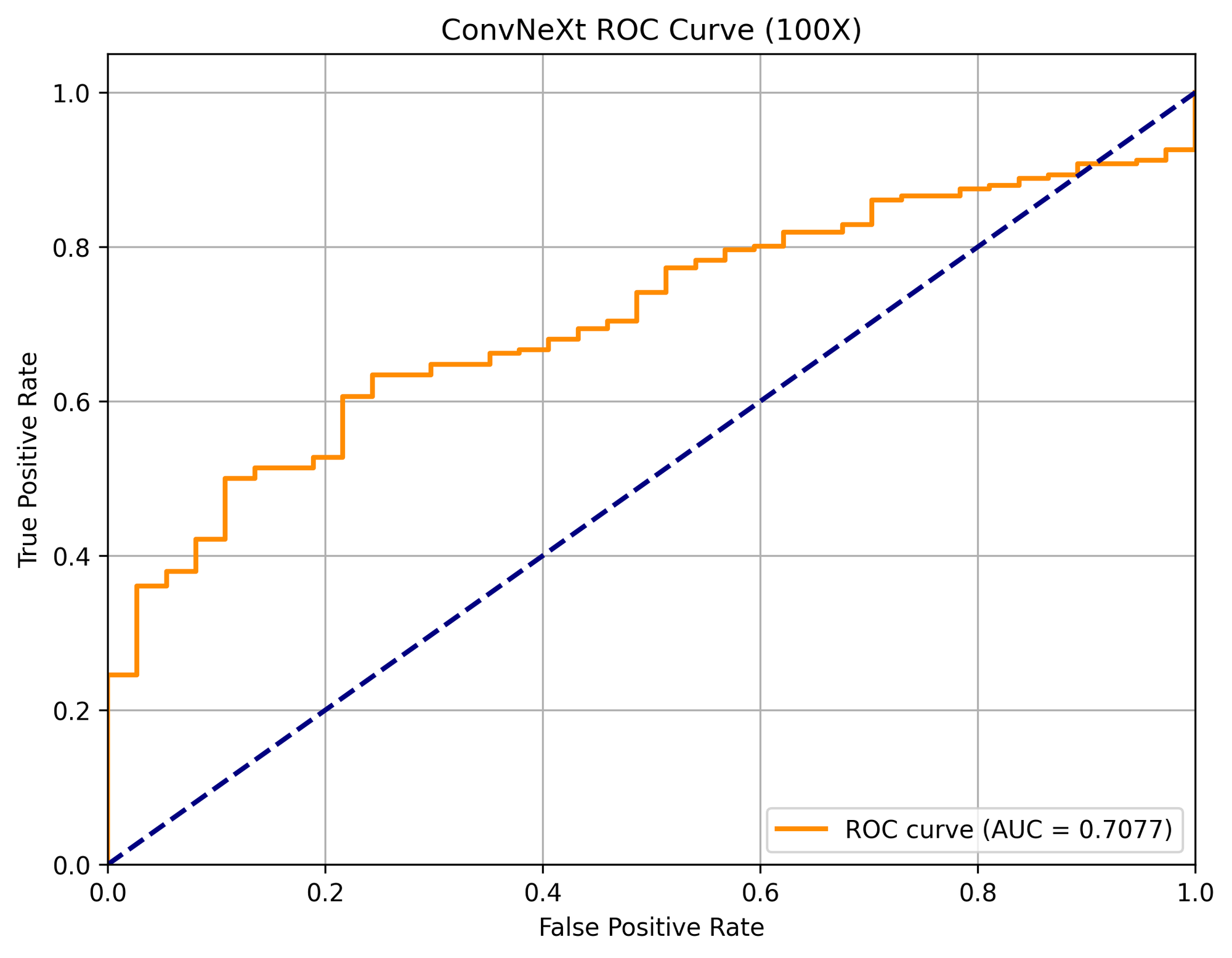}
}\hfill
\subfloat[ROC 200X]{%
 \includegraphics[width=0.23\textwidth]{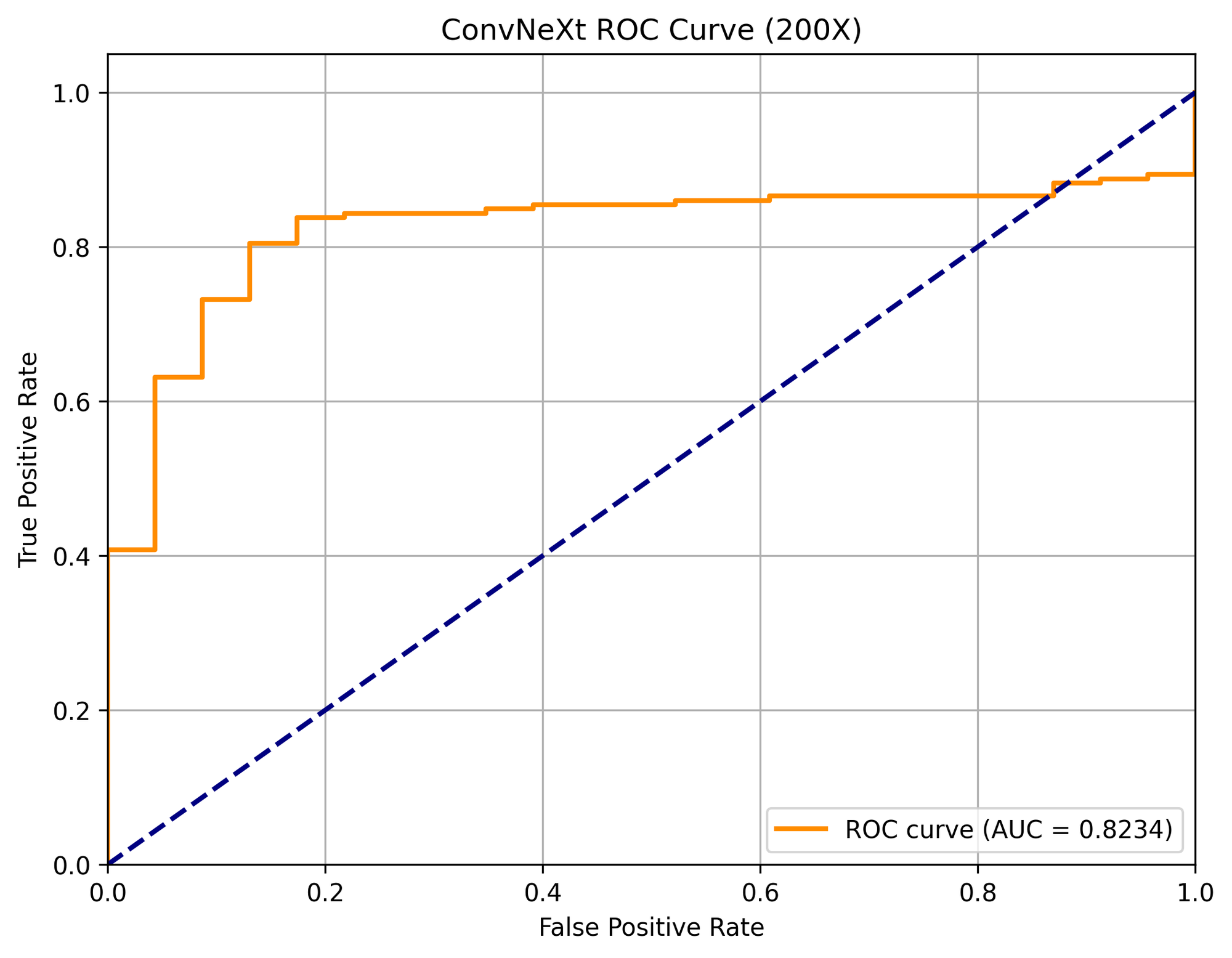}
}\hfill
\subfloat[ROC 400X]{%
 \includegraphics[width=0.23\textwidth]{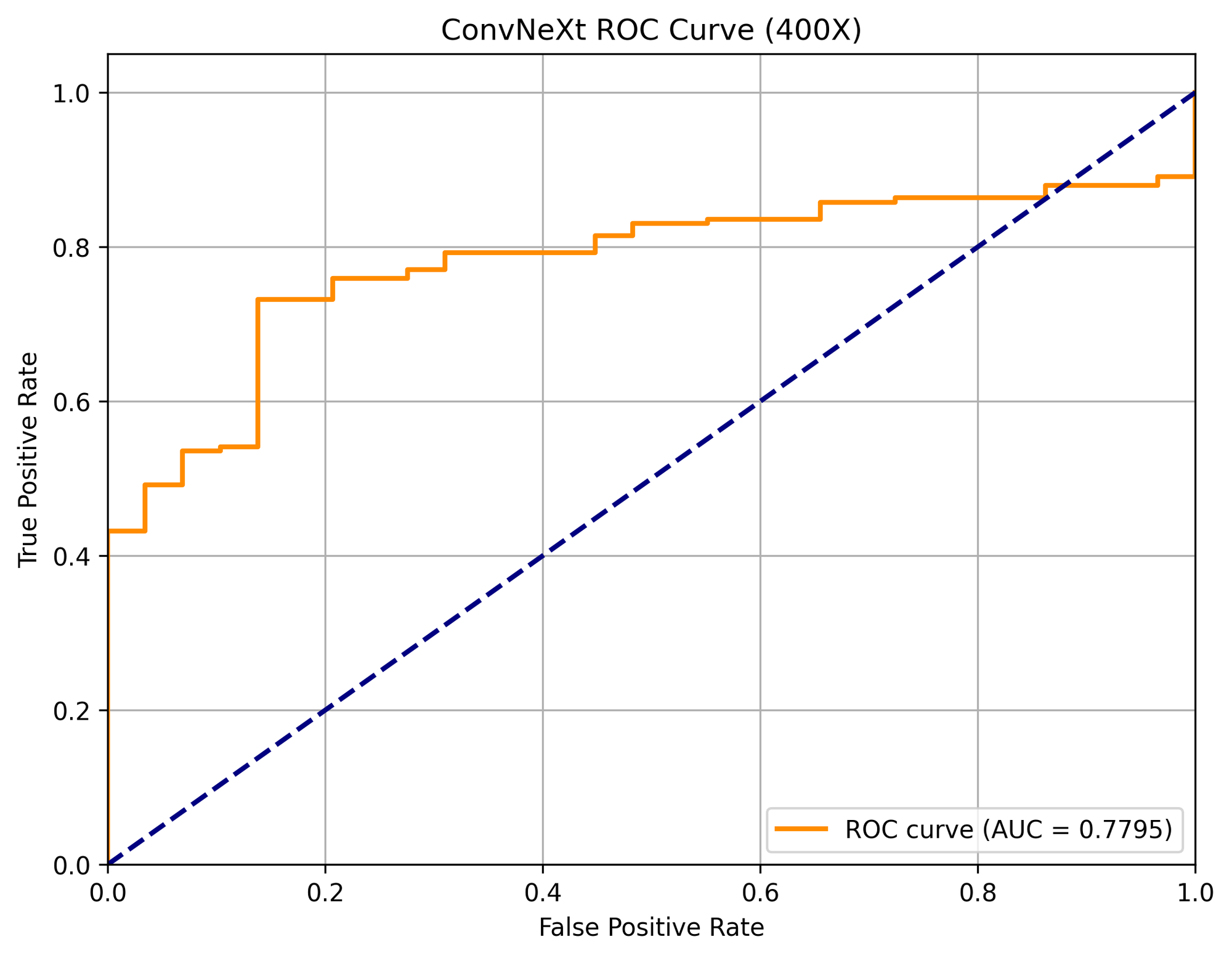}
}

\vspace{0.6ex}

\subfloat[PR 40X]{%
 \includegraphics[width=0.23\textwidth]{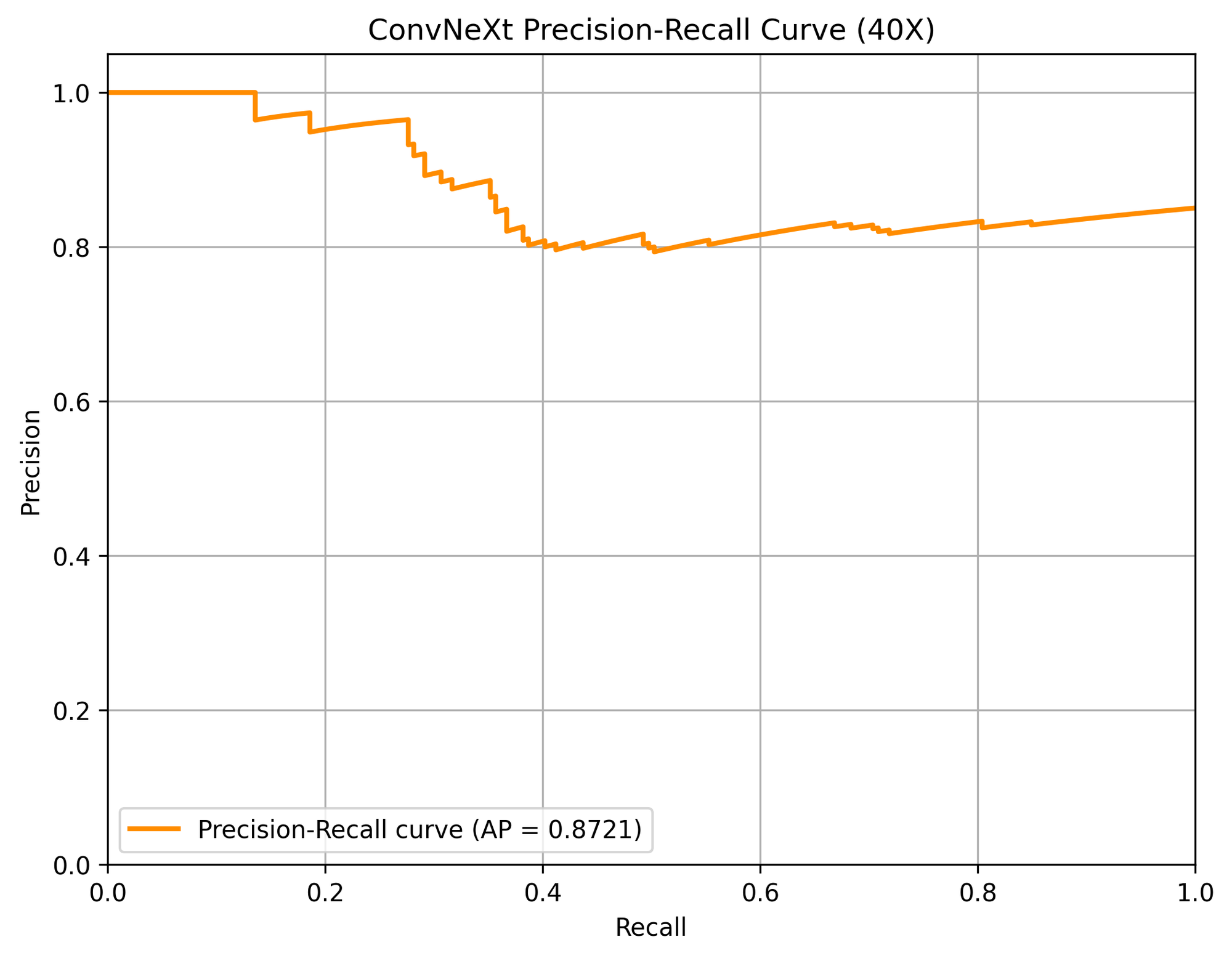}
}\hfill
\subfloat[PR 100X]{%
 \includegraphics[width=0.23\textwidth]{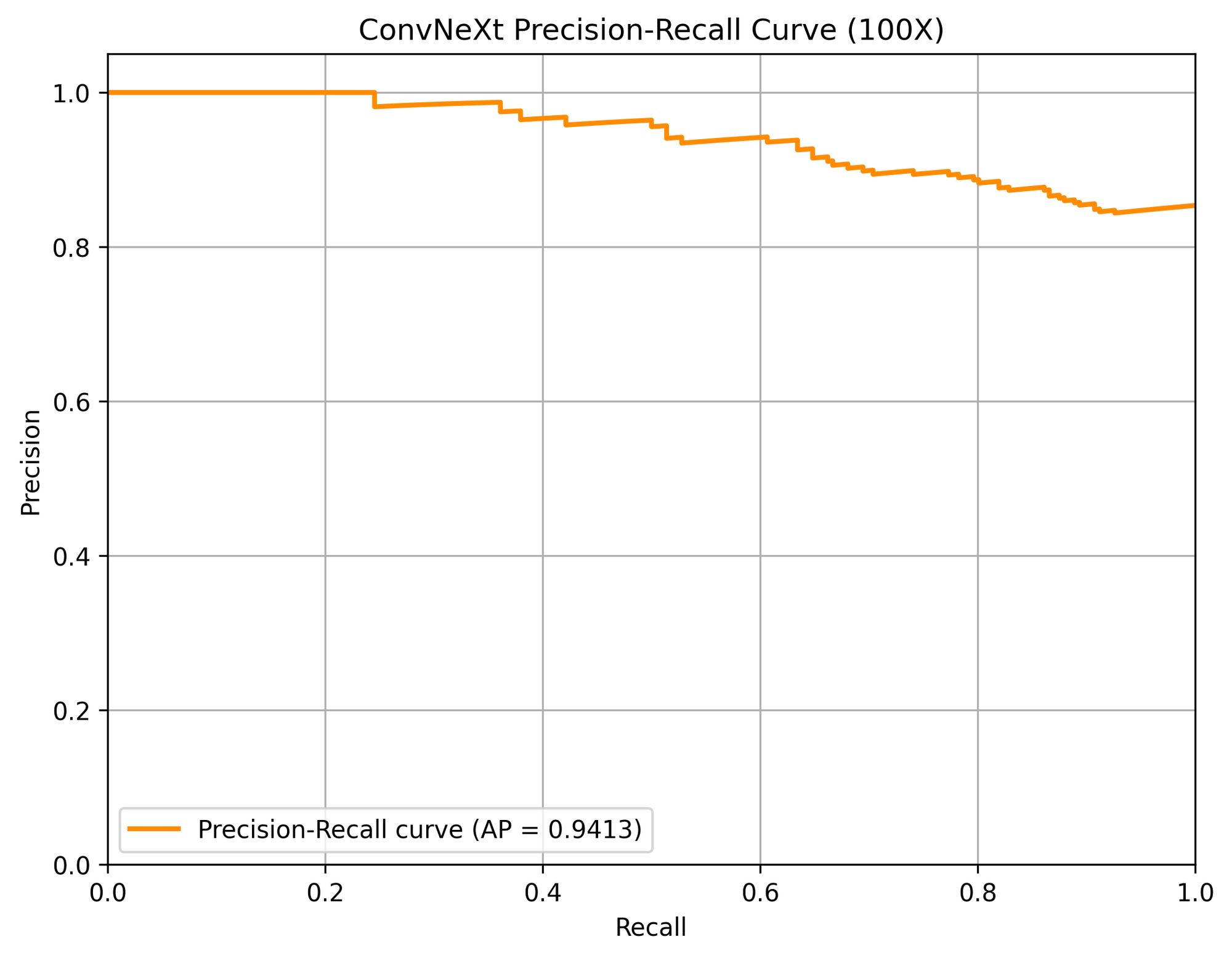}
}\hfill
\subfloat[PR 200X]{%
 \includegraphics[width=0.23\textwidth]{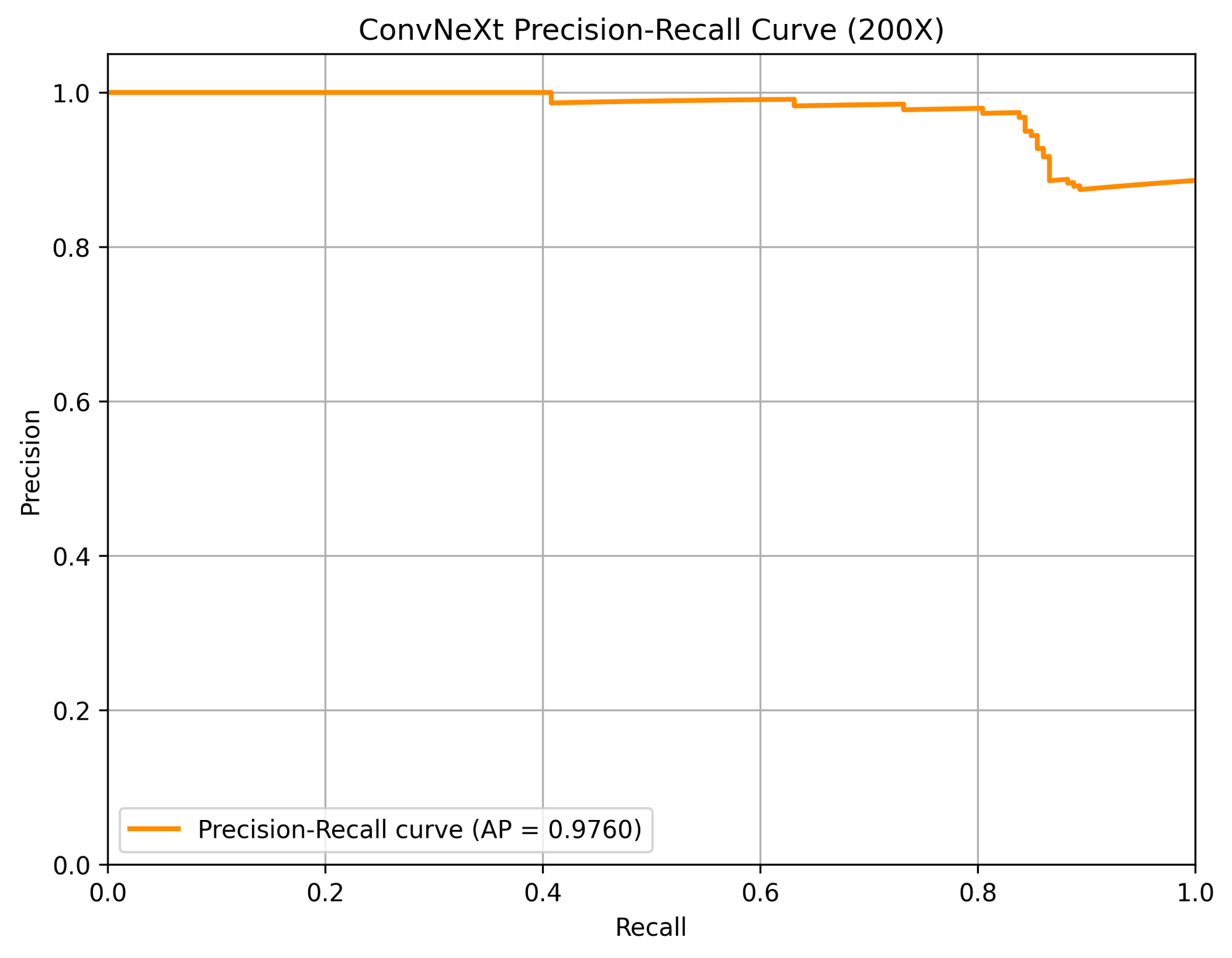}
}\hfill
\subfloat[PR 400X]{%
 \includegraphics[width=0.23\textwidth]{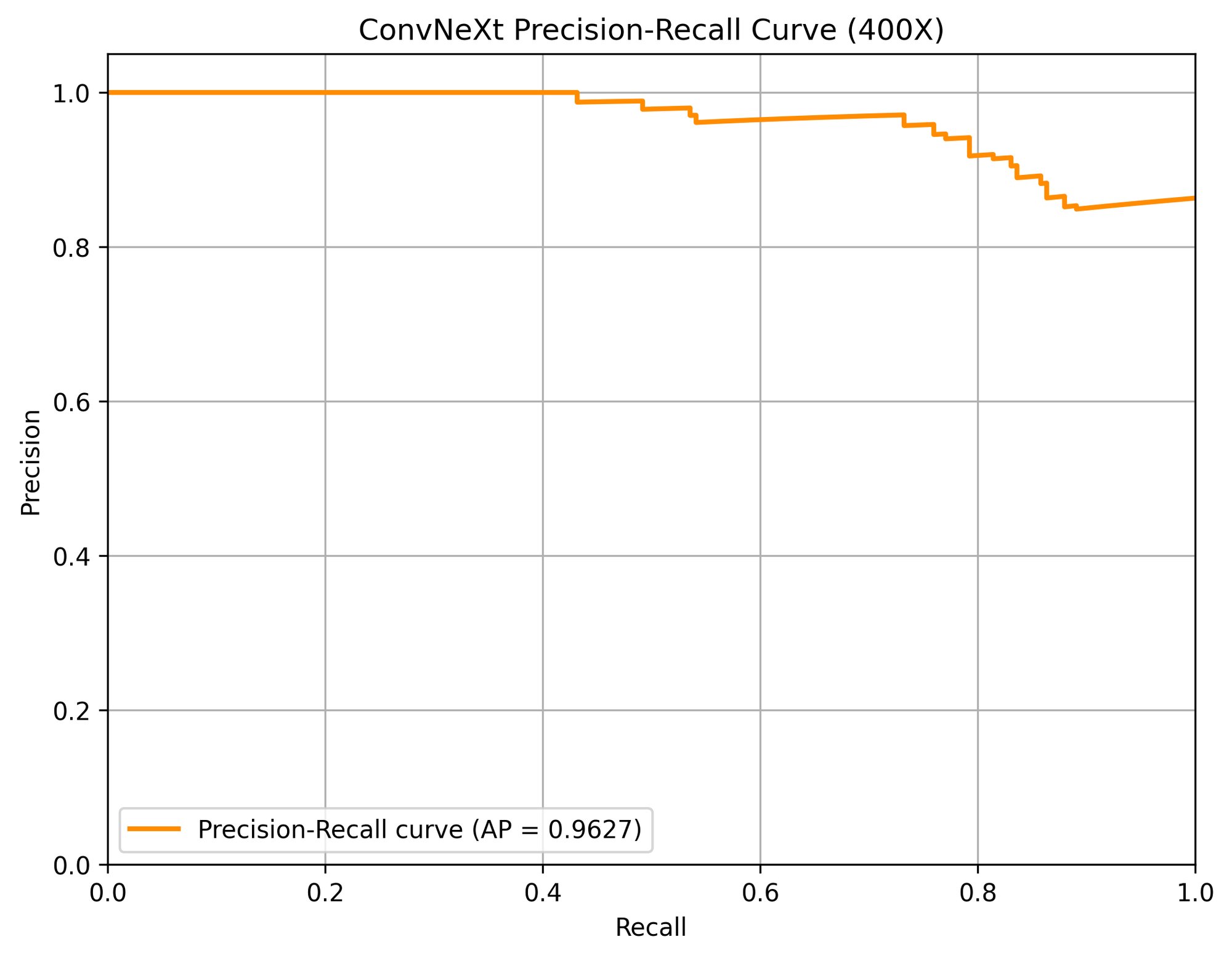}
}

\caption{Receiver Operating Characteristic (ROC) curves (a–d) and Precision–Recall (PR) curves (e–h) for the ConvNeXt model evaluated on the BreakHis dataset across 40X, 100X, 200X, and 400X magnifications.}
\label{fig:convnext_roc_pr}
\end{figure}

\begin{table}[ht]
\caption{Performance of InceptionV3 on the BreakHis dataset}
\label{tab:inceptionv3}
\centering
\begin{tabular}{@{}lcccccc@{}}
\toprule
\textbf{Dataset} 
& \textbf{Accuracy (\%)} 
& \textbf{Precision (\%)} 
& \textbf{Recall (\%)} 
& \textbf{F1-score (\%)} 
& \textbf{Specificity (\%)} 
& \textbf{AUC} \\
\midrule
Validation (All Mag.) 
& 94.3 
& 96.2 
& 95.9 
& 96.1 
& 89.9 
& 0.980 \\

Test (All Mag.)    
& 84.6 
& 79.5 
& 88.4 
& 83.7 
& 81.5 
& 0.933 \\

Test (40$\times$)      
& 87.2 
& 86.0 
& 86.0 
& 86.0 
& 88.2 
& 0.937 \\

Test (100$\times$)      
& 81.2 
& 77.0 
& 85.9 
& 81.3 
& 76.9 
& 0.924 \\

Test (200$\times$)      
& 85.1 
& 80.4 
& 86.3 
& 83.3 
& 84.1 
& 0.938 \\

Test (400$\times$)      
& 86.9 
& 78.3 
& 95.9 
& 86.2 
& 80.2 
& 0.943 \\
\bottomrule
\end{tabular}
\end{table}

Inception V3 showed strong results on the BreakHis validation set, achieving 94.29\% accuracy, 96.22\% precision, 95.91\% recall, 89.96\% specificity, 96.06\% F1-score, and 0.9803 AUC-ROC, indicating exceptional generalization and a well-balanced sensitivity–specificity ratio (Table 4). Nonetheless, performance on the test set (across all magnifications) diminished to 84.56\% accuracy, 79.49\% precision, 88.36\% recall, 81.47\% specificity, 83.69\% F1-score, and 0.933 AUC-ROC, suggesting a propensity to misclassify benign instances (Figure 13, 14 \& 15). The model attained accuracies of 87.18\% at 40X magnification (AUC 0.9366), 81.18\% at 100X (AUC 0.9243), and 85.07\% at 200X (AUC 0.9383), with recall continuously surpassing precision, indicating effective detection of malignant cases but a higher incidence of false positives (Figure 16). These results underscore Inception V3's efficacy in extracting multi-scale histopathological features, while simultaneously revealing its vulnerability to magnification-related variability and diminished specificity on previously unreported test data.

\begin{figure}[H]
 \centering
 \includegraphics[width=0.75\linewidth]{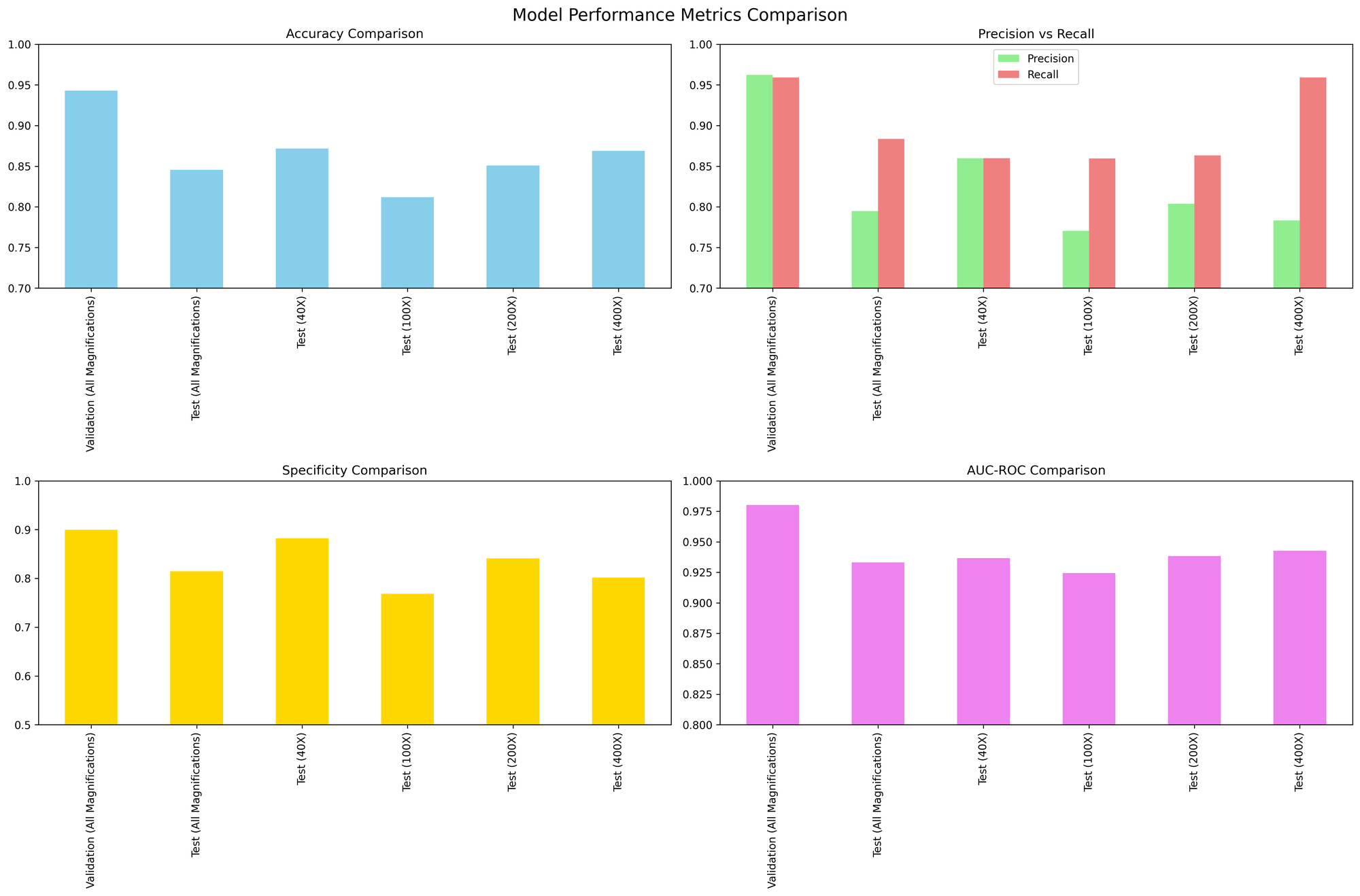}
 \caption{Performance of the Inception model shows high validation accuracy (94.5\%, AUC = 0.98) and stable test results (81–87\% accuracy, AUC = 0.93–0.95), with recall generally exceeding precision across magnifications.}
 \label{fig:single3}
\end{figure}

\begin{figure}[H]
\centering

\subfloat[40X]{%
 \includegraphics[width=0.23\textwidth]{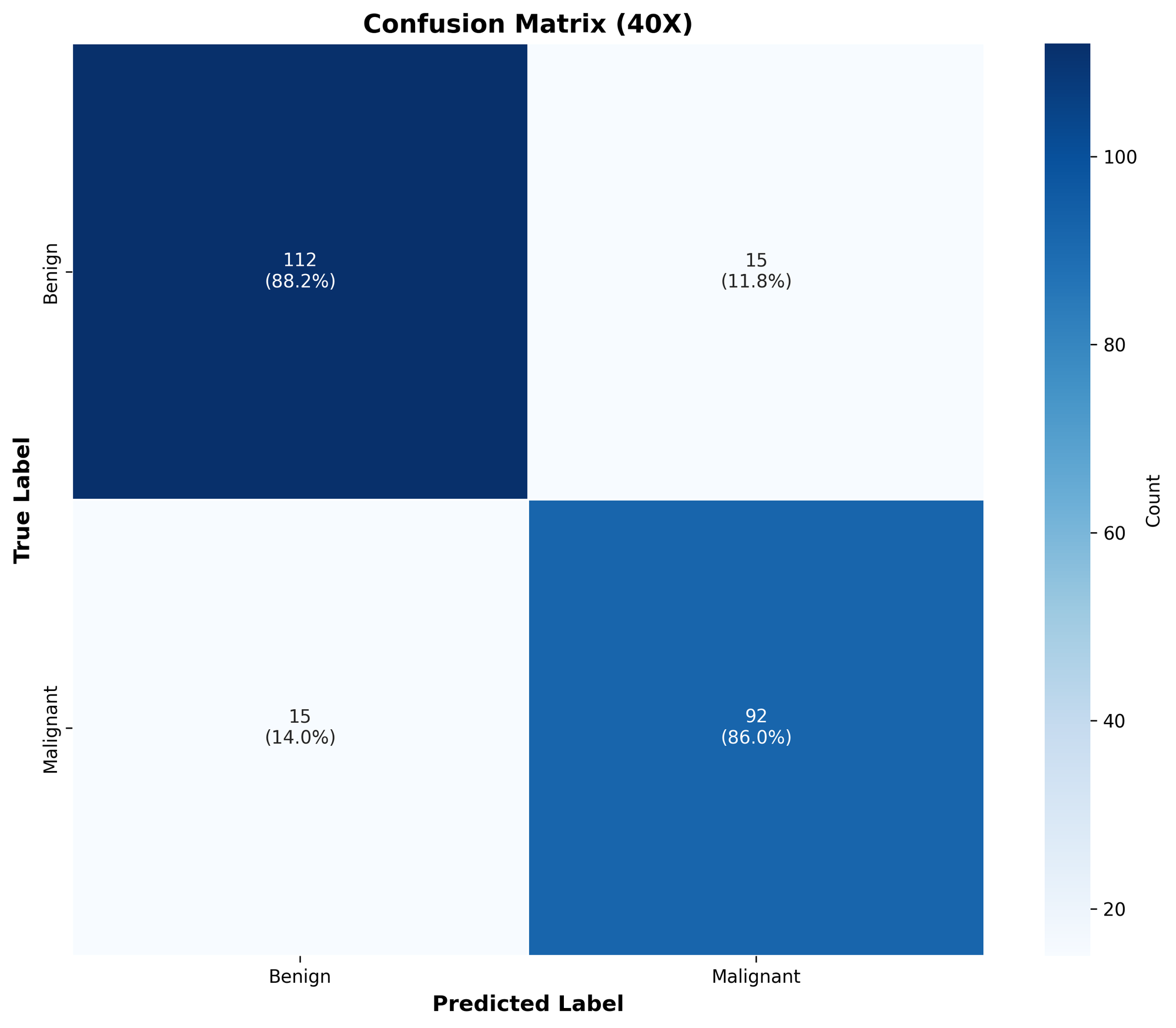}
}\hfill
\subfloat[100X]{%
 \includegraphics[width=0.23\textwidth]{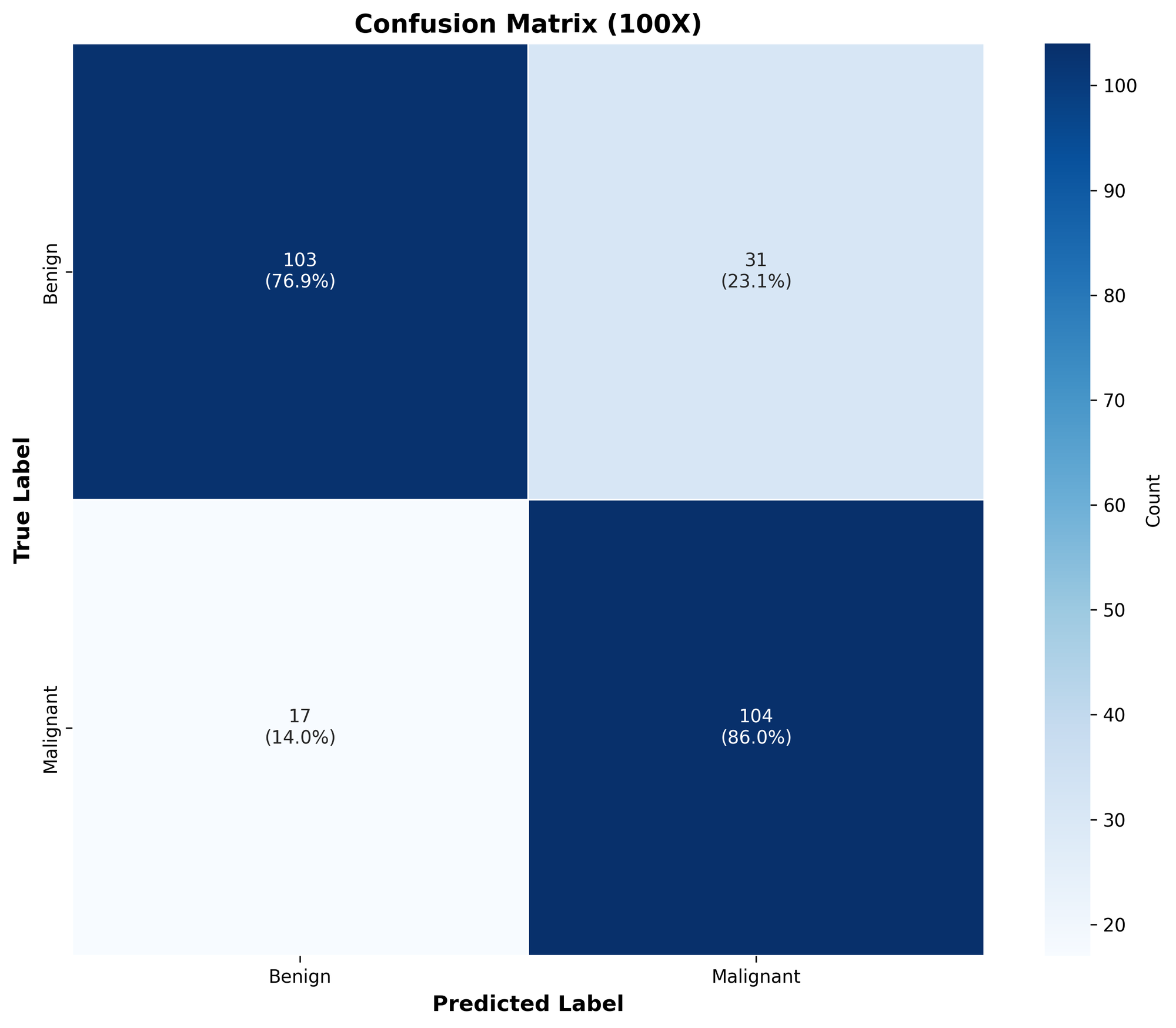}
}\hfill
\subfloat[200X]{%
 \includegraphics[width=0.23\textwidth]{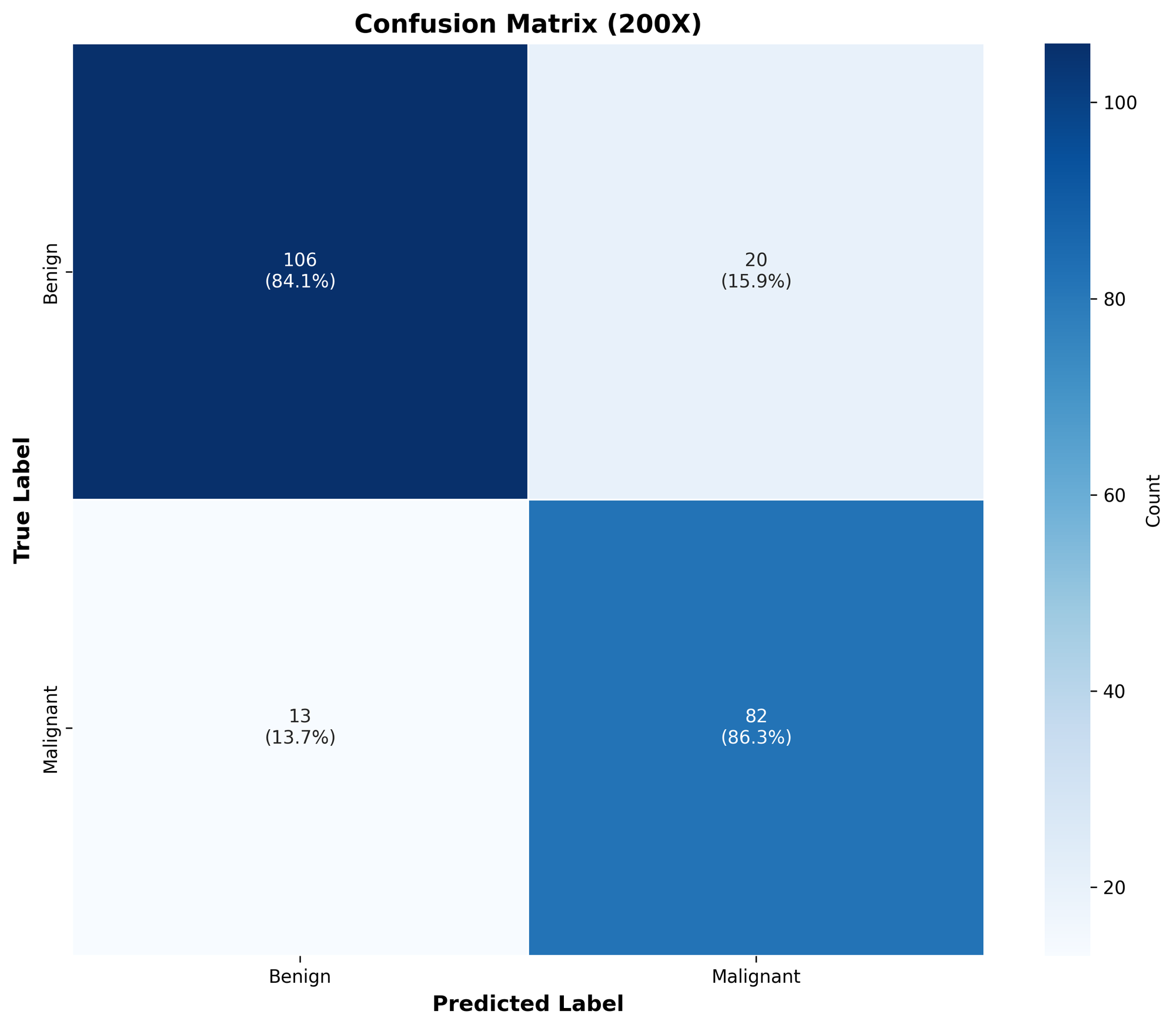}
}\hfill
\subfloat[400X]{%
 \includegraphics[width=0.23\textwidth]{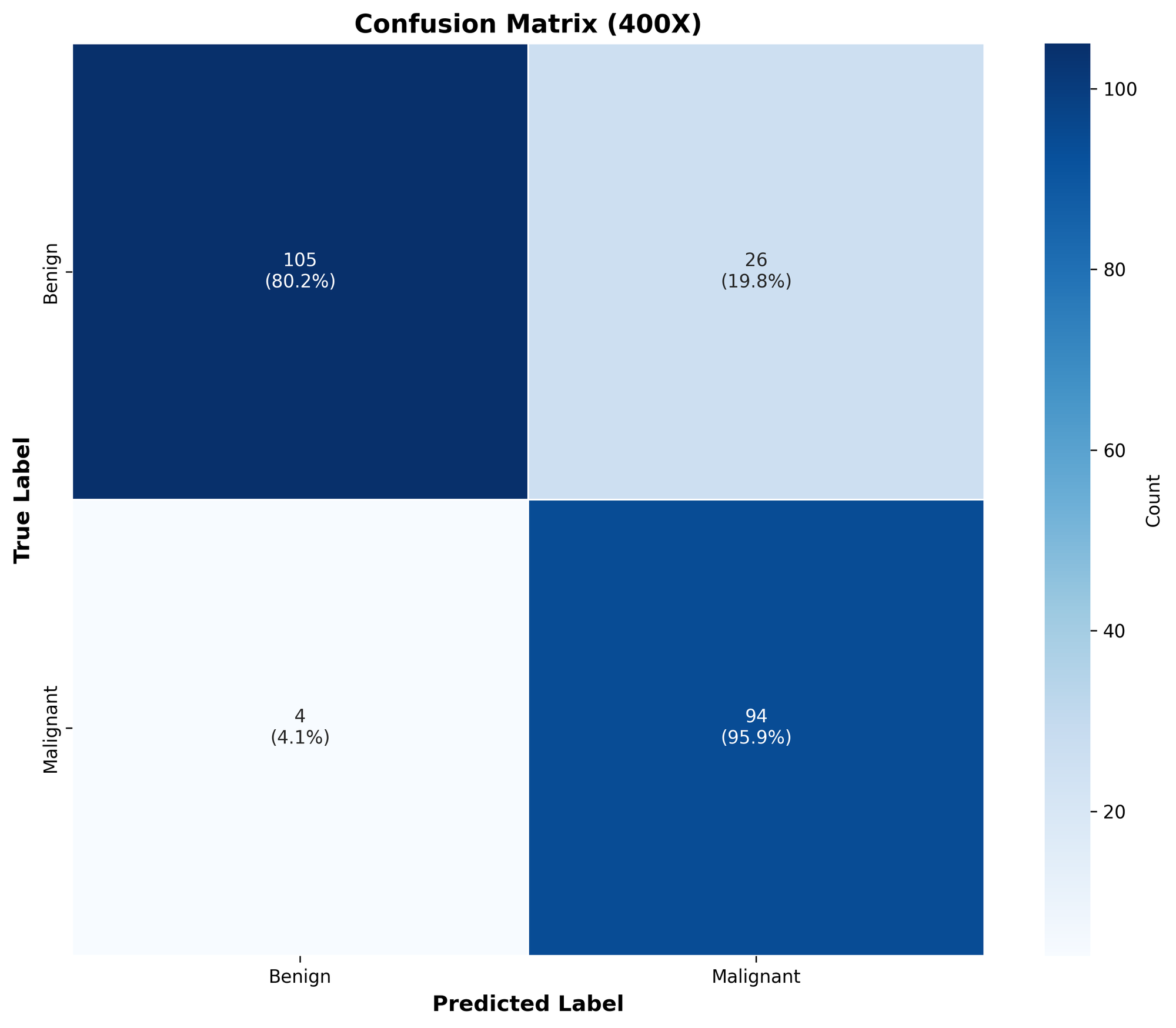}
}

\caption{Confusion matrices of the Inception model across 40X, 100X, 200X, and 400X magnifications}
\label{fig:inception_confusion_matrices}
\end{figure}

\begin{figure}[H]
\centering

\subfloat[Validation ROC]{%
 \includegraphics[width=0.23\textwidth]{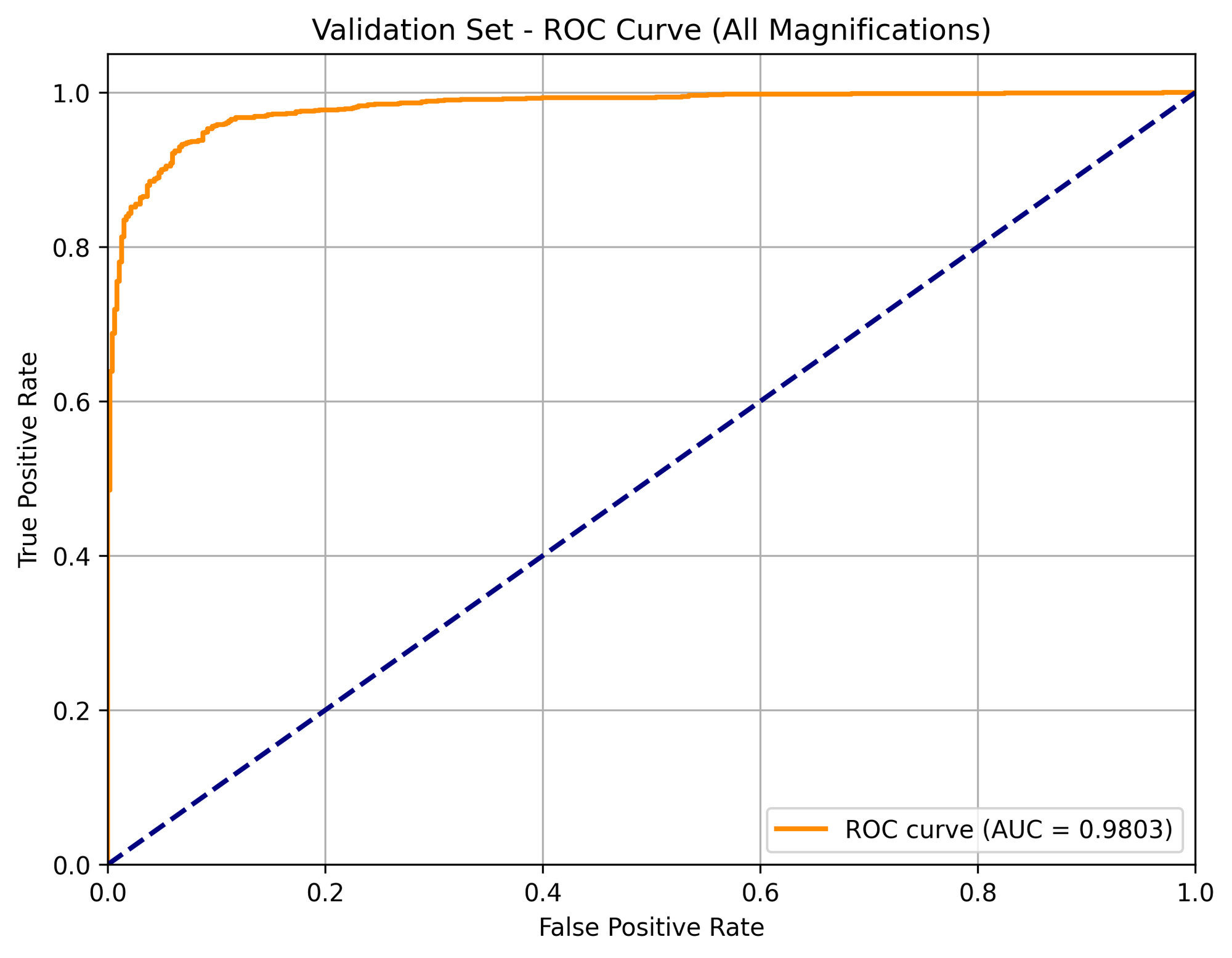}
}\hfill
\subfloat[Validation Confusion Matrix]{%
 \includegraphics[width=0.23\textwidth]{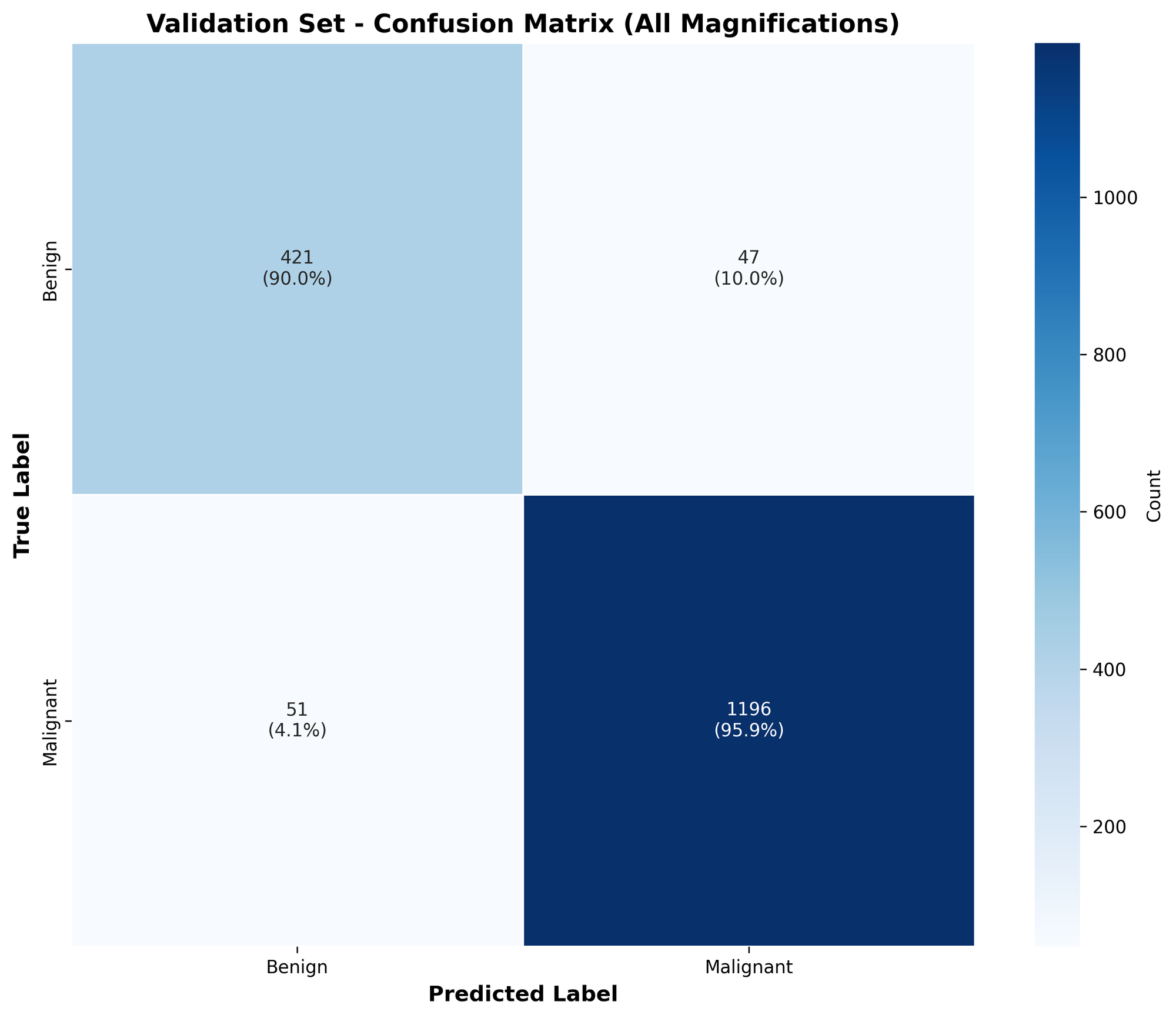}
}\hfill
\subfloat[Test ROC]{%
 \includegraphics[width=0.23\textwidth]{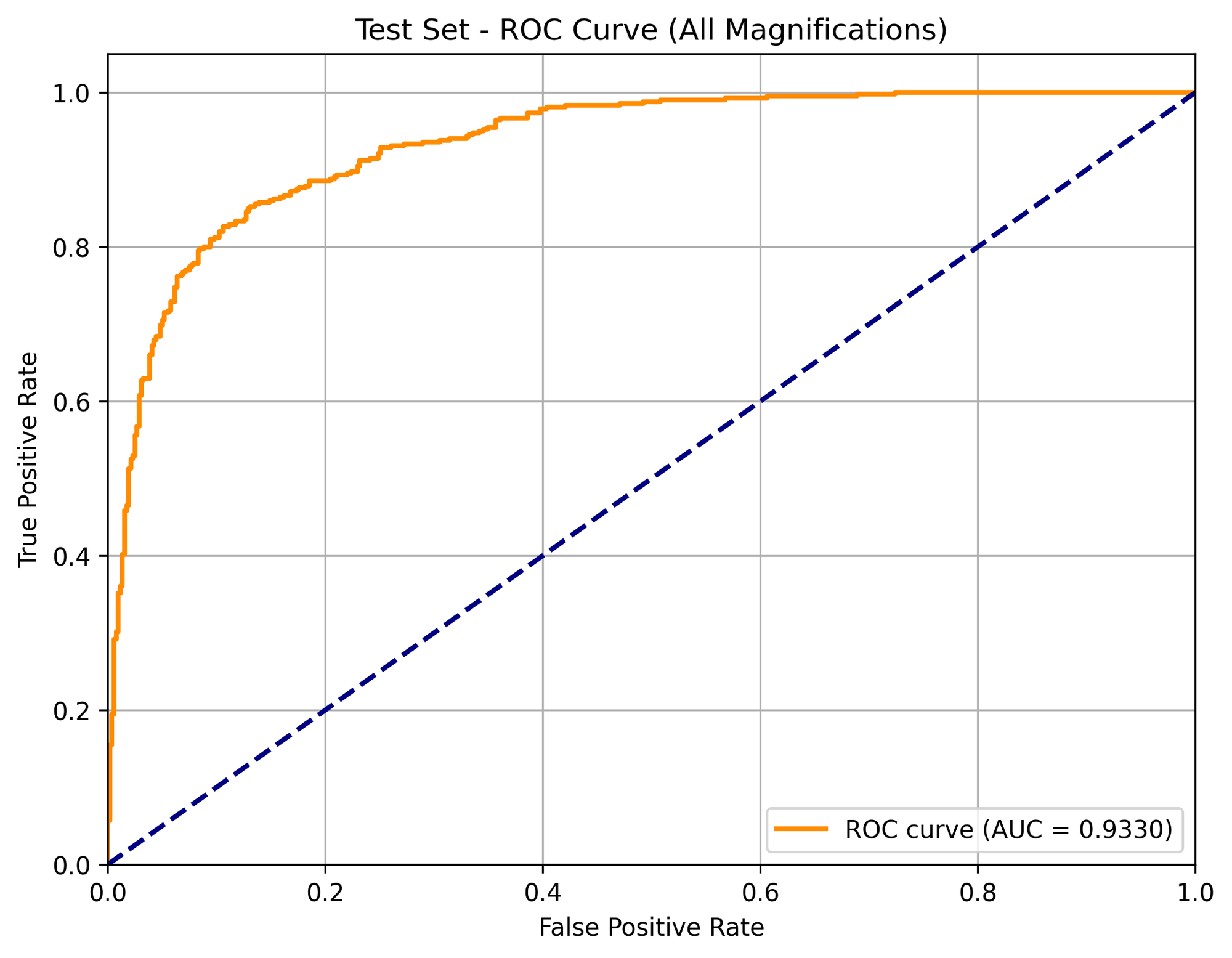}
}\hfill
\subfloat[Test Confusion Matrix]{%
 \includegraphics[width=0.23\textwidth]{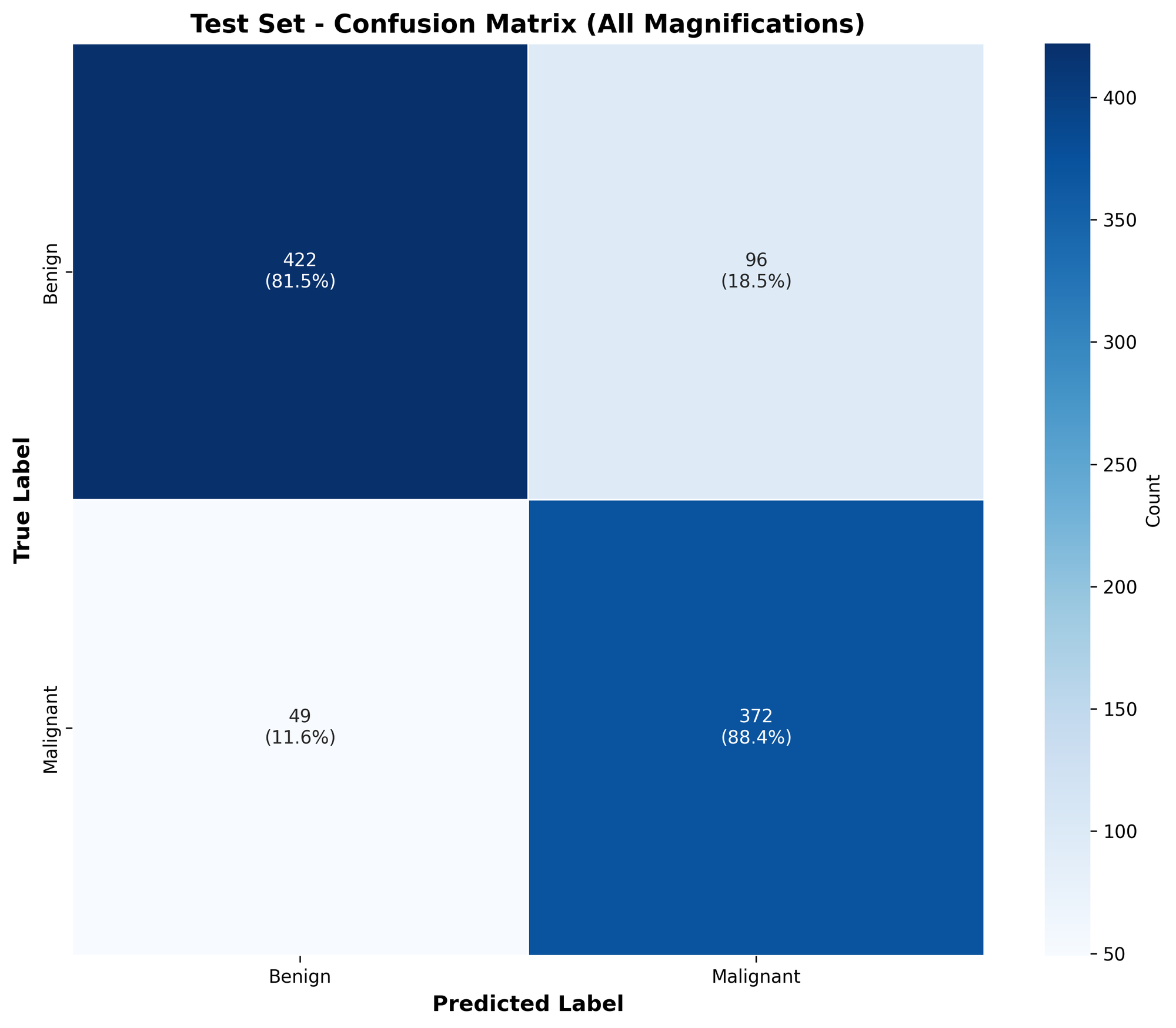}
}

\caption{Validation and test performance of the Inception model on the BreakHis dataset across all magnifications.}
\label{fig:inception_val_test}
\end{figure}

\begin{figure}[H]
\centering

\subfloat[ROC 40X]{%
 \includegraphics[width=0.23\textwidth]{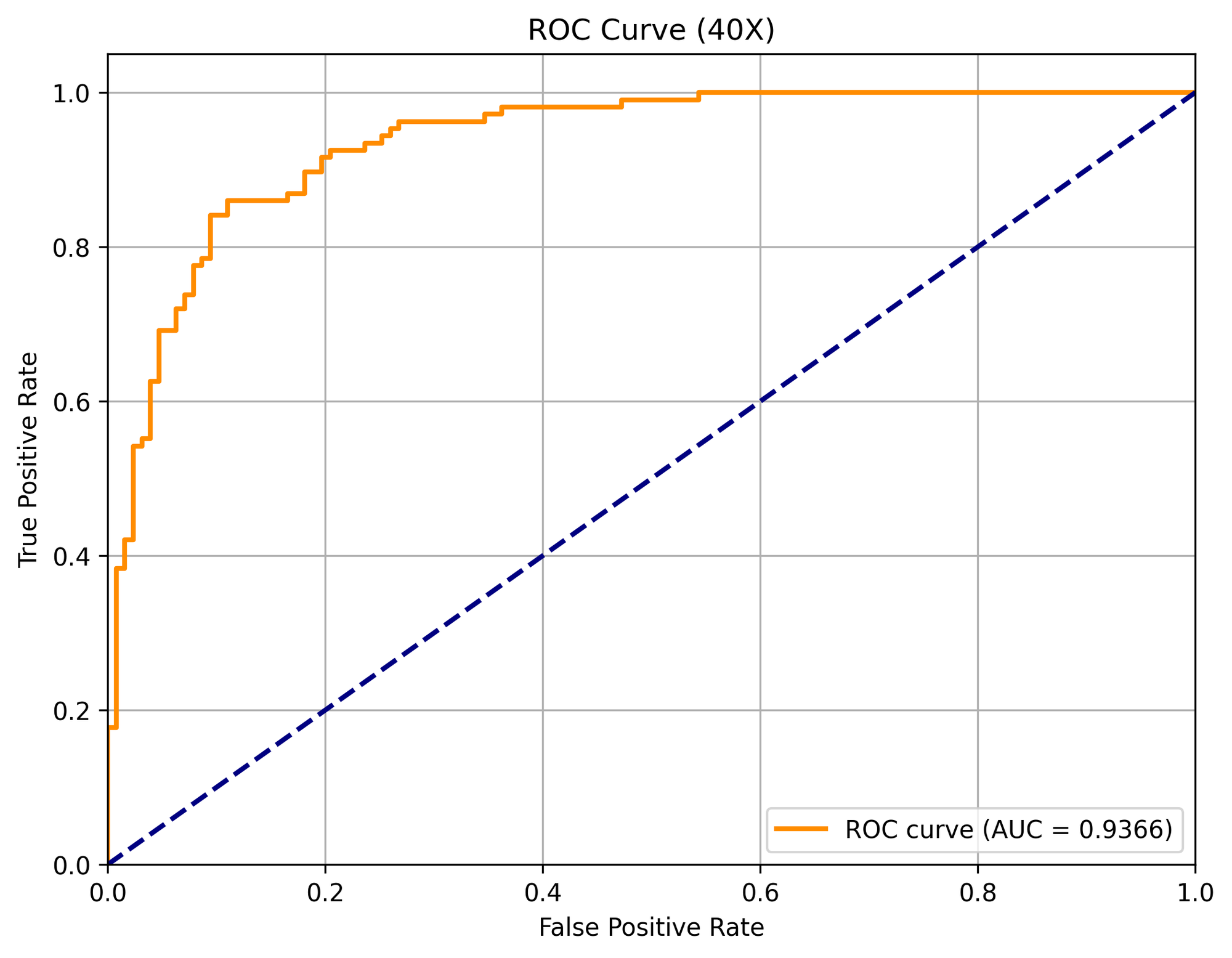}
}\hfill
\subfloat[ROC 100X]{%
 \includegraphics[width=0.23\textwidth]{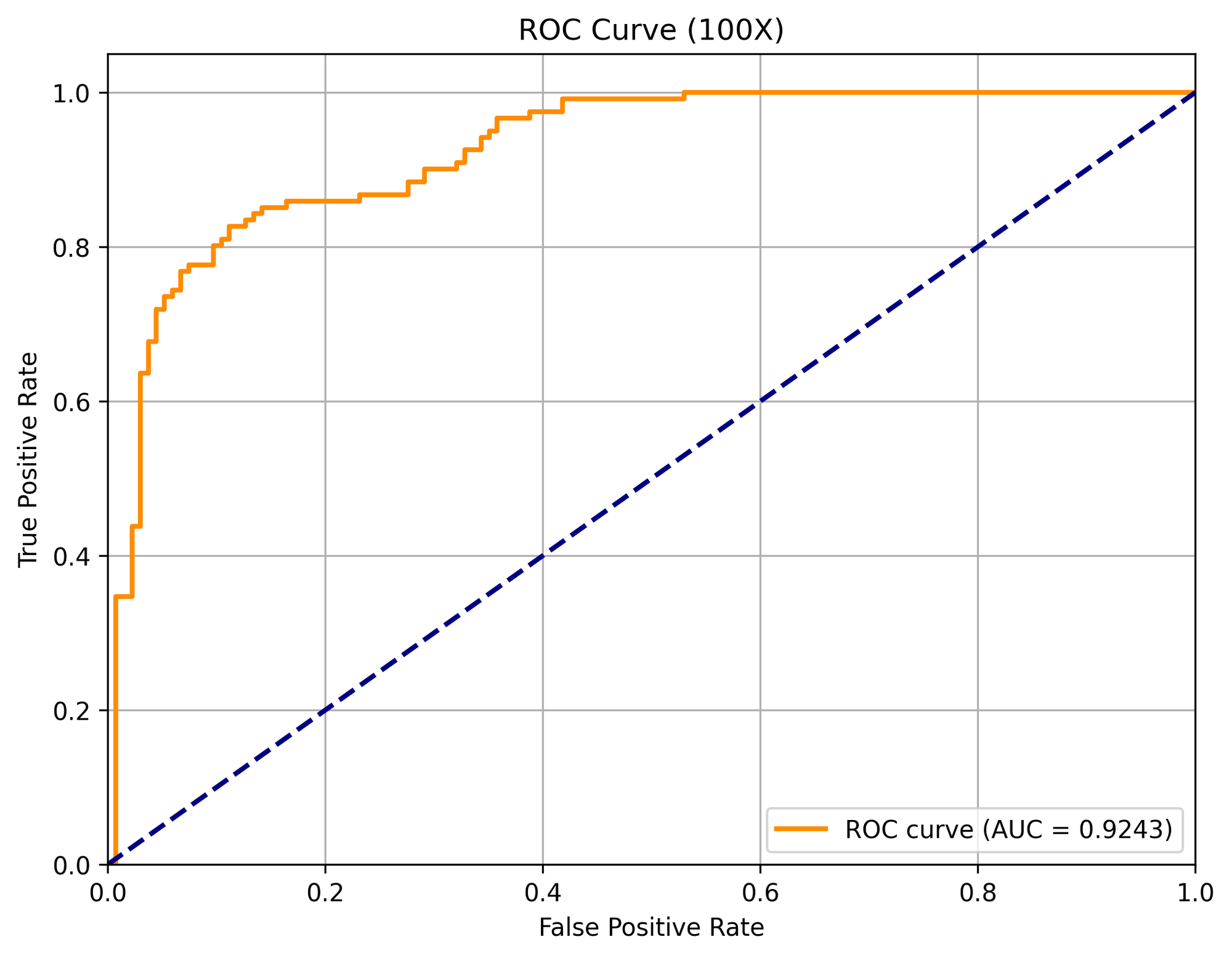}
}\hfill
\subfloat[ROC 200X]{%
 \includegraphics[width=0.23\textwidth]{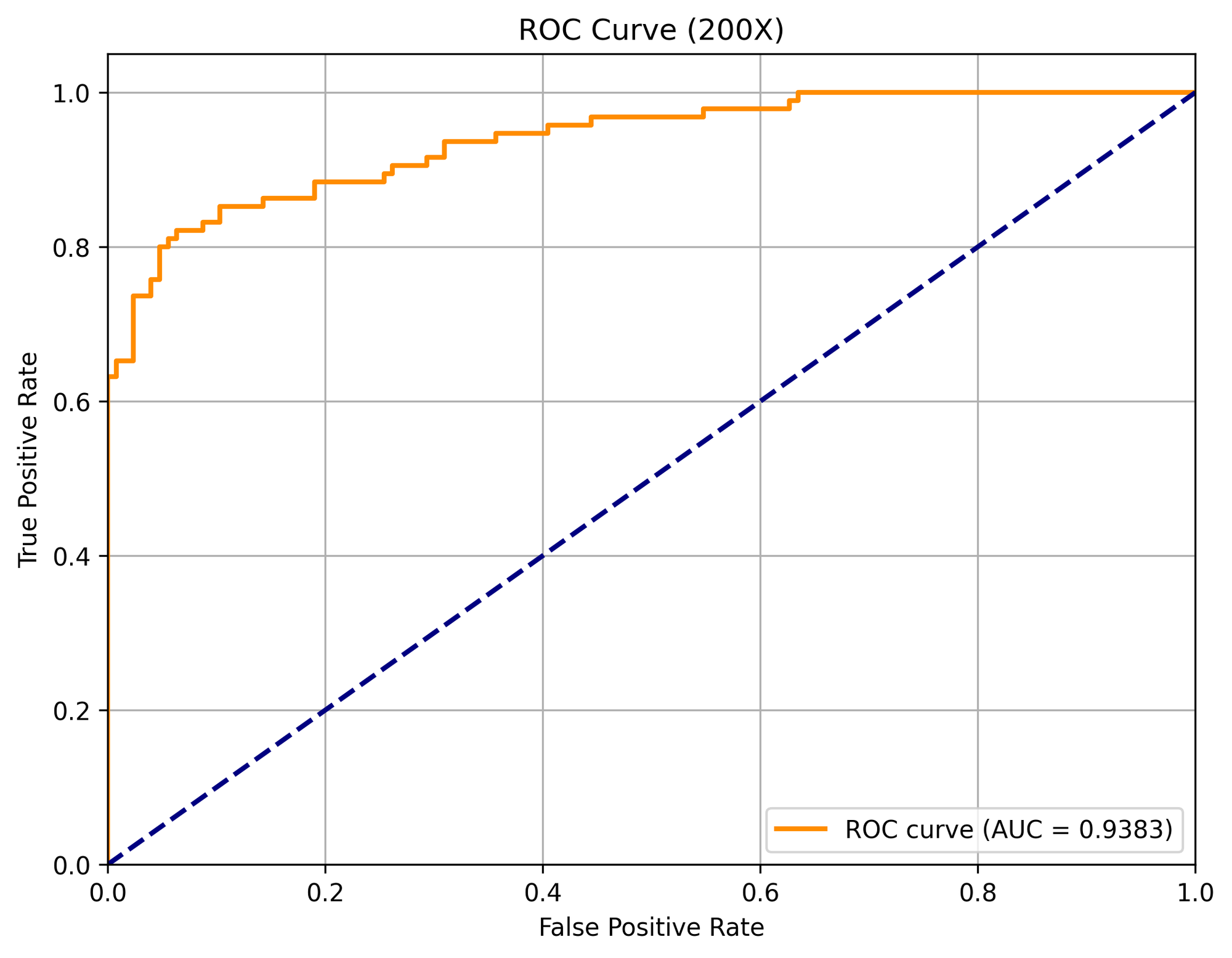}
}\hfill
\subfloat[ROC 400X]{%
 \includegraphics[width=0.23\textwidth]{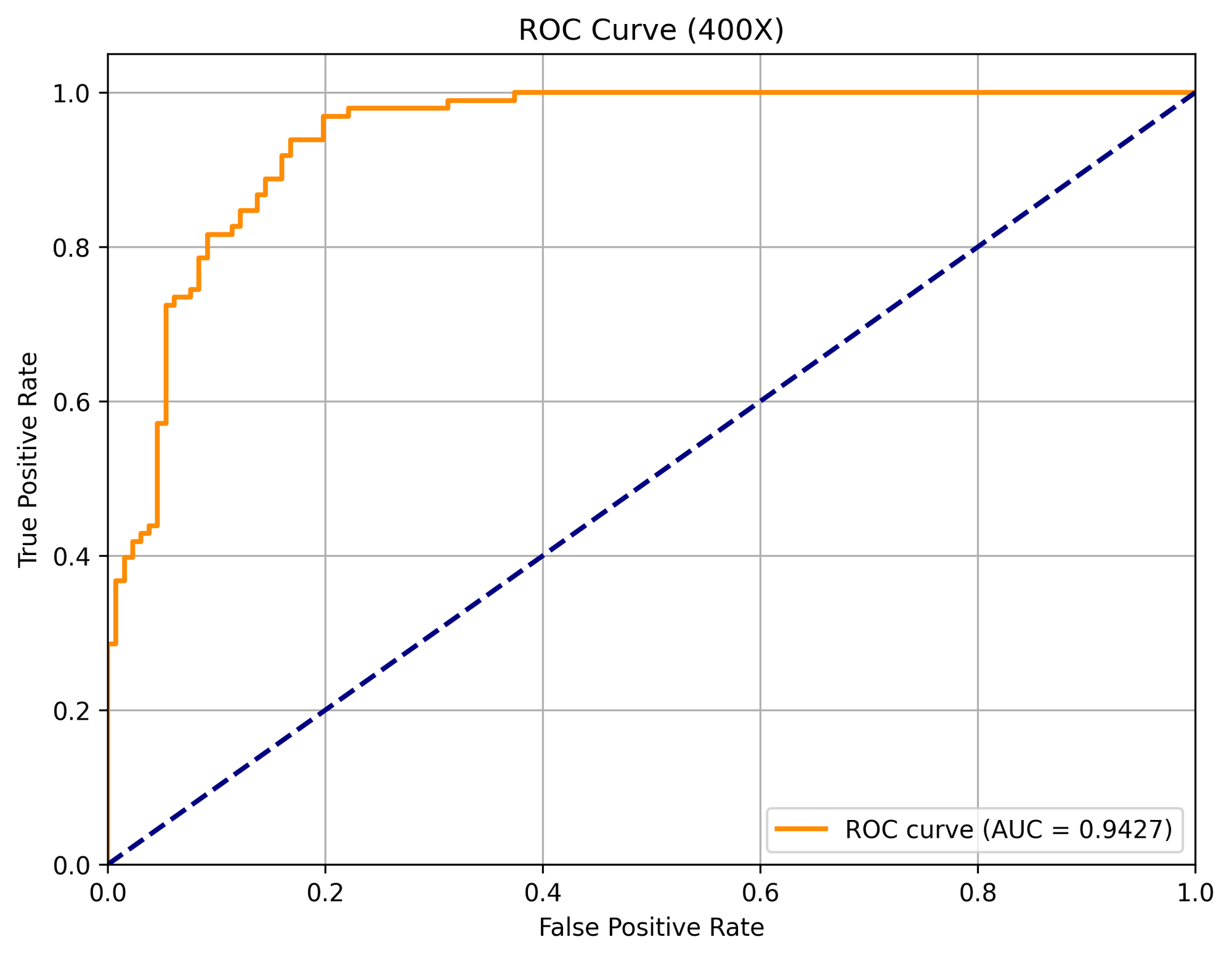}
}

\vspace{0.6ex}

\subfloat[PR 40X]{%
 \includegraphics[width=0.23\textwidth]{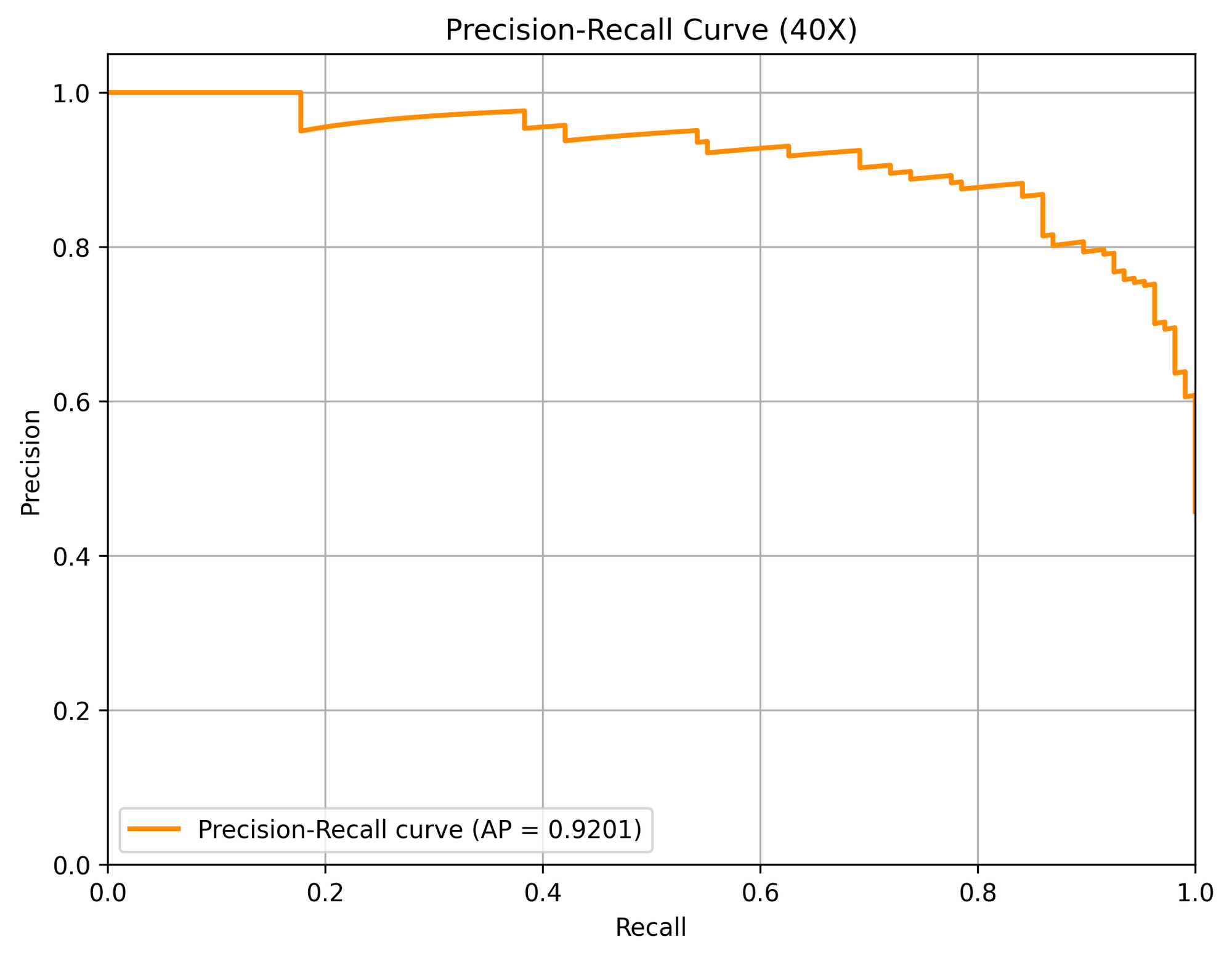}
}\hfill
\subfloat[PR 100X]{%
 \includegraphics[width=0.23\textwidth]{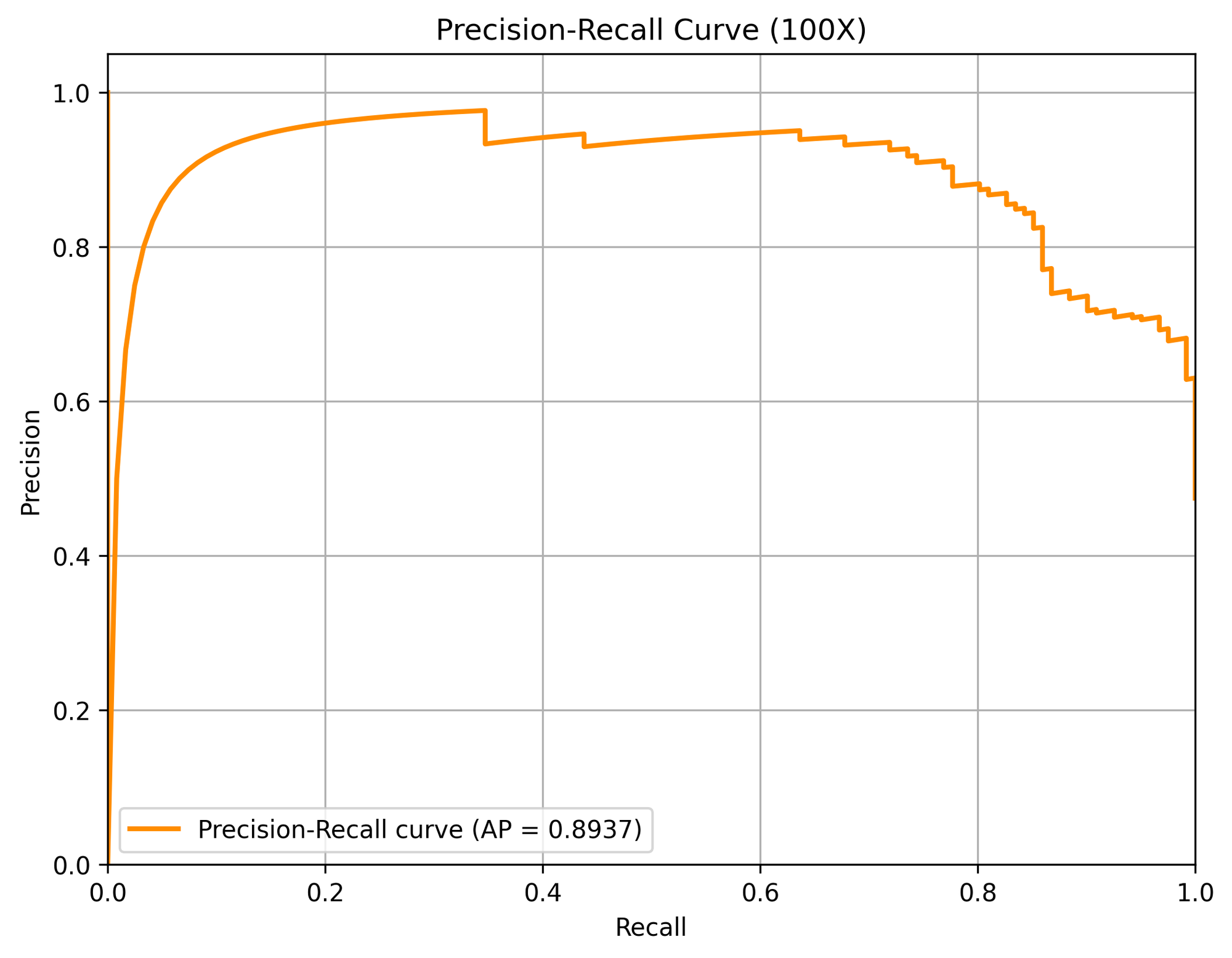}
}\hfill
\subfloat[PR 200X]{%
 \includegraphics[width=0.23\textwidth]{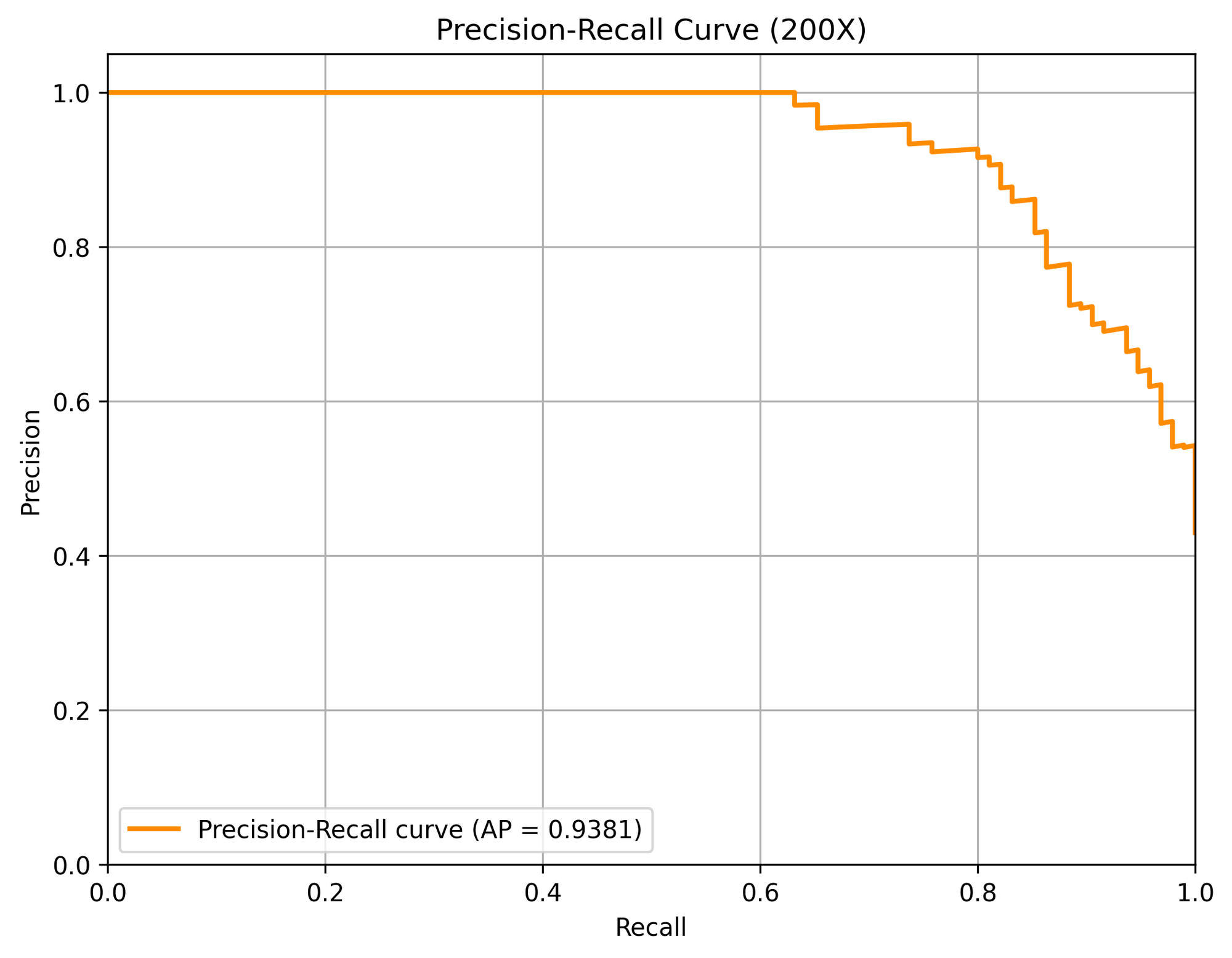}
}\hfill
\subfloat[PR 400X]{%
 \includegraphics[width=0.23\textwidth]{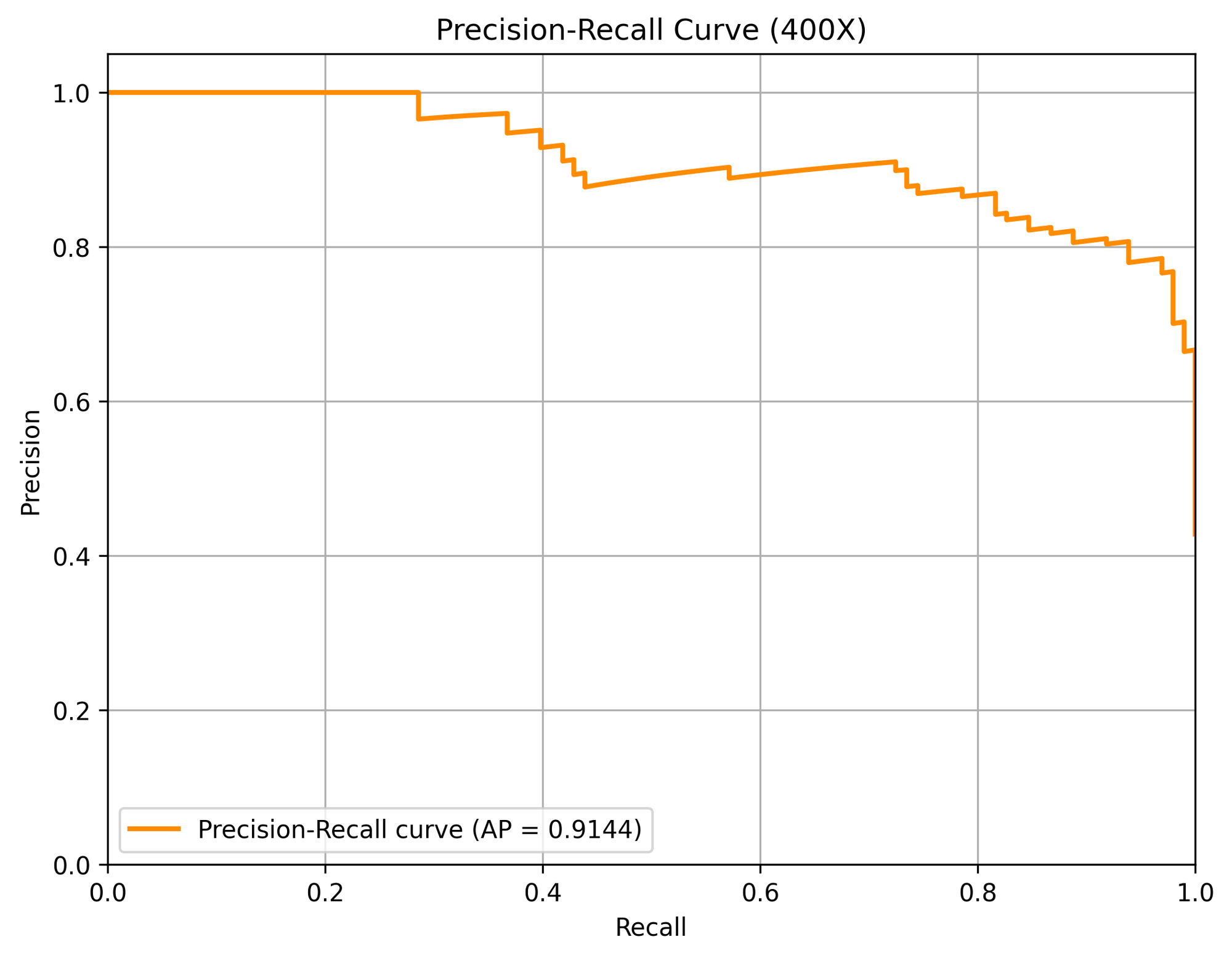}
}

\caption{Receiver Operating Characteristic (ROC) curves (a–d) and Precision–Recall (PR) curves (e–h) for the Inception model evaluated on the BreakHis dataset across 40X, 100X, 200X, and 400X magnifications.}
\label{fig:inception_roc_pr}
\end{figure}

\begin{table}[ht]
\caption{Performance of MobileNetV2 on the BreakHis dataset}
\label{tab:mobilenetv2}
\centering
\begin{tabular}{@{}lcccccc@{}}
\toprule
\textbf{Dataset} 
& \textbf{Accuracy (\%)} 
& \textbf{Precision (\%)} 
& \textbf{Recall (\%)} 
& \textbf{F1-score (\%)} 
& \textbf{Specificity (\%)} 
& \textbf{AUC} \\
\midrule
Validation (All Mag.) 
& 91.8 
& 95.9 
& 92.7 
& 94.3 
& 89.5 
& 0.975 \\

Test (All Mag.)    
& 84.7 
& 83.5 
& 81.9 
& 82.7 
& 86.9 
& 0.933 \\

Test (40$\times$)      
& 81.6 
& 84.0 
& 73.8 
& 78.6 
& 88.2 
& 0.923 \\

Test (100$\times$)      
& 84.7 
& 82.0 
& 86.8 
& 84.3 
& 82.8 
& 0.928 \\

Test (200$\times$)      
& 86.4 
& 85.0 
& 83.2 
& 84.0 
& 88.9 
& 0.956 \\

Test (400$\times$)      
& 84.3 
& 81.6 
& 81.6 
& 81.6 
& 86.3 
& 0.927 \\
\bottomrule
\end{tabular}
\end{table}

MobileNetV2 showed strong efficacy on the BreakHis validation set, attaining 91.84\% accuracy, 95.93\% precision, 92.70\% recall, 89.53\% specificity, 94.29\% F1-score, and 0.9746 AUC-ROC, so affirming effective generalization during training (Table 5). In the test set (across all magnifications), the findings diminished to 84.66\% accuracy, 83.54\% precision, 81.95\% recall, 86.87\% specificity, 82.73\% F1-score, and 0.9332 AUC-ROC, indicating a marginal reduction in sensitivity while maintaining balanced specificity (Figure 17, 18 \& 19). At several magnifications, performance exhibited variability: 40X (81.62\% accuracy, 73.83\% recall, AUC 0.9229) shown diminished sensitivity, whereas 100X (84.71\%, AUC 0.9284), 200X (86.43\%, AUC 0.9558), and 400X (84.28\%, AUC 0.9271) yielded more consistent outcomes, with recall and specificity ranging from 82\% to 89\% (Figure 20). The results demonstrate that MobileNetV2, utilizing lightweight depthwise separable convolutions, achieves competitive accuracy and efficiency; however, its diminished capacity relative to deeper architectures leads to decreased recall and heightened magnification sensitivity in test data.

\begin{figure}[H]
 \centering
 \includegraphics[width=0.85\linewidth]{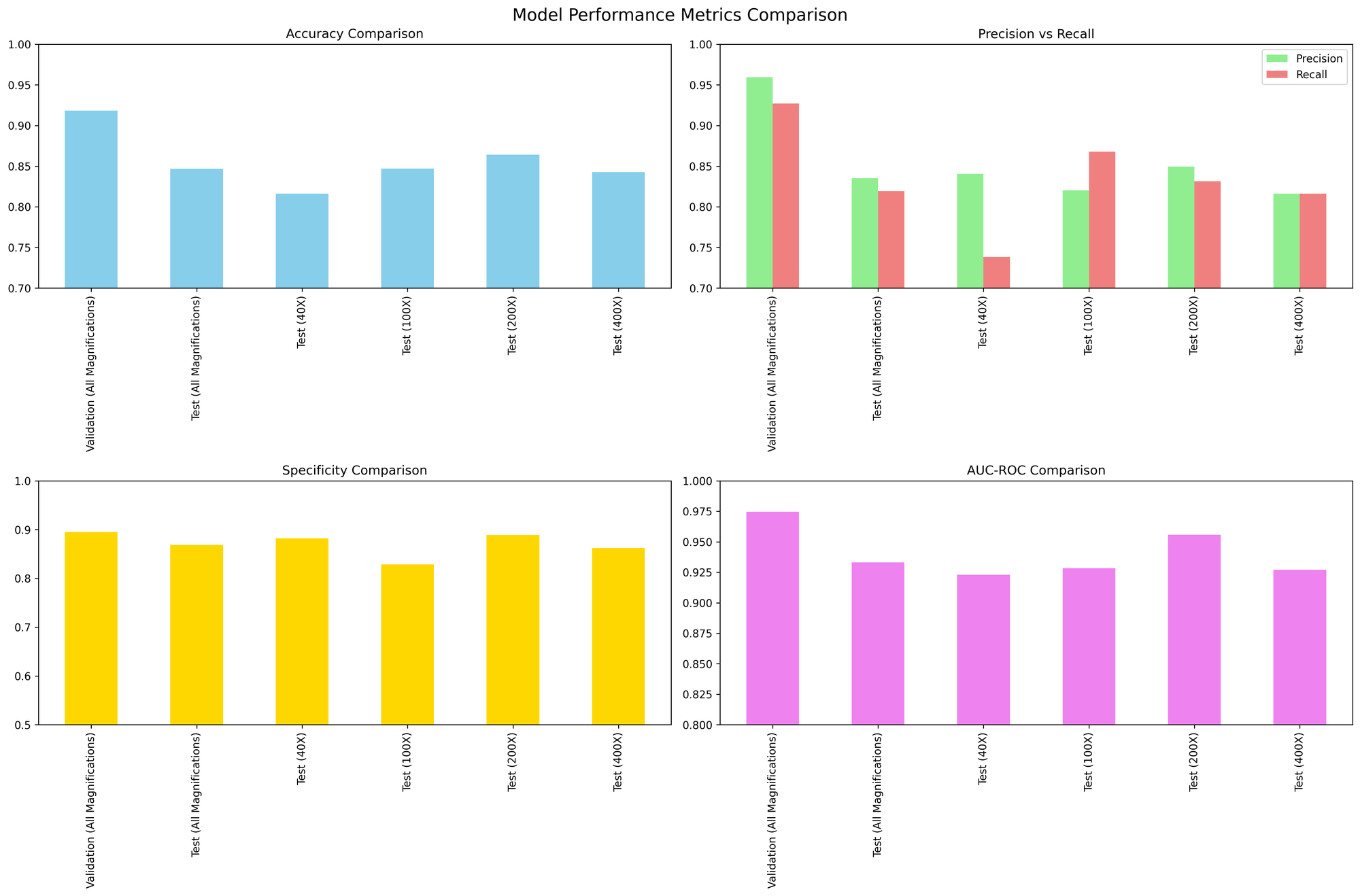}
 \caption{MobileNet performance across validation and test sets at multiple magnifications}
 \label{fig:single4}
\end{figure}

\begin{figure}[H]
\centering

\subfloat[40X]{%
 \includegraphics[width=0.23\textwidth]{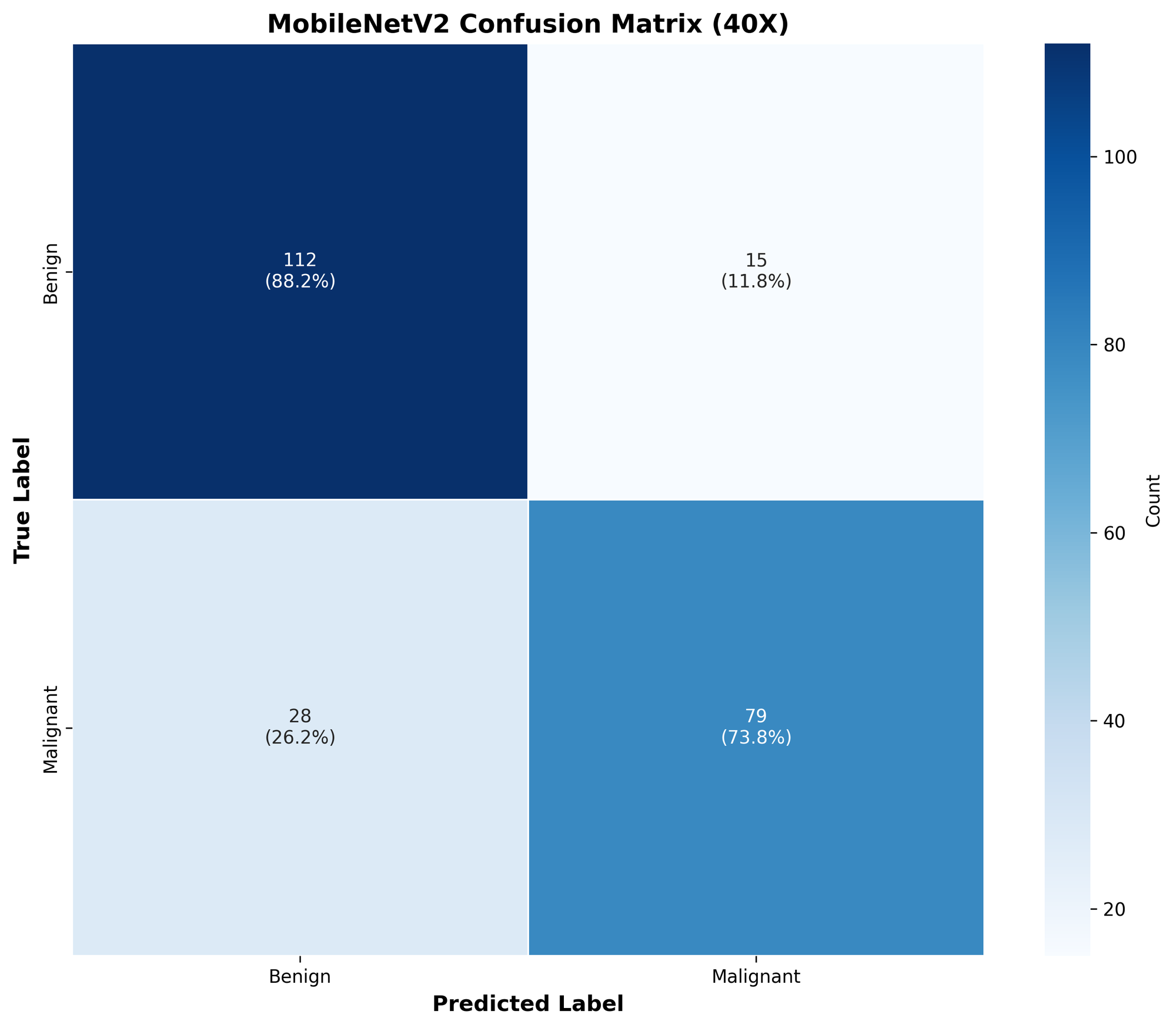}
}\hfill
\subfloat[100X]{%
 \includegraphics[width=0.23\textwidth]{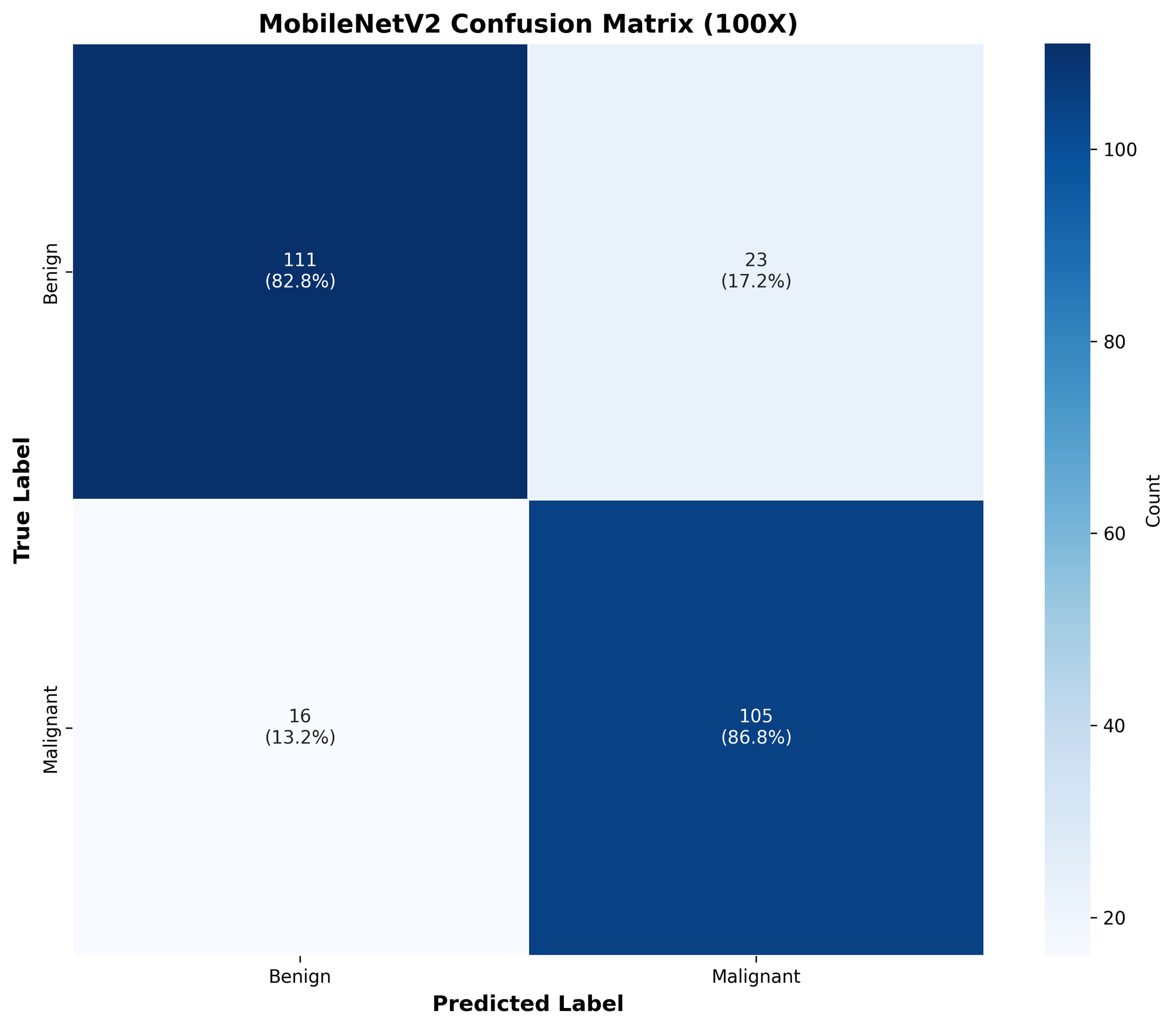}
}\hfill
\subfloat[200X]{%
 \includegraphics[width=0.23\textwidth]{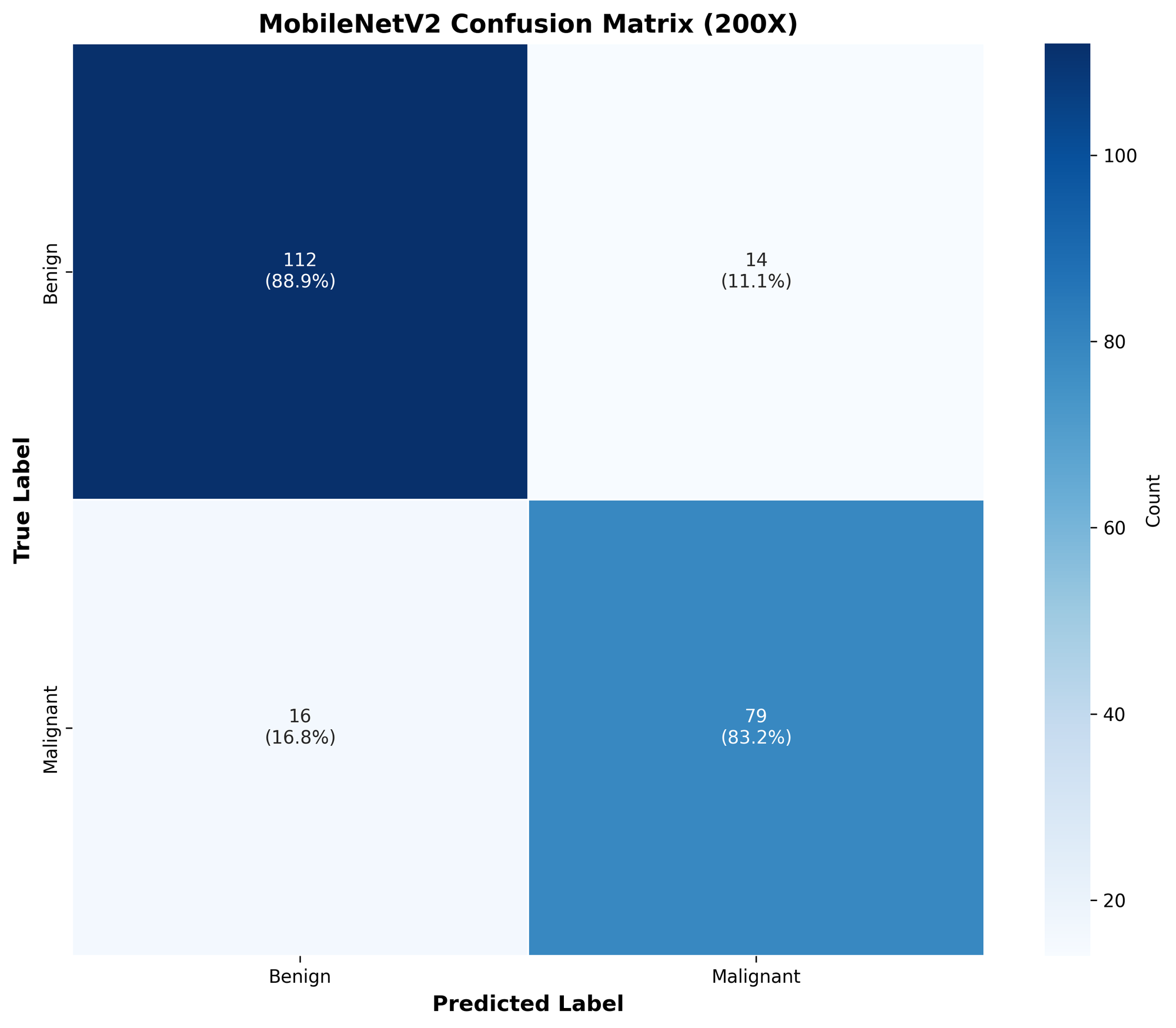}
}\hfill
\subfloat[400X]{%
 \includegraphics[width=0.23\textwidth]{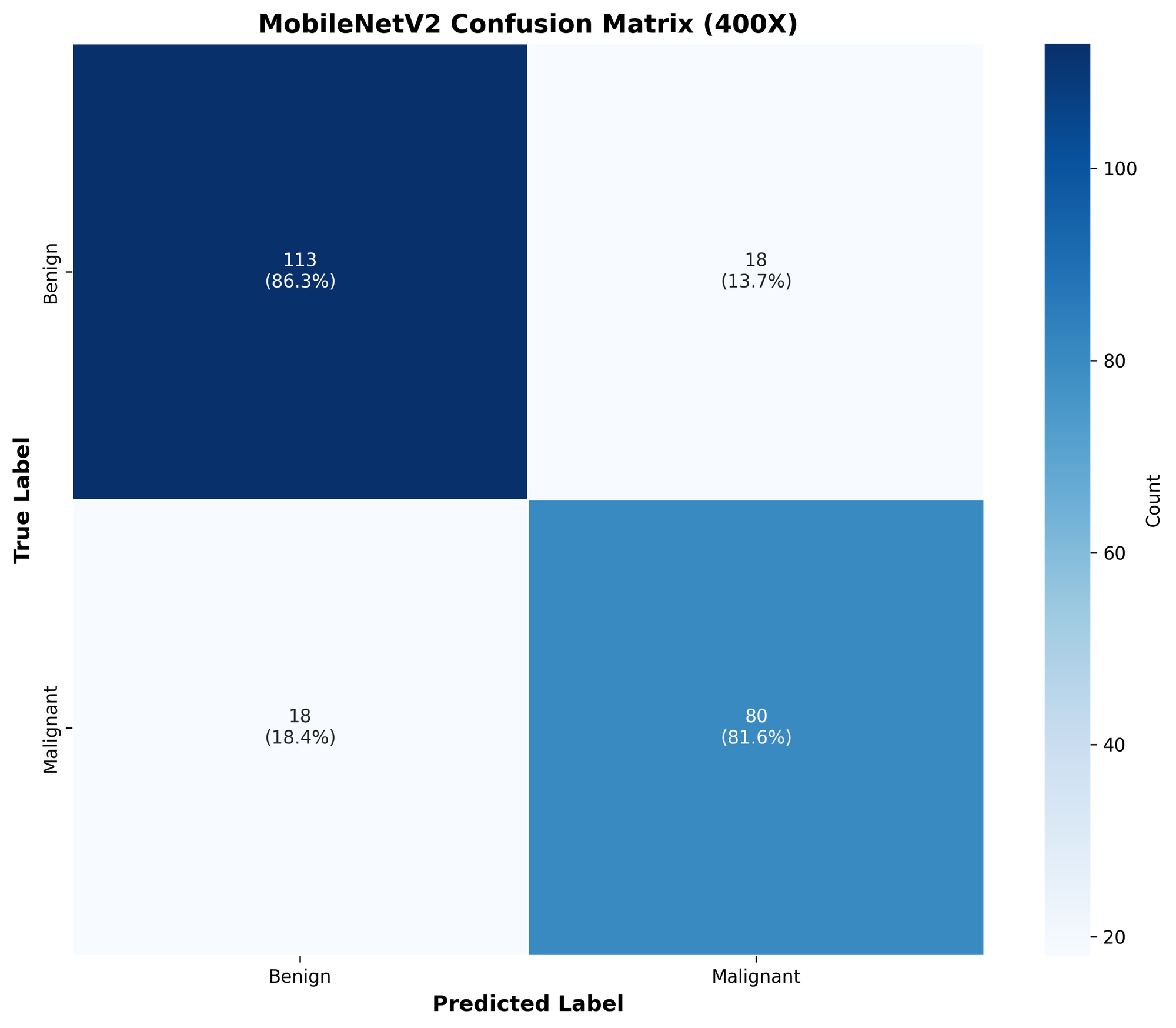}
}

\caption{Confusion matrices of the MobileNetV2 model across 40X, 100X, 200X, and 400X magnifications on the BreakHis dataset.}
\label{fig:mobilenetv2_confusion_matrices}
\end{figure}

\begin{figure}[H]
\centering

\subfloat[Validation ROC]{%
 \includegraphics[width=0.23\textwidth]{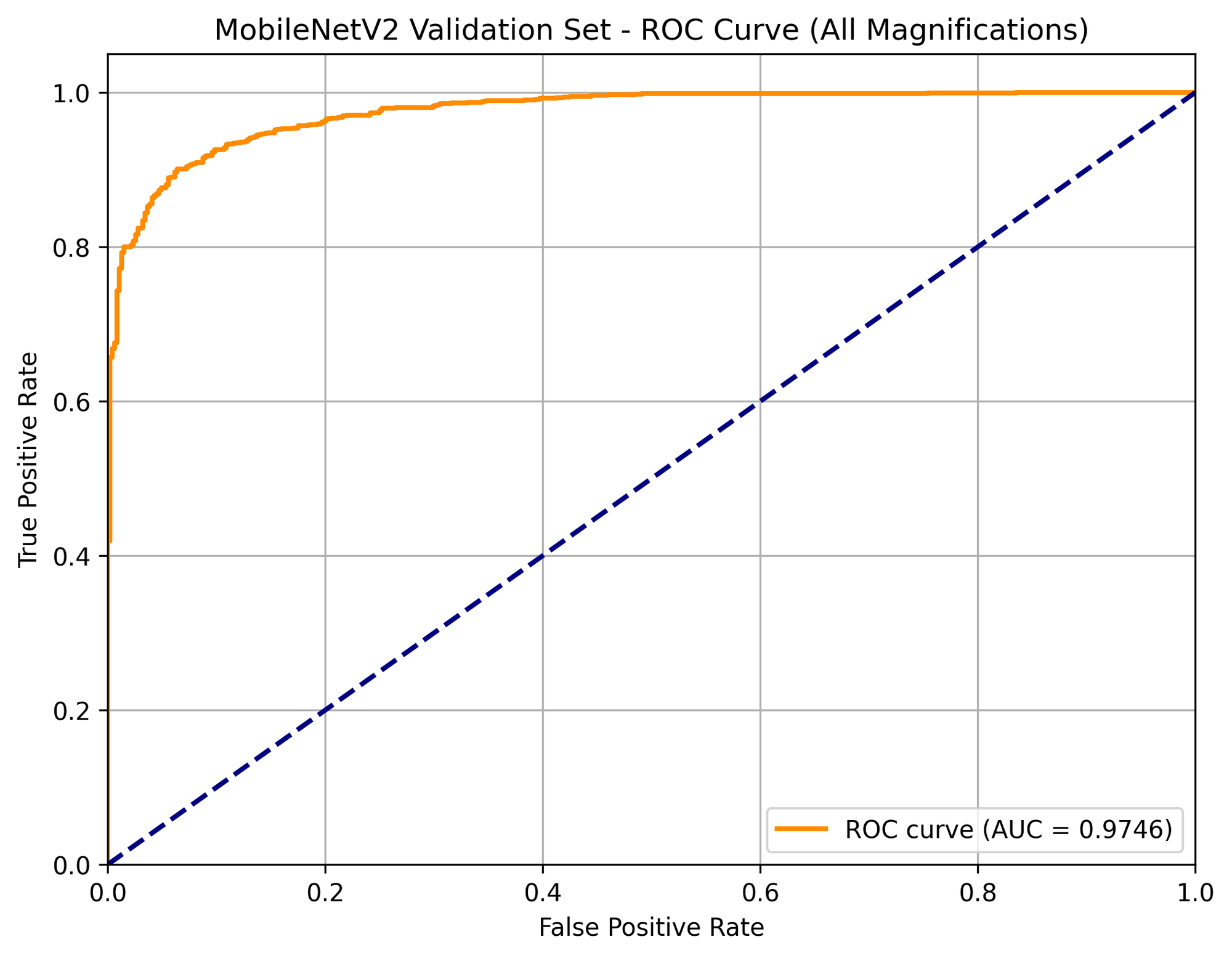}
}\hfill
\subfloat[Validation Confusion Matrix]{%
 \includegraphics[width=0.23\textwidth]{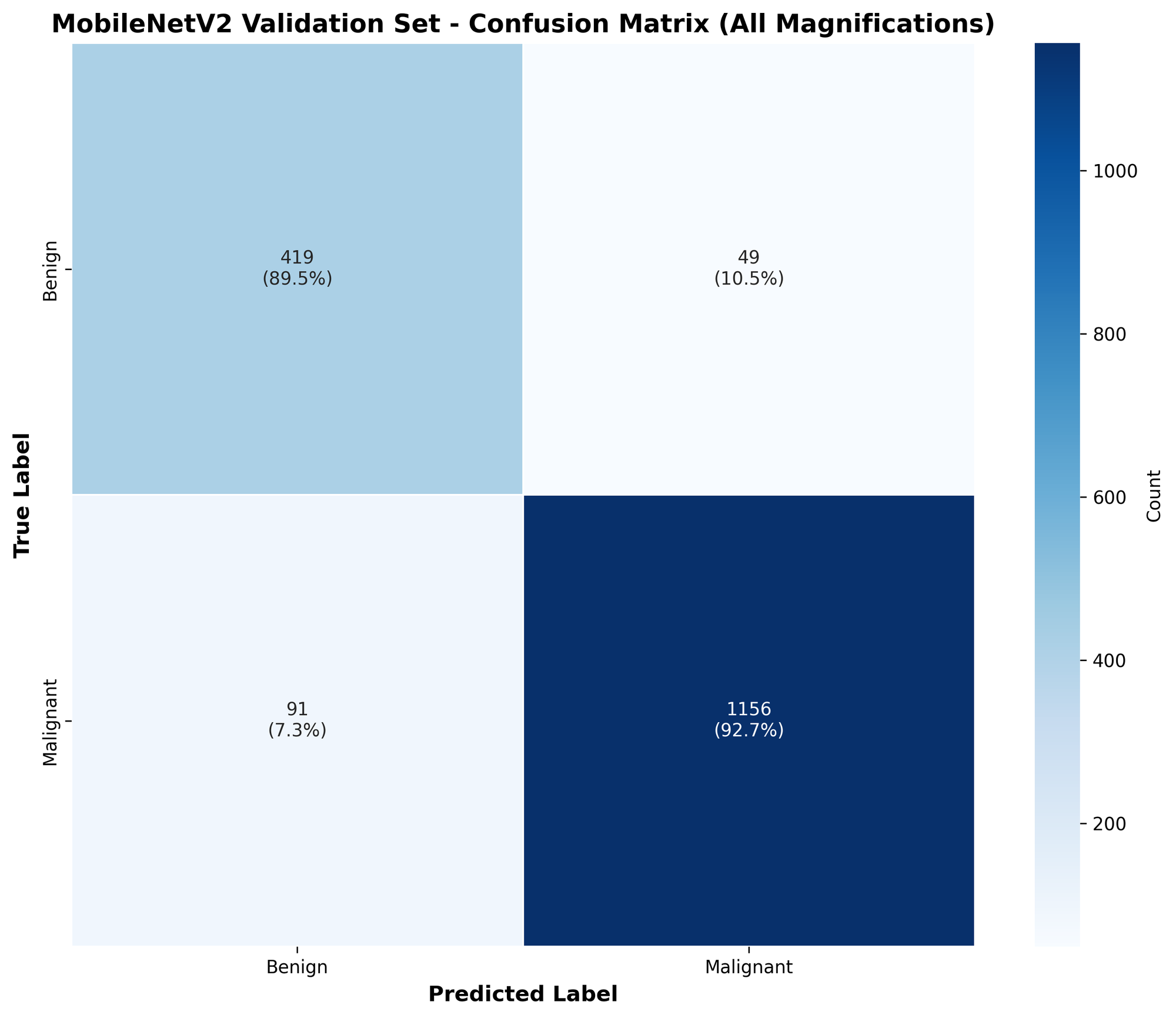}
}\hfill
\subfloat[Test ROC]{%
 \includegraphics[width=0.23\textwidth]{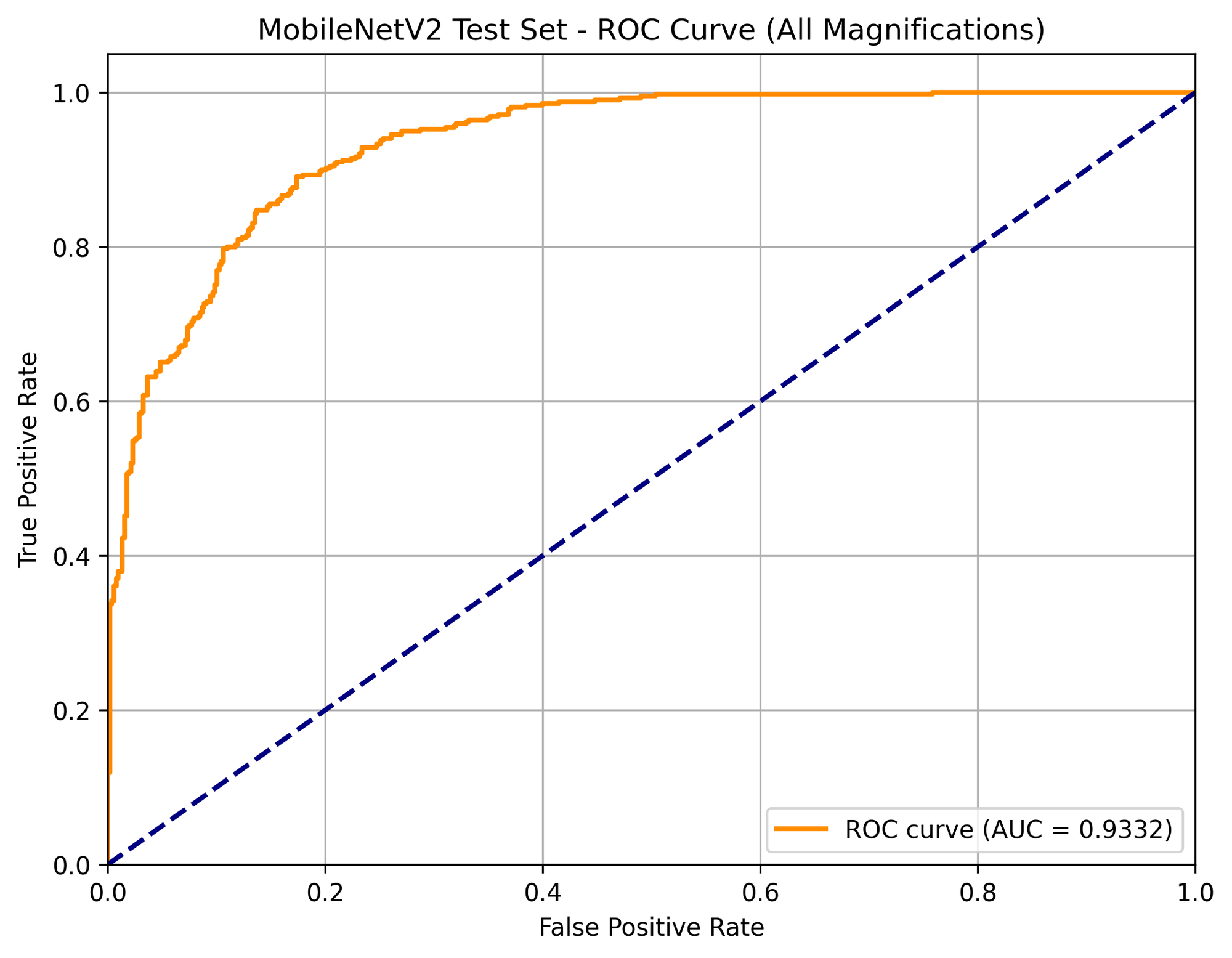}
}\hfill
\subfloat[Test Confusion Matrix]{%
 \includegraphics[width=0.23\textwidth]{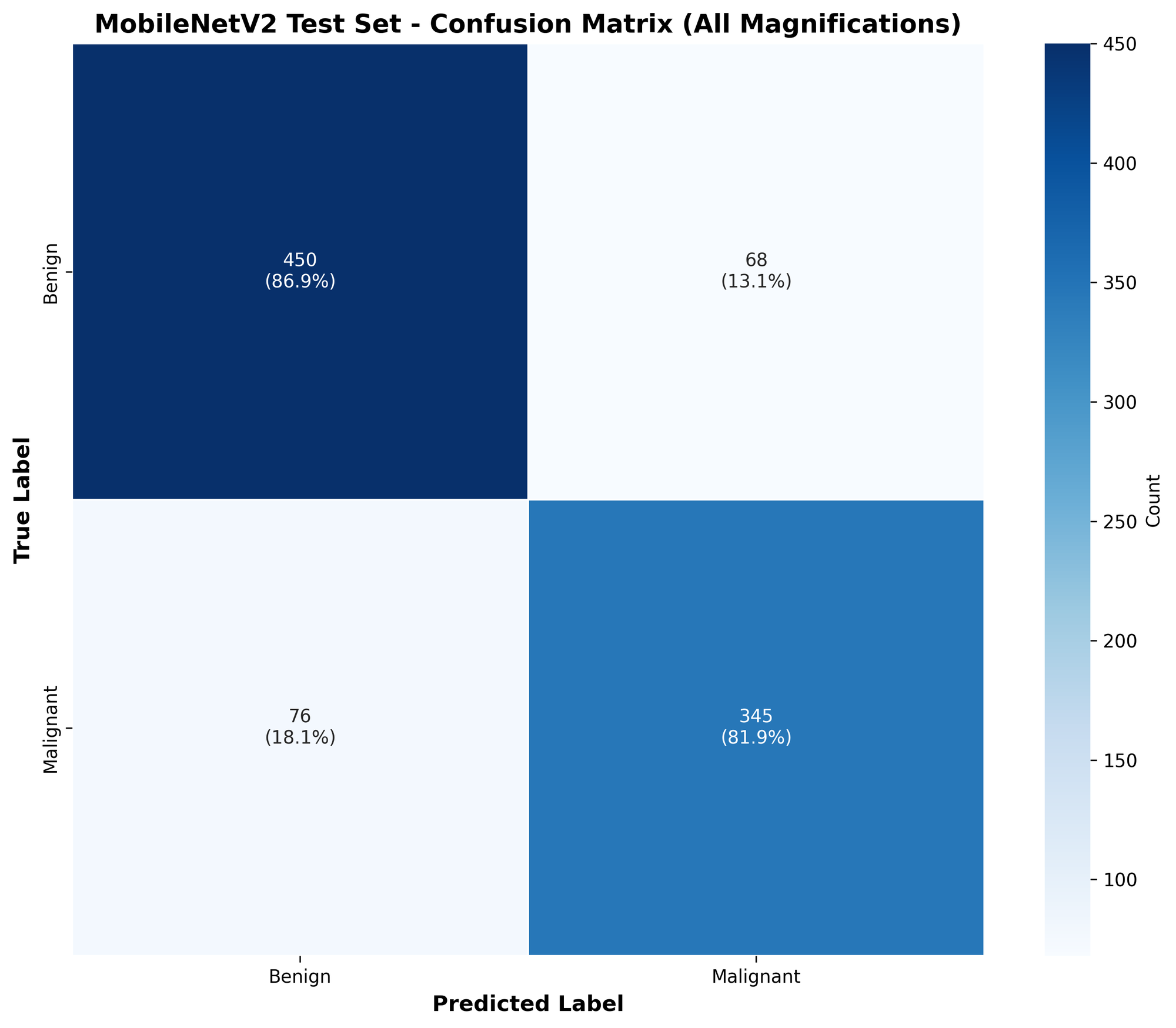}
}

\caption{Validation and test performance of the MobileNetV2 model}
\label{fig:mobilenetv2_val_test}
\end{figure}

\begin{figure}[H]
\centering

\subfloat[ROC 40X]{%
 \includegraphics[width=0.23\textwidth]{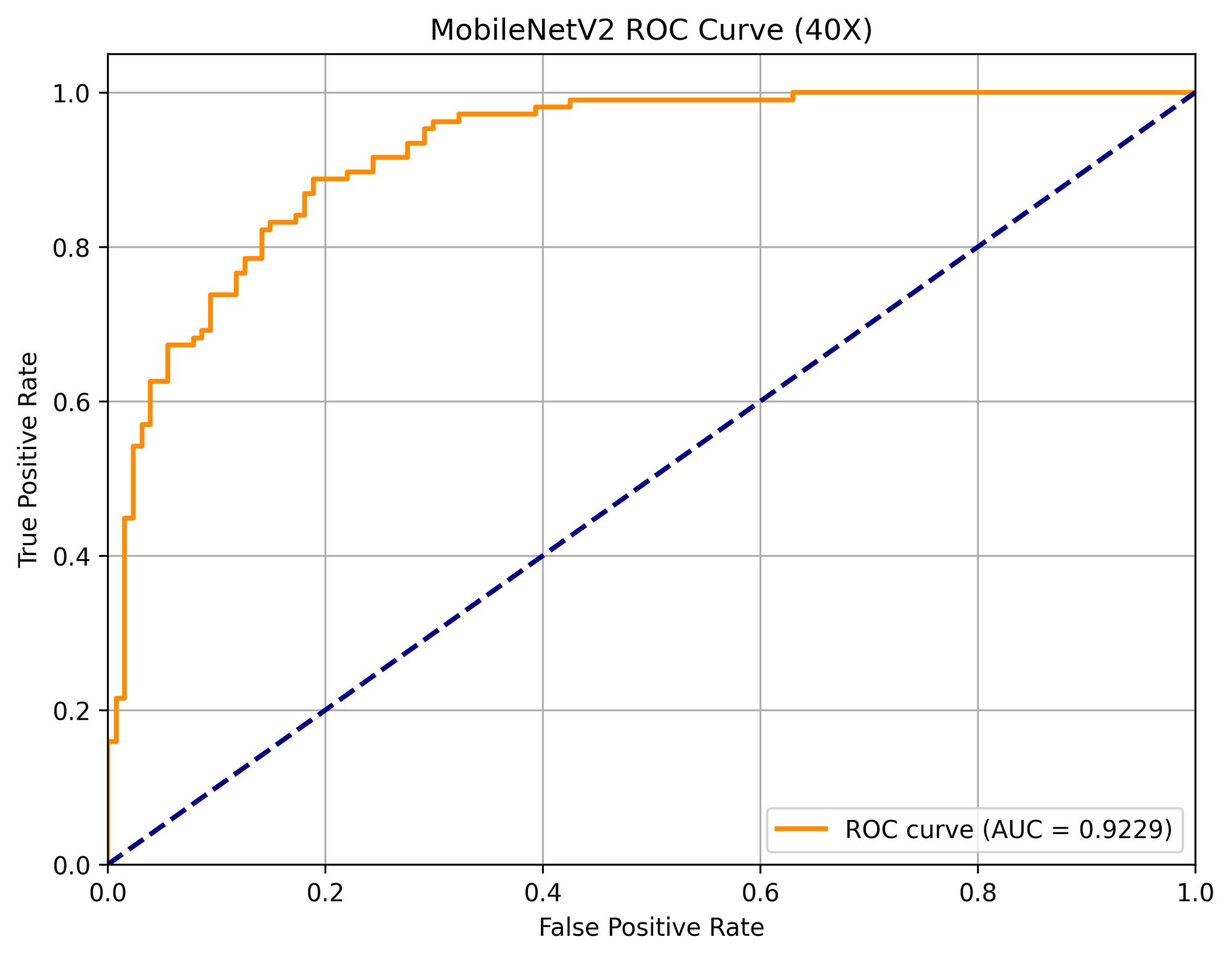}
}\hfill
\subfloat[ROC 100X]{%
 \includegraphics[width=0.23\textwidth]{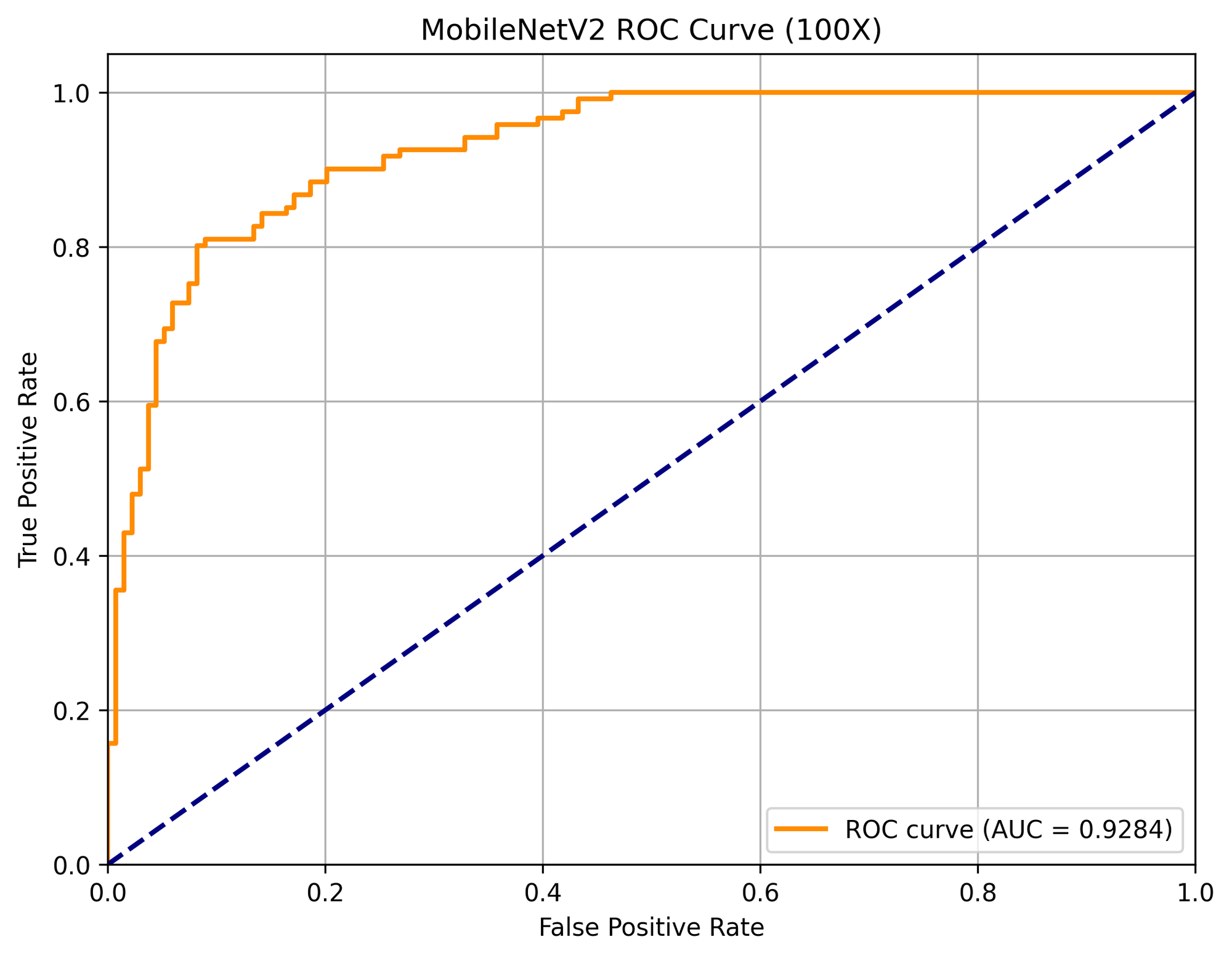}
}\hfill
\subfloat[ROC 200X]{%
 \includegraphics[width=0.23\textwidth]{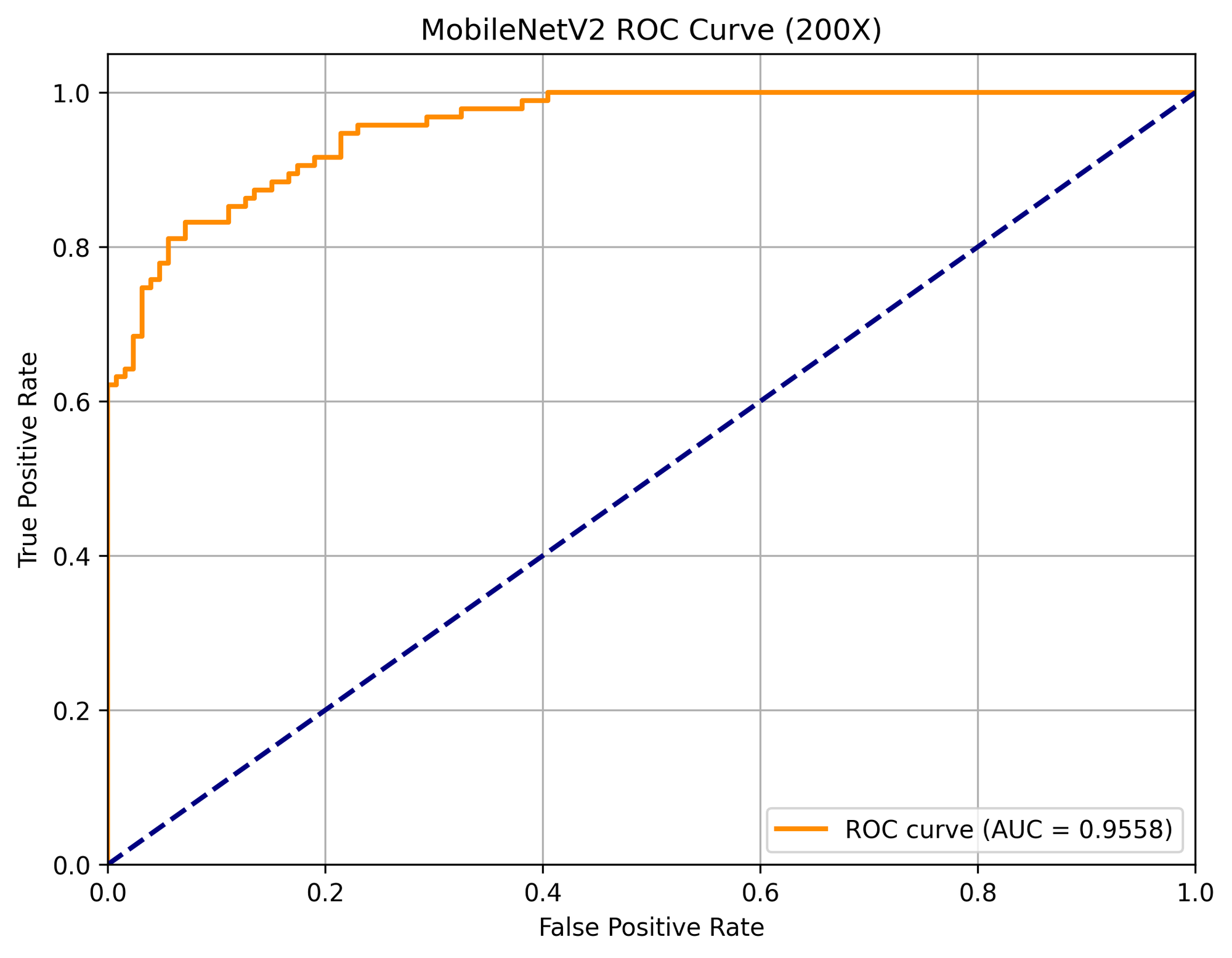}
}\hfill
\subfloat[ROC 400X]{%
 \includegraphics[width=0.23\textwidth]{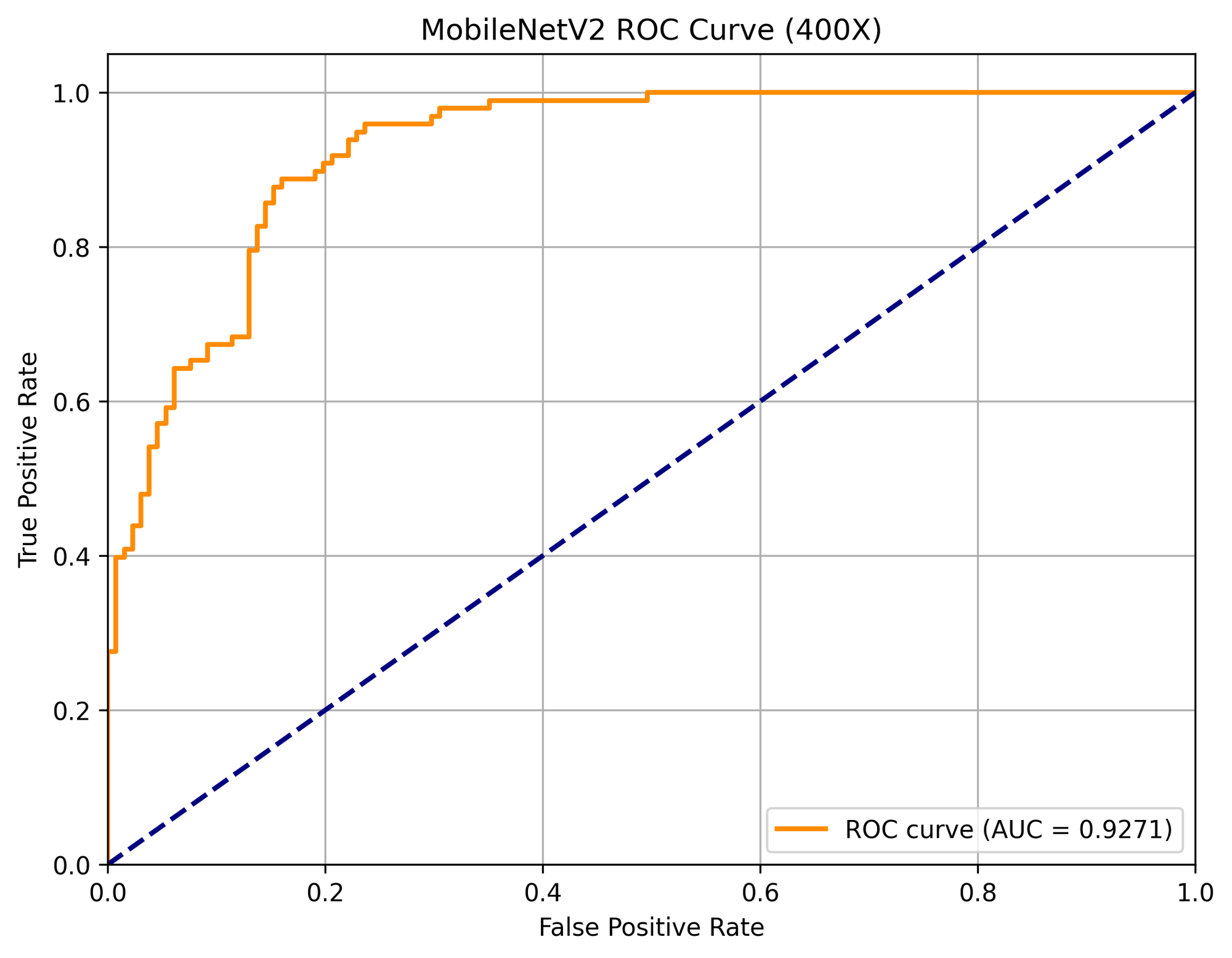}
}

\vspace{0.6ex}

\subfloat[PR 40X]{%
 \includegraphics[width=0.23\textwidth]{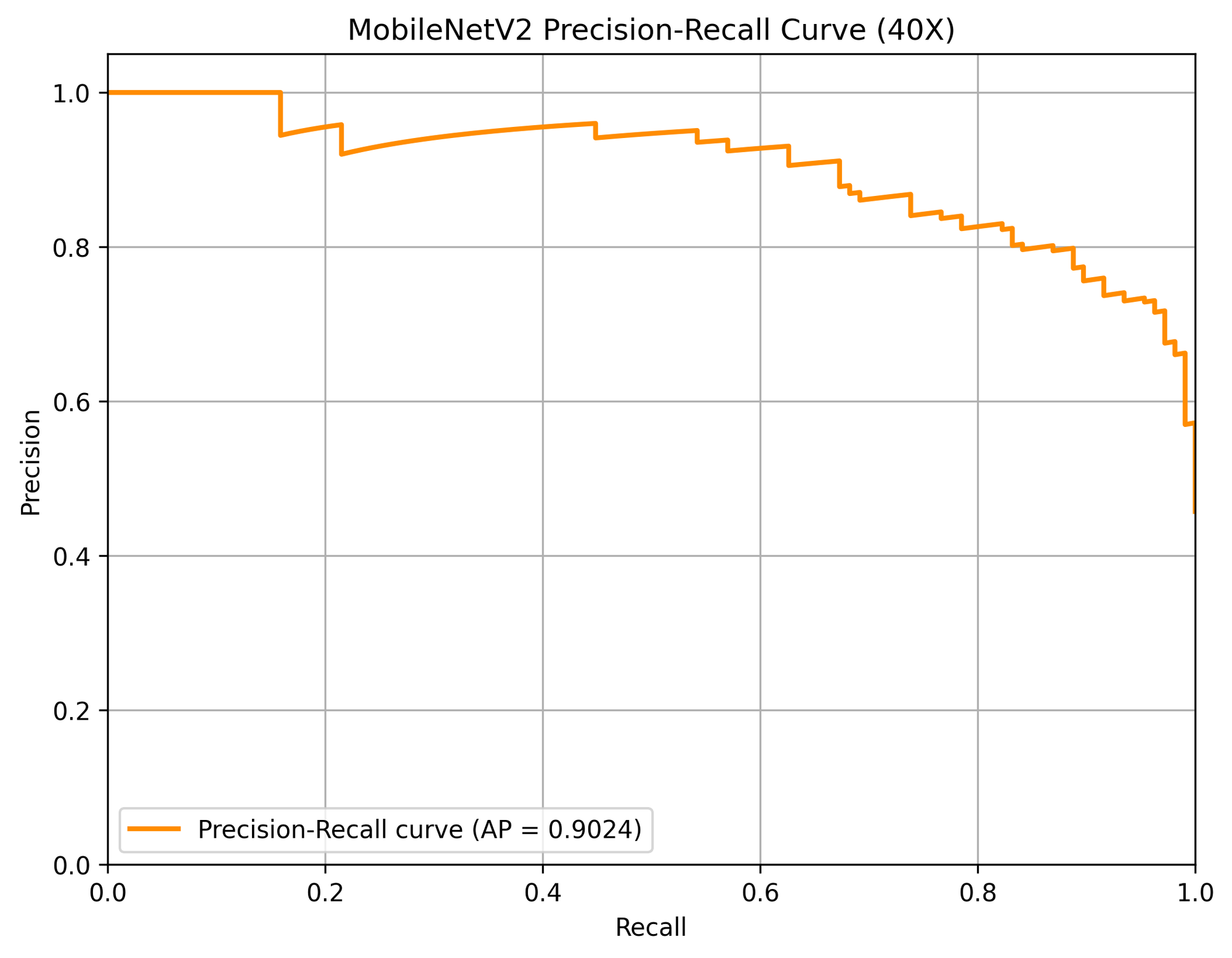}
}\hfill
\subfloat[PR 100X]{%
 \includegraphics[width=0.23\textwidth]{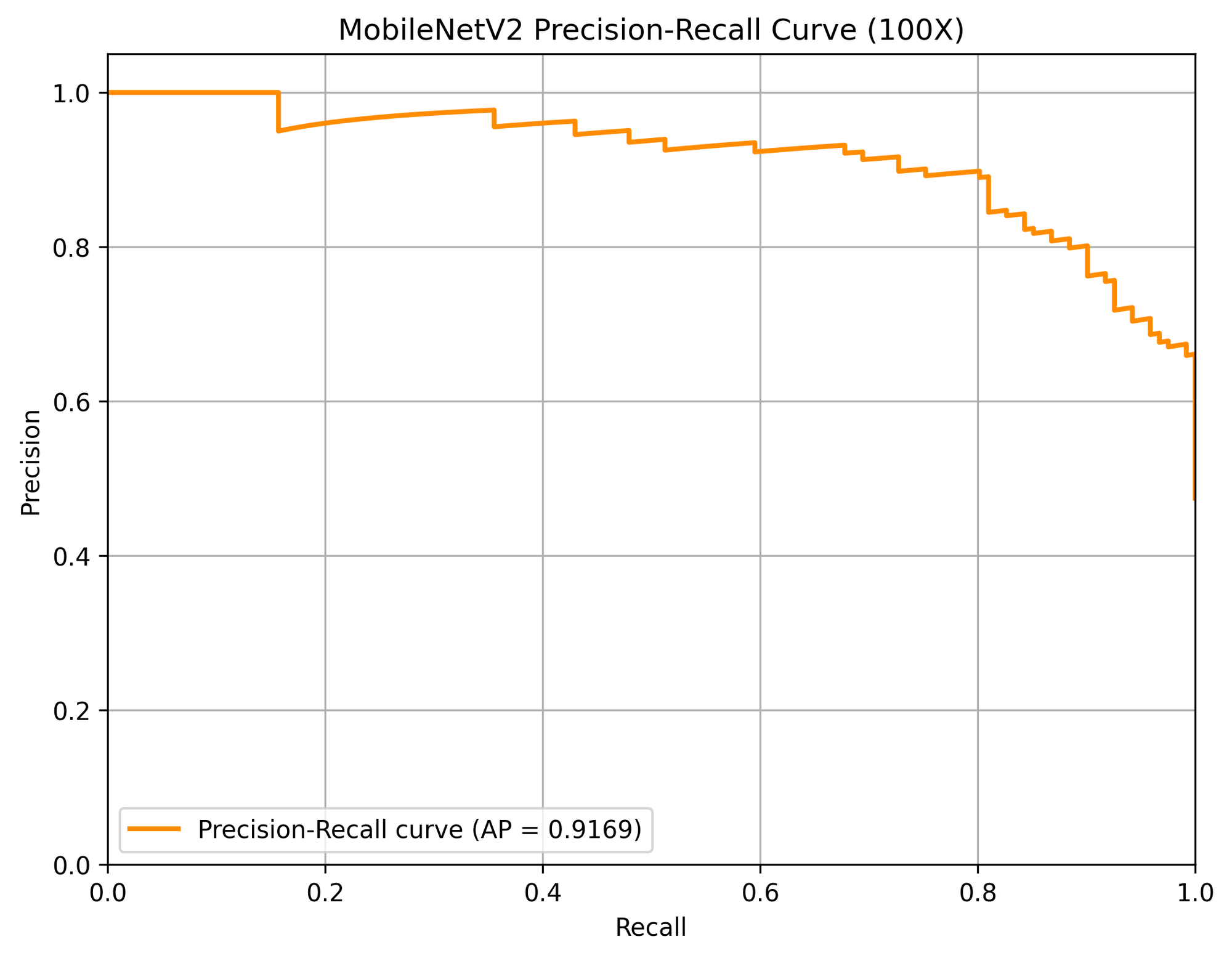}
}\hfill
\subfloat[PR 200X]{%
 \includegraphics[width=0.23\textwidth]{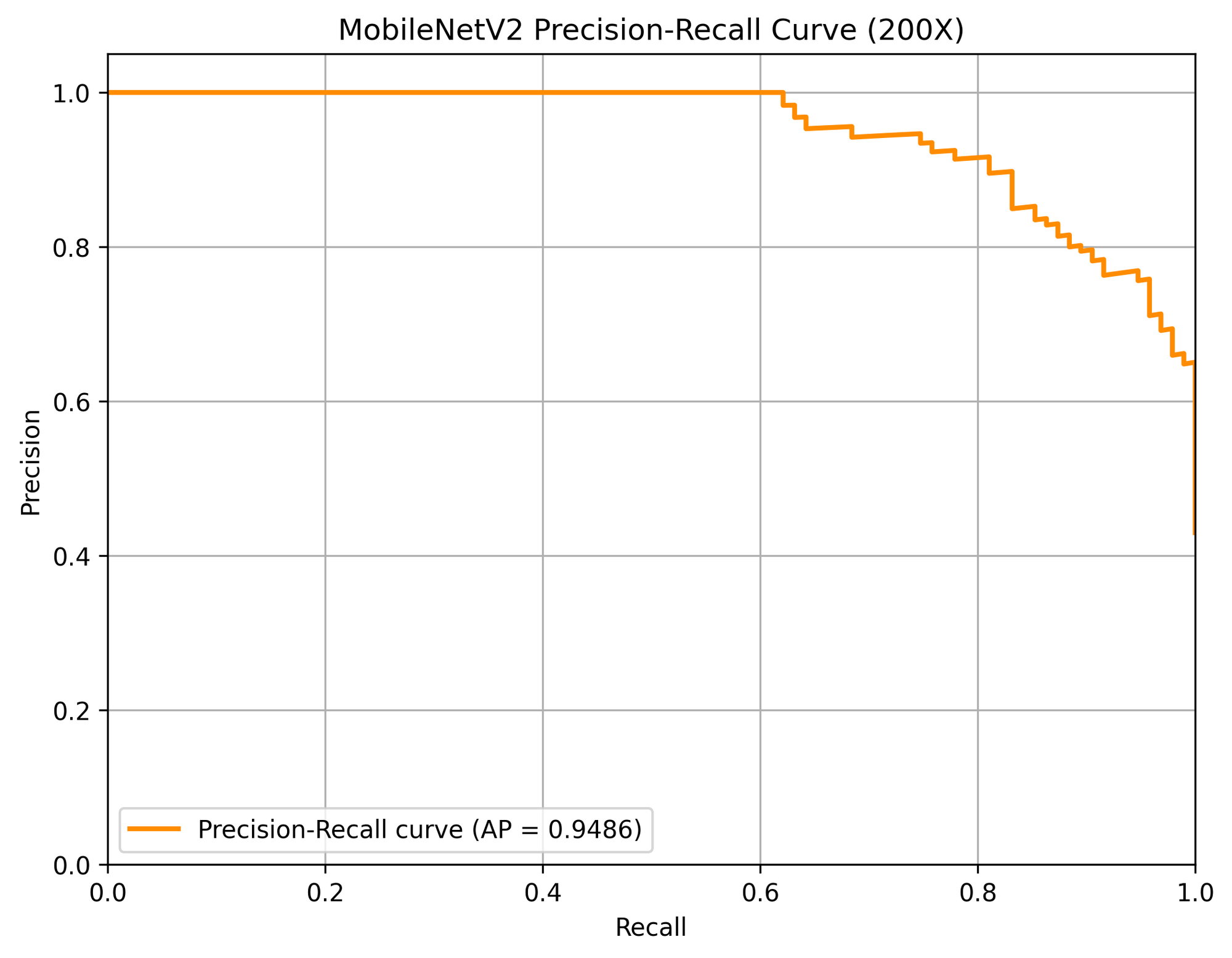}
}\hfill
\subfloat[PR 400X]{%
 \includegraphics[width=0.23\textwidth]{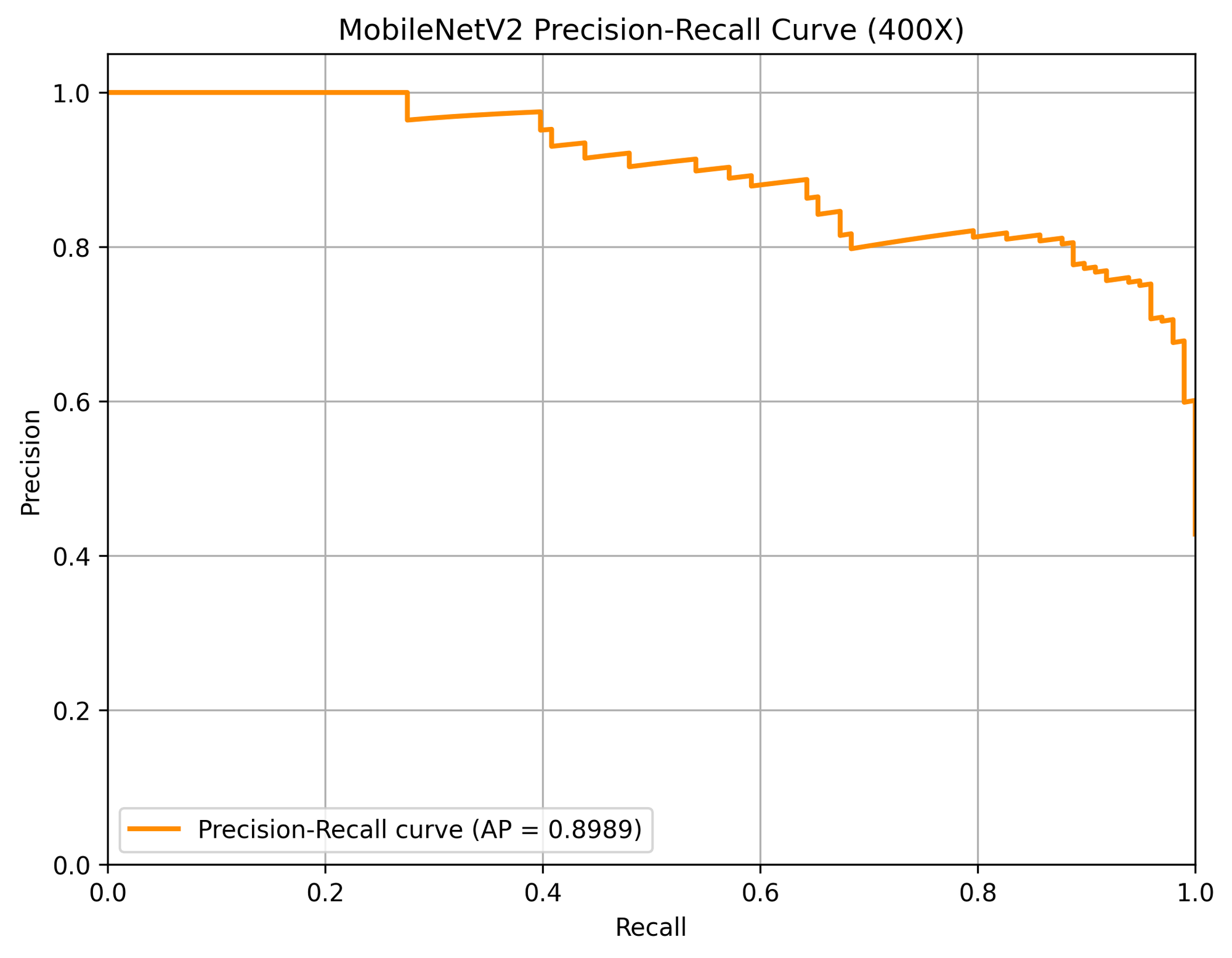}
}

\caption{Receiver Operating Characteristic (ROC) curves (a–d) and Precision–Recall (PR) curves (e–h) for the MobileNetV2 model evaluated on the BreakHis dataset across 40X, 100X, 200X, and 400X magnifications.}
\label{fig:mobilenetv2_roc_pr}
\end{figure}

\begin{table}[ht]
\caption{Performance of ResNet50 on the BreakHis dataset}
\label{tab:resnet50}
\centering
\begin{tabular}{@{}lcccccc@{}}
\toprule
\textbf{Dataset} 
& \textbf{Accuracy (\%)} 
& \textbf{Precision (\%)} 
& \textbf{Recall (\%)} 
& \textbf{F1-score (\%)} 
& \textbf{Specificity (\%)} 
& \textbf{AUC} \\
\midrule
Validation (All Mag.) 
& 93.2 
& 96.2 
& 94.4 
& 95.3 
& 90.2 
& 0.978 \\

Test (All Mag.)    
& 83.6 
& 79.1 
& 86.2 
& 82.5 
& 81.5 
& 0.919 \\

Test (40$\times$)      
& 87.2 
& 87.4 
& 84.1 
& 85.7 
& 89.8 
& 0.914 \\

Test (100$\times$)      
& 82.4 
& 80.6 
& 82.6 
& 81.6 
& 82.1 
& 0.925 \\

Test (200$\times$)      
& 83.3 
& 76.4 
& 88.4 
& 81.9 
& 79.4 
& 0.928 \\

Test (400$\times$)      
& 80.8 
& 73.3 
& 86.7 
& 79.4 
& 76.3 
& 0.907 \\
\bottomrule
\end{tabular}
\end{table}

ResNet50 achieved excellent generalization on the BreakHis validation set, achieving 93.24\% accuracy, 96.24\% precision, 94.39\% recall, 90.17\% specificity, 95.30\% F1-score, and 0.9784 AUC-ROC, indicating an adequate sensitivity–specificity balance during training (Table 6). In the test set (across all magnifications), performance decreased to 83.60\% accuracy, 79.08\% precision, 86.22\% recall, 81.47\% specificity, 82.50\% F1-score, and 0.9185 AUC-ROC, signifying diminished precision and an increase in false positives (Figure 21, 22 \& 23). At various magnifications, the model exhibited optimal performance at 40X, achieving 87.18\% accuracy, 87.38\% precision, 84.11\% recall, 89.76\% specificity, and an AUC of 0.9135. In contrast, the results at 100X yielded 82.35\% accuracy with a 92.51 AUC, and at 200X, 83.26\% accuracy with a 92.82 AUC, demonstrating commendable recall but diminished precision (Figure 24). Performance was suboptimal at 400X (80.79\% accuracy, 73.28\% precision, 86.73\% recall, AUC 0.9065), indicating difficulties in collecting intricate elements at elevated magnification. ResNet50 exhibited dependable malignant diagnosis with excellent recall; however, diminished precision and variability across magnifications underscore its sensitivity to scale variations in histopathological pictures.

\begin{figure}[H]
 \centering
 \includegraphics[width=0.85\linewidth]{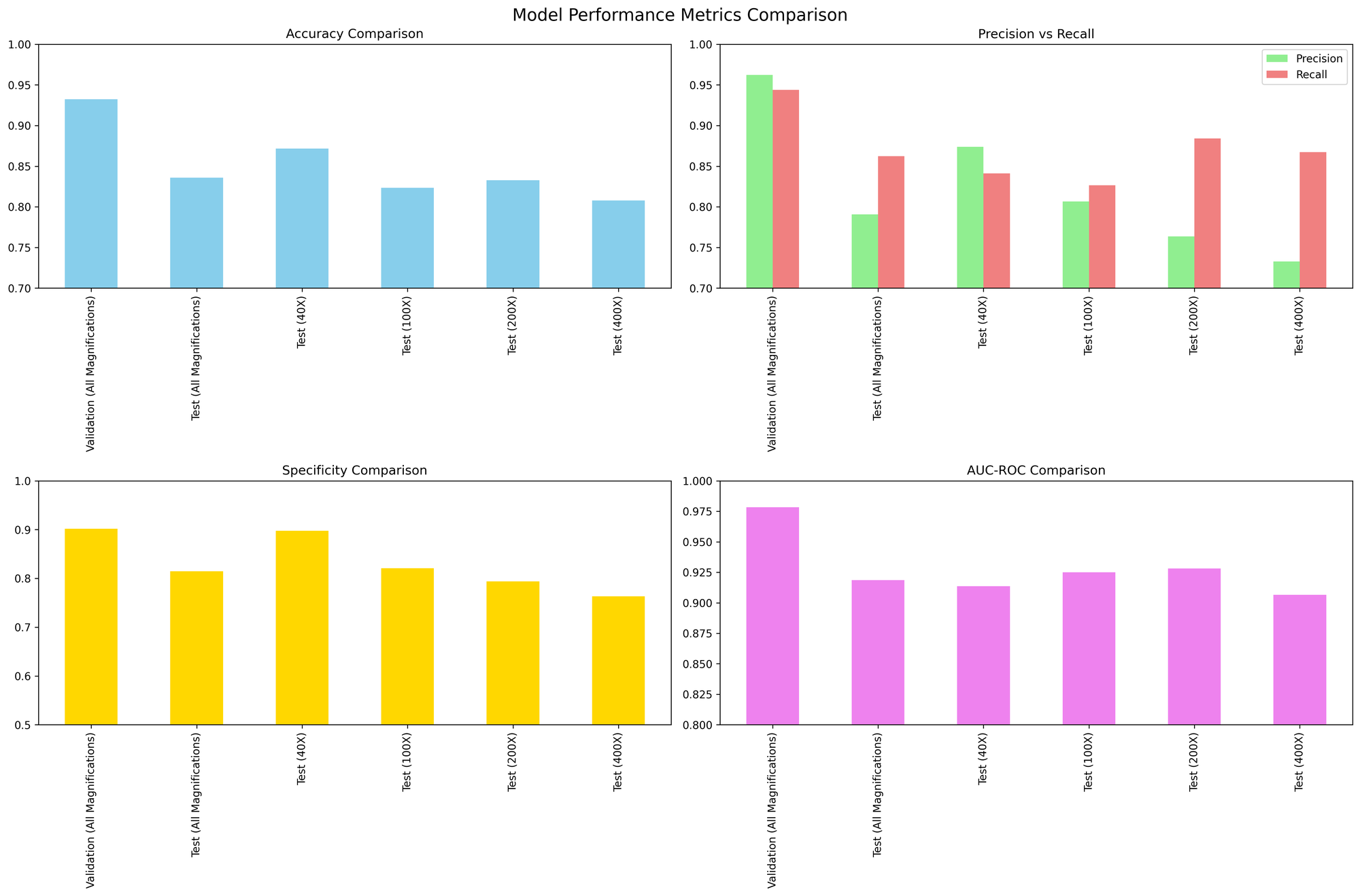}
 \caption{Performance metrics of ResNet across validation and test sets at multiple magnifications.}
 \label{fig:single5}
\end{figure}

\begin{figure}[H]
\centering

\subfloat[40X]{%
 \includegraphics[width=0.23\textwidth]{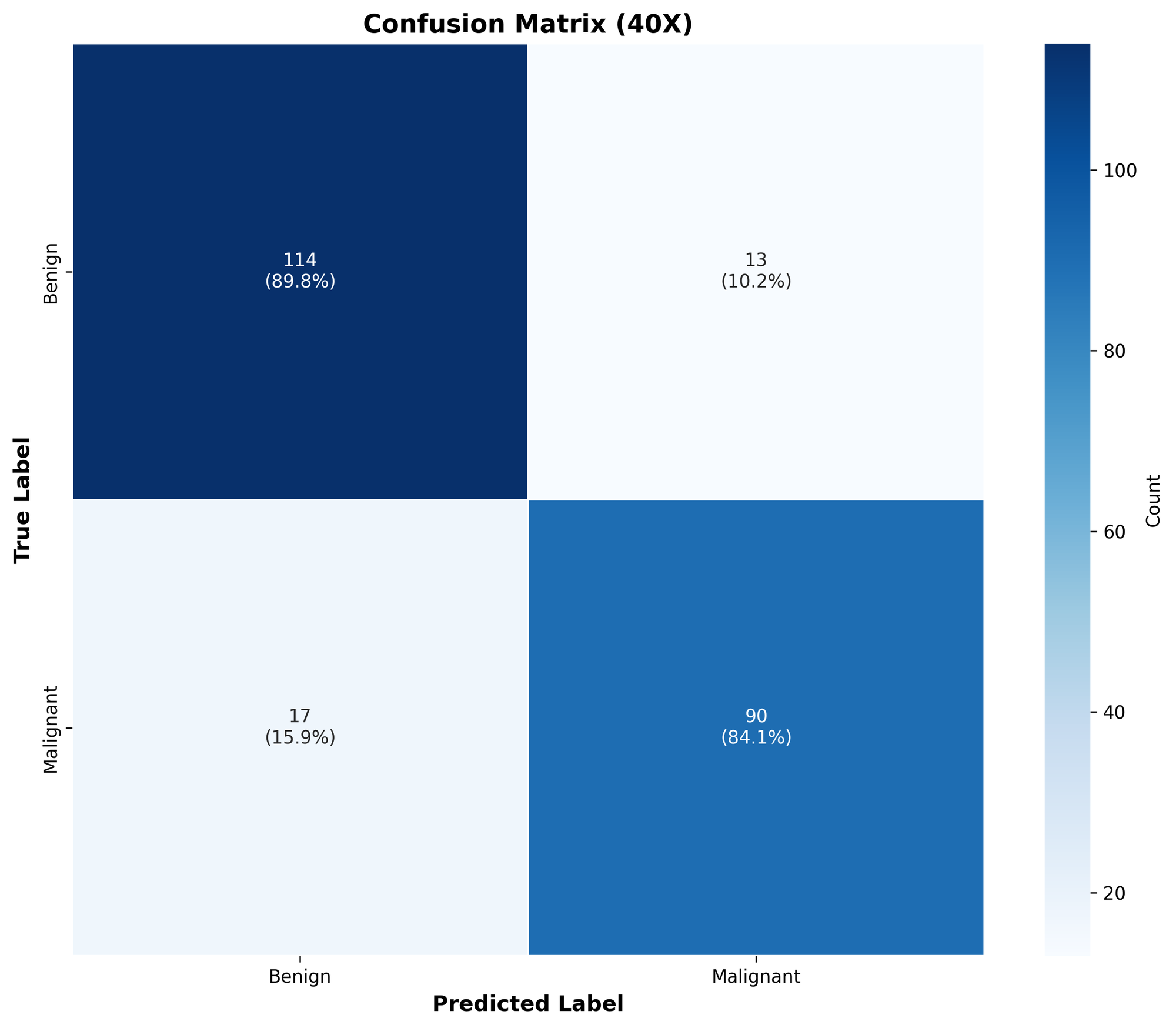}
}\hfill
\subfloat[100X]{%
 \includegraphics[width=0.23\textwidth]{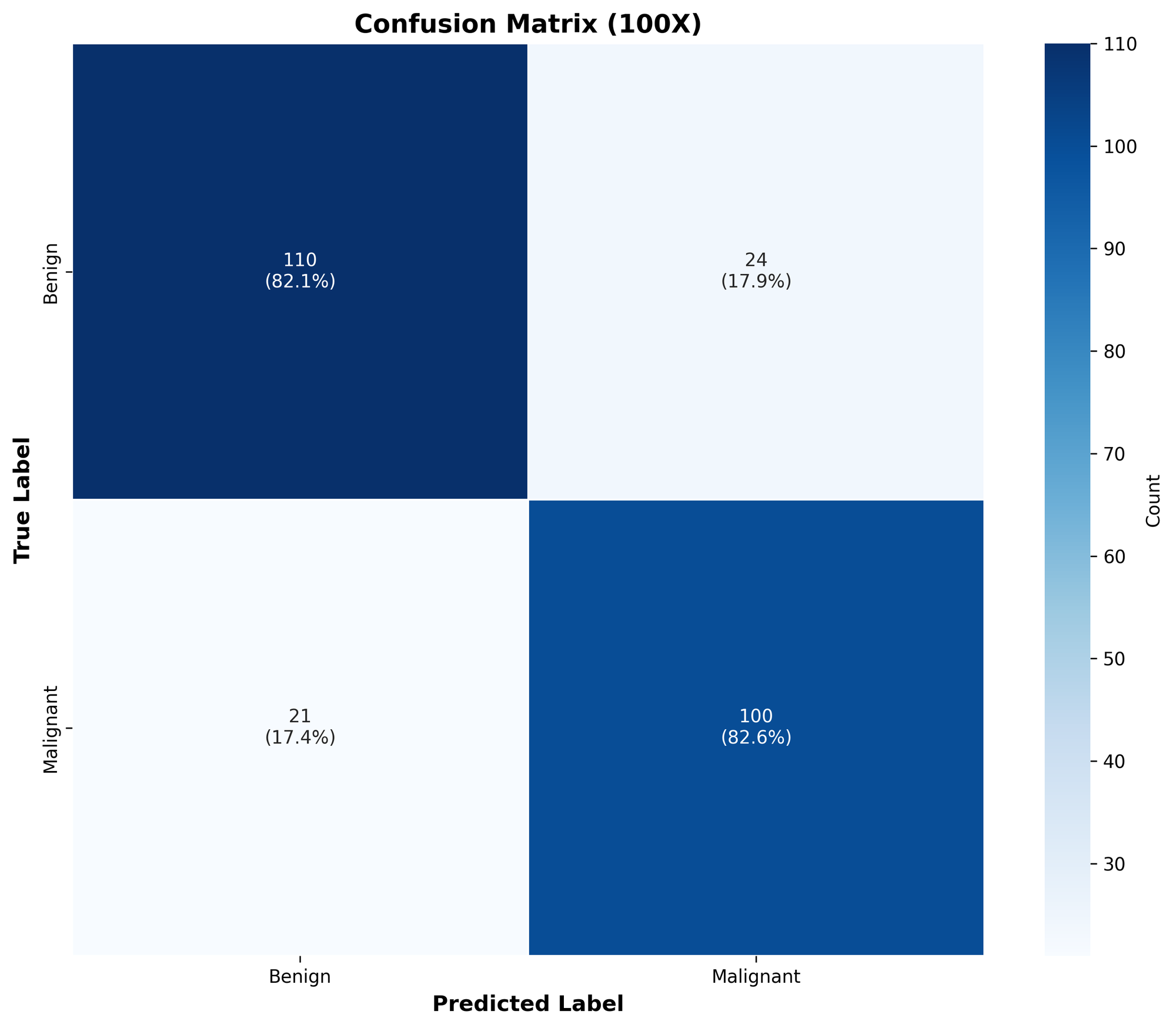}
}\hfill
\subfloat[200X]{%
 \includegraphics[width=0.23\textwidth]{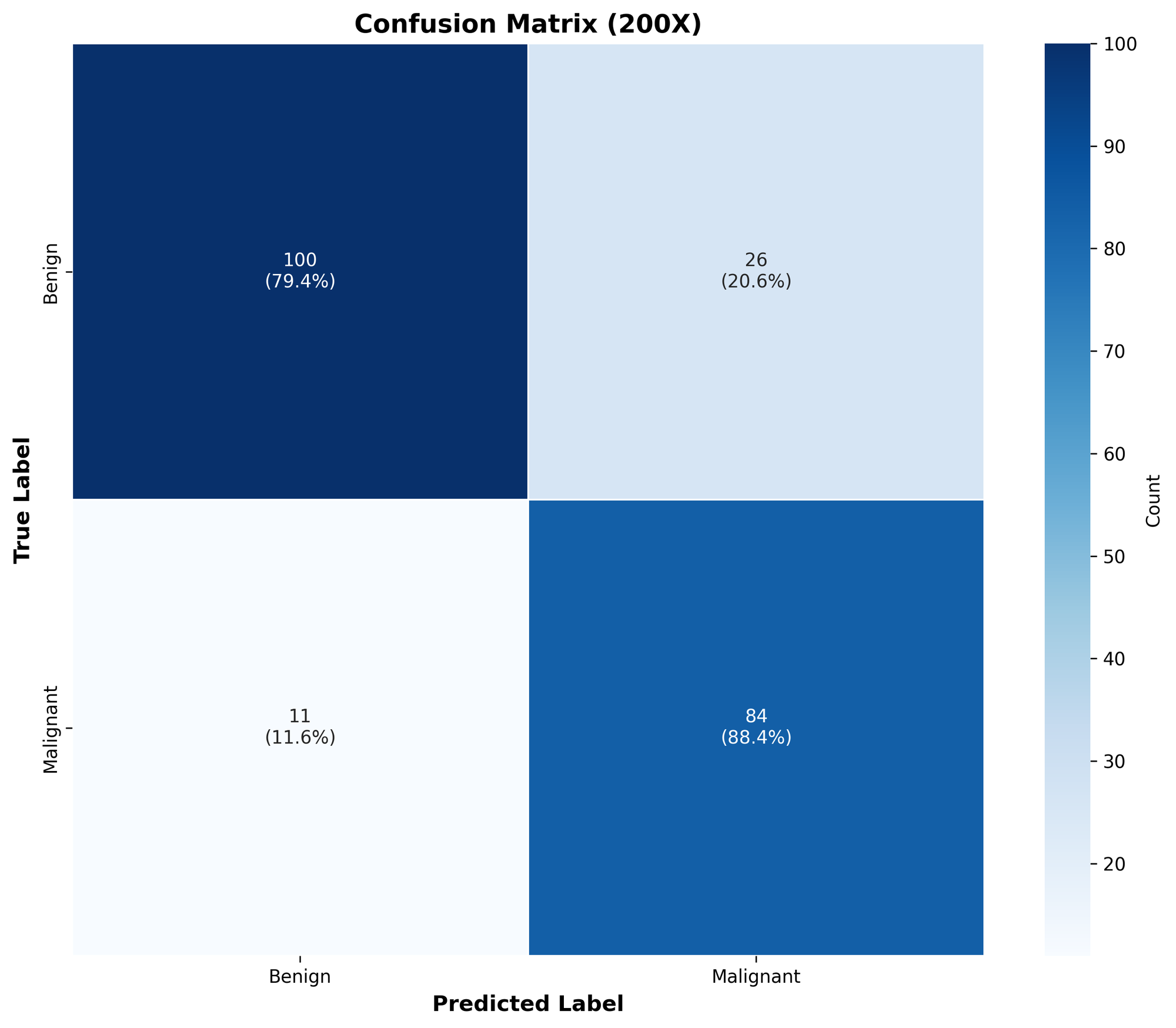}
}\hfill
\subfloat[400X]{%
 \includegraphics[width=0.23\textwidth]{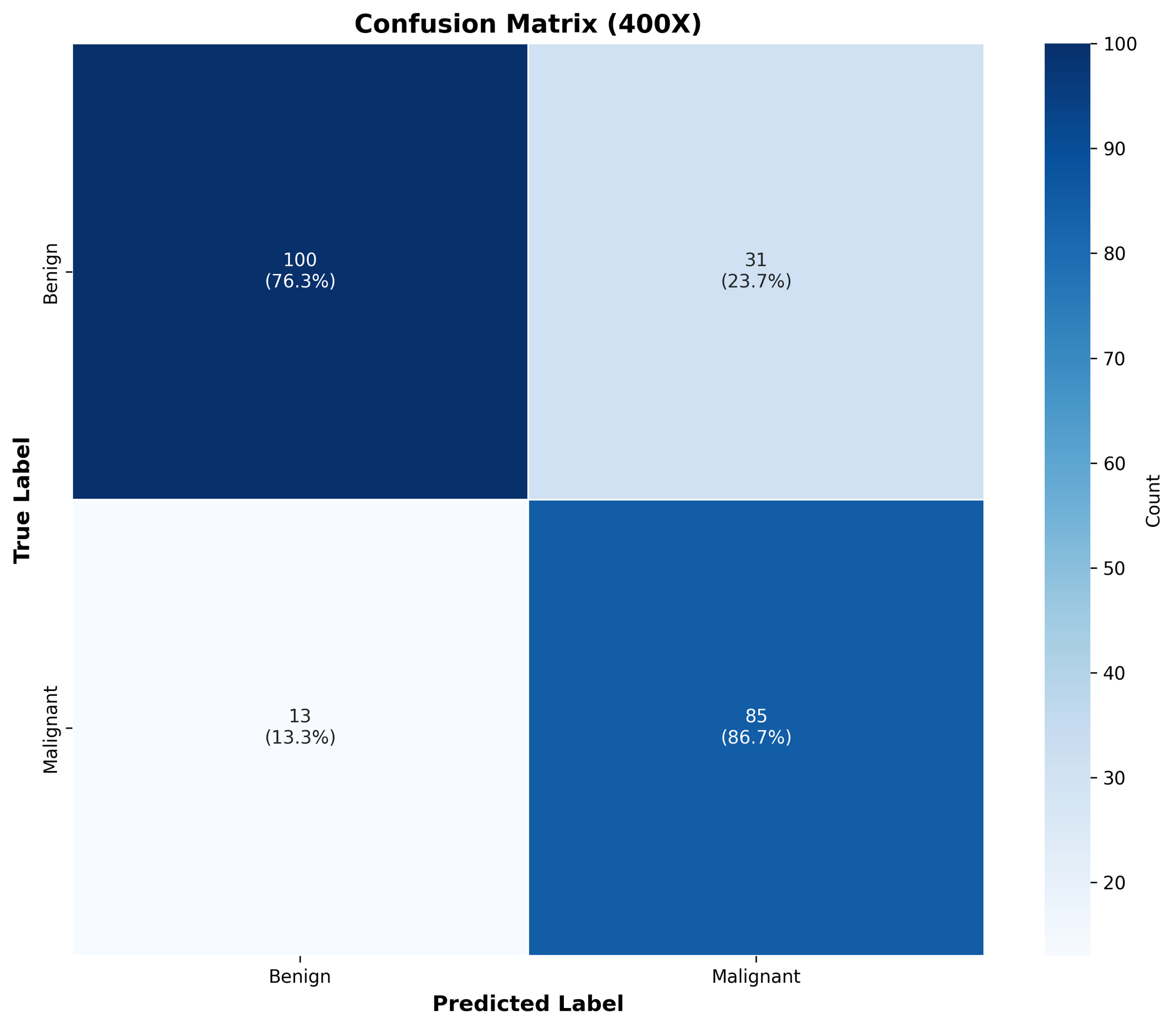}
}

\caption{Confusion matrices of the ResNet model across 40X, 100X, 200X, and 400X magnifications}
\label{fig:resnet_confusion_matrices}
\end{figure}

\begin{figure}[H]
\centering

\subfloat[Validation ROC]{%
 \includegraphics[width=0.23\textwidth]{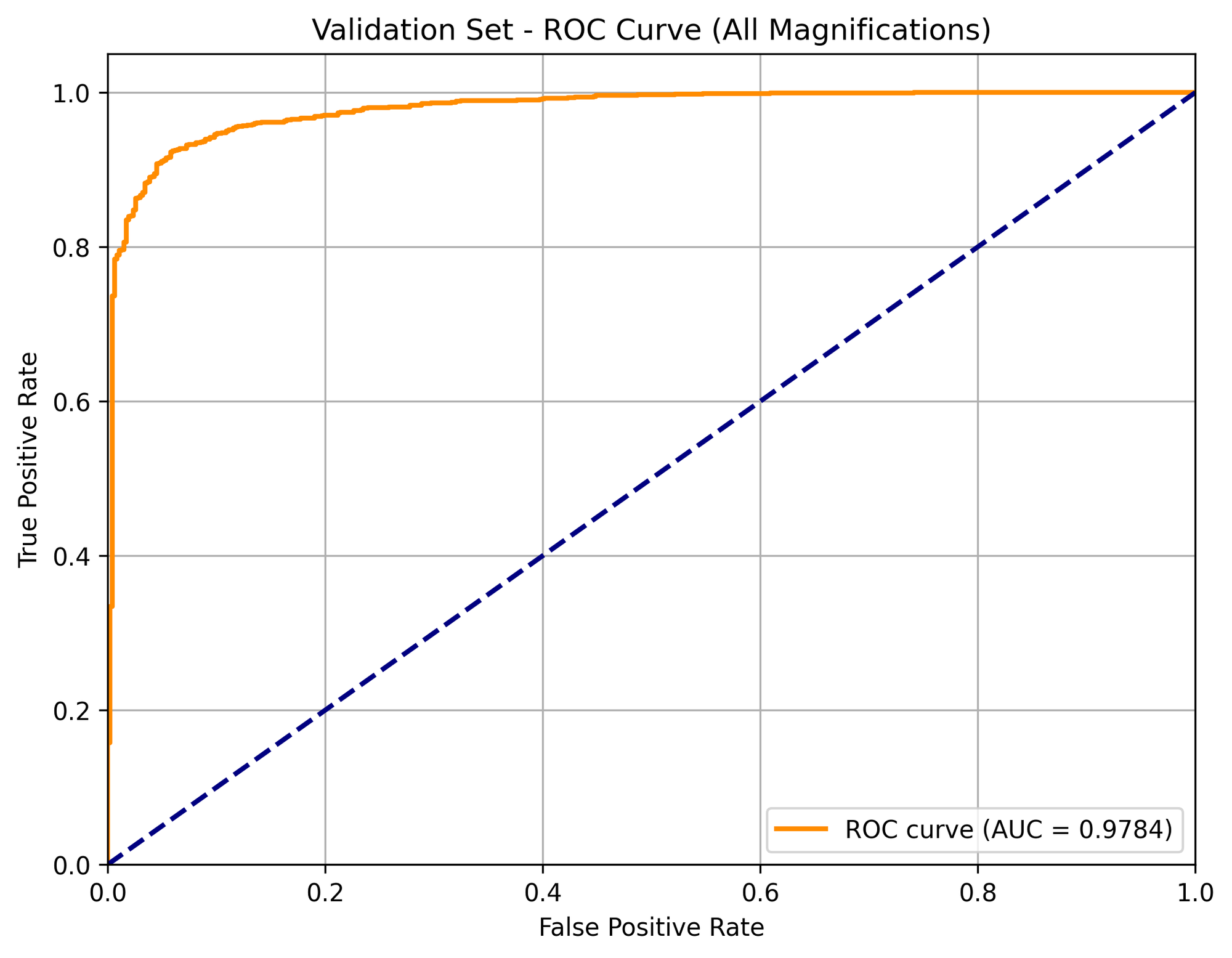}
}\hfill
\subfloat[Validation Confusion Matrix]{%
 \includegraphics[width=0.23\textwidth]{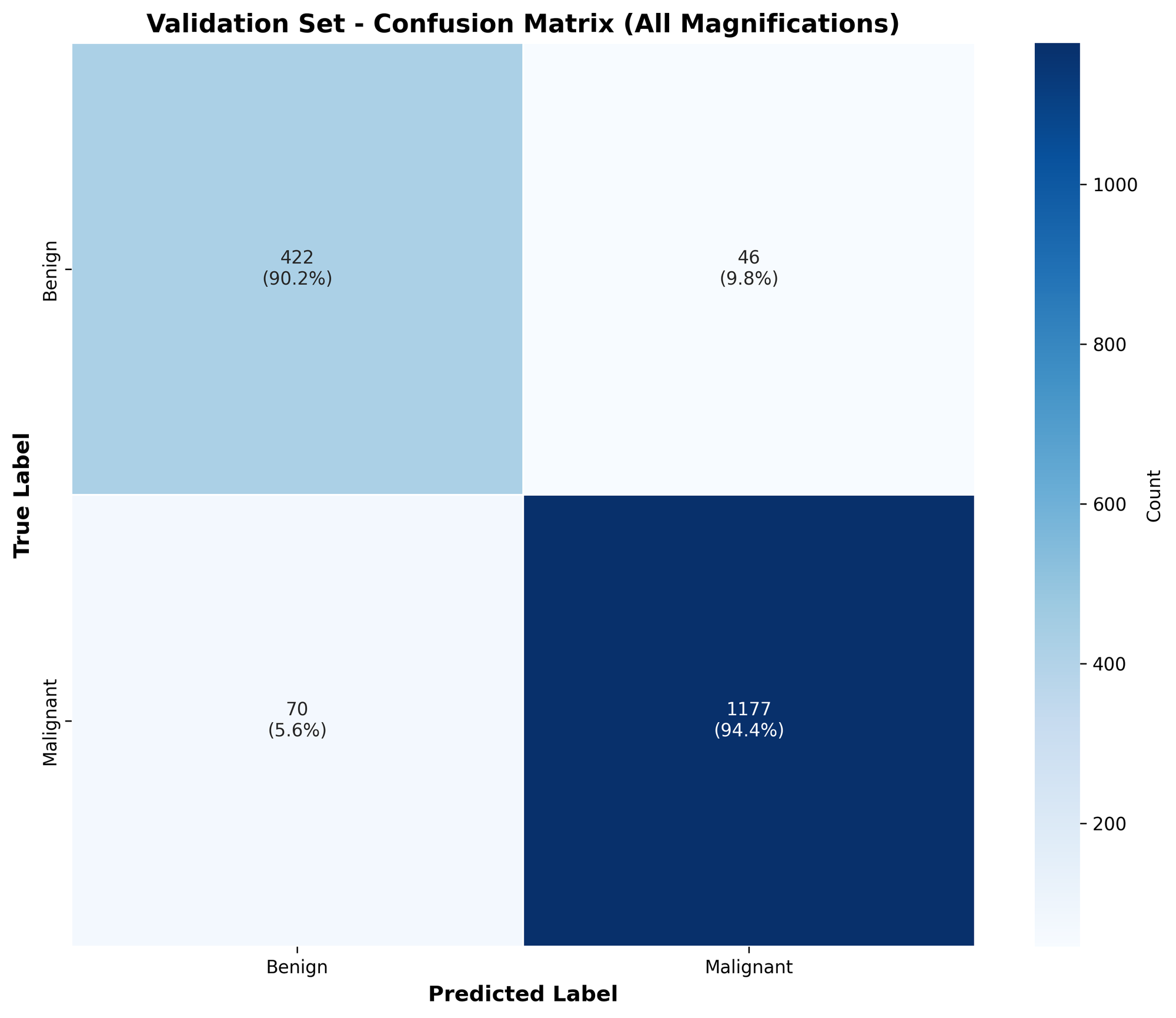}
}\hfill
\subfloat[Test ROC]{%
 \includegraphics[width=0.23\textwidth]{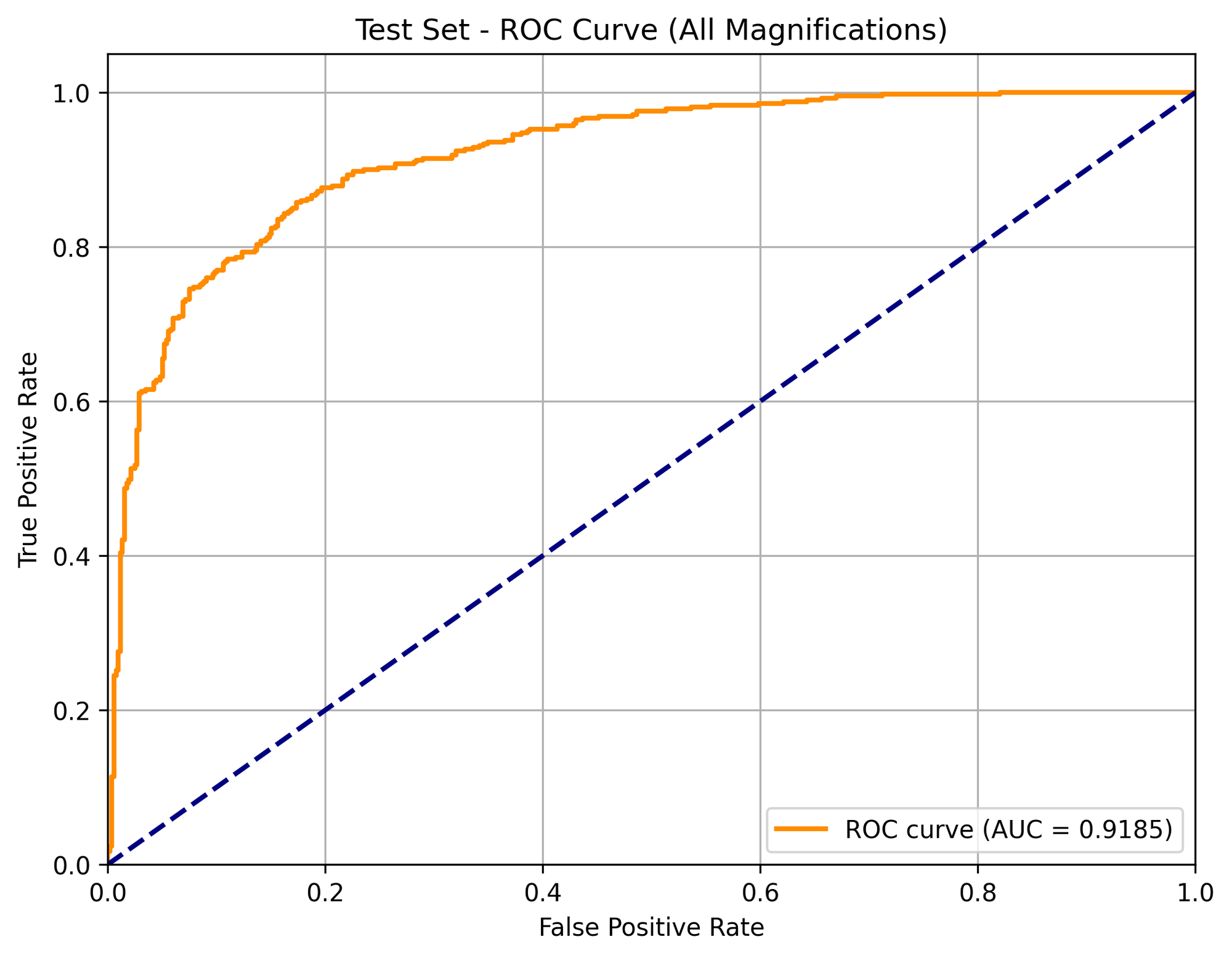}
}\hfill
\subfloat[Test Confusion Matrix]{%
 \includegraphics[width=0.23\textwidth]{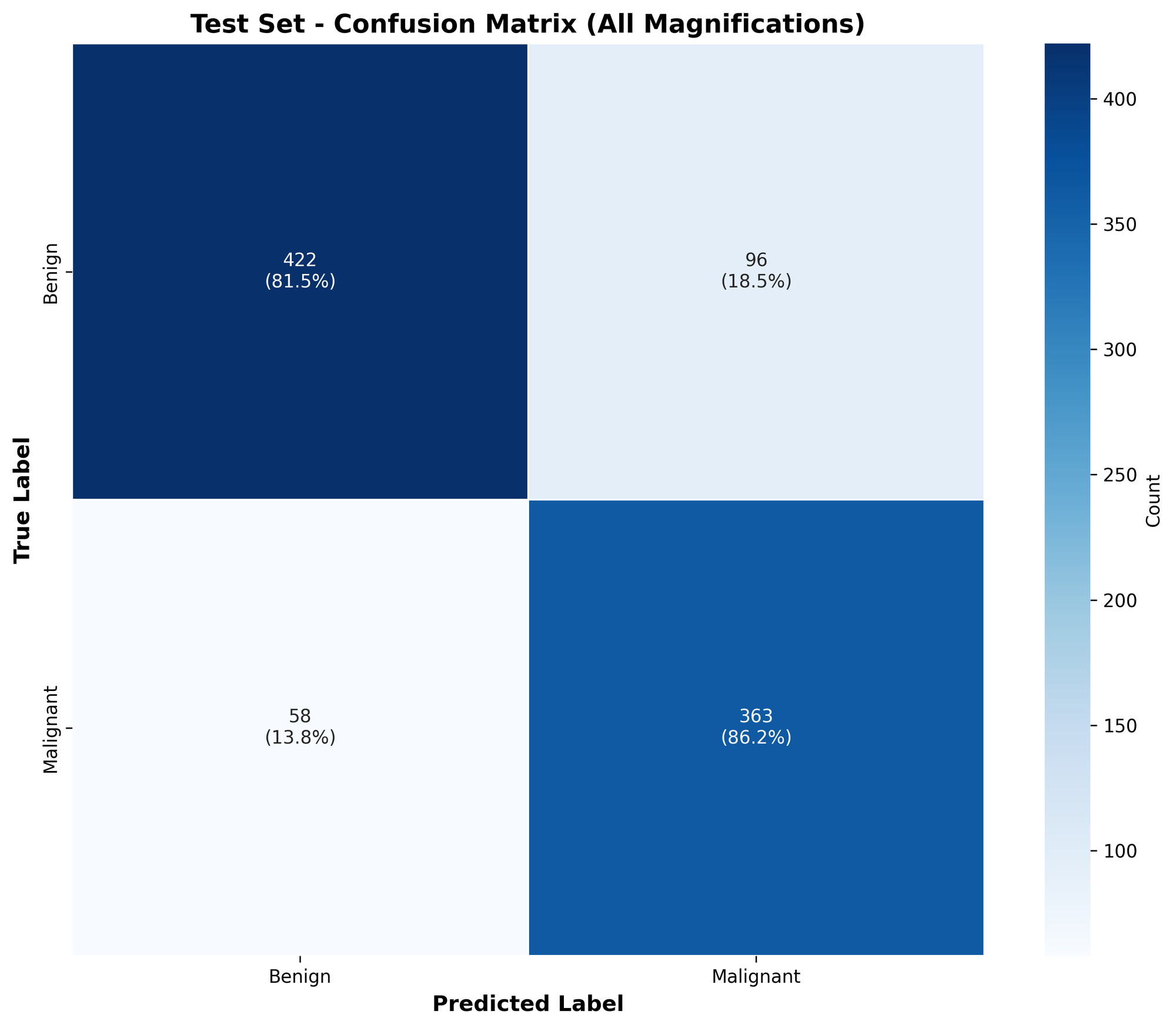}
}

\caption{Validation and test performance of the ResNet model on the BreakHis dataset across all magnifications.}
\label{fig:resnet_val_test}
\end{figure}

\begin{figure}[H]
\centering

\subfloat[ROC 40X]{%
 \includegraphics[width=0.23\textwidth]{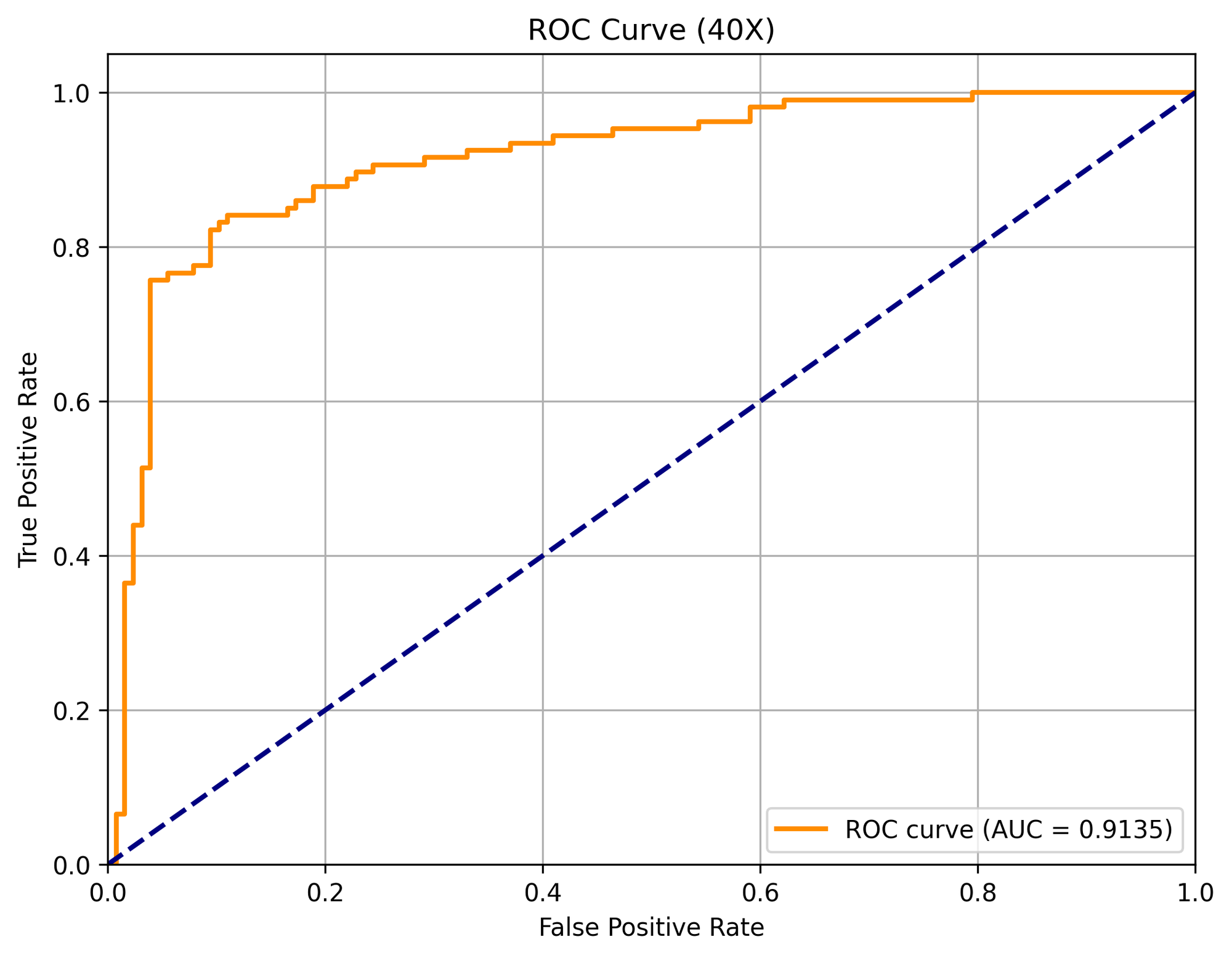}
}\hfill
\subfloat[ROC 100X]{%
 \includegraphics[width=0.23\textwidth]{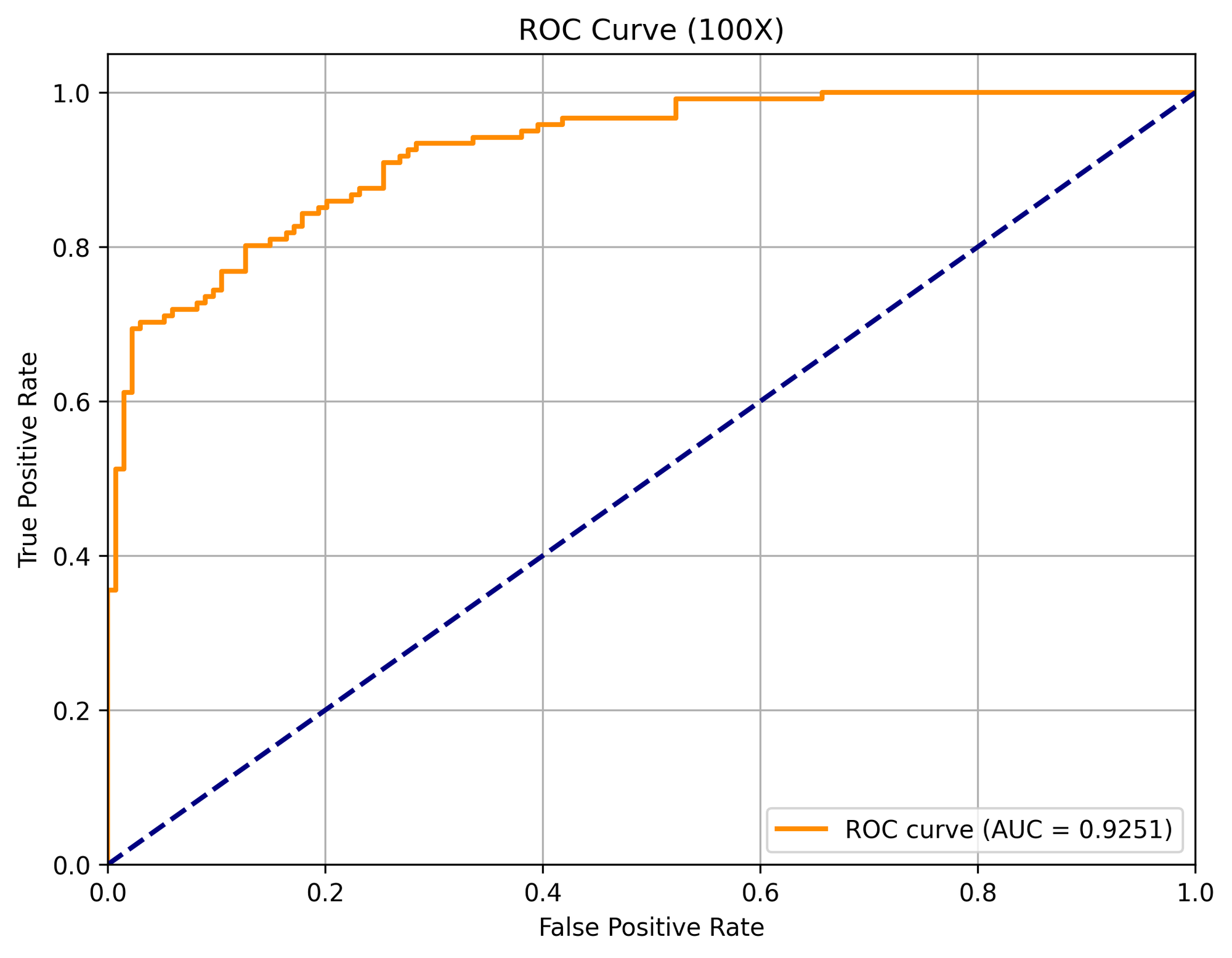}
}\hfill
\subfloat[ROC 200X]{%
 \includegraphics[width=0.23\textwidth]{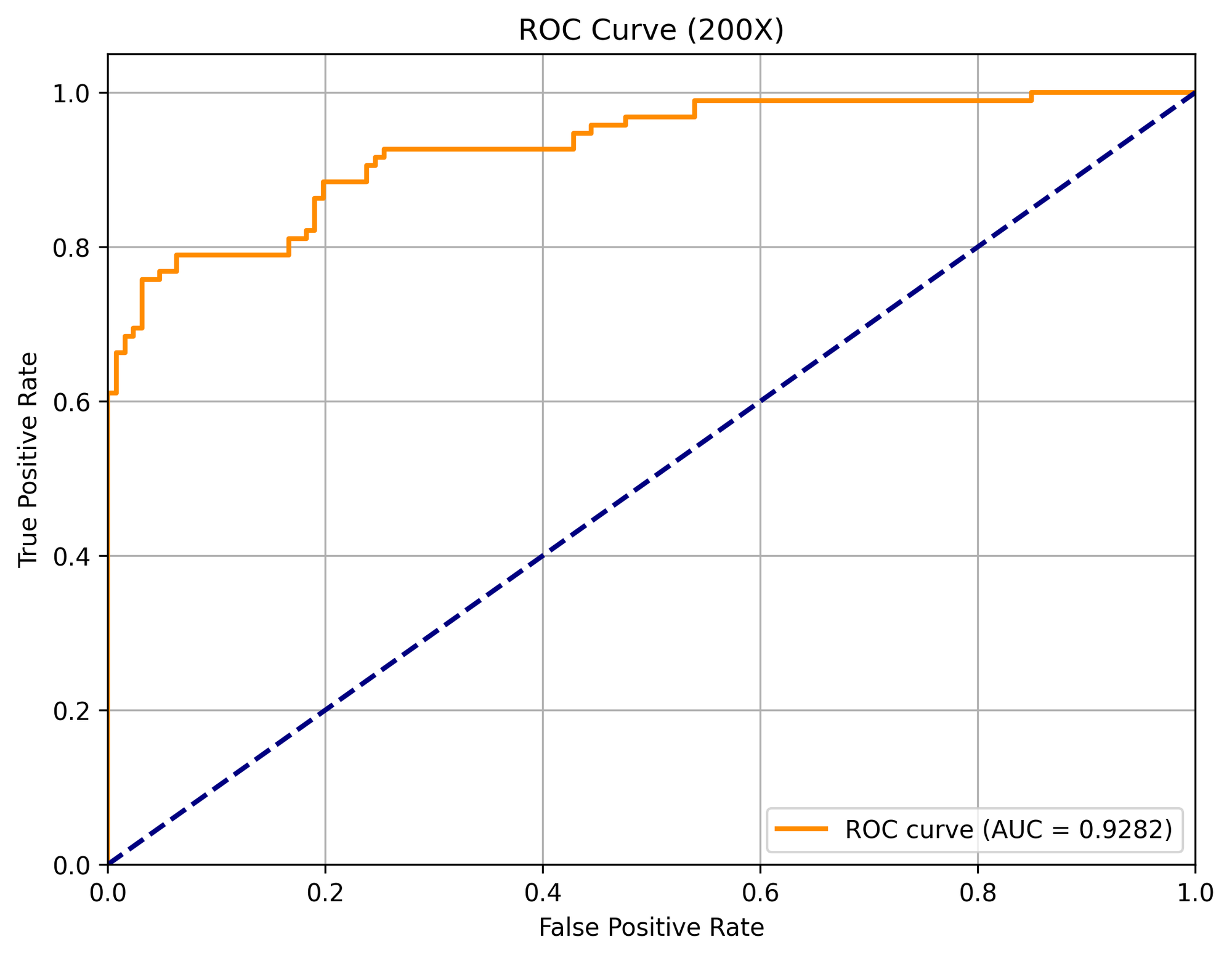}
}\hfill
\subfloat[ROC 400X]{%
 \includegraphics[width=0.23\textwidth]{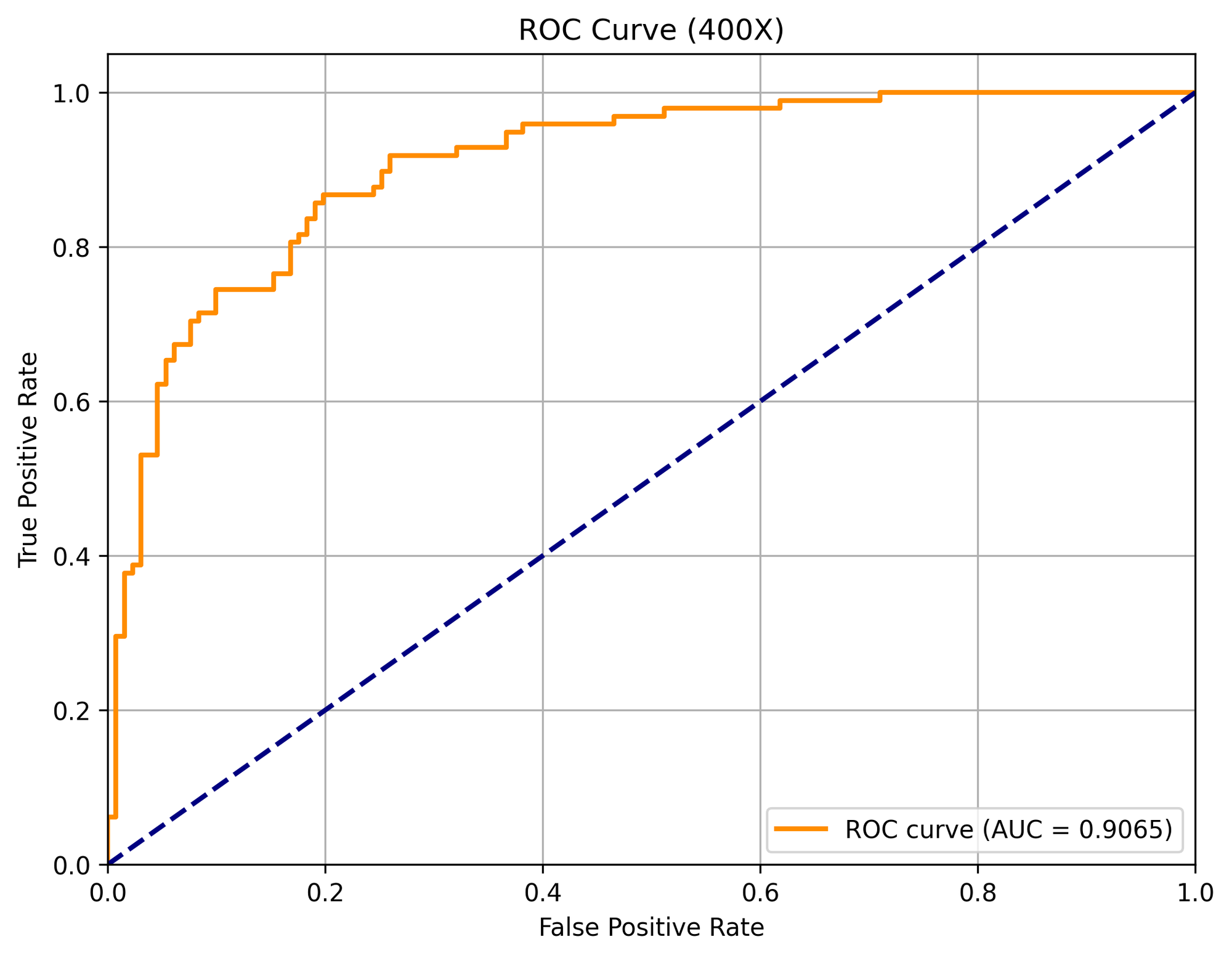}
}

\vspace{0.6ex}

\subfloat[PR 40X]{%
 \includegraphics[width=0.23\textwidth]{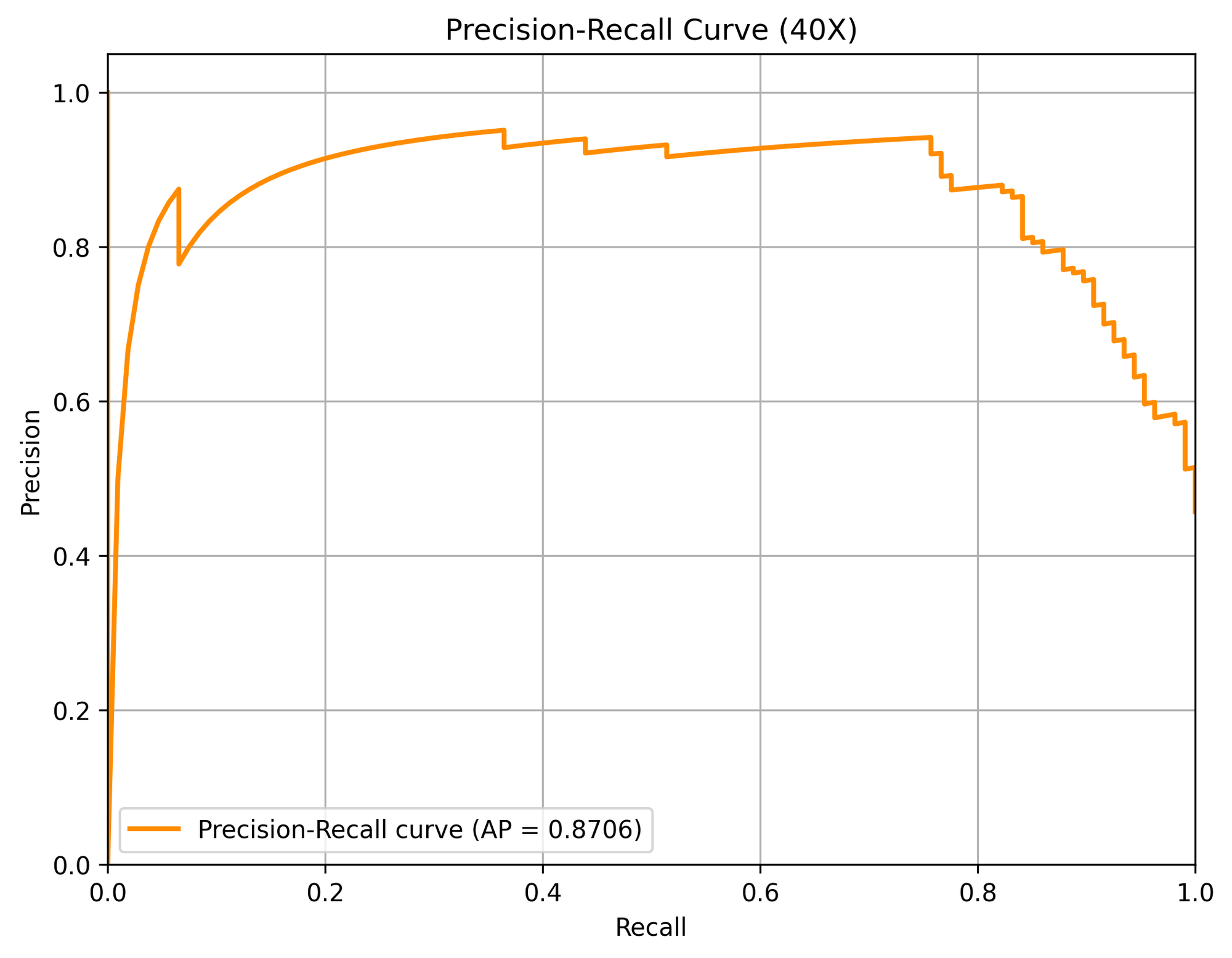}
}\hfill
\subfloat[PR 100X]{%
 \includegraphics[width=0.23\textwidth]{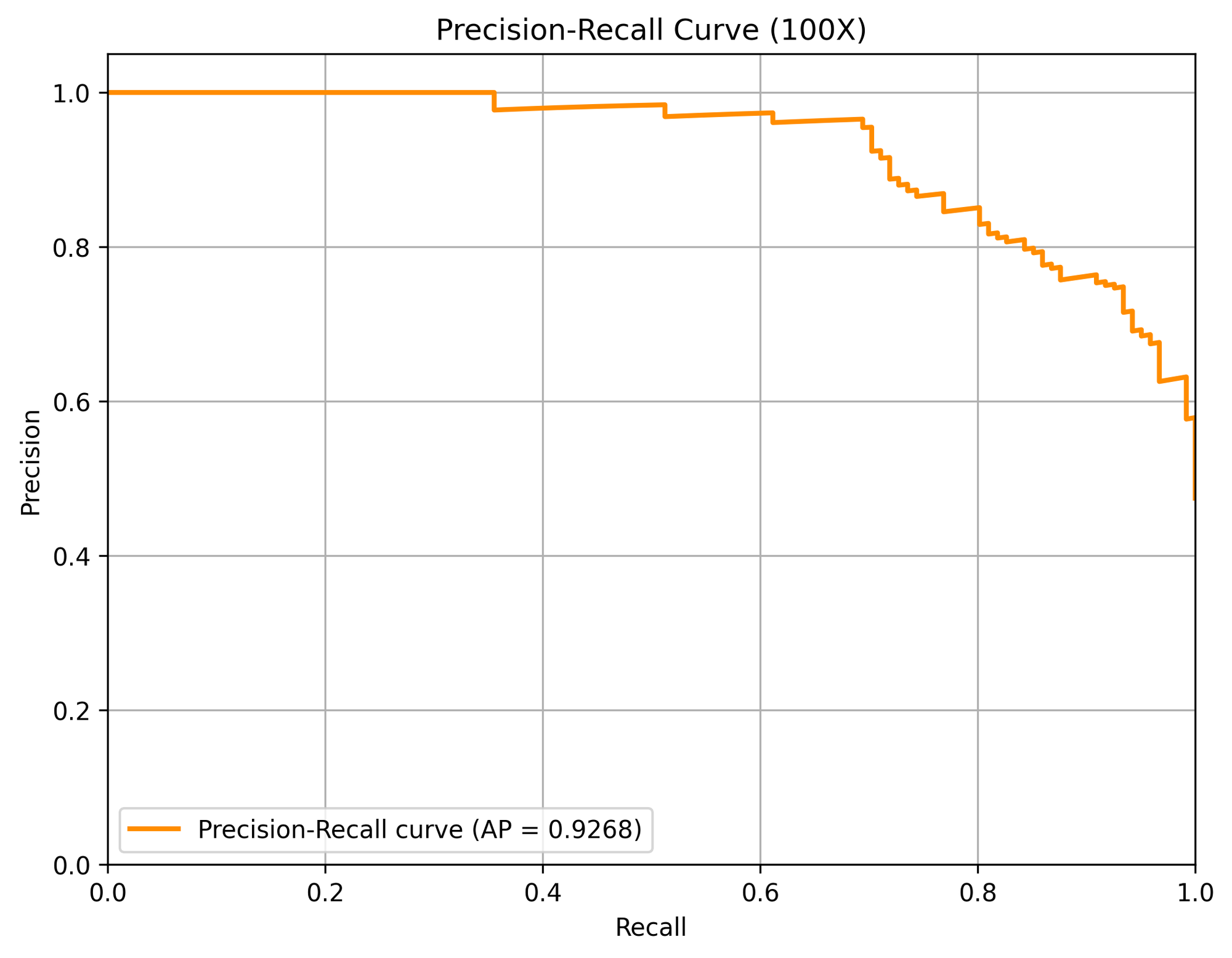}
}\hfill
\subfloat[PR 200X]{%
 \includegraphics[width=0.23\textwidth]{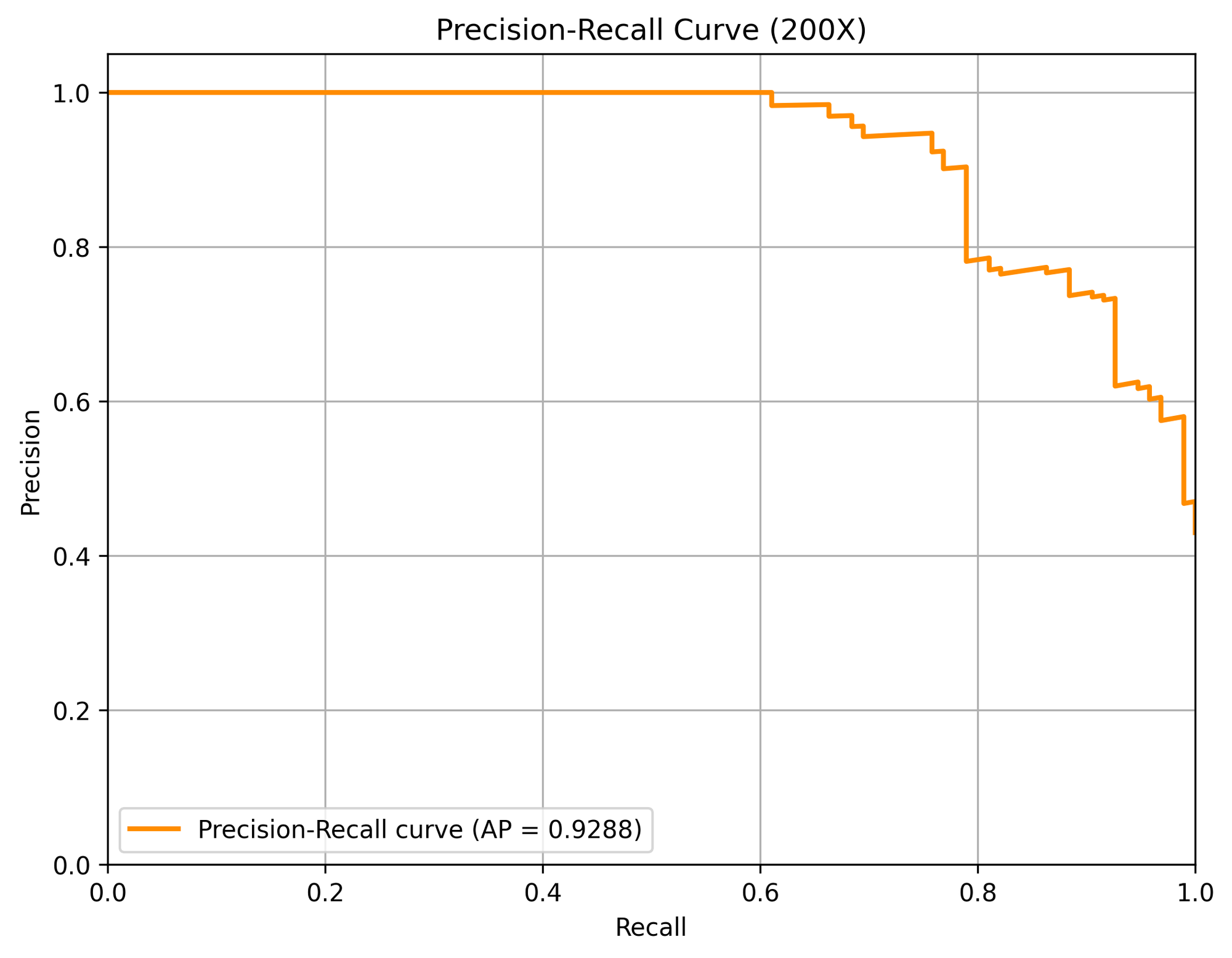}
}\hfill
\subfloat[PR 400X]{%
 \includegraphics[width=0.23\textwidth]{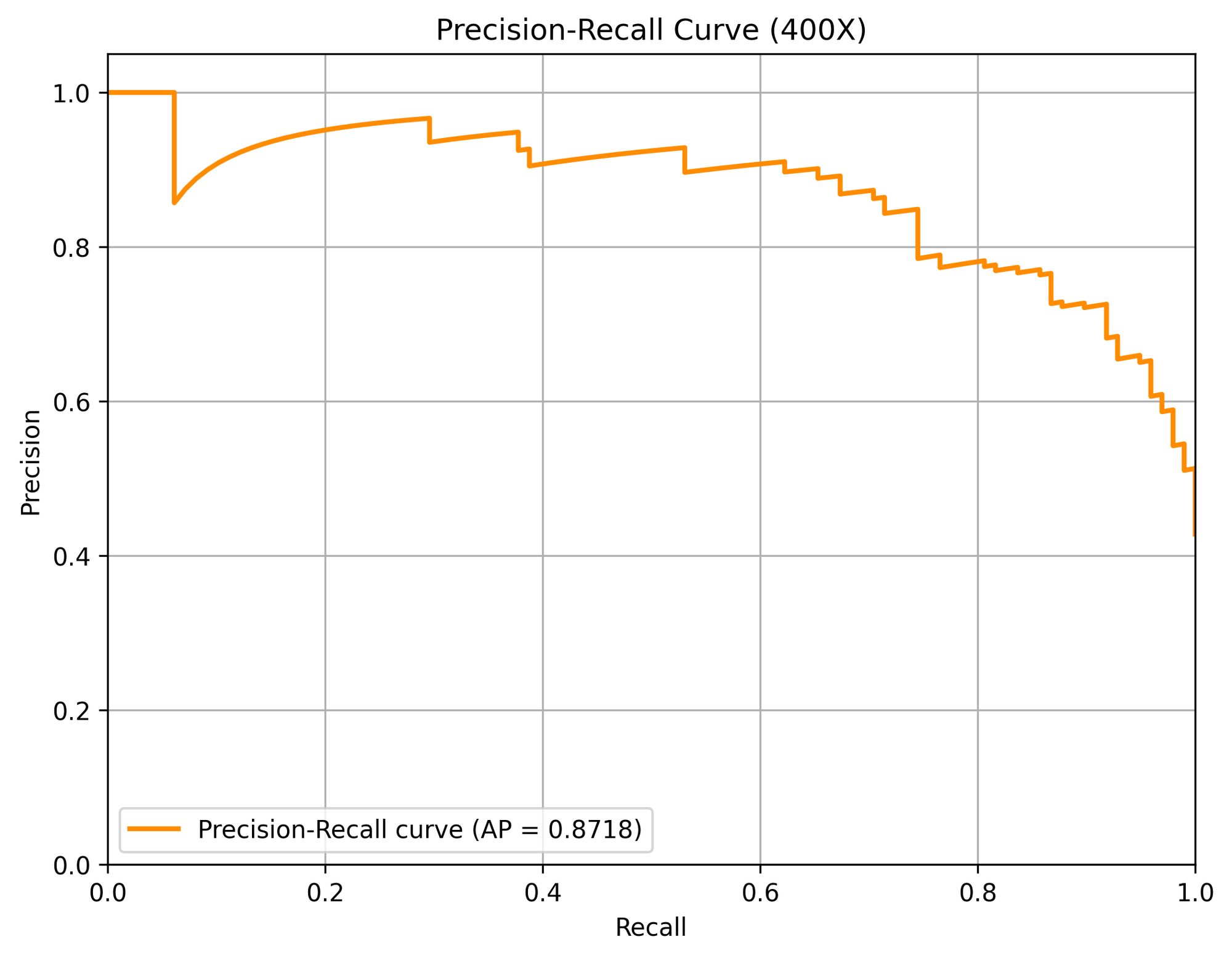}
}

\caption{Receiver Operating Characteristic (ROC) curves (a–d) and Precision–Recall (PR) curves (e–h).}
\label{fig:resnet_roc_pr}
\end{figure}

\begin{table}[ht]
\caption{Performance of Swin Transformer Tiny on the BreakHis dataset}
\label{tab:swintiny}
\centering
\begin{tabular}{@{}lcccccc@{}}
\toprule
\textbf{Dataset} 
& \textbf{Accuracy (\%)} 
& \textbf{Precision (\%)} 
& \textbf{Recall (\%)} 
& \textbf{F1-score (\%)} 
& \textbf{Specificity (\%)} 
& \textbf{AUC} \\
\midrule
Test (All Mag.) 
& 79.5 
& 73.6 
& 84.6 
& 78.7 
& 75.3 
& 0.860 \\

Test (40$\times$)      
& 85.0 
& 81.6 
& 86.9 
& 84.2 
& 83.5 
& 0.929 \\

Test (100$\times$)      
& 79.2 
& 74.6 
& 85.1 
& 79.5 
& 73.9 
& 0.874 \\

Test (200$\times$)      
& 77.8 
& 69.8 
& 85.3 
& 76.8 
& 72.2 
& 0.834 \\

Test (400$\times$)      
& 77.7 
& 70.1 
& 83.7 
& 76.3 
& 73.3 
& 0.821 \\
\bottomrule
\end{tabular}
\end{table}

The Swin Transformer Tiny attained an accuracy of 79.45\%, precision of 73.55\%, recall of 84.56\%, specificity of 75.29\%, F1-score of 78.67\%, and an AUC-ROC of 0.8595 across all magnifications on BreakHis, indicating robust malignancy identification but moderate precision and specificity (Table 7). Performance peaked at 40X magnification, achieving 85.04\% accuracy, 81.58\% precision, 86.92\% recall, 83.46\% specificity, an F1 score of 84.16\%, and an AUC of 0.9285. In contrast, findings at 100X yielded 79.22\% accuracy and an AUC of 87.36, while 200X produced 77.83\% accuracy and an AUC of 83.38, both demonstrating commendable recall (>85\%) but diminished precision (<75\%) (Figure 25, 26 \& 27). At 400X magnification, accuracy decreased to 77.73\%, with precision at 70.09\%, recall at 83.67\%, specificity at 73.28\%, and an AUC-ROC of 0.8212, suggesting challenges in discerning fine-grained patterns at elevated magnification levels. The Swin Transformer exhibited strong recall and stable F1-scores across many scales; however, its significantly lower accuracy and specificity indicate a propensity for false positives, despite its efficacy in detecting cancerous regions.

\begin{figure}[H]
 \centering
 \includegraphics[width=0.85\linewidth]{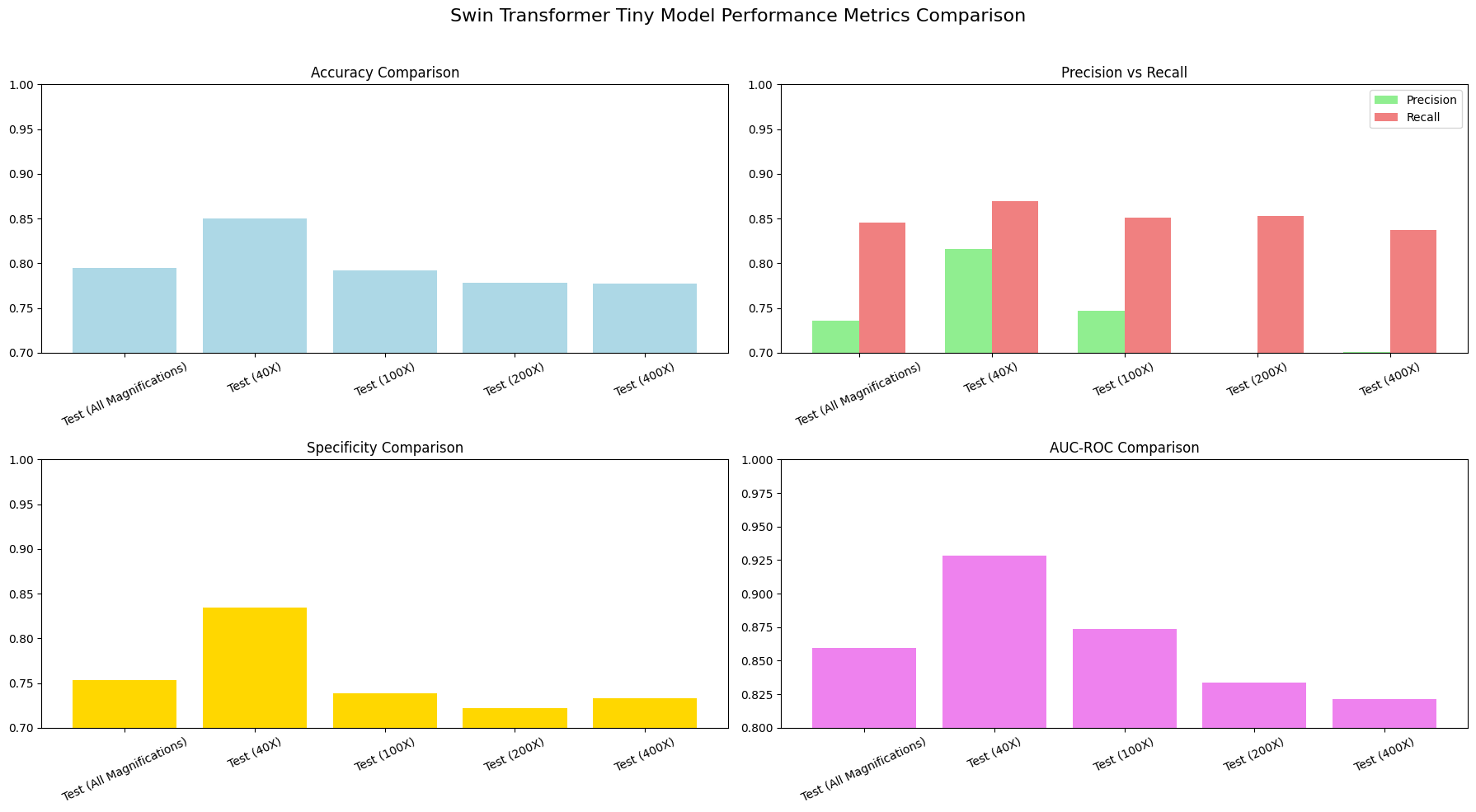}
 \caption{Performance metrics of the Swin Transformer Tiny across different magnifications}
 \label{fig:single6}
\end{figure}

\begin{figure}[H]
 \centering
 \includegraphics[width=1\linewidth]{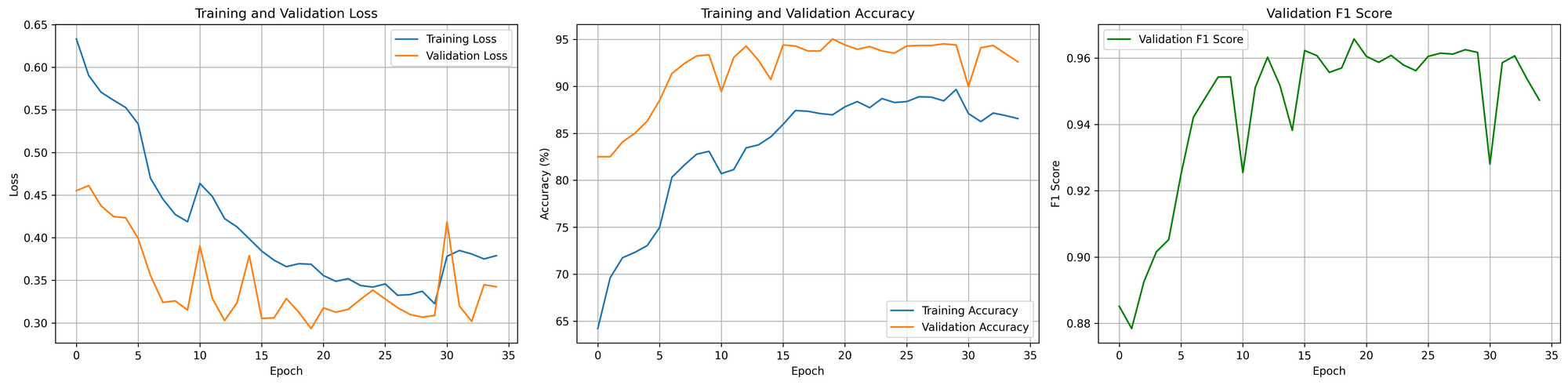}
 \caption{Training curves of Swin Transformer Tiny}
 \label{fig:single7}
\end{figure}

\begin{figure}[H]
\centering

\subfloat[(a) 40X]{%
 \includegraphics[width=0.23\textwidth]{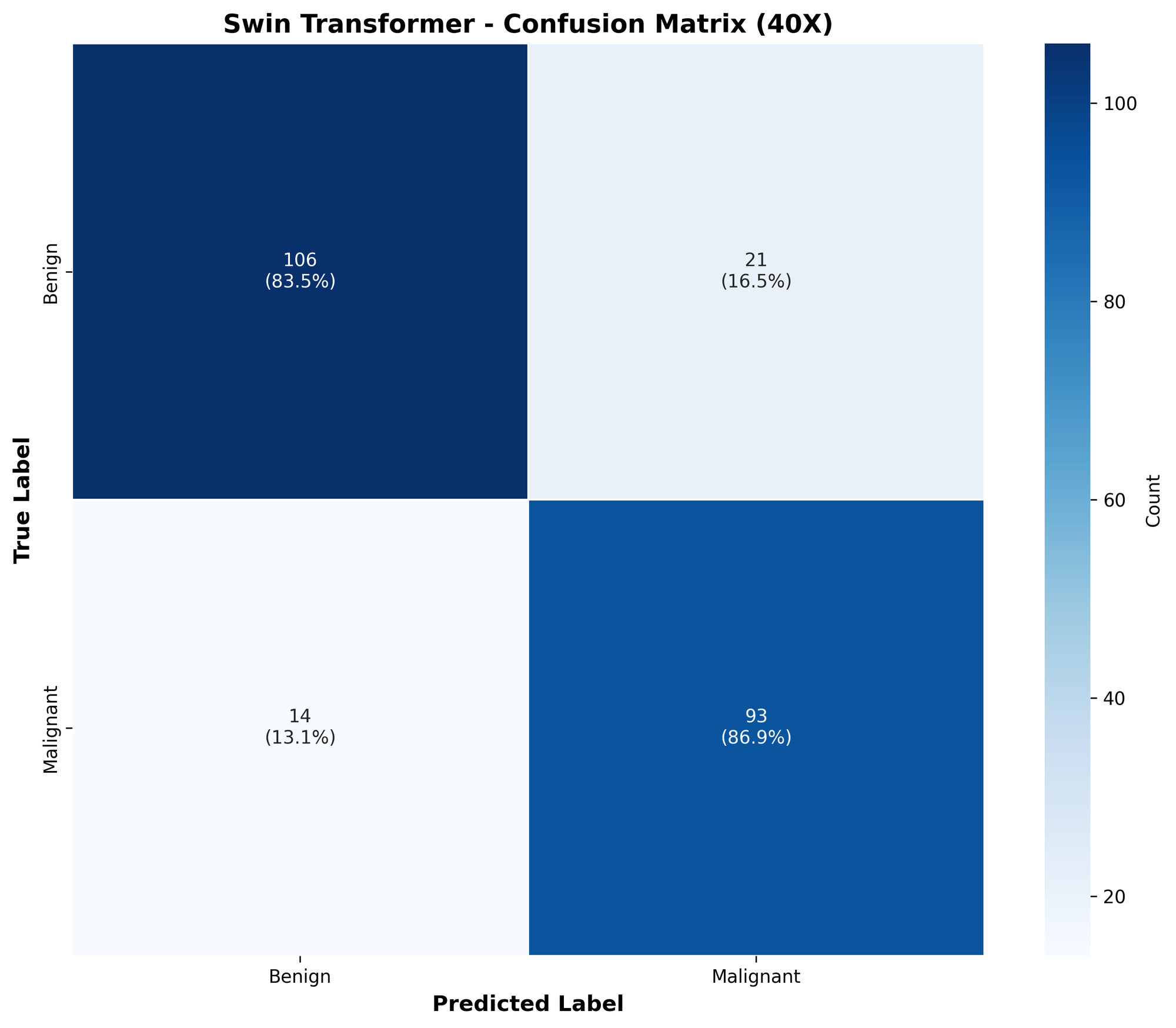}
}\hfill
\subfloat[(b) 100X]{%
 \includegraphics[width=0.23\textwidth]{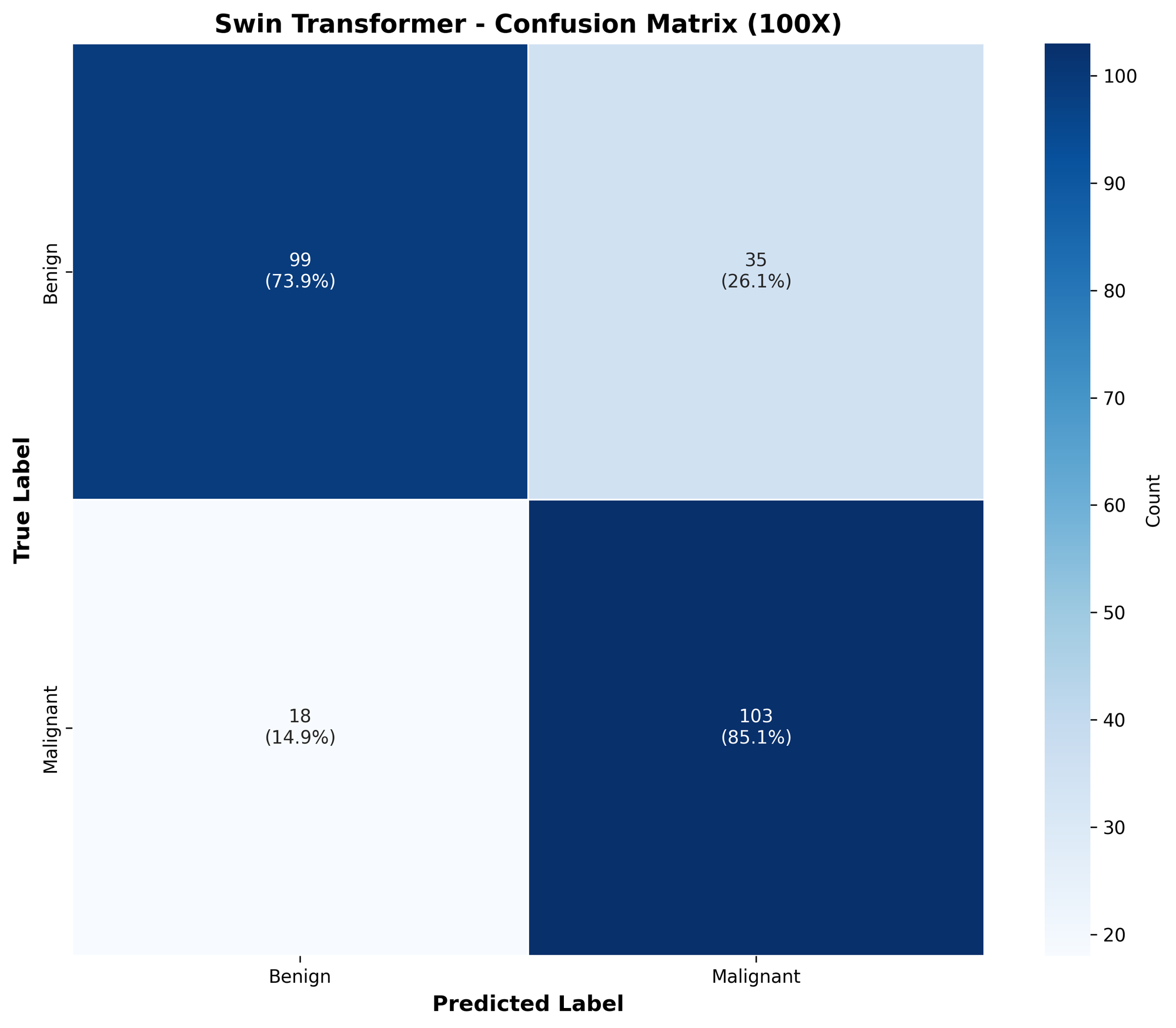}
}\hfill
\subfloat[(c) 200X]{%
 \includegraphics[width=0.23\textwidth]{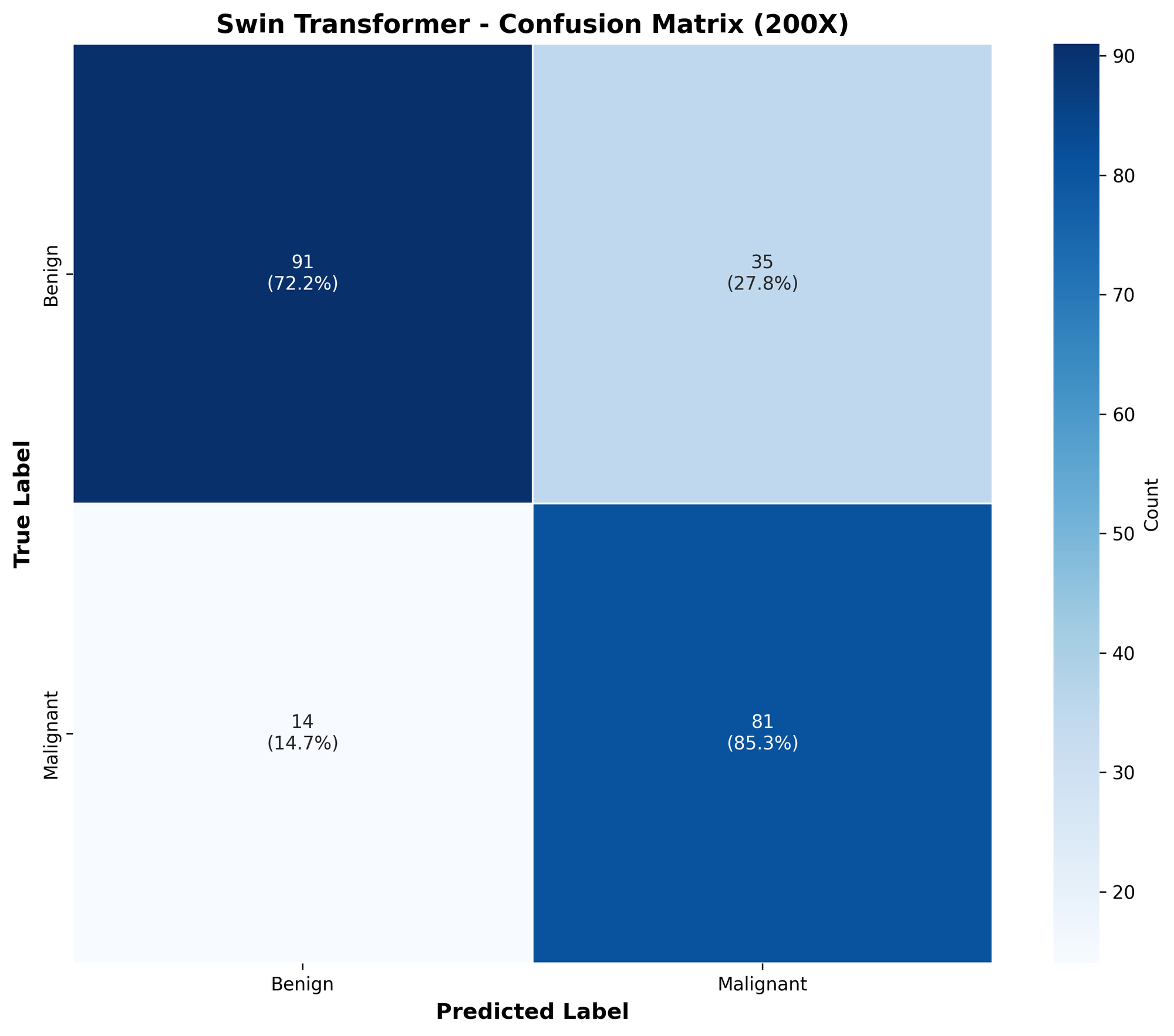}
}\hfill
\subfloat[(d) 400X]{%
 \includegraphics[width=0.23\textwidth]{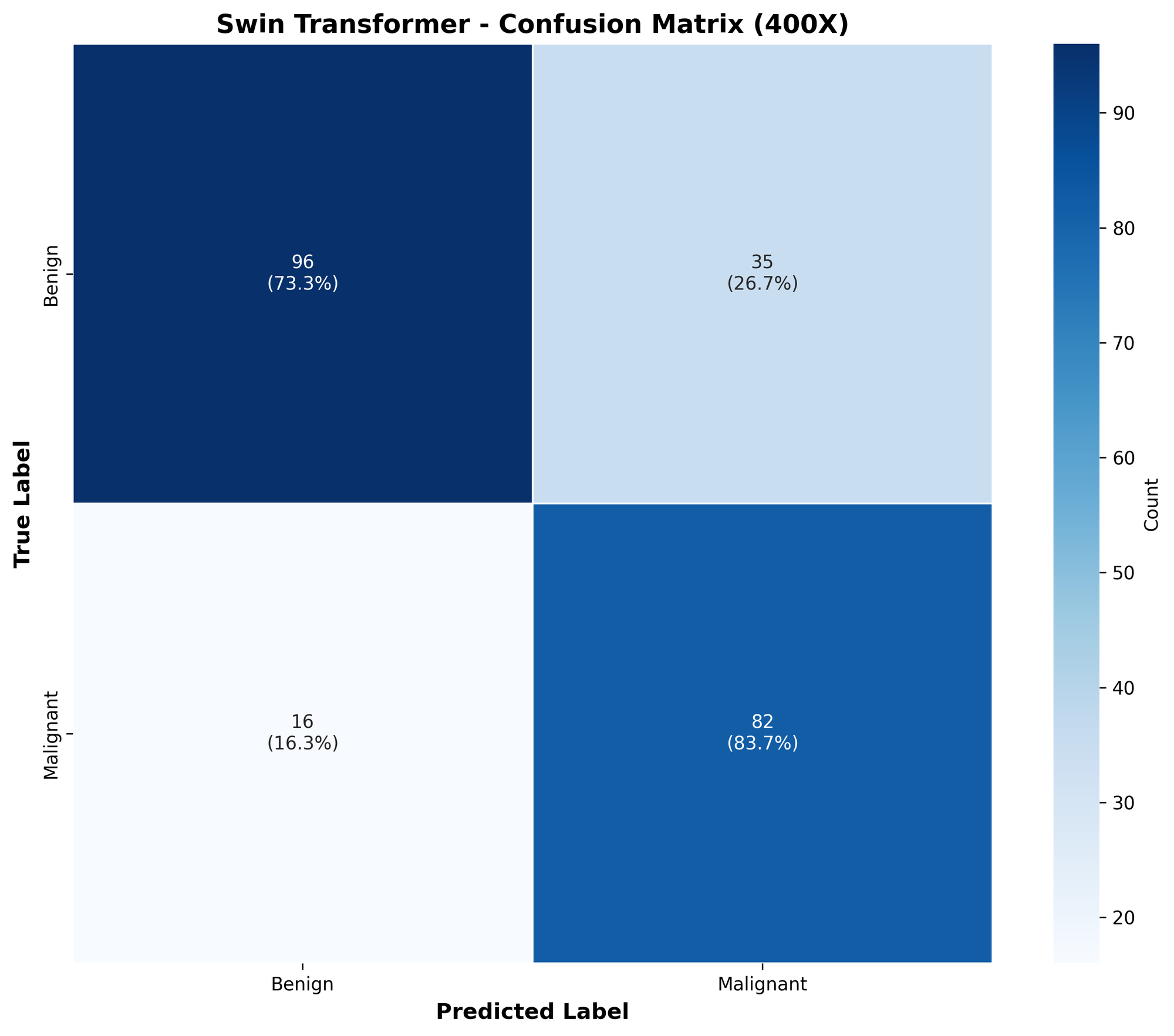}
}

\caption{Confusion matrices of the Swin Transformer Tiny model across 40X, 100X, 200X, and 400X magnifications}
\label{fig:swin_confusion_matrices}
\end{figure}

\begin{figure}[H]
 \centering
 \includegraphics[width=0.75\linewidth]{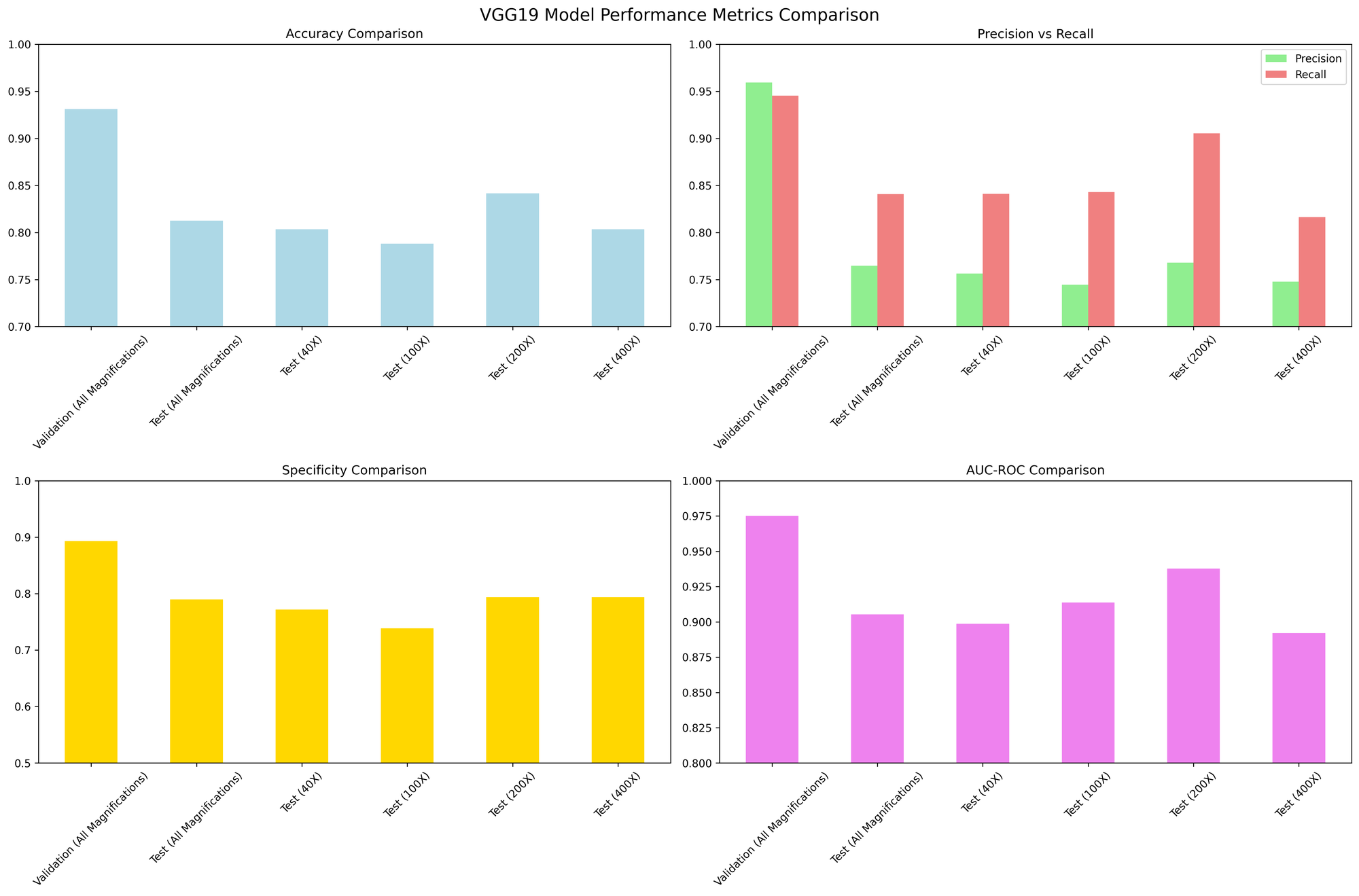}
 \caption{Performance metrics of the VGG19 model across validation and test sets at multiple magnifications.}
 \label{fig:single8}
\end{figure}

\begin{table}[ht]
\caption{Performance of VGG19 on the BreakHis dataset}
\label{tab:vgg19}
\centering
\begin{tabular}{@{}lcccccc@{}}
\toprule
\textbf{Dataset} 
& \textbf{Accuracy (\%)} 
& \textbf{Precision (\%)} 
& \textbf{Recall (\%)} 
& \textbf{F1-score (\%)} 
& \textbf{Specificity (\%)} 
& \textbf{AUC} \\
\midrule
Validation (All Mag.) 
& 93.1 
& 95.9 
& 94.6 
& 95.2 
& 89.3 
& 0.975 \\

Test (All Mag.)    
& 81.3 
& 76.5 
& 84.1 
& 80.1 
& 79.0 
& 0.905 \\

Test (40$\times$)      
& 80.3 
& 75.6 
& 84.1 
& 79.6 
& 77.2 
& 0.899 \\

Test (100$\times$)      
& 78.8 
& 74.5 
& 84.3 
& 79.1 
& 73.9 
& 0.914 \\

Test (200$\times$)      
& 84.2 
& 76.8 
& 90.5 
& 83.1 
& 79.4 
& 0.938 \\

Test (400$\times$)      
& 80.4 
& 74.8 
& 81.6 
& 78.1 
& 79.4 
& 0.892 \\
\bottomrule
\end{tabular}
\end{table}

VGG19 had outstanding results on the validation set, achieving 93.12\% accuracy, 95.93\% precision, 94.55\% recall, 89.32\% specificity, 95.23\% F1-score, and 0.975 AUC-ROC, so proving its strong generalization throughout training (Table 8). Test performance declined to 81.26\% accuracy, 76.46\% precision, 84.09\% recall, 78.96\% specificity, 80.09\% F1-score, and 0.9053 AUC-ROC, signifying diminished precision and an increase in false positives (Figure 28, 29 \& 30). At certain magnifications, the results were consistent: 40X (80.34\% accuracy, AUC 0.8987), 100X (78.82\%, AUC 0.9138), 200X (84.16\%, recall 90.53\%, AUC 0.9378), and 400X (80.35\%, AUC 0.892). The model exhibited strong recall at various magnifications ($\geq$.84\%), achieving the optimal balance at 200X; however, precision was comparatively lower (74–77\%), indicating a predisposition towards malignant predictions (Figure 31). VGG19 exhibited significant sensitivity and dependable detection of malignant cases; nevertheless, diminished accuracy and unpredictability in magnification underscore the limitations of its deeper, sequential CNN design in comparison to contemporary alternatives.

\begin{figure}[H]
\centering

\subfloat[40X]{%
 \includegraphics[width=0.23\textwidth]{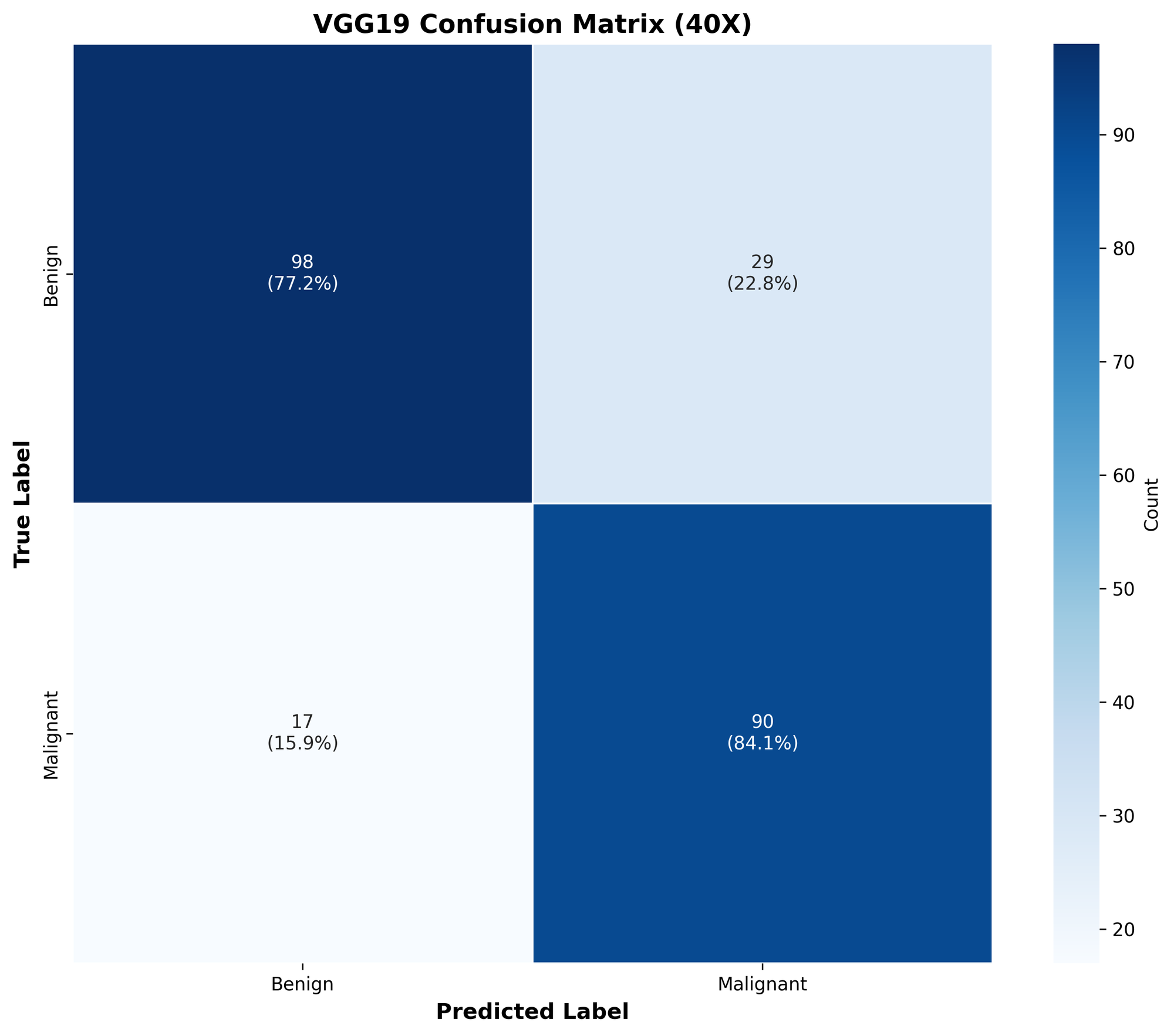}
}\hfill
\subfloat[100X]{%
 \includegraphics[width=0.23\textwidth]{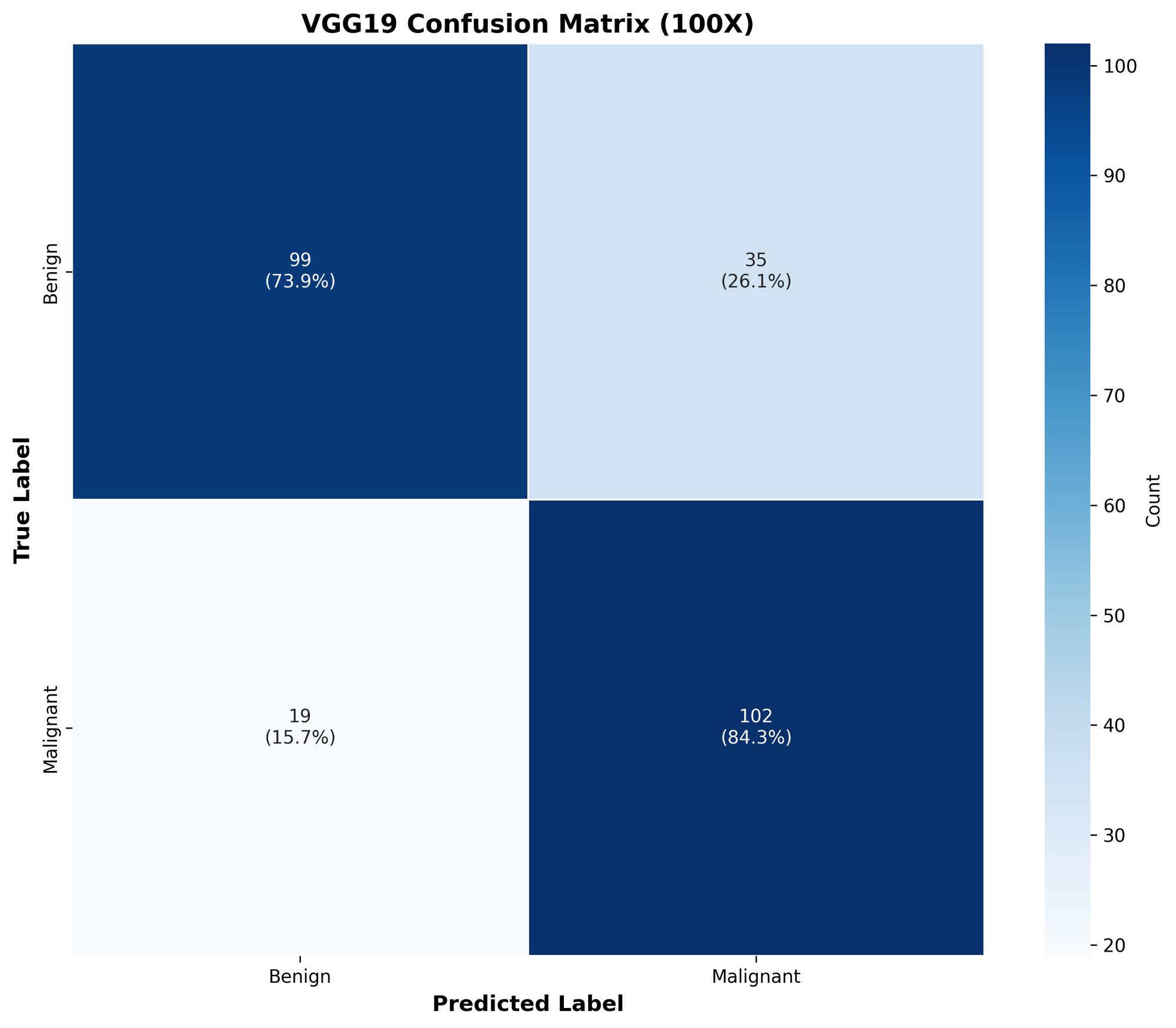}
}\hfill
\subfloat[200X]{%
 \includegraphics[width=0.23\textwidth]{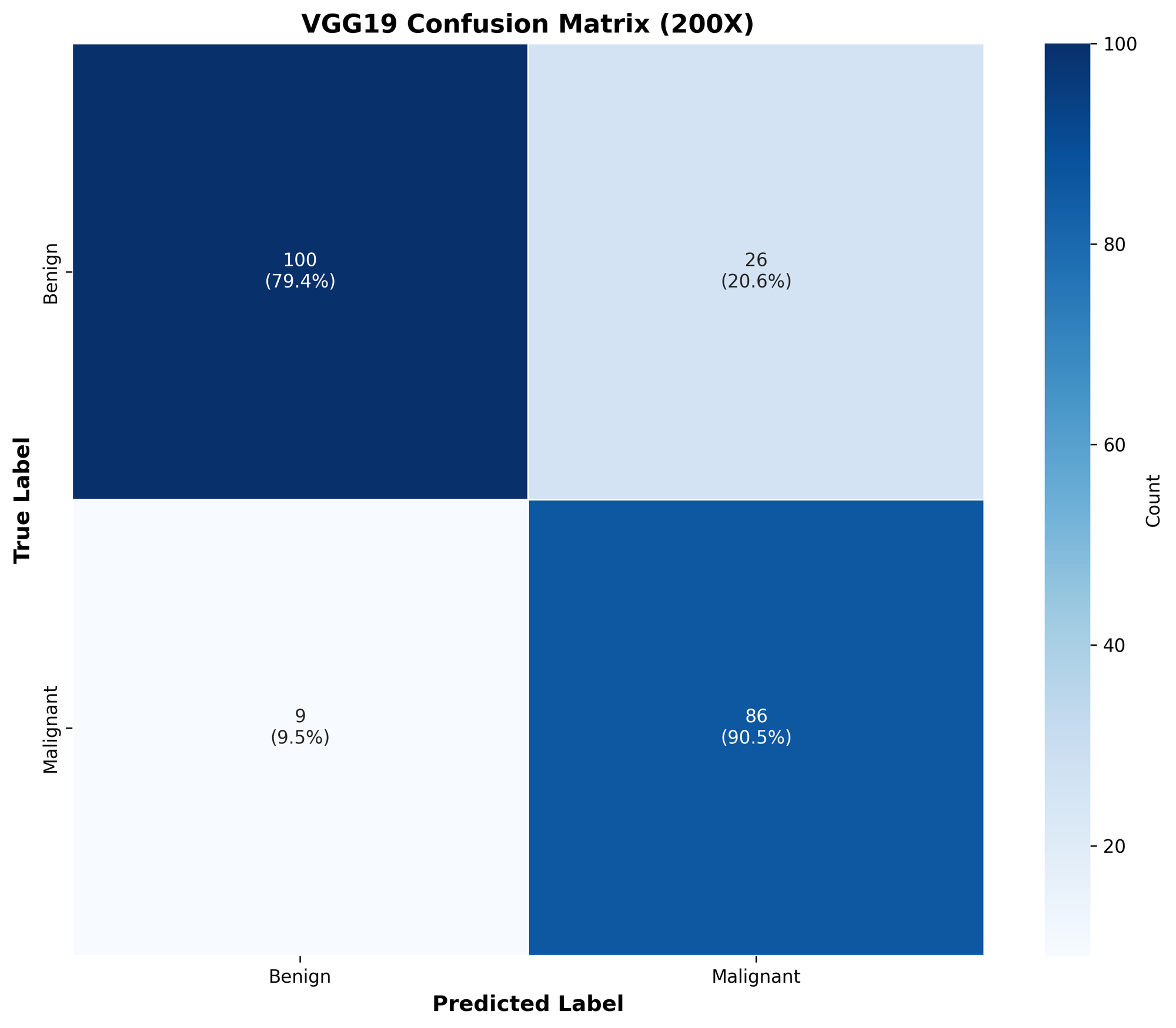}
}\hfill
\subfloat[400X]{%
 \includegraphics[width=0.23\textwidth]{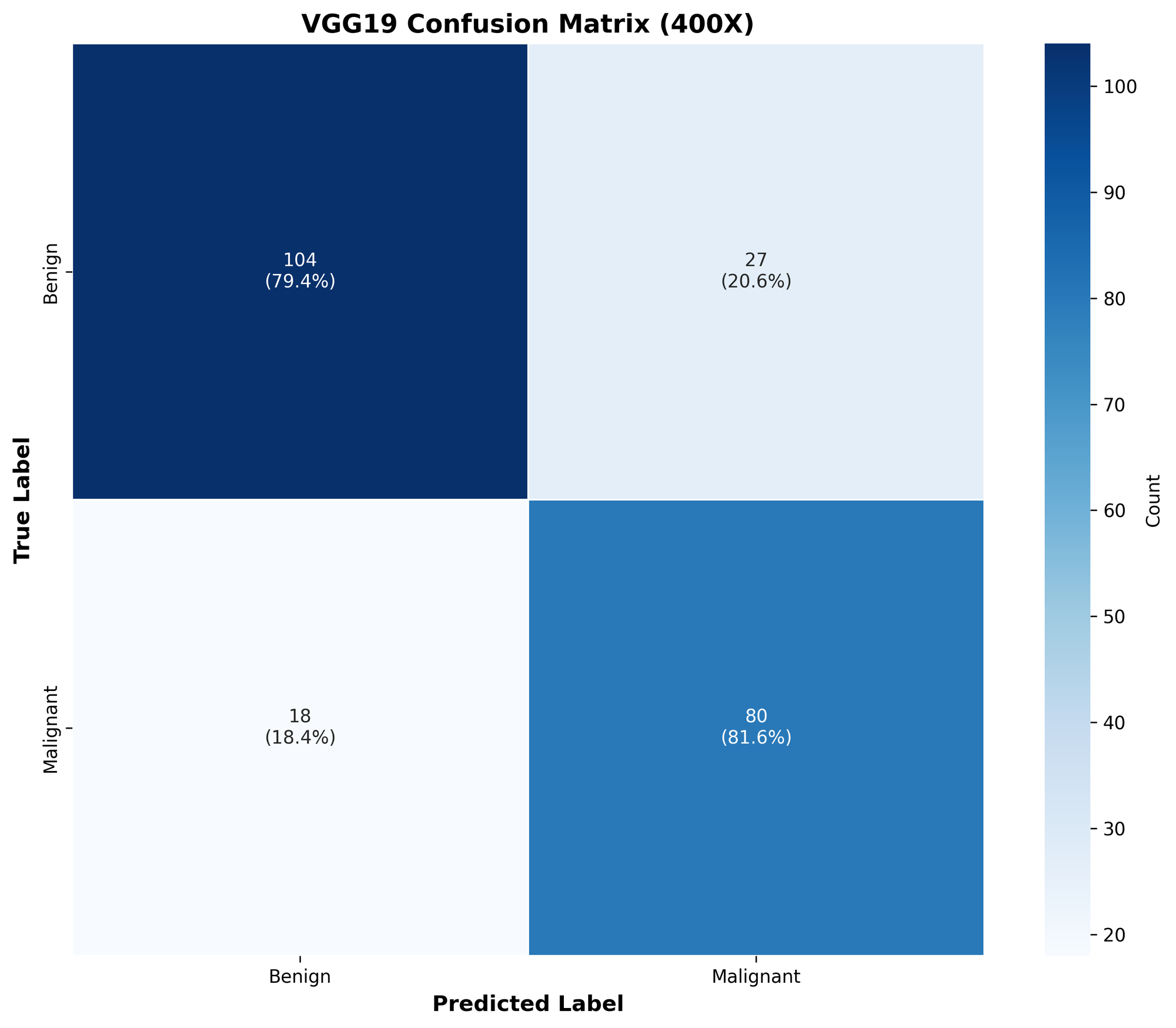}
}

\caption{Confusion matrices of the VGG19 model across 40X, 100X, 200X, and 400X magnifications on the BreakHis dataset.}
\label{fig:vgg19_confusion_matrices}
\end{figure}

\begin{figure}[H]
\centering

\subfloat[Validation ROC]{%
 \includegraphics[width=0.23\textwidth]{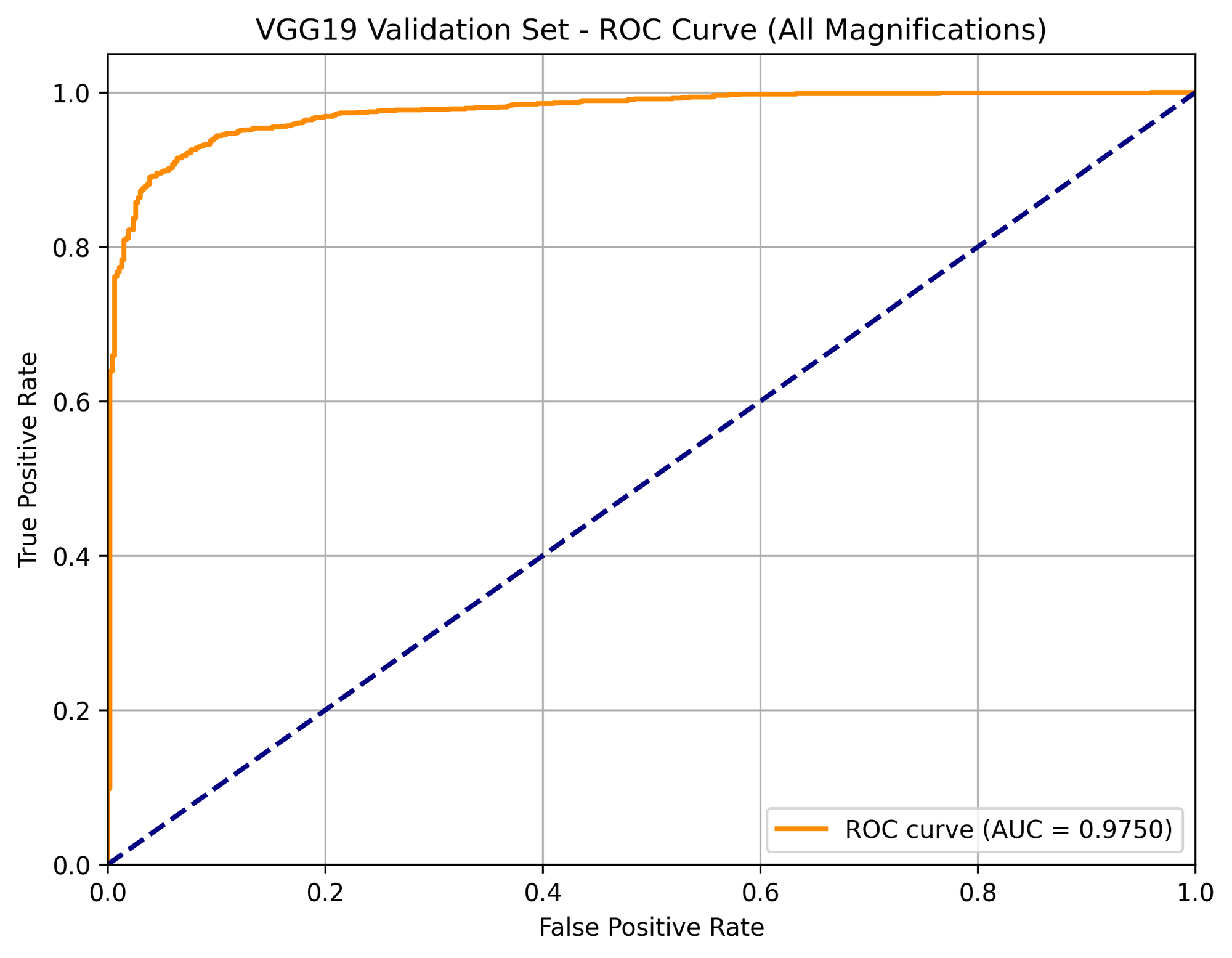}
}\hfill
\subfloat[Validation Confusion Matrix]{%
 \includegraphics[width=0.23\textwidth]{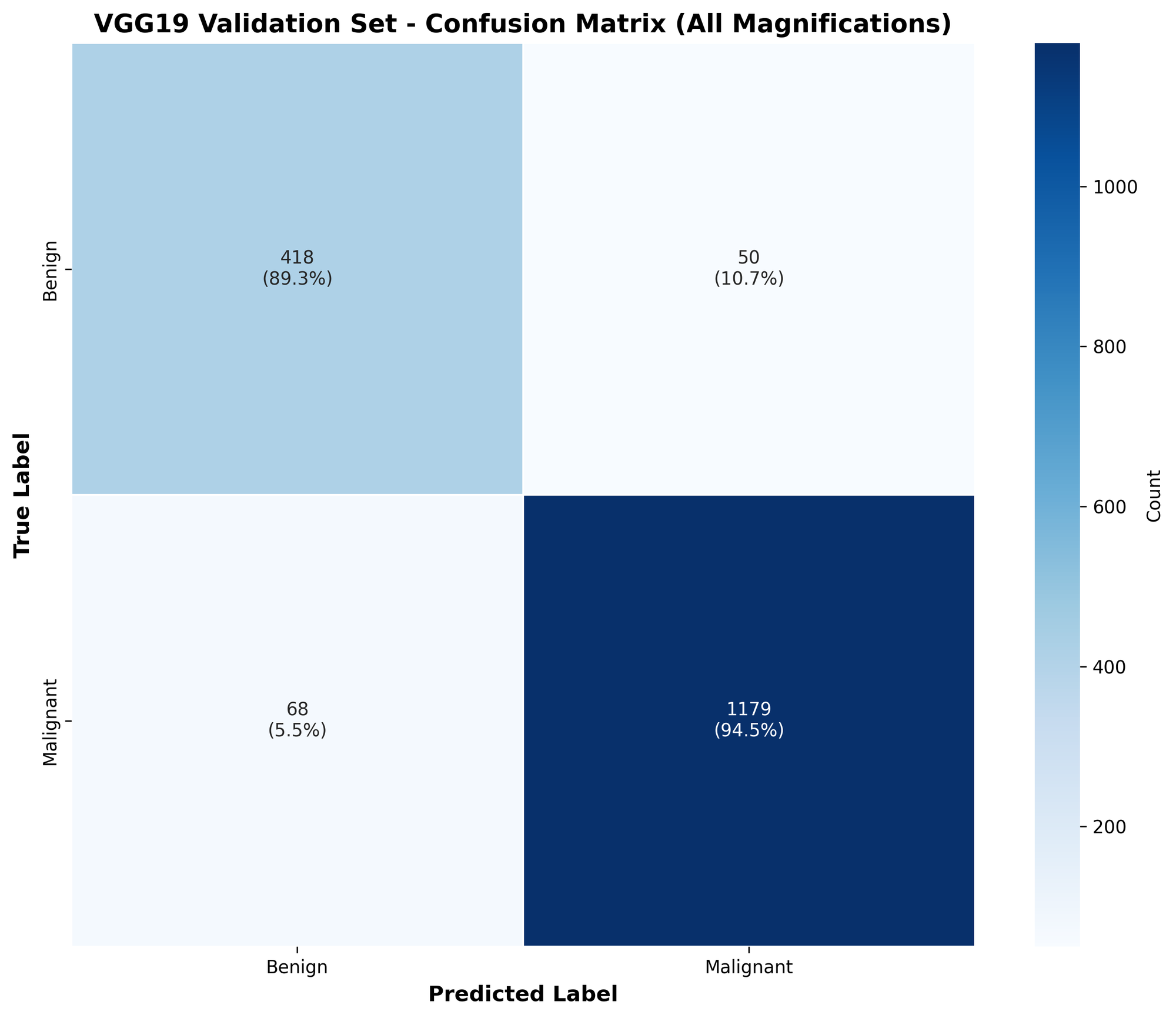}
}\hfill
\subfloat[Test ROC]{%
 \includegraphics[width=0.23\textwidth]{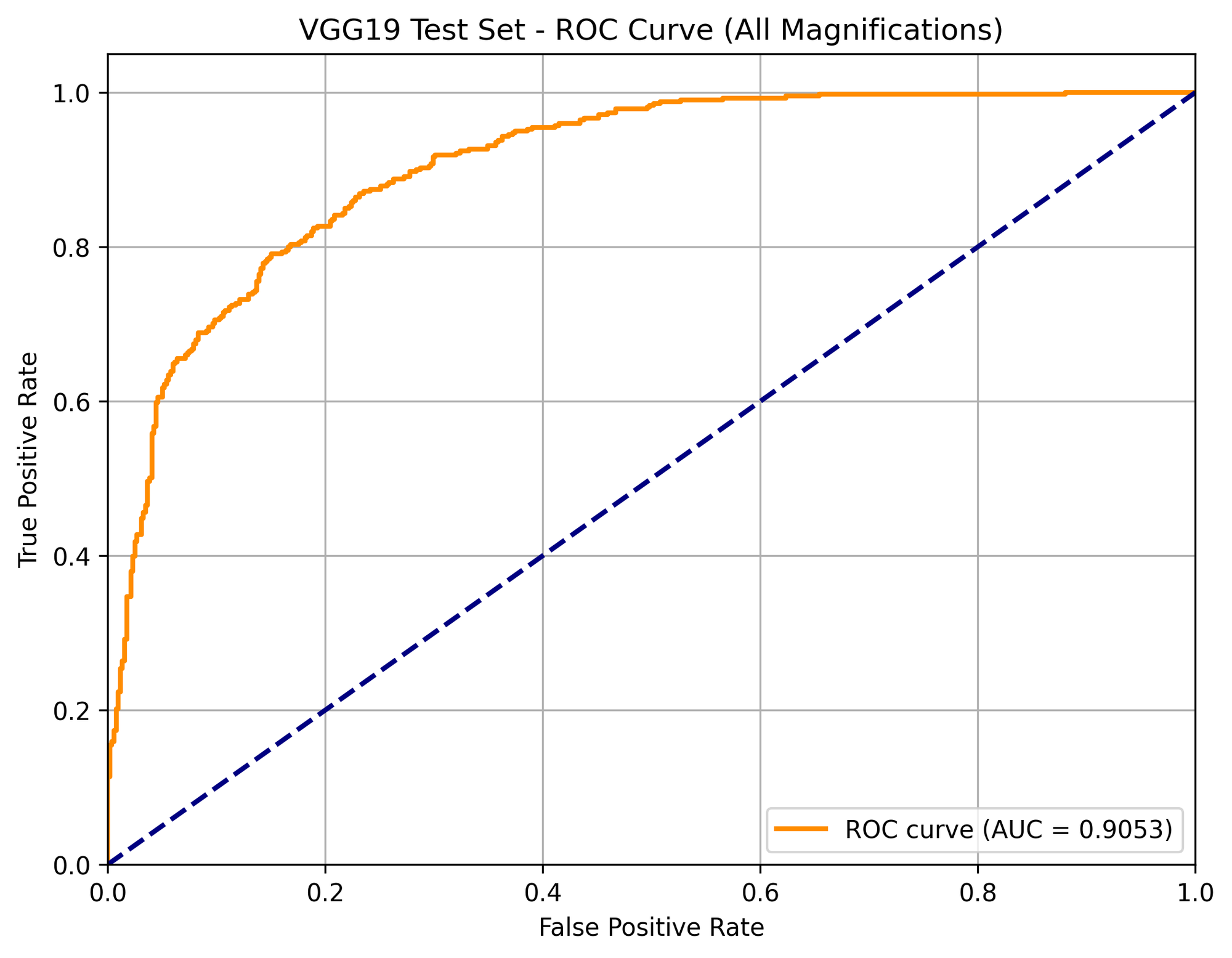}
}\hfill
\subfloat[Test Confusion Matrix]{%
 \includegraphics[width=0.23\textwidth]{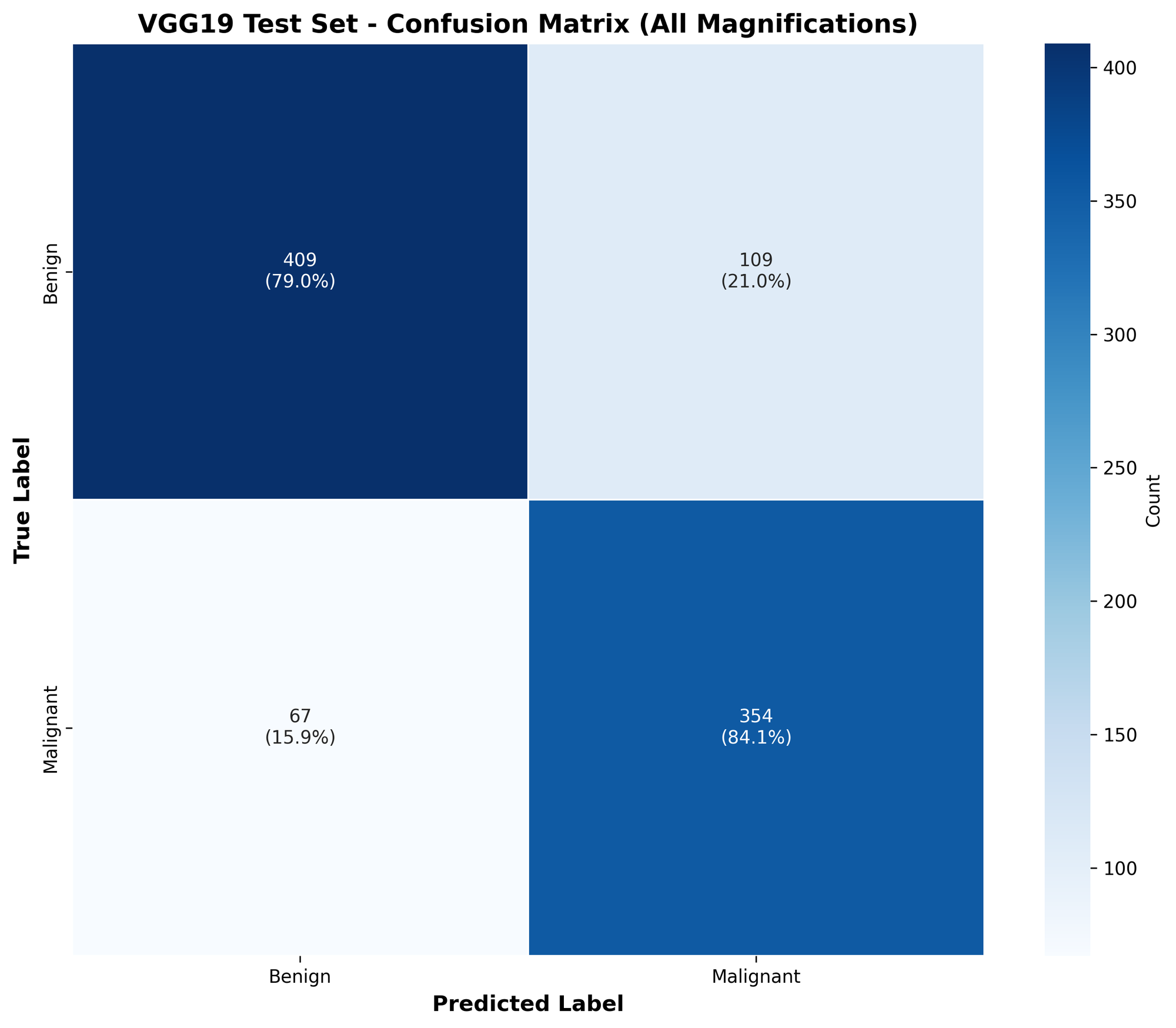}
}

\caption{Validation and test performance of the VGG19 model on the BreakHis dataset across all magnifications.}
\label{fig:vgg19_val_test}
\end{figure}

\begin{figure}[H]
\centering

\subfloat[ROC 40X]{%
 \includegraphics[width=0.23\textwidth]{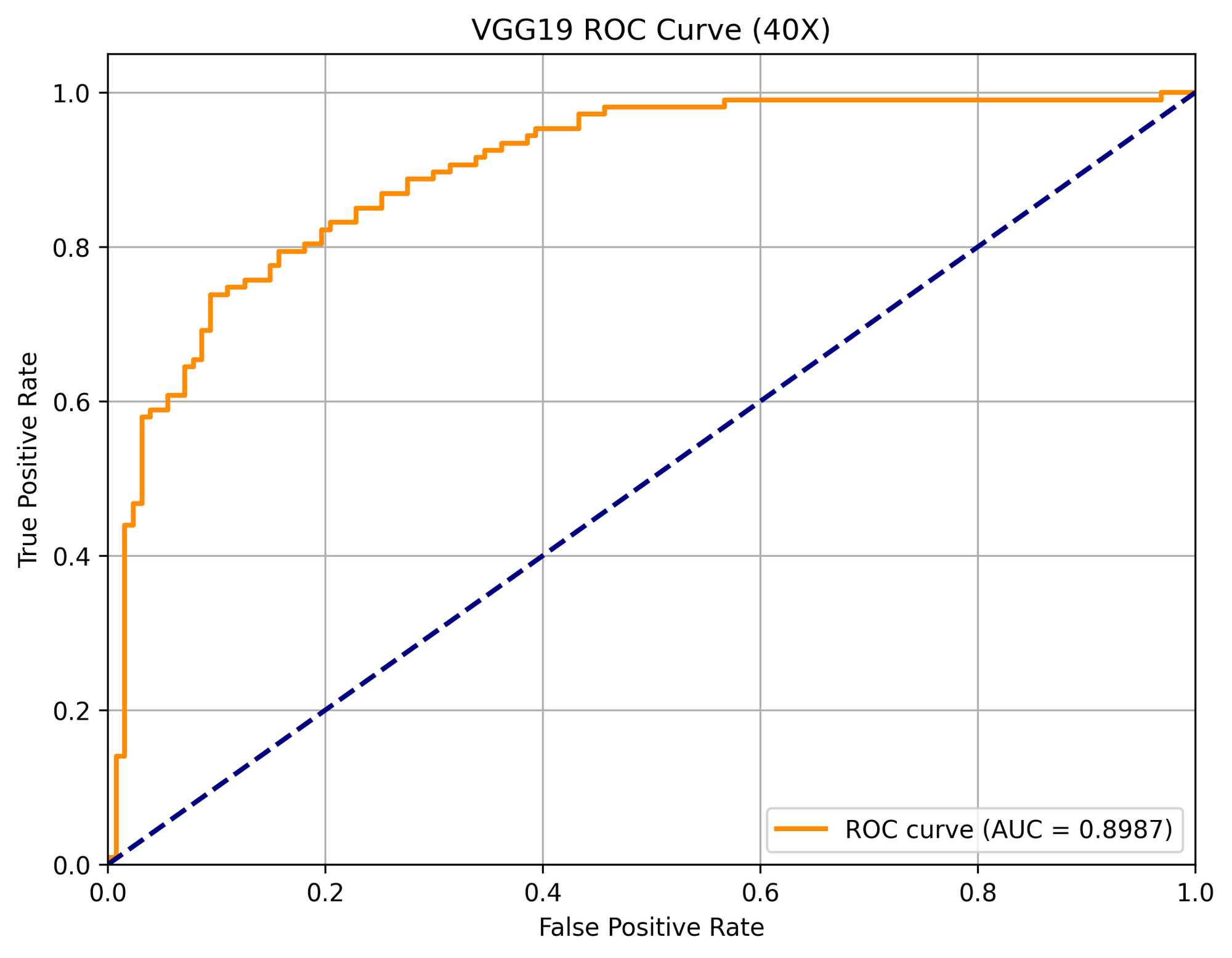}
}\hfill
\subfloat[ROC 100X]{%
 \includegraphics[width=0.23\textwidth]{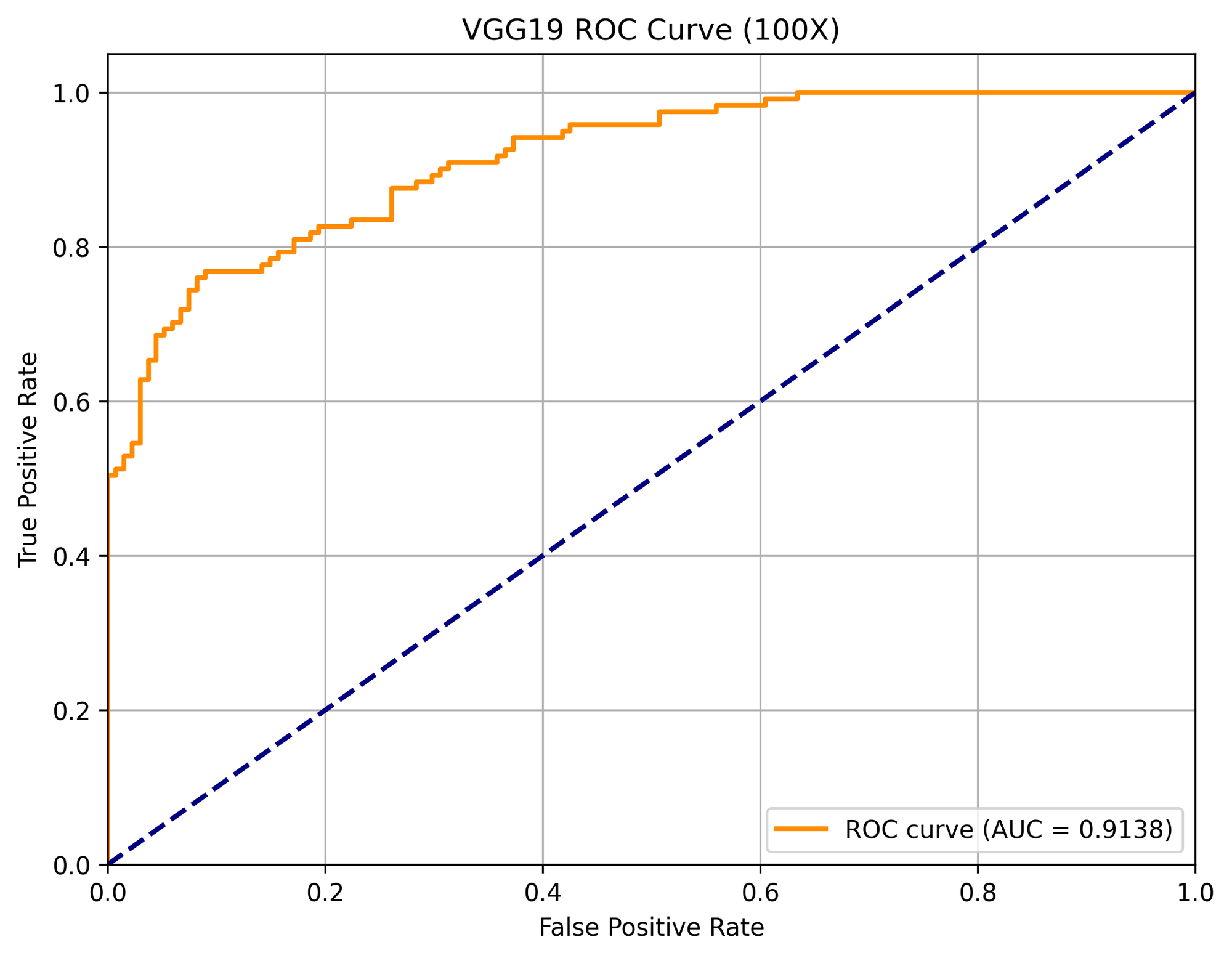}
}\hfill
\subfloat[ROC 200X]{%
 \includegraphics[width=0.23\textwidth]{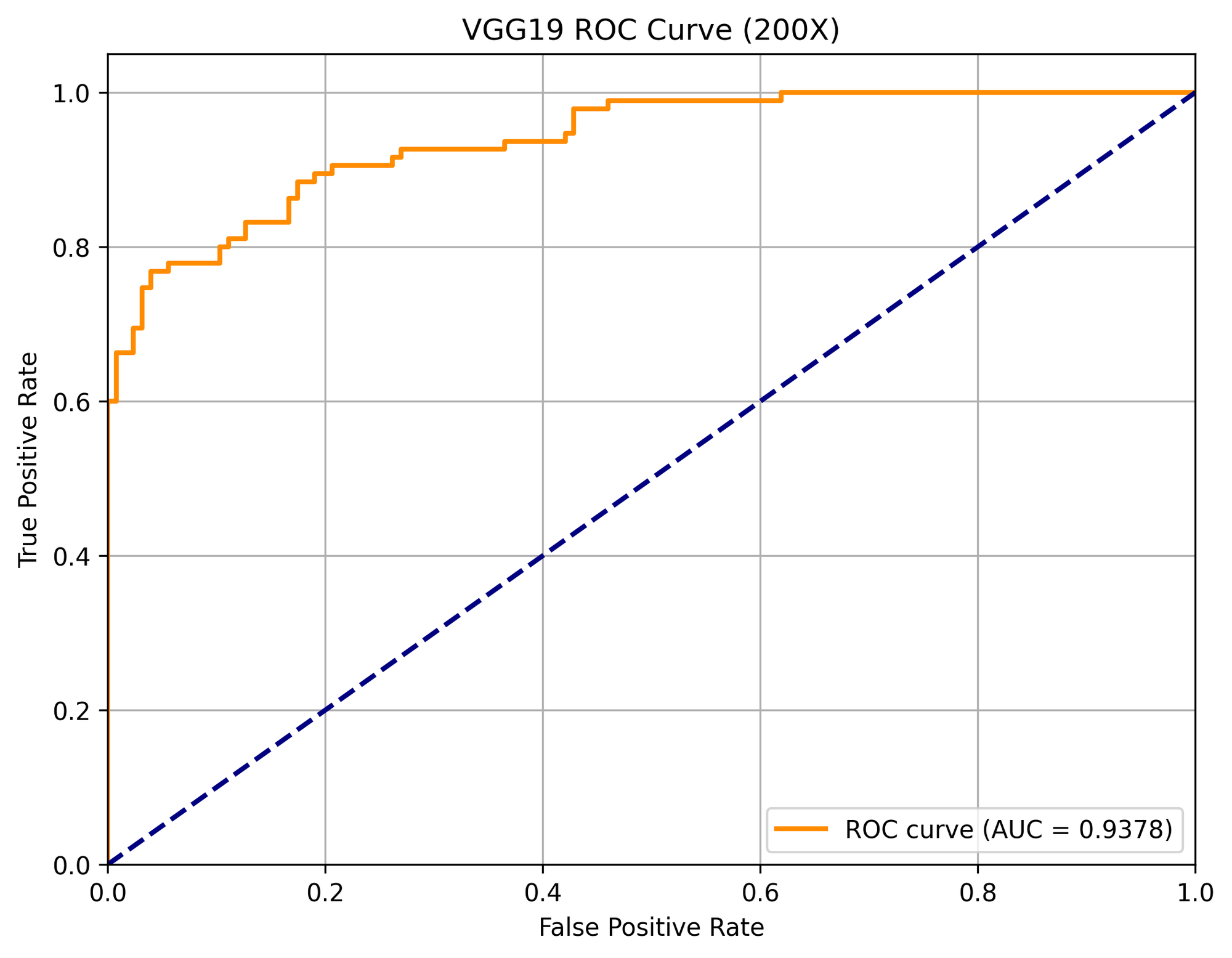}
}\hfill
\subfloat[ROC 400X]{%
 \includegraphics[width=0.23\textwidth]{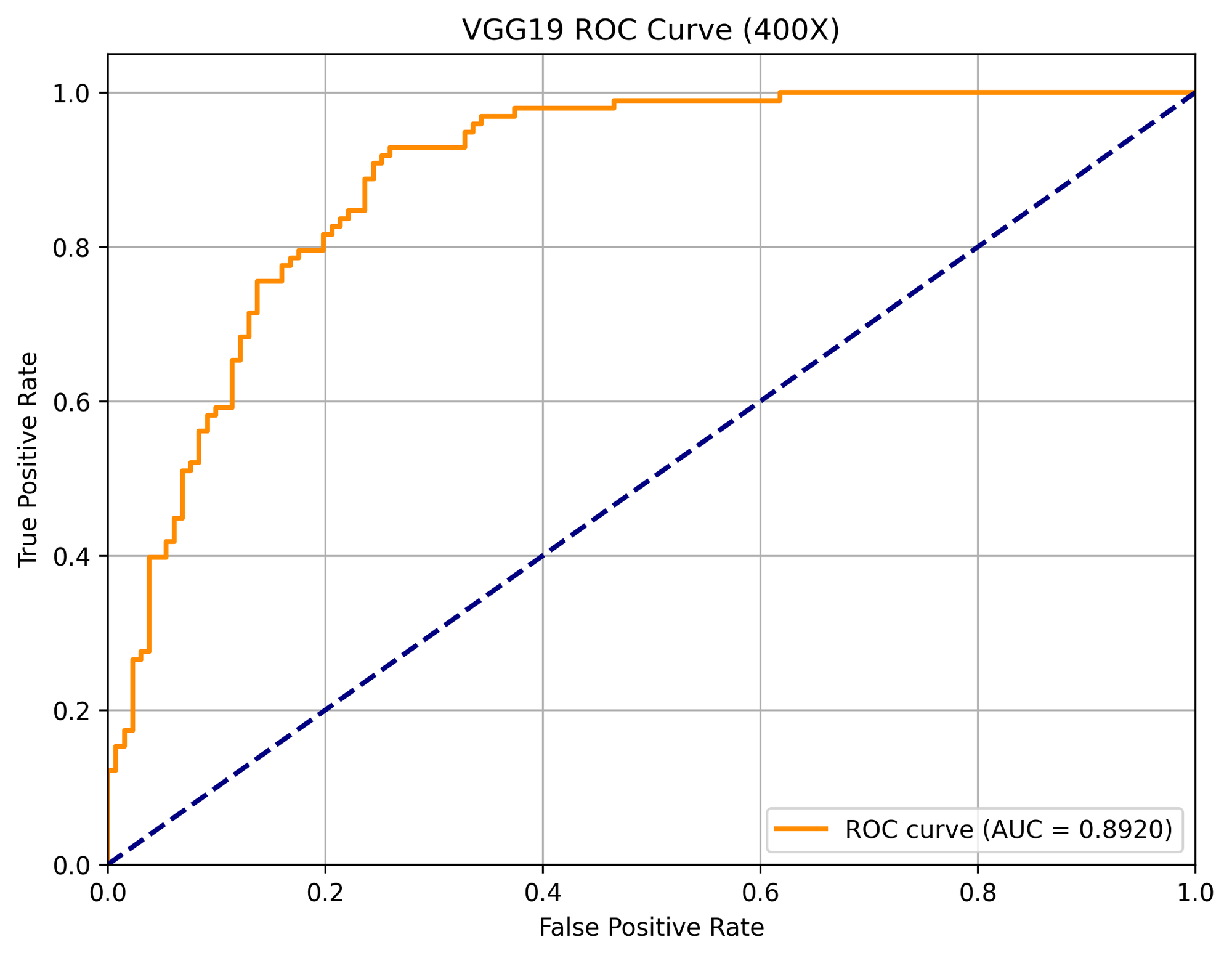}
}

\vspace{0.6ex}

\subfloat[PR 40X]{%
 \includegraphics[width=0.23\textwidth]{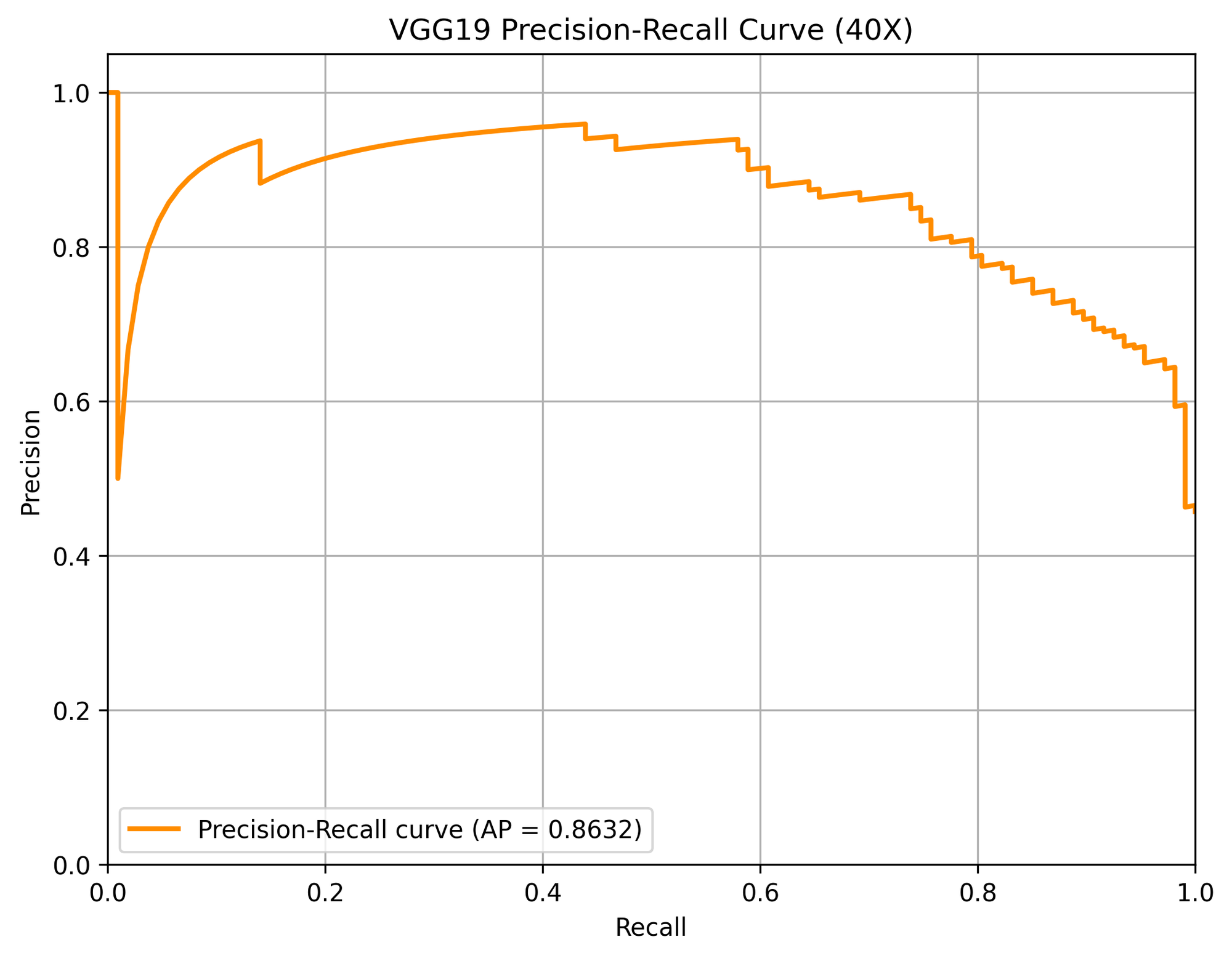}
}\hfill
\subfloat[PR 100X]{%
 \includegraphics[width=0.23\textwidth]{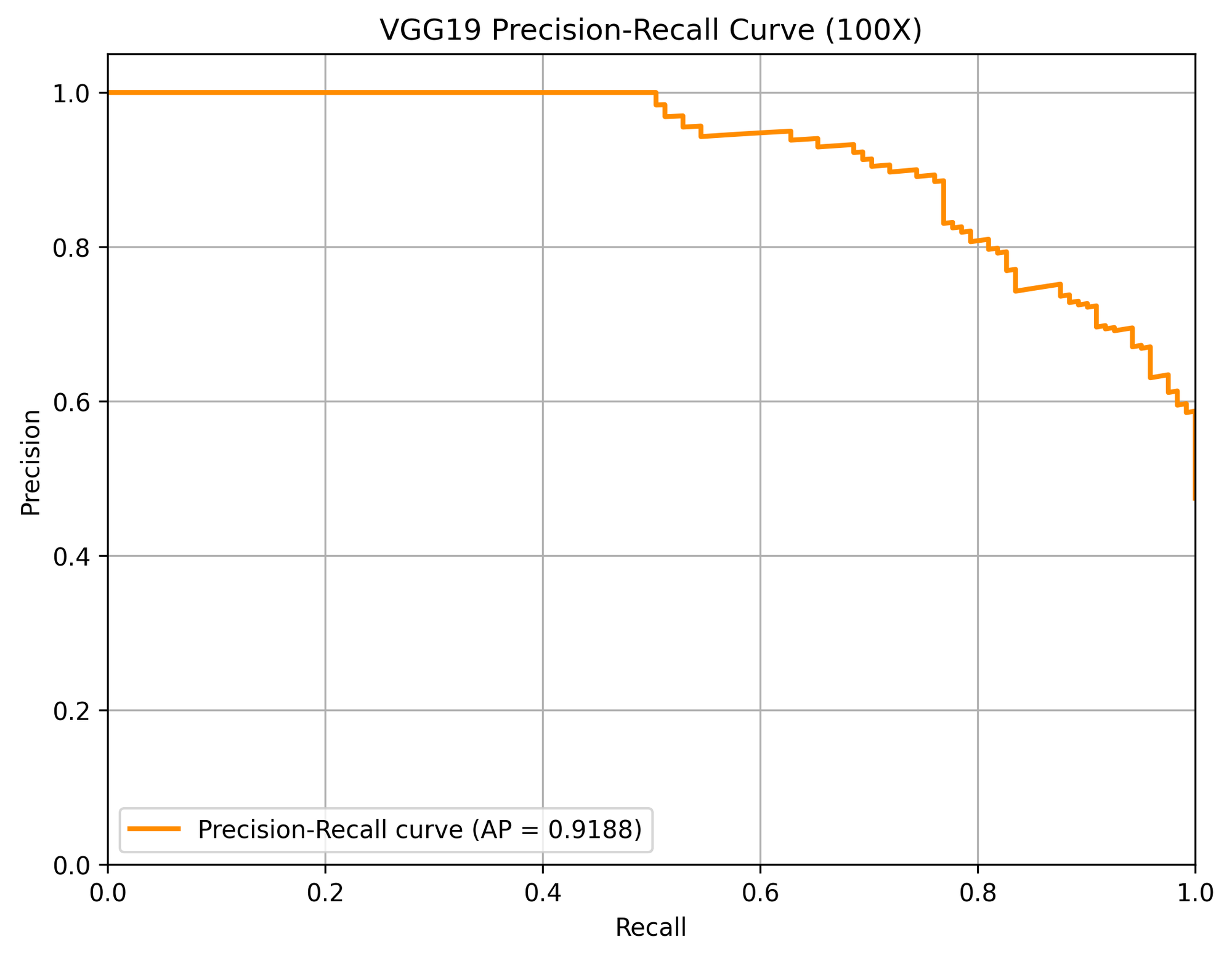}
}\hfill
\subfloat[PR 200X]{%
 \includegraphics[width=0.23\textwidth]{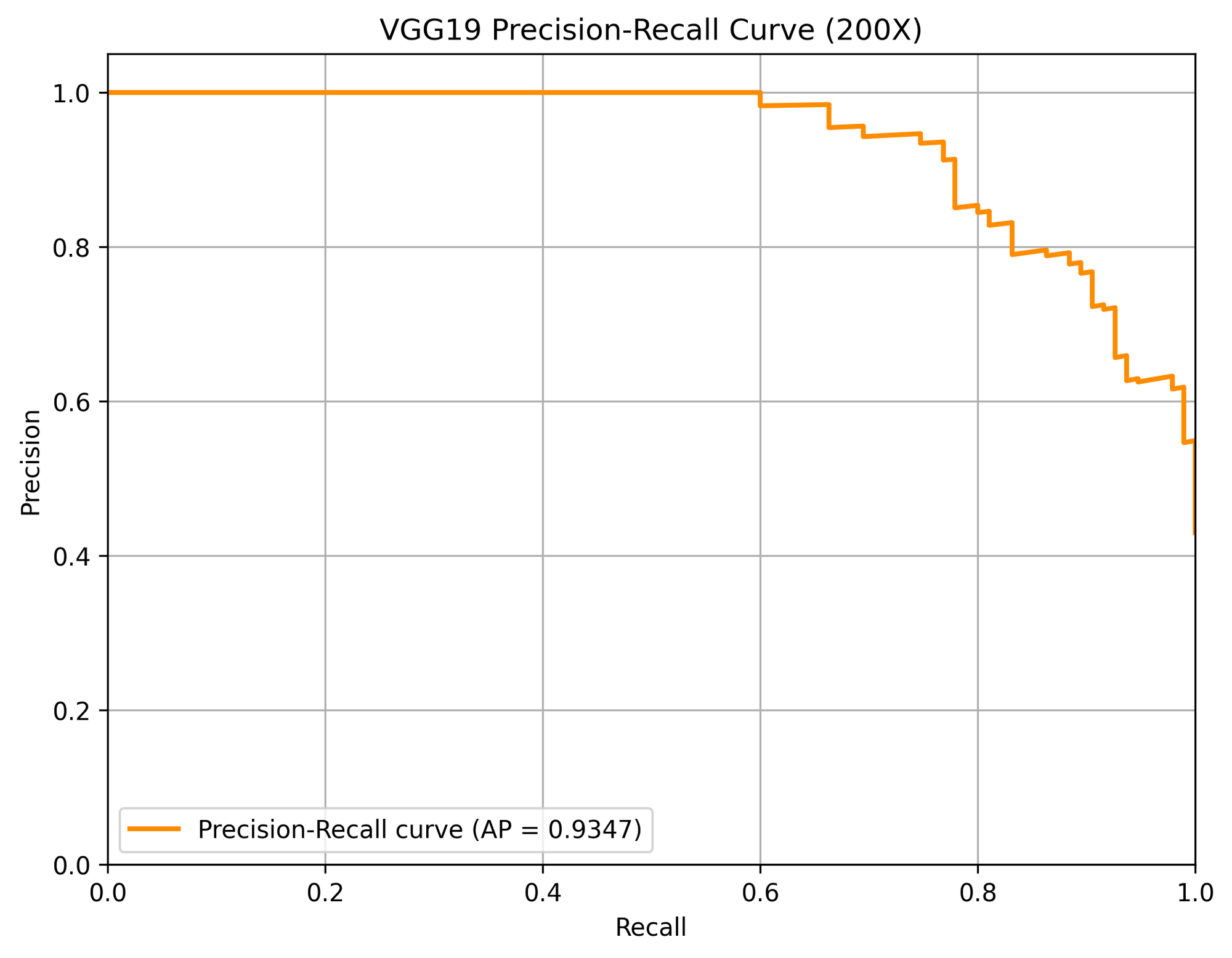}
}\hfill
\subfloat[PR 400X]{%
 \includegraphics[width=0.23\textwidth]{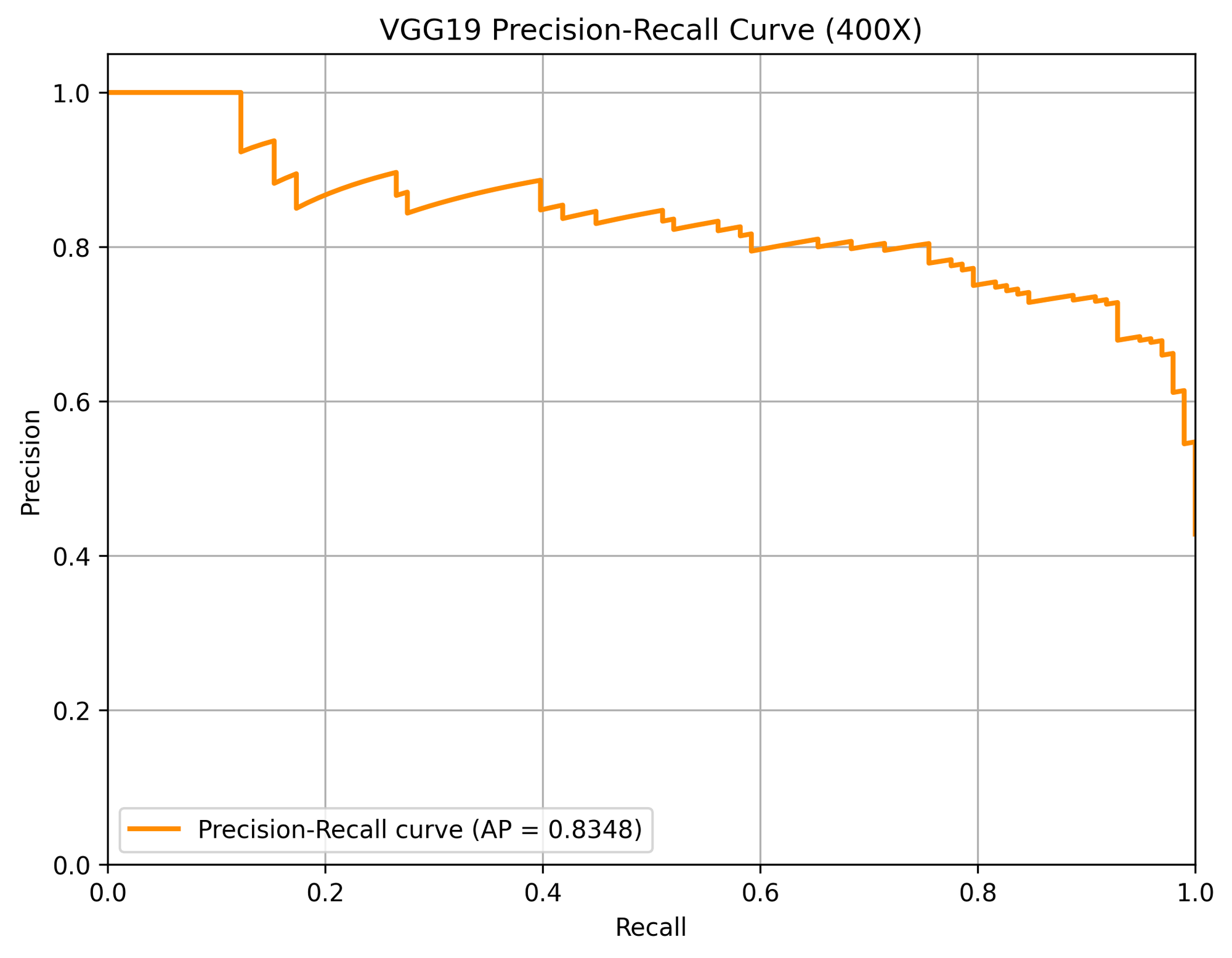}
}

\caption{Receiver Operating Characteristic (ROC) curves (a–d) and Precision–Recall (PR) curves (e–h) for the VGG19 model evaluated on the BreakHis dataset across 40X, 100X, 200X, and 400X magnifications.}
\label{fig:vgg19_roc_pr}
\end{figure}

\begin{table}[ht]
\caption{Performance of Vision Transformer (ViT) on the BreakHis dataset}
\label{tab:vit}
\centering
\begin{tabular}{@{}lcccccc@{}}
\toprule
\textbf{Dataset} 
& \textbf{Accuracy (\%)} 
& \textbf{Precision (\%)} 
& \textbf{Recall (\%)} 
& \textbf{F1-score (\%)} 
& \textbf{Specificity (\%)} 
& \textbf{AUC} \\
\midrule
Validation (All Mag.) 
& 86.5 
& 90.9 
& 90.5 
& 90.7 
& 75.9 
& 0.892 \\

Test (All Mag.)    
& 77.6 
& 71.1 
& 84.3 
& 77.2 
& 72.2 
& 0.858 \\

Test (40$\times$)      
& 79.1 
& 71.0 
& 91.6 
& 80.0 
& 68.5 
& 0.856 \\

Test (100$\times$)      
& 72.6 
& 68.4 
& 78.5 
& 73.1 
& 67.2 
& 0.789 \\

Test (200$\times$)      
& 81.0 
& 73.9 
& 86.3 
& 79.6 
& 77.0 
& 0.893 \\

Test (400$\times$)      
& 79.0 
& 71.9 
& 83.7 
& 77.4 
& 75.6 
& 0.862 \\
\bottomrule
\end{tabular}

\end{table}

The Vision Transformer (ViT) showed an accuracy of 86.47\%, precision of 90.89\%, recall of 90.46\%, specificity of 75.85\%, F1-score of 90.68\%, and an AUC-ROC of 0.8918 on the validation set, showing robust sensitivity and balanced performance throughout training (Table 9). In the test set (across all magnifications), performance decreased to 77.64\% accuracy, 71.14\% precision, 84.32\% recall, 72.20\% specificity, 77.17\% F1-score, and 0.8578 AUC-ROC, signifying diminished precision and an elevated false-positive rate (Figure 32, 33 \& 34). At particular magnifications, ViT exhibited optimal performance at 200X, achieving 81.0\% accuracy, 73.87\% precision, 86.32\% recall, 76.98\% specificity, 79.61\% F1-score, and an AUC of 0.8928. In contrast, both 40X and 400X attained approximately 79\% accuracy with recall exceeding 83\%, while performance was least effective at 100X, yielding 72.55\% accuracy and an AUC of 0.7885 (Figure 35). The results underscore ViT's proficiency in recall across various magnifications, affirming its capacity to identify malignant patients; nevertheless, diminished precision and magnification-dependent variability indicate difficulties in discerning intricate histological patterns relative to deeper CNN-based models.

\begin{figure}[H]
 \centering
 \includegraphics[width=0.75\linewidth]{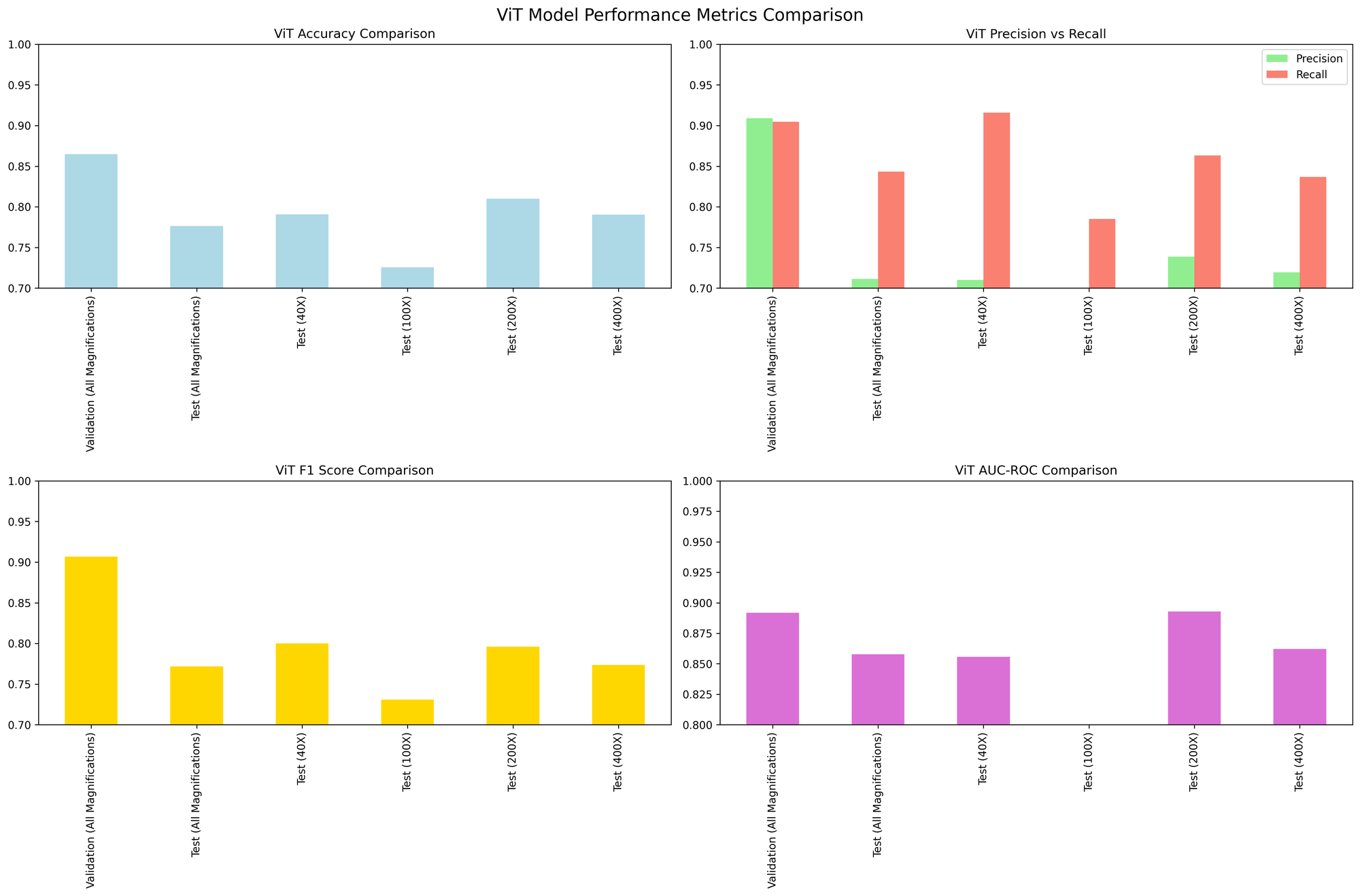}
 \caption{ViT model performance across validation and test sets at multiple magnifications. Validation achieved strong results (Accuracy = 0.86, Precision = 0.91, Recall = 0.90, F1 = 0.91, AUC-ROC = 0.89).}
 \label{fig:single9}
\end{figure}

\begin{figure}[H]
\centering

\subfloat[40X]{%
 \includegraphics[width=0.23\textwidth]{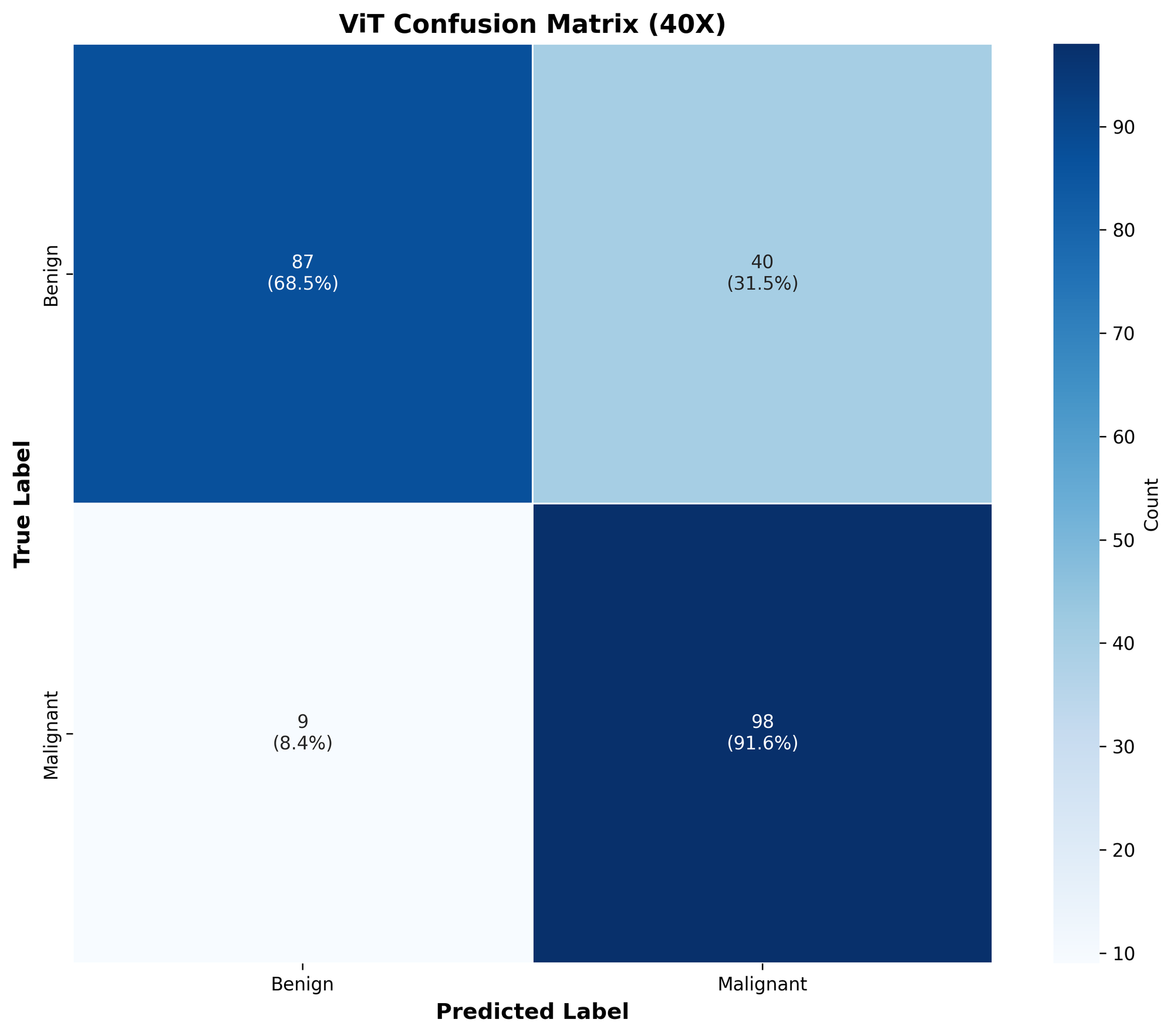}
}\hfill
\subfloat[100X]{%
 \includegraphics[width=0.23\textwidth]{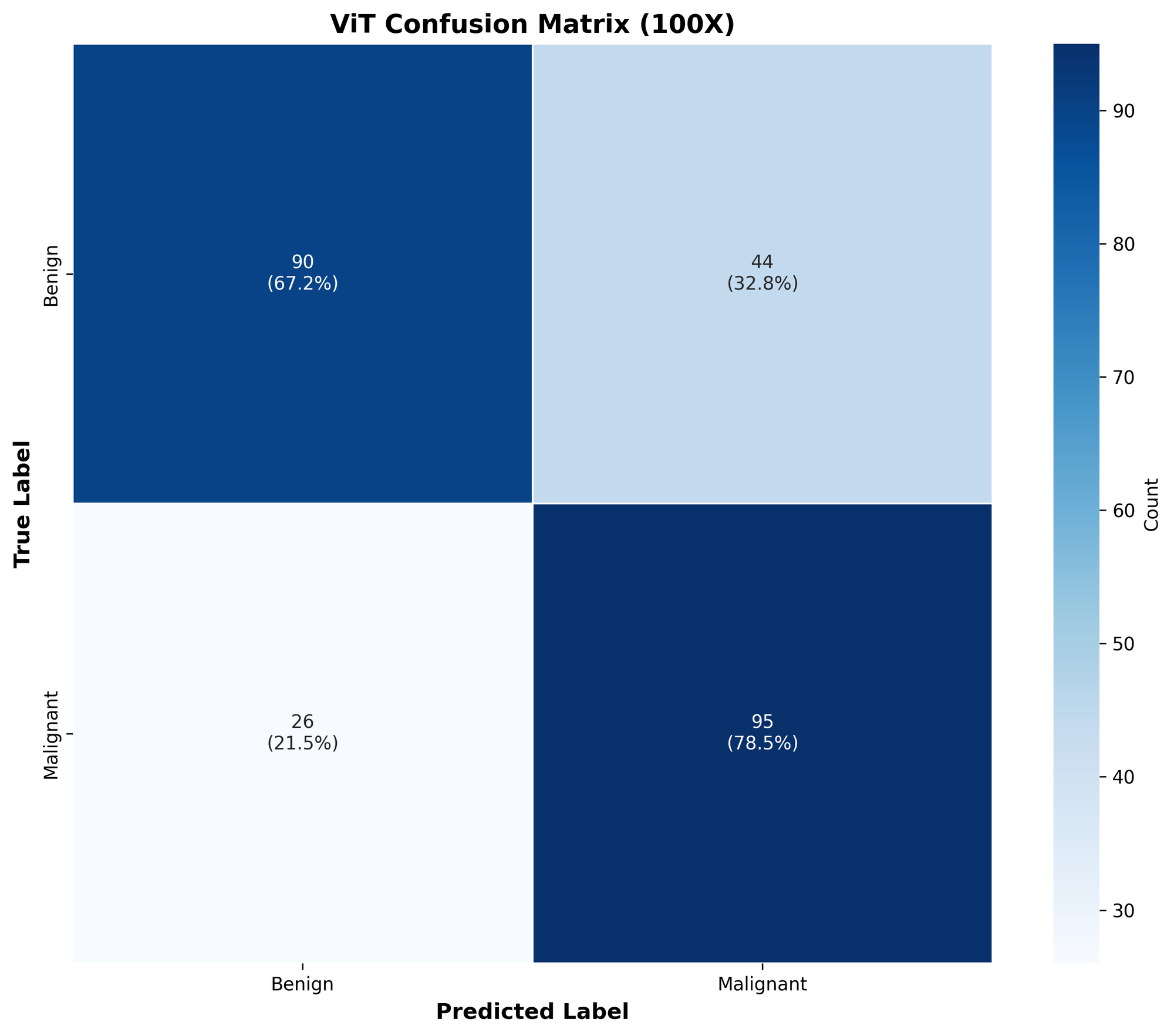}
}\hfill
\subfloat[200X]{%
 \includegraphics[width=0.23\textwidth]{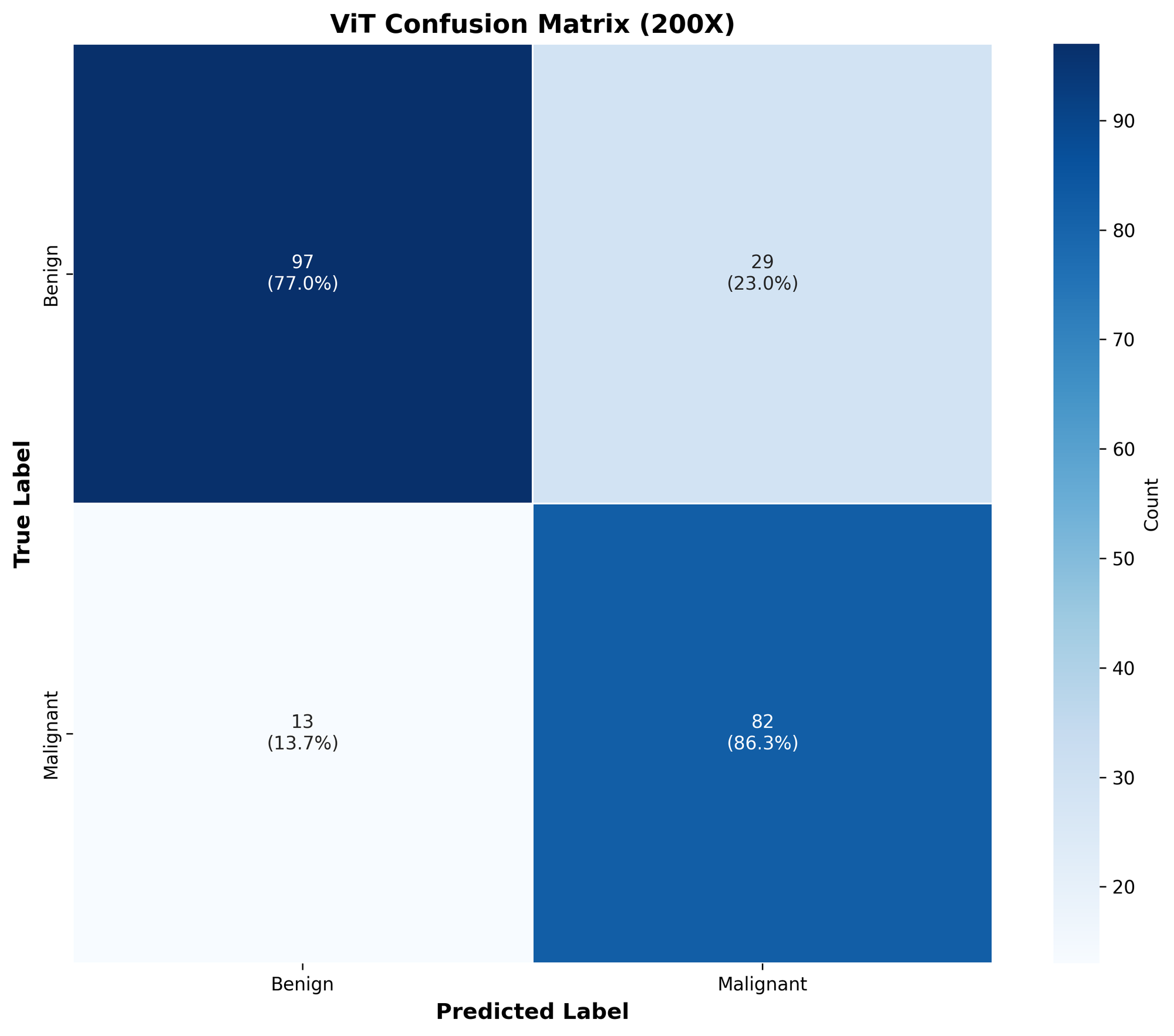}
}\hfill
\subfloat[400X]{%
 \includegraphics[width=0.23\textwidth]{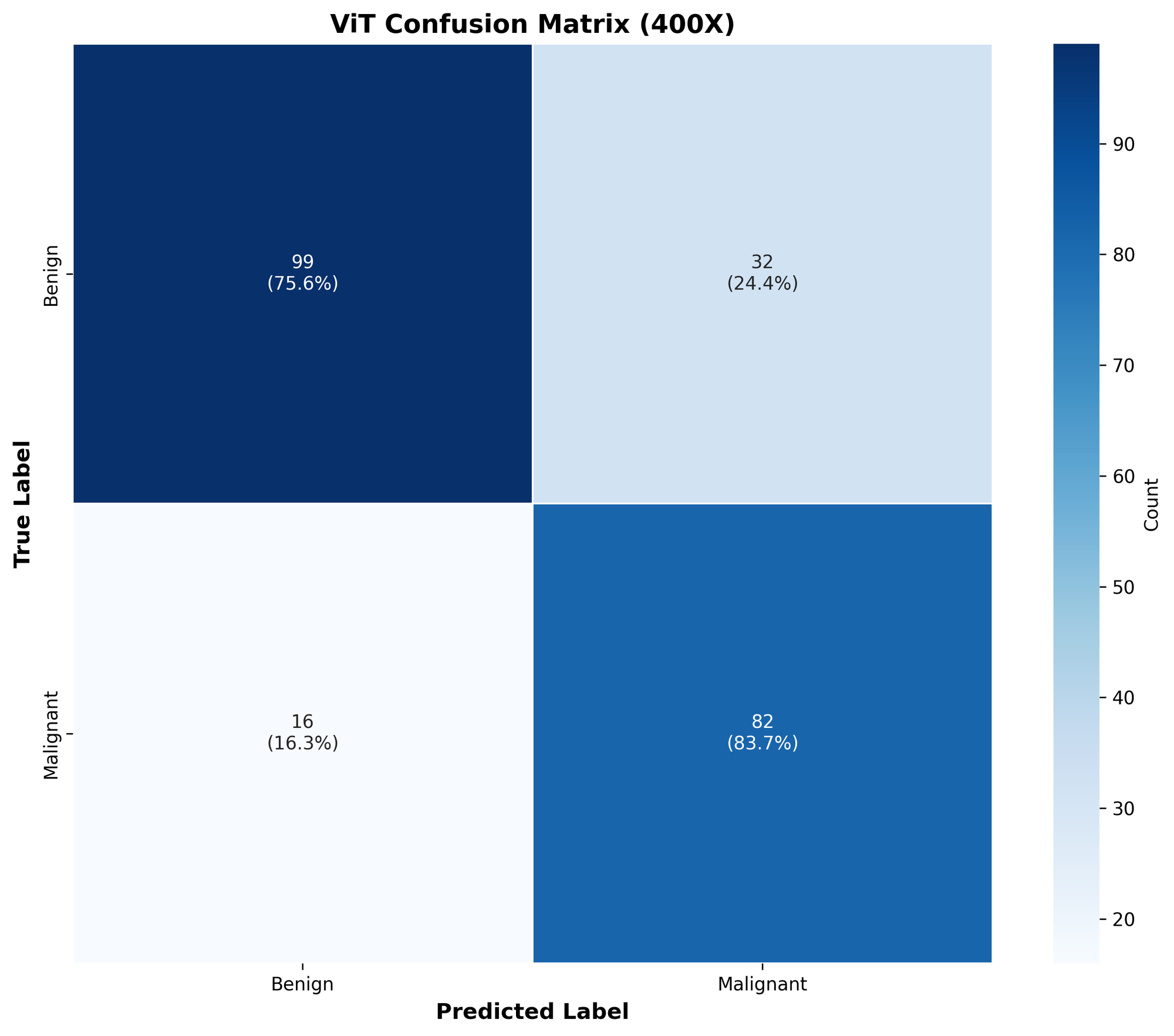}
}

\caption{Confusion matrices of the Vision Transformer (ViT) model across 40X, 100X, 200X, and 400X magnifications on the BreakHis dataset.}
\label{fig:vit_confusion_matrices}
\end{figure}

\begin{figure}[H]
\centering

\subfloat[Validation ROC]{%
 \includegraphics[width=0.23\textwidth]{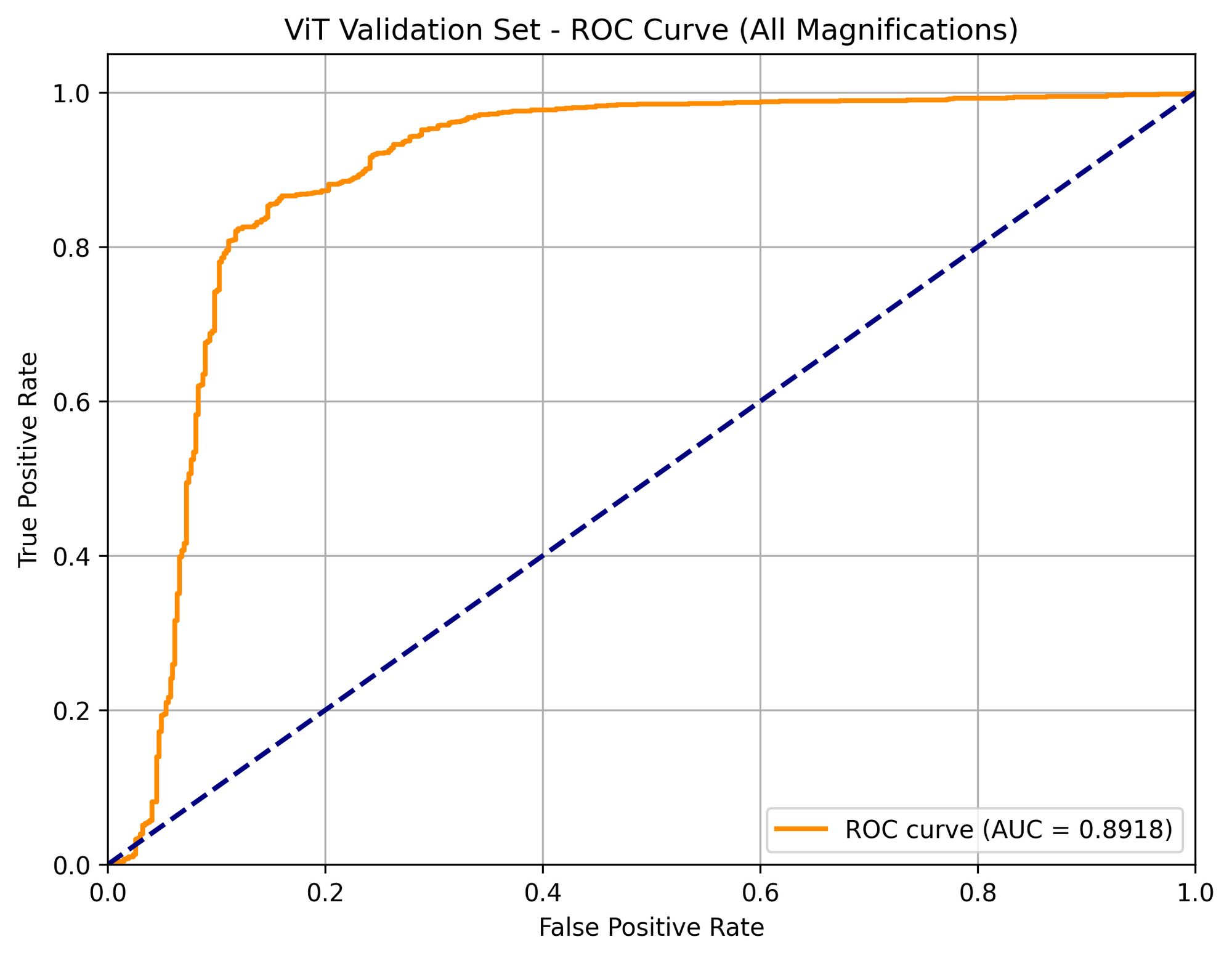}
}\hfill
\subfloat[Validation Confusion Matrix]{%
 \includegraphics[width=0.23\textwidth]{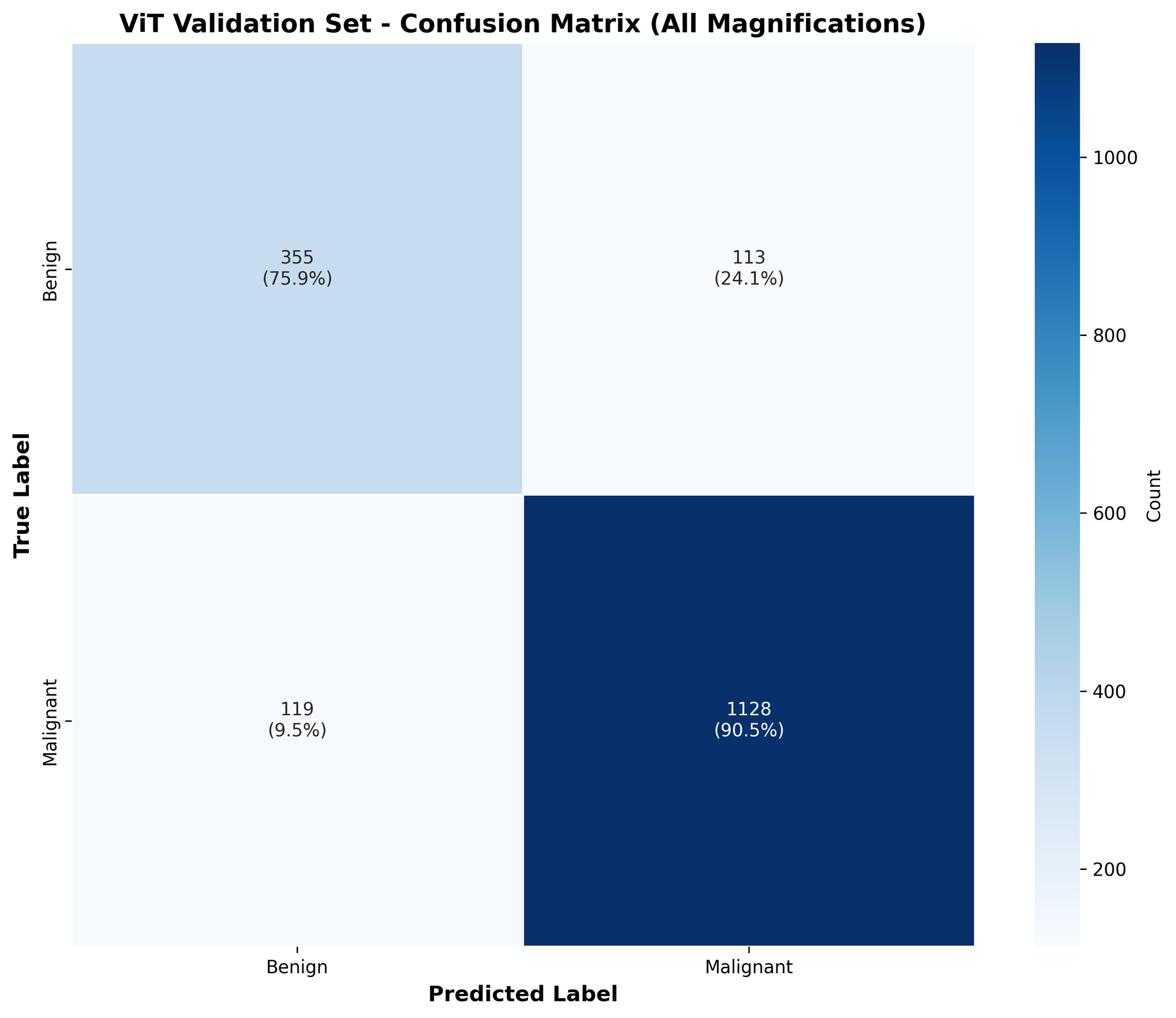}
}\hfill
\subfloat[Test ROC]{%
 \includegraphics[width=0.23\textwidth]{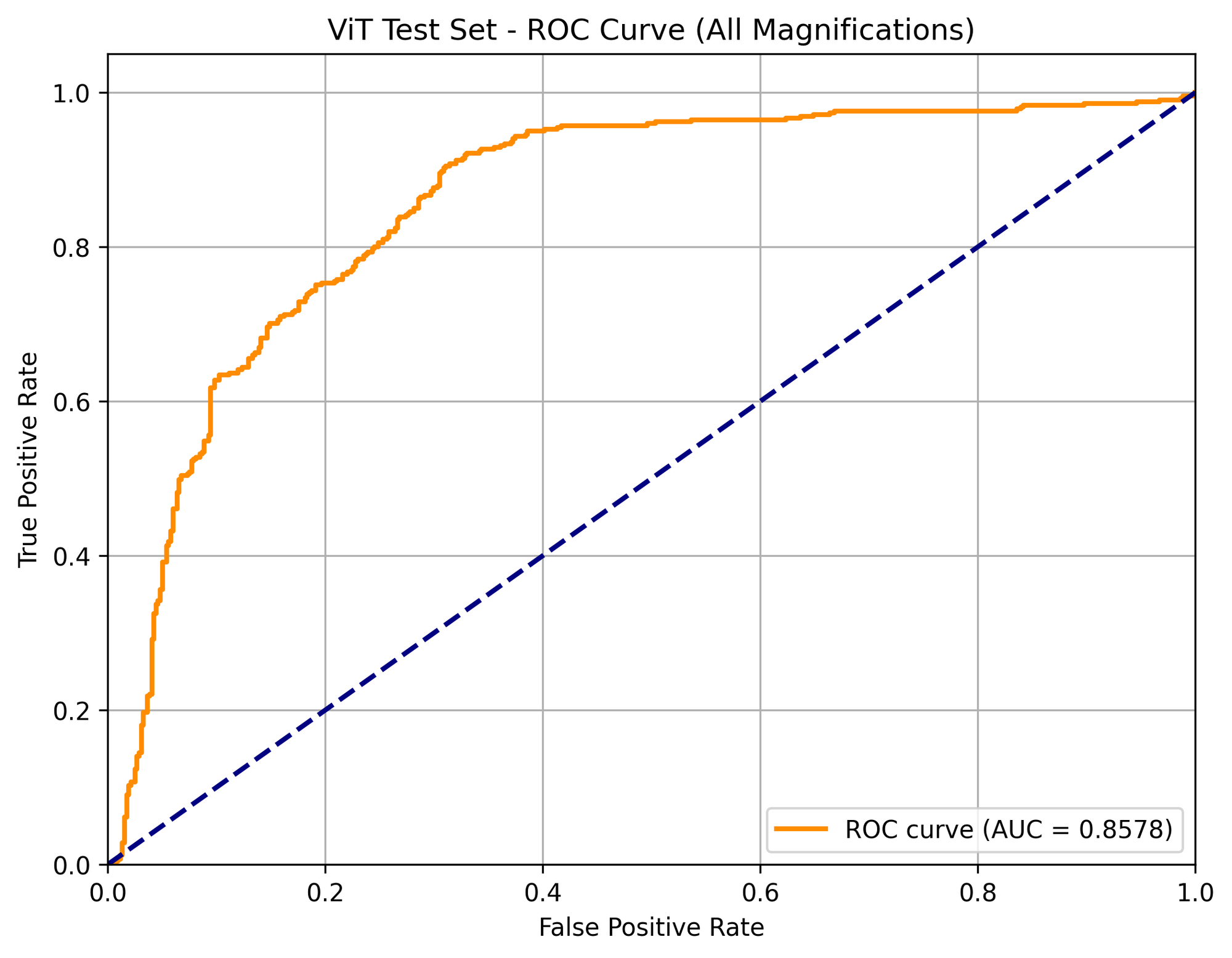}
}\hfill
\subfloat[Test Confusion Matrix]{%
 \includegraphics[width=0.23\textwidth]{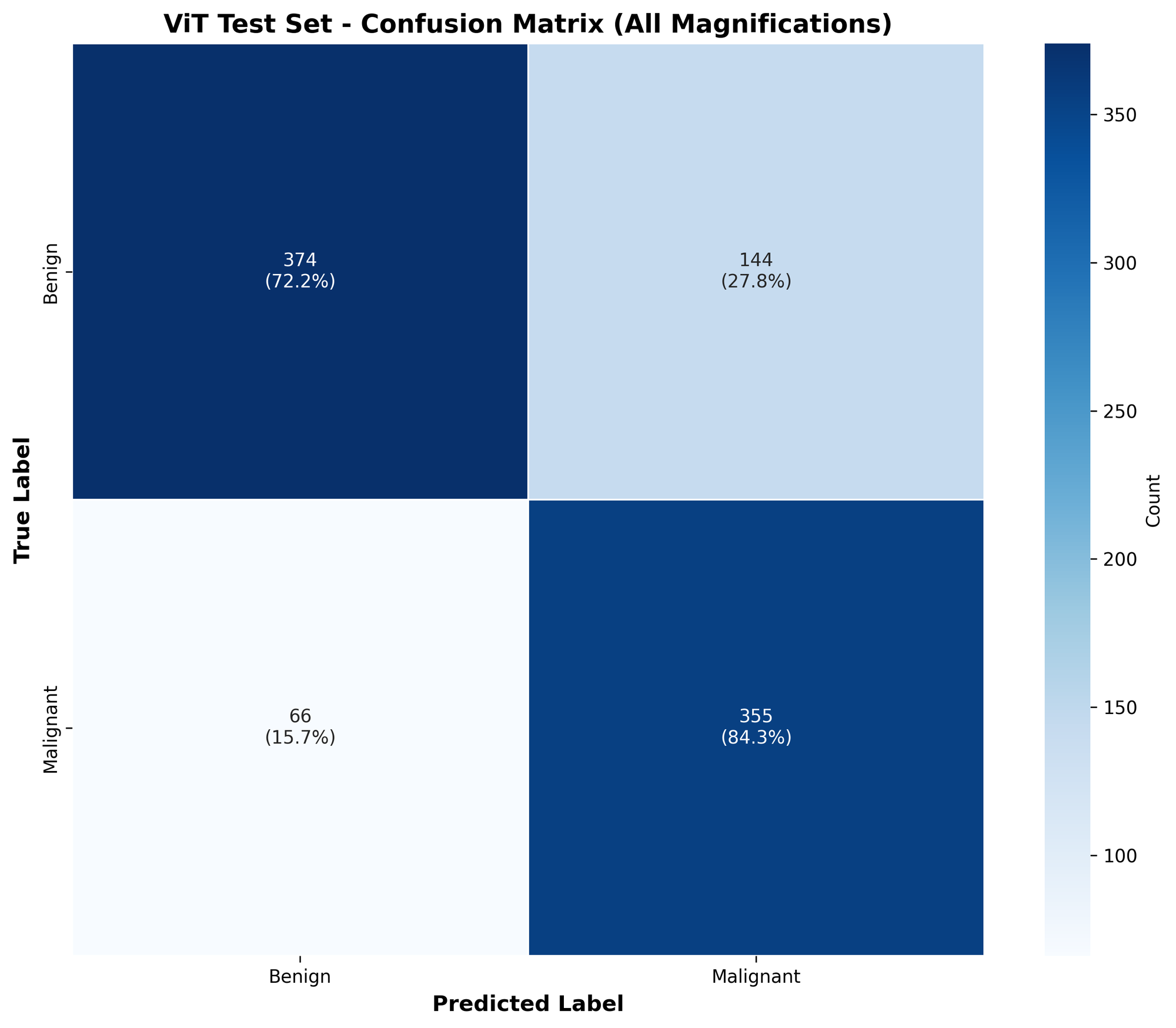}
}

\caption{Validation and test performance of the Vision Transformer (ViT) model on the BreakHis dataset across all magnifications.}
\label{fig:vit_val_test}
\end{figure}

\begin{figure}[H]
\centering

\subfloat[ROC 40X]{%
 \includegraphics[width=0.23\textwidth]{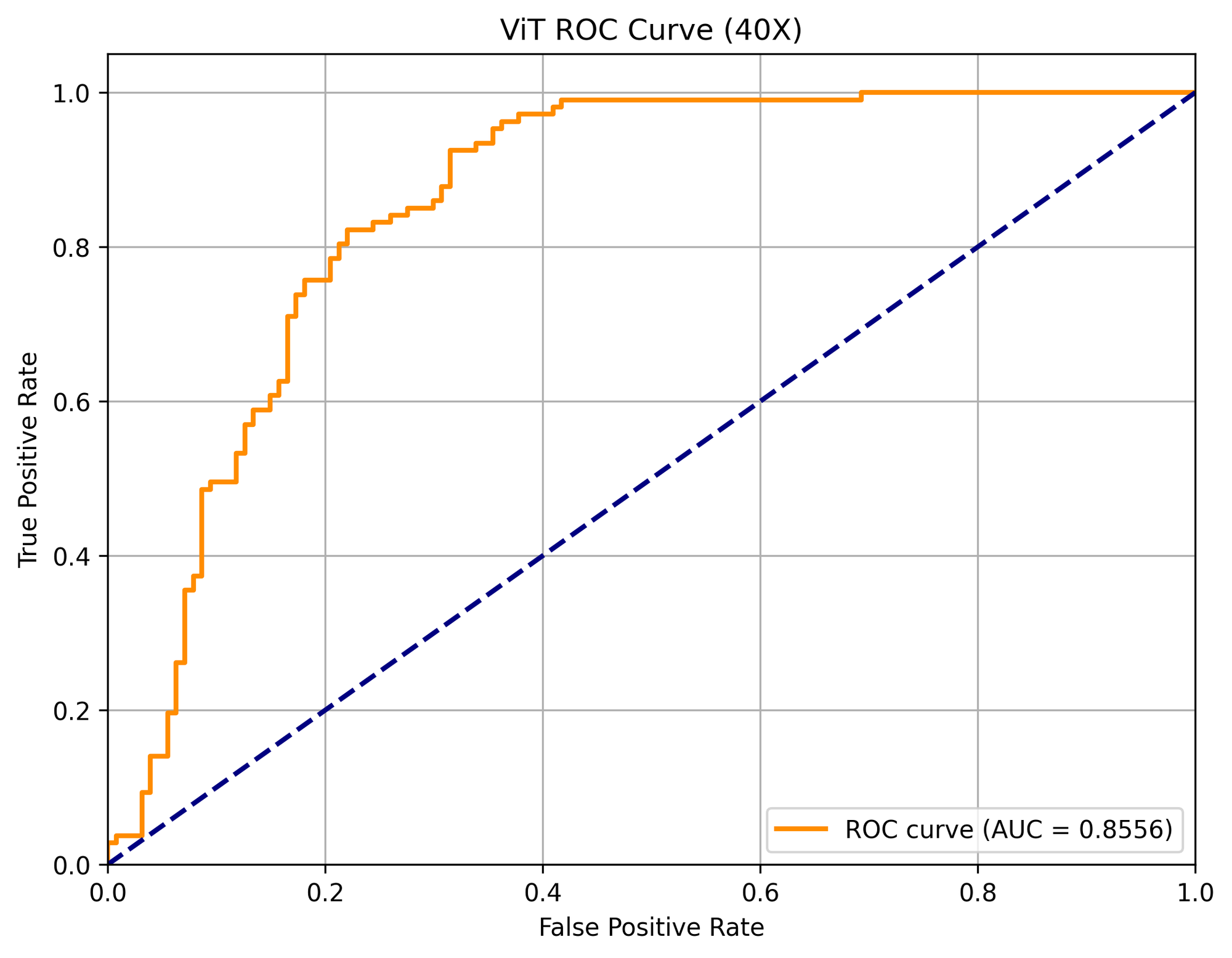}
}\hfill
\subfloat[ROC 100X]{%
 \includegraphics[width=0.23\textwidth]{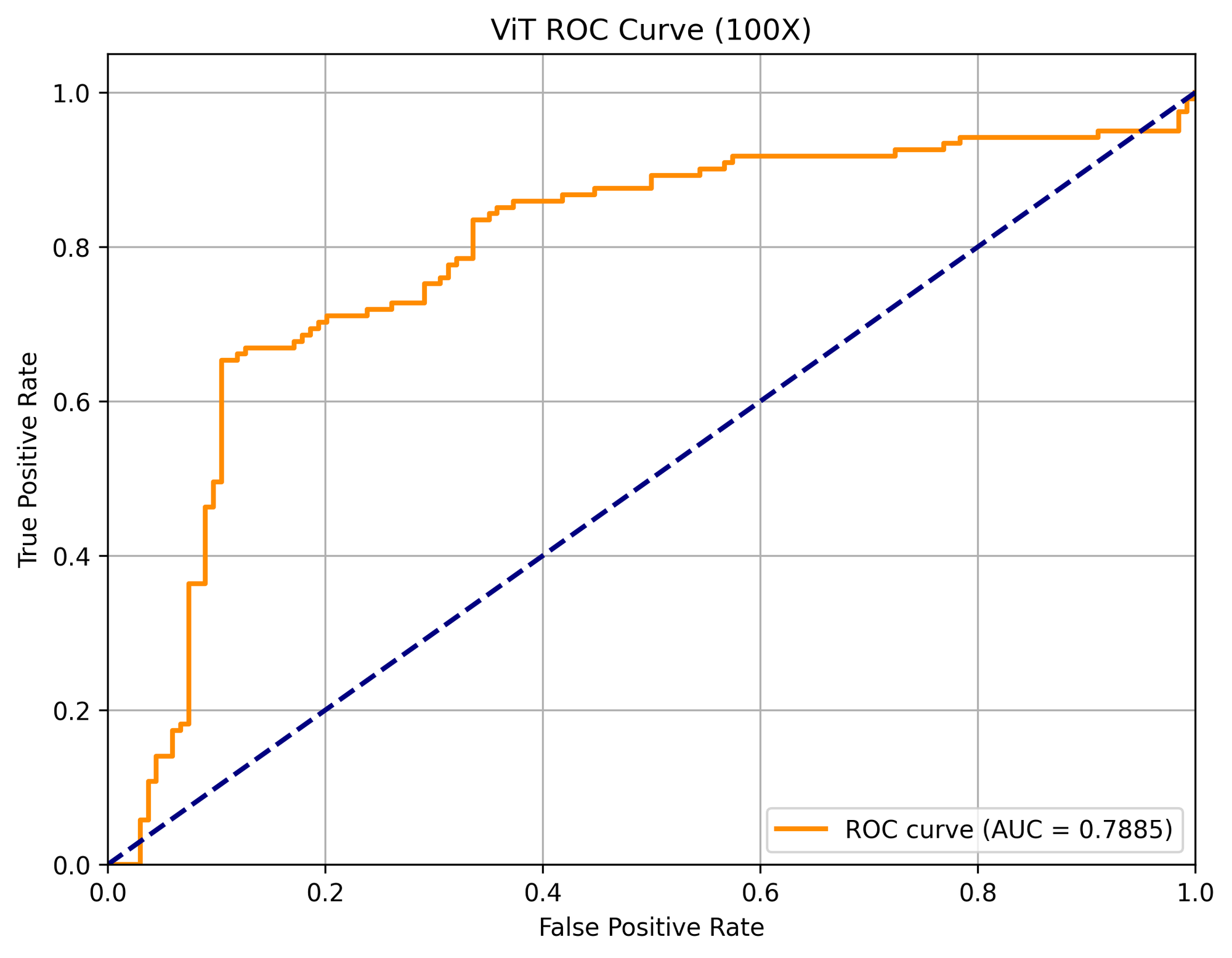}
}\hfill
\subfloat[ROC 200X]{%
 \includegraphics[width=0.23\textwidth]{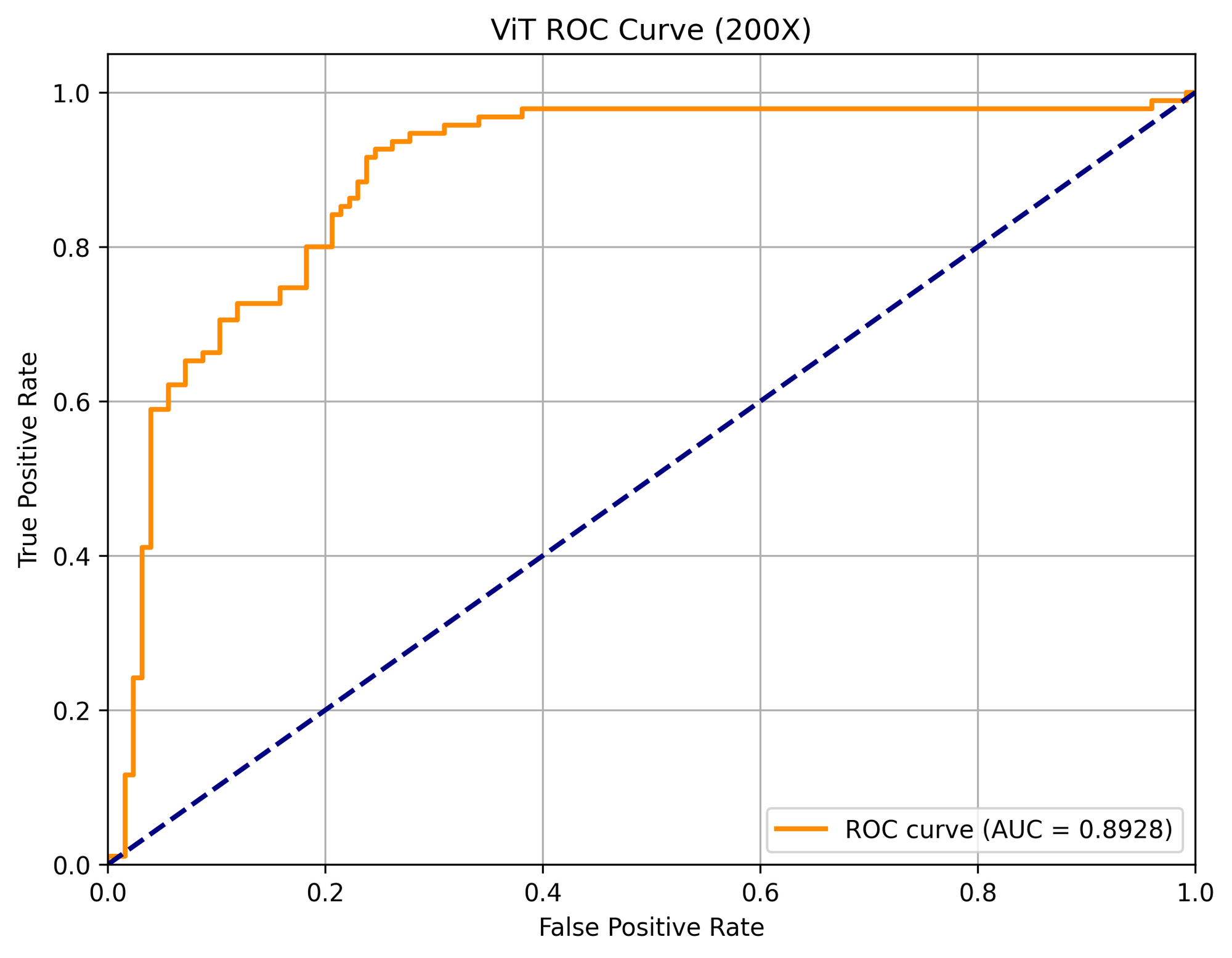}
}\hfill
\subfloat[ROC 400X]{%
 \includegraphics[width=0.23\textwidth]{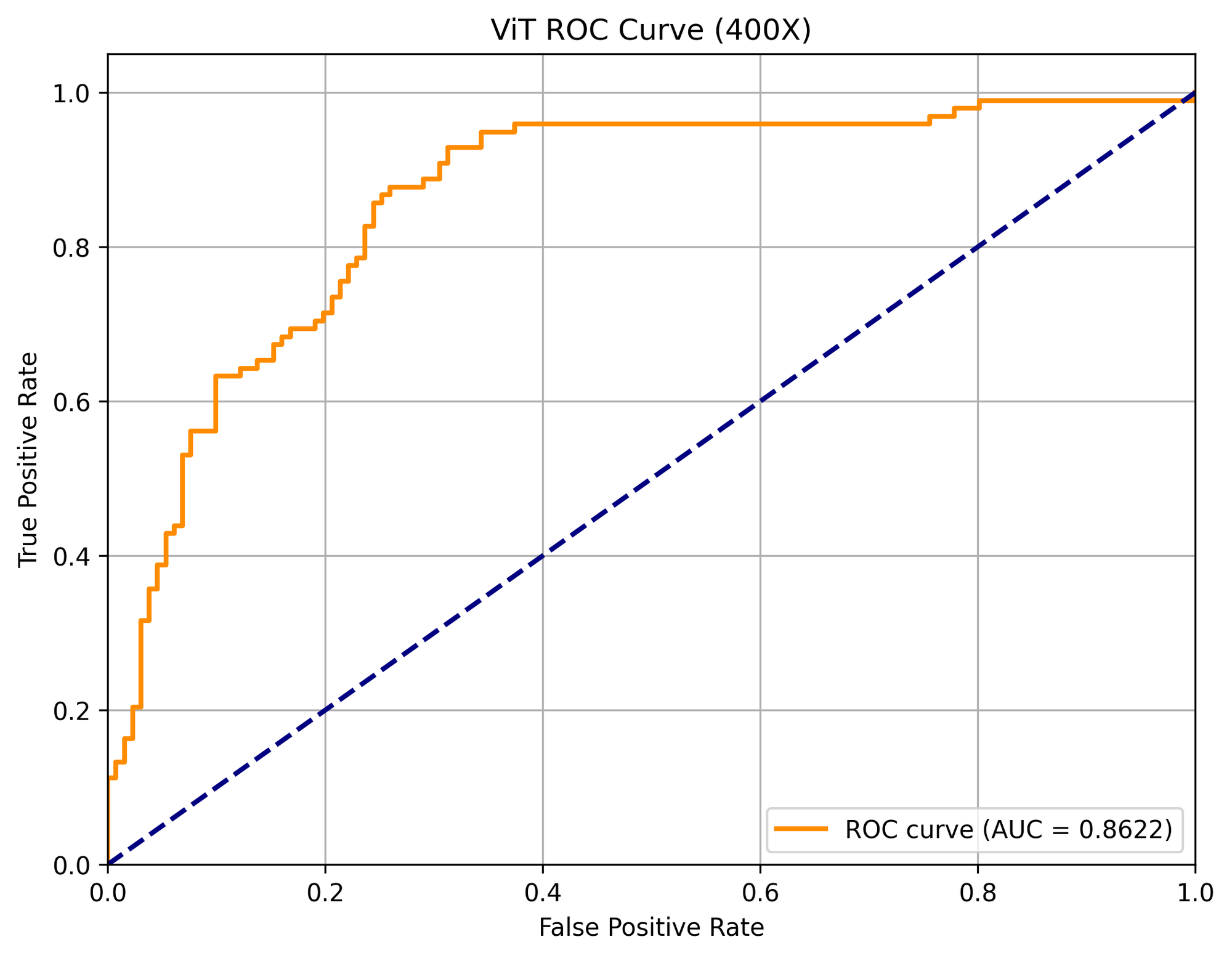}
}

\vspace{1ex}

\subfloat[PR 40X]{%
 \includegraphics[width=0.23\textwidth]{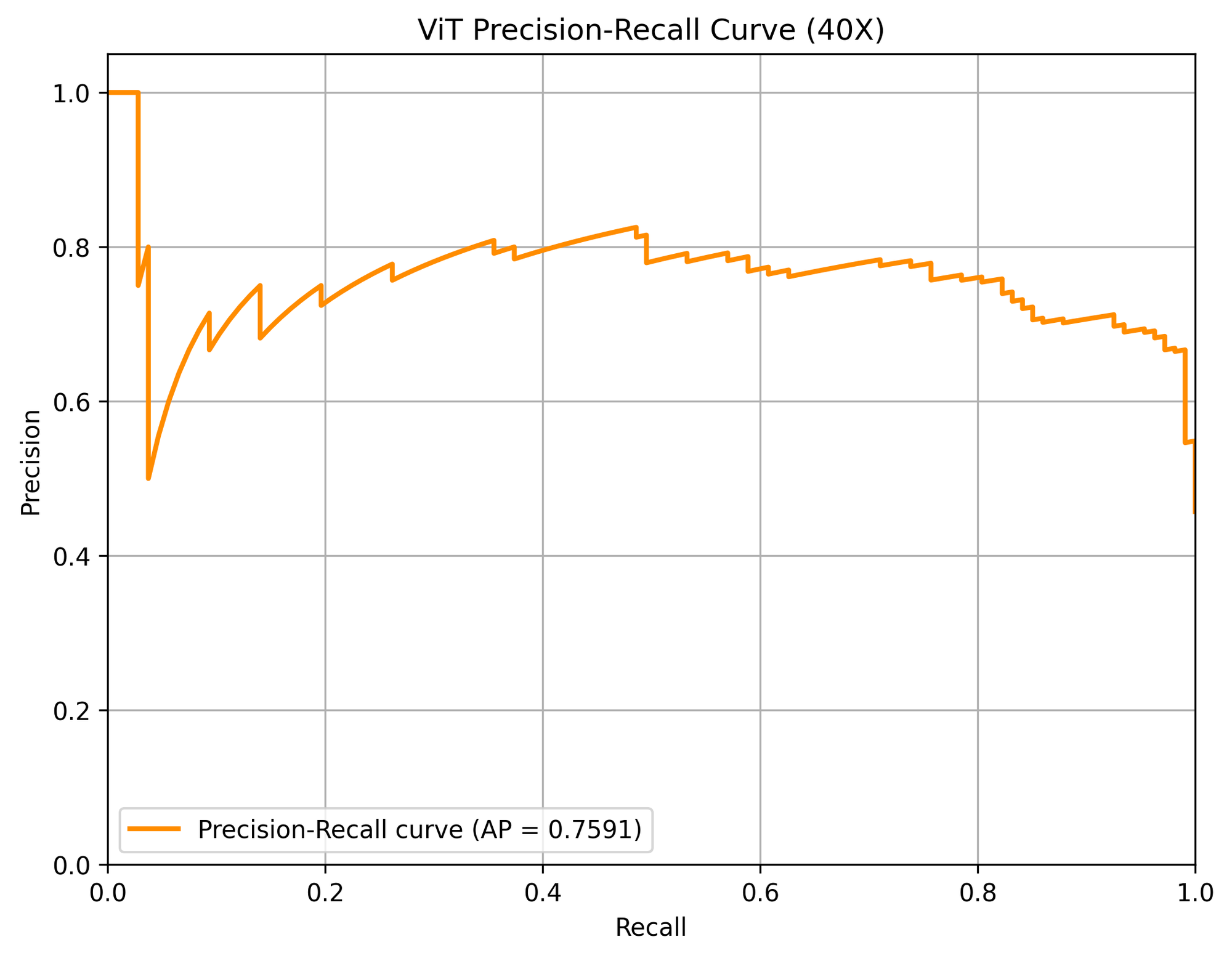}
}\hfill
\subfloat[PR 100X]{%
 \includegraphics[width=0.23\textwidth]{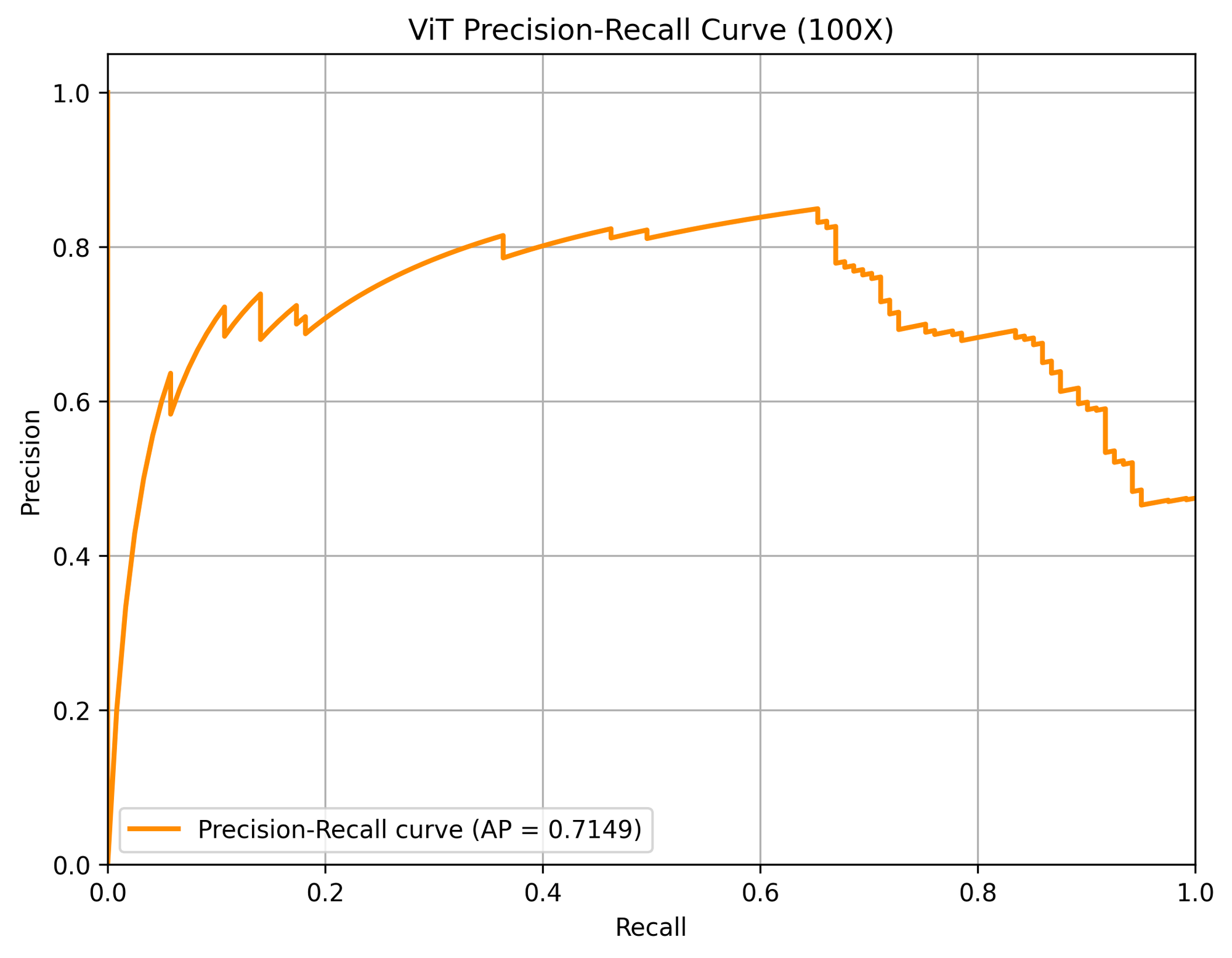}
}\hfill
\subfloat[PR 200X]{%
 \includegraphics[width=0.23\textwidth]{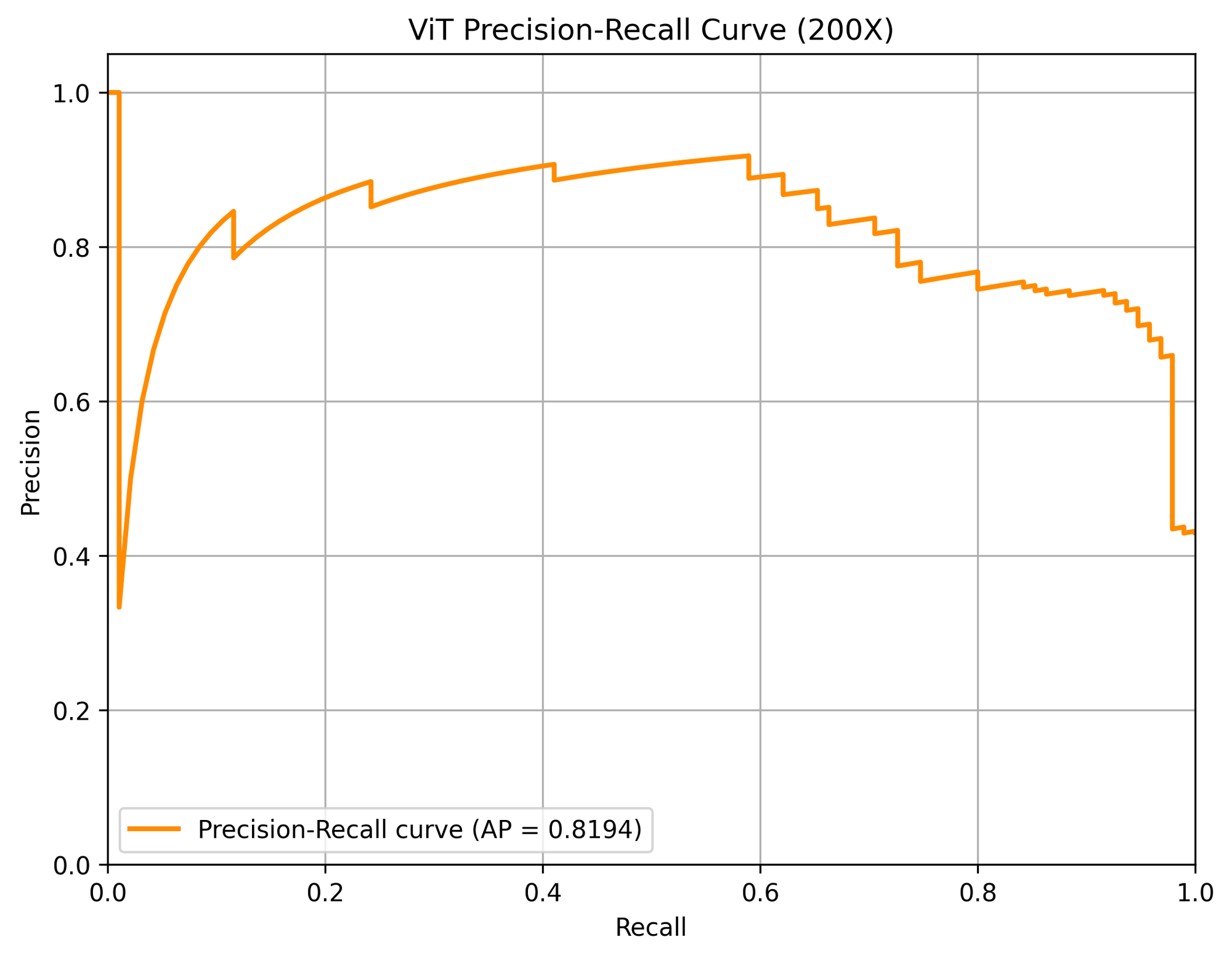}
}\hfill
\subfloat[PR 400X]{%
 \includegraphics[width=0.23\textwidth]{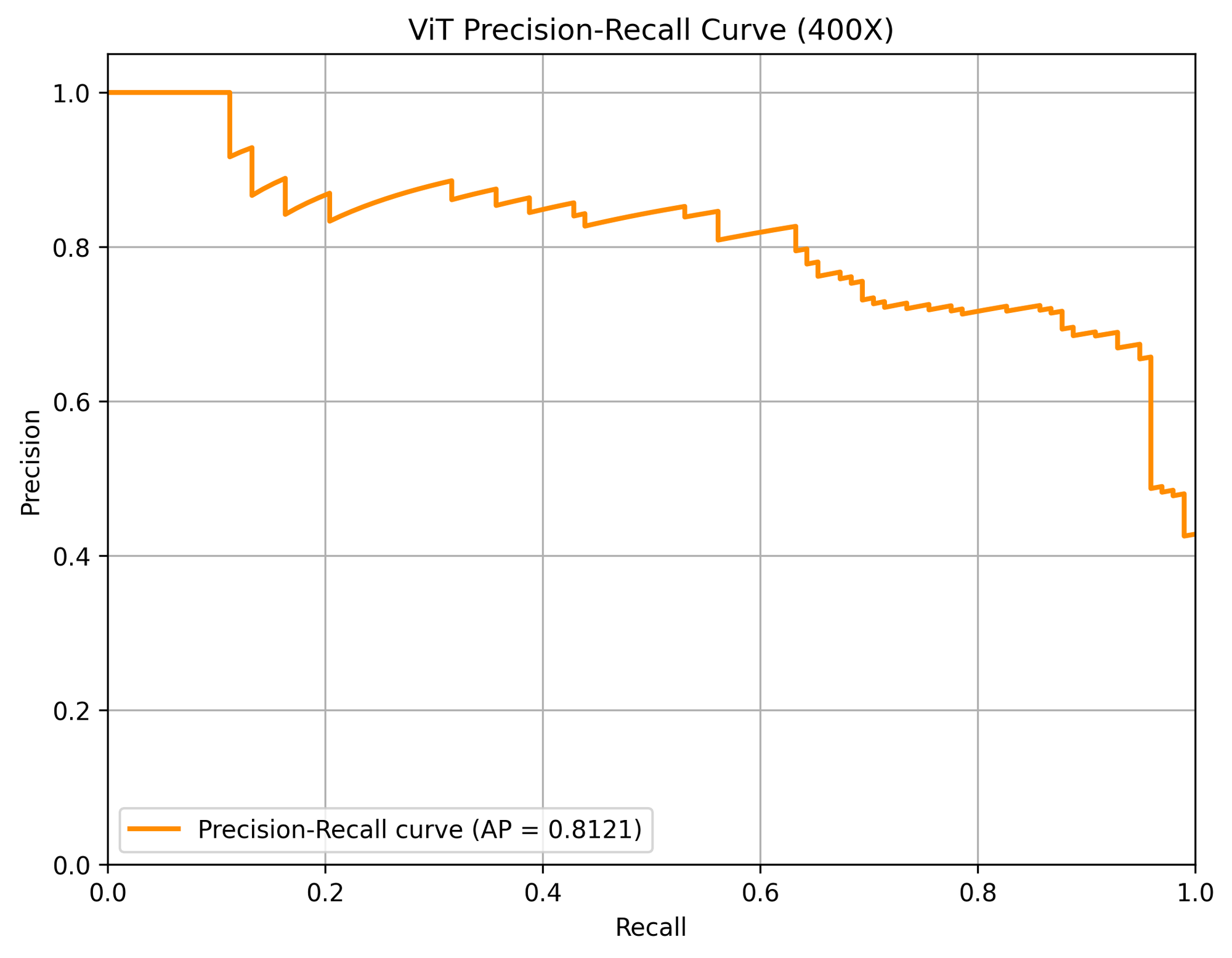}
}

\caption{Receiver Operating Characteristic (ROC) curves (a–d) and Precision–Recall (PR) curves (e–h) for the Vision Transformer (ViT) model evaluated on the BreakHis dataset across 40X, 100X, 200X, and 400X magnifications.}
\label{fig:vit_roc_pr}
\end{figure}

\begin{table}[ht]
\caption{Performance of Xception on the BreakHis dataset}
\label{tab:xception}
\centering
\begin{tabular}{@{}lcccccc@{}}
\toprule
\textbf{Dataset} 
& \textbf{Accuracy (\%)} 
& \textbf{Precision (\%)} 
& \textbf{Recall (\%)} 
& \textbf{F1-score (\%)} 
& \textbf{Specificity (\%)} 
& \textbf{AUC} \\
\midrule
Validation (All Mag.) 
& 94.5 
& 95.7 
& 96.9 
& 96.3 
& 88.3 
& 0.985 \\

Test (All Mag.)    
& 85.8 
& 80.0 
& 91.2 
& 85.2 
& 81.5 
& 0.939 \\

Test (40$\times$)      
& 87.2 
& 82.9 
& 90.7 
& 86.6 
& 84.3 
& 0.945 \\

Test (100$\times$)      
& 84.3 
& 80.9 
& 87.6 
& 84.1 
& 81.3 
& 0.935 \\

Test (200$\times$)      
& 87.3 
& 80.2 
& 93.7 
& 86.4 
& 82.5 
& 0.961 \\

Test (400$\times$)      
& 83.8 
& 74.0 
& 95.9 
& 83.6 
& 74.8 
& 0.928 \\
\bottomrule
\end{tabular}

\end{table}

The Xception model showed remarkable performance on the validation set, achieving 94.52\% accuracy, 95.65\% precision, 96.87\% recall, 88.25\% specificity, 96.25\% F1-score, and 0.9849 AUC-ROC, showing a robust sensitivity-specificity balance during training (Table 10). In the test set (across all magnifications), performance dropped to 85.84\% accuracy, 80.0\% precision, 91.21\% recall, 81.47\% specificity, 85.24\% F1-score, and 0.9392 AUC-ROC, indicating elevated recall but diminished precision attributable to false positives. Results were comparable across magnifications: 40X (87.18\% accuracy, 86.61\% F1, AUC 0.9452), 100X (84.31\% accuracy, 84.13\% F1, AUC 0.9352), 200X (87.33\% accuracy, 86.41\% F1, AUC 0.9609), and 400X (83.84\% accuracy, 83.56\% F1, AUC 0.9278) (Figure 36). Recall was consistently elevated ($\geq$.87\%, culminating at 93.68\% at 200X and 95.92\% at 400X), affirming Xception's efficacy in detecting malignant cases; nonetheless, its diminished precision throughout test sets underscores a trade-off between sensitivity and false-positive rates.

\begin{figure}[H]
 \centering
 \includegraphics[width=0.75\linewidth]{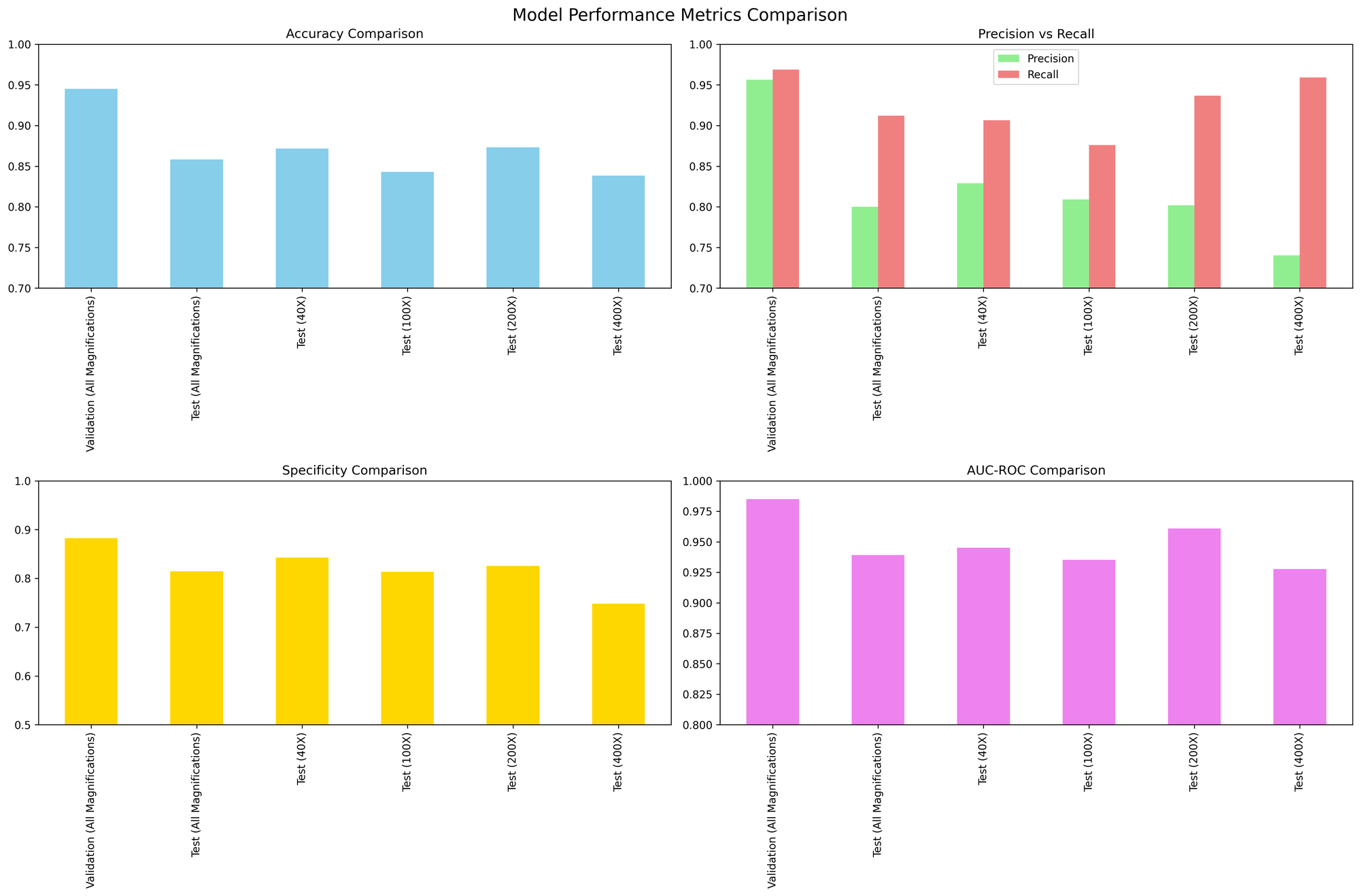}
 \caption{Performance metrics comparison of the Xception model across validation and test sets at multiple magnifications (40X, 100X, 200X, 400X).}
 \label{fig:single10}
\end{figure}

\begin{figure}[H]
 \centering
 \includegraphics[width=0.75\linewidth]{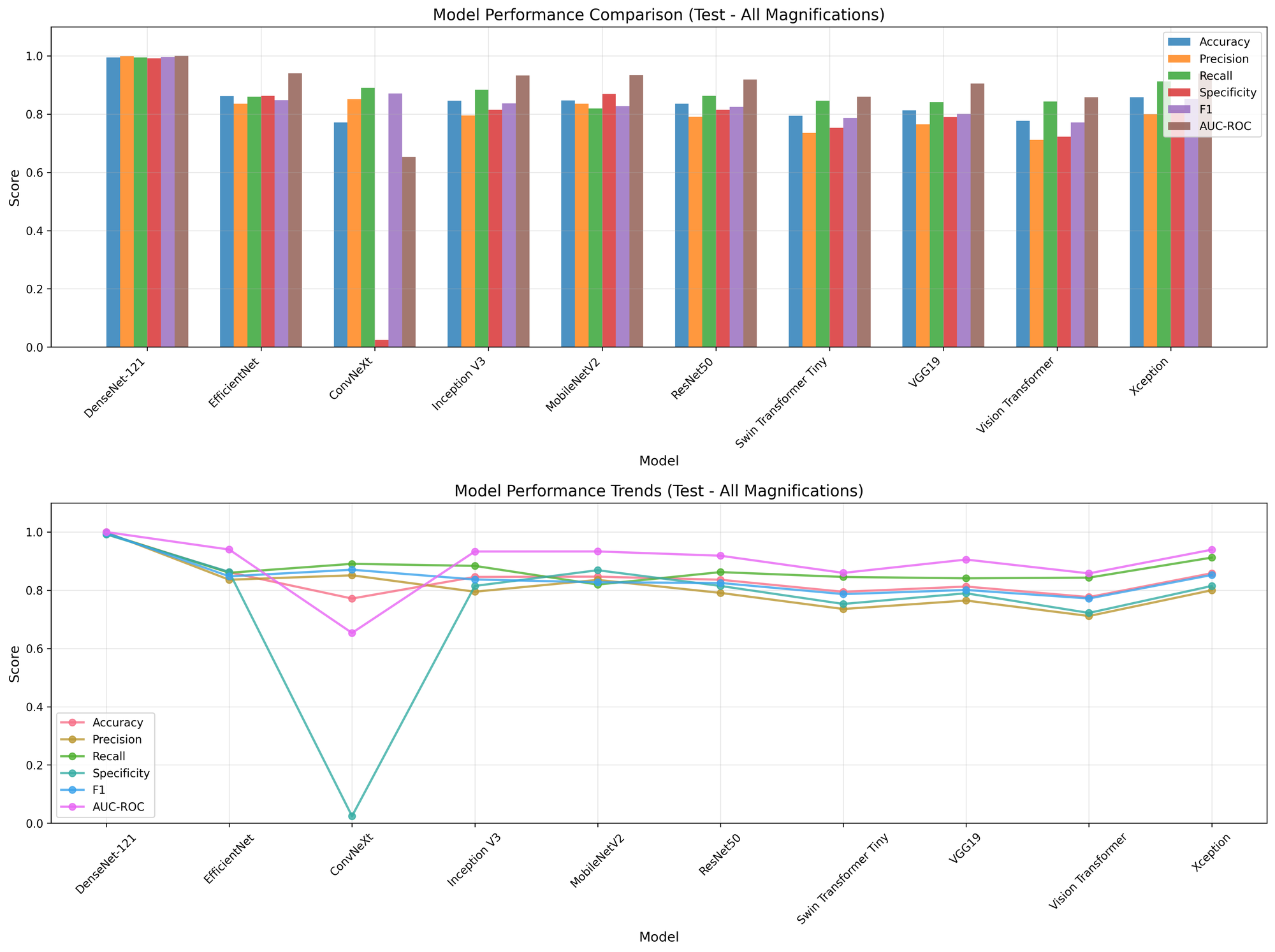}
 \caption{Comparison of ten deep learning models on the test set (all magnifications) across accuracy, precision, recall, specificity, F1, and AUC-ROC.}
 \label{fig:single11}
\end{figure}

\begin{figure}[H]
 \centering
 \includegraphics[width=0.75\linewidth]{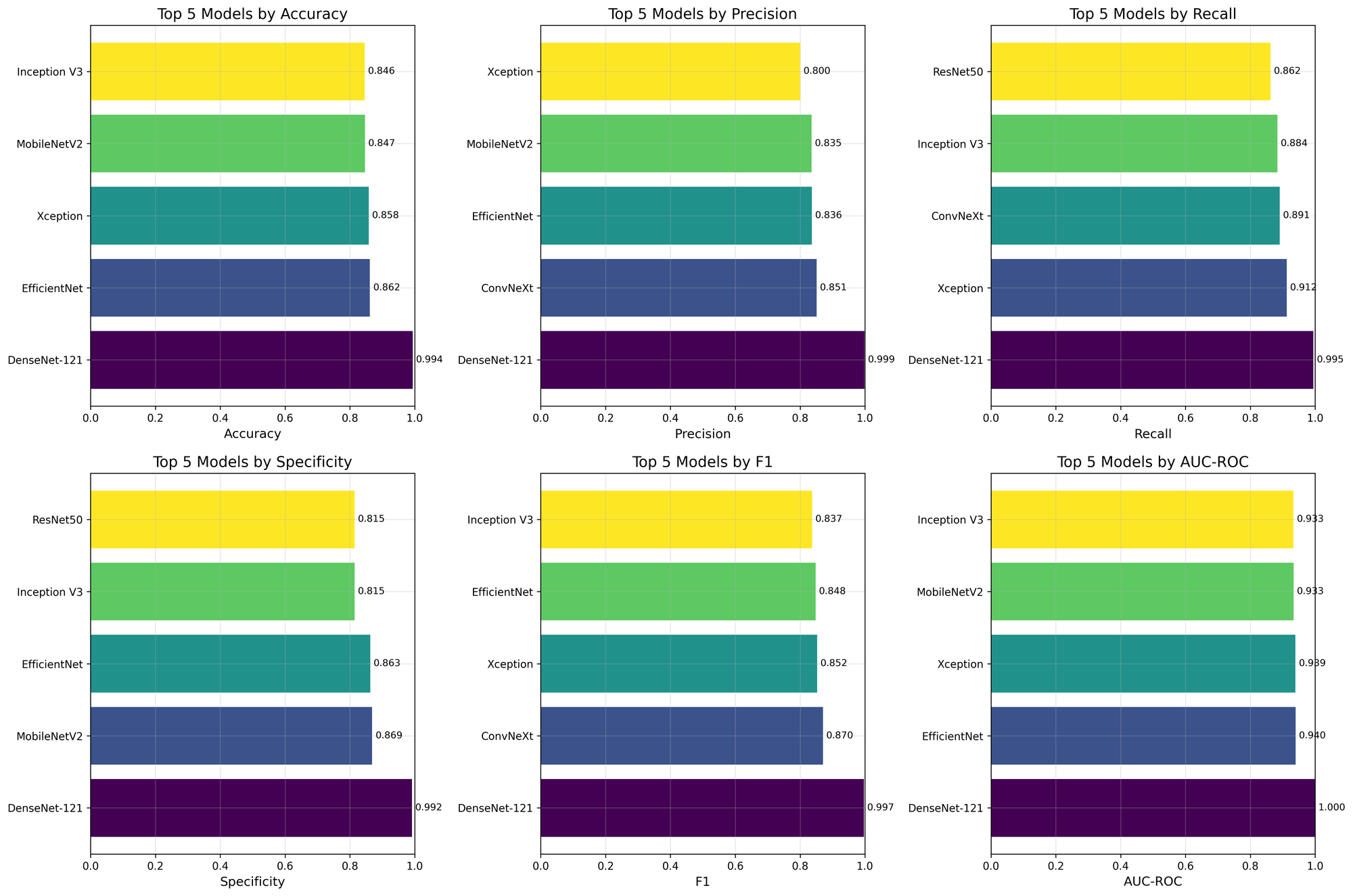}
 \caption{Top-5 performing models across evaluation metrics on the test set (all magnifications).}
 \label{fig:single12}
\end{figure}

Among all models, DenseNet-121 showed outstanding results, attaining near-perfect generalization on the BreakHis test set with an accuracy of 99.41\%, precision of 99.86\%, recall of 99.45\%, specificity of 99.17\%, F1-score of 99.65\%, and an AUC-ROC of 0.9997, while achieving 100\% performance at 40X, 100X, and 200X magnifications. Other CNN-based models achieved competitive performance: Xception showed impressive performance with an accuracy of 85.84\%, a recall of 91.21\%, and an AUC-ROC of 0.9392, showing superior sensitivity across magnifications (recall $\geq$.90\% at 200X and 400X) (Figure 37). Inception V3 and ResNet50 demonstrated comparable performance (accuracy ~84–86\%, AUC ~0.92–0.94) with uniform recall across magnifications, albeit with marginally reduced precision, suggesting a propensity for false positives. EfficientNet and MobileNetV2 preserved efficiency while yielding consistent outcomes (accuracy approximately 84–86\%, AUC approximately 0.93–0.95), albeit with some variability in recall contingent upon magnification (Figure 38). VGG19, despite exhibiting robust validation performance (93.12\% accuracy, 0.975 AUC), demonstrated inferior generalization on test sets (about 81\% accuracy, AUC around 0.91), highlighting the constraints of its antiquated sequential design. Conversely, transformer-based models like the Vision Transformer (ViT) and Swin Transformer Tiny exhibited inferior performance relative to CNNs, achieving test accuracies of approximately 77–79\% and lower AUC values (~0.85–0.89). Nevertheless, they demonstrated relatively high recall (>83\%) across various magnifications, indicating potential in malignant case detection, albeit with diminished precision and specificity. ConvNeXt showed the lowest performance overall, achieving a test accuracy of 77.14\% and an exceedingly low specificity of 2.42\%, despite a high recall rate of 89\%, which suggests a significant bias towards malignant diagnoses.

The results show that DenseNet-121 had the most robust and generalizable performance, whilst Xception, Inception V3, and EfficientNet achieved a commendable balance between accuracy and sensitivity. Transformer-based models showed potential for high recall but encountered difficulties with precision and generalization in histopathology data. These findings underscore the advantages of densely connected CNNs in breast cancer histopathology image classification, while emphasizing the necessity for additional improvement of vision transformers to address domain-specific heterogeneity.

\section{Explainability}
The DenseNet121 model achieved outstanding performance on the BreakHis dataset (test accuracy: 0.994, F1 score: 0.997, AUC-ROC: 1.0), with prediction confidence consistently above 0.98. To interpret its decisions, multiple XAI techniques were employed. Grad-CAM revealed progressive localization of discriminative regions, with deeper layers focusing on nuclei and glandular structures. Integrated Gradients consistently highlighted nuclei-dense areas and boundaries across different baselines, ensuring attribution robustness (Figure 39). LIME identified superpixel-level regions supporting benign predictions, while LRP decomposed evidence into positive contributions from epithelial and nuclei structures and negative relevance in background areas. Saliency maps further confirmed fine-grained sensitivity to relevant tissue features. Collectively, these results demonstrate that DenseNet121 not only achieves state-of-the-art performance but also grounds its predictions in clinically meaningful histopathological patterns.

\begin{figure}[H]
\centering

\subfloat[Grad-CAM]{%
 \includegraphics[width=0.45\textwidth]{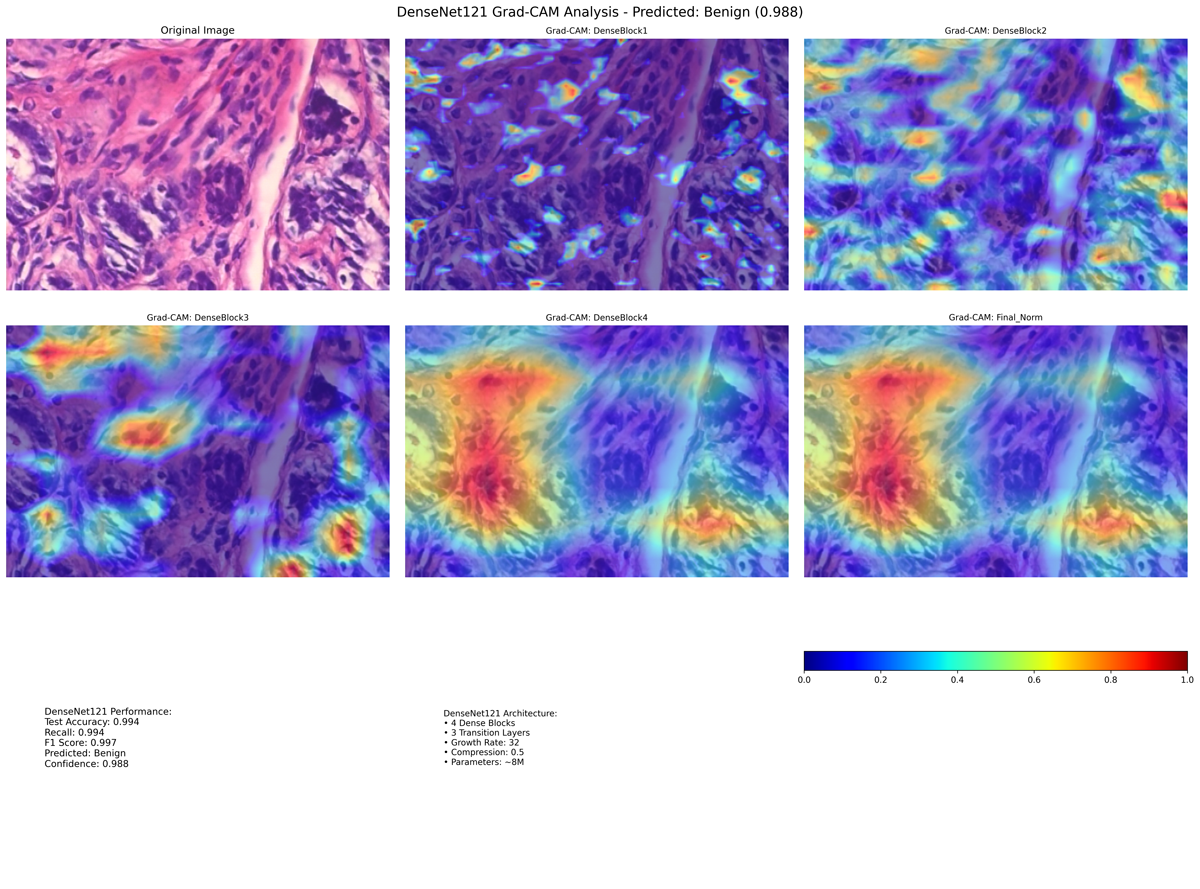}
}\hfill
\subfloat[Integrated Gradients]{%
 \includegraphics[width=0.45\textwidth]{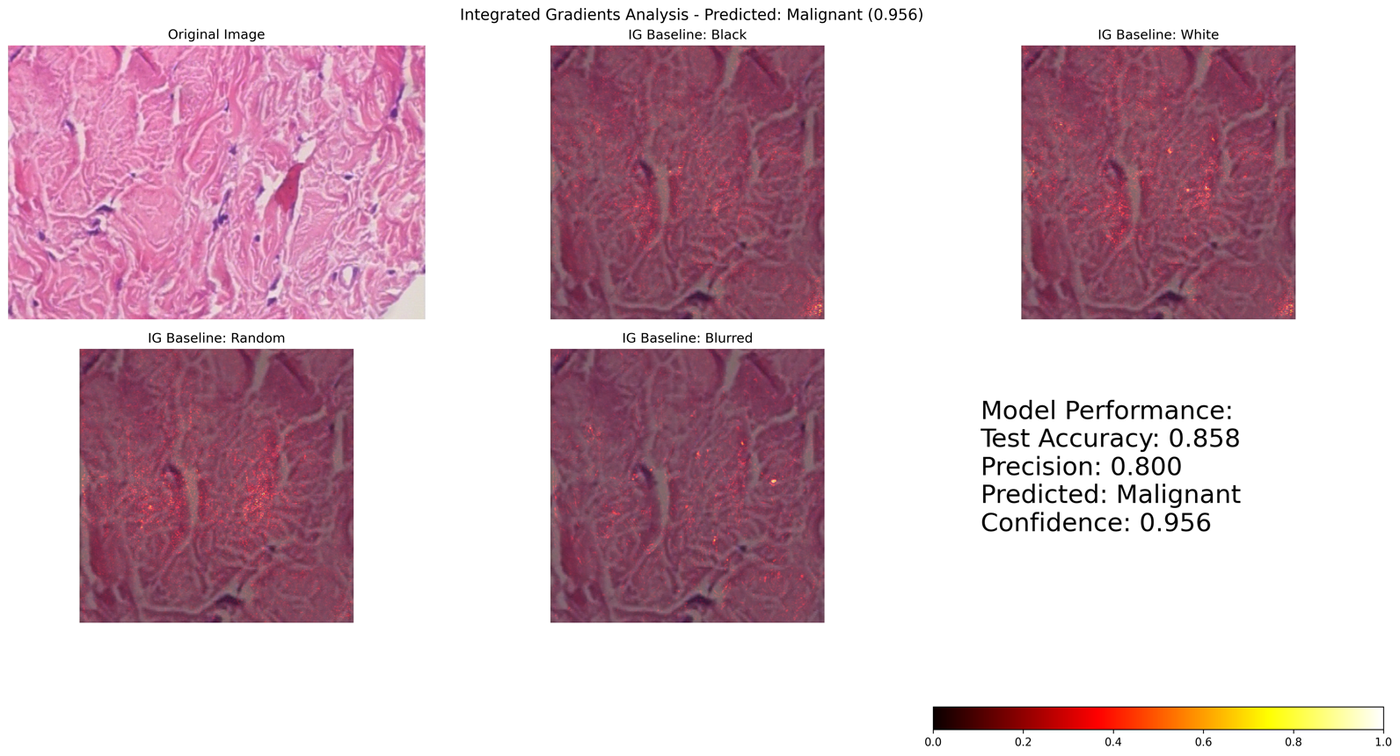}
}

\vspace{0.4cm}

\subfloat[LIME]{%
 \includegraphics[width=0.45\textwidth]{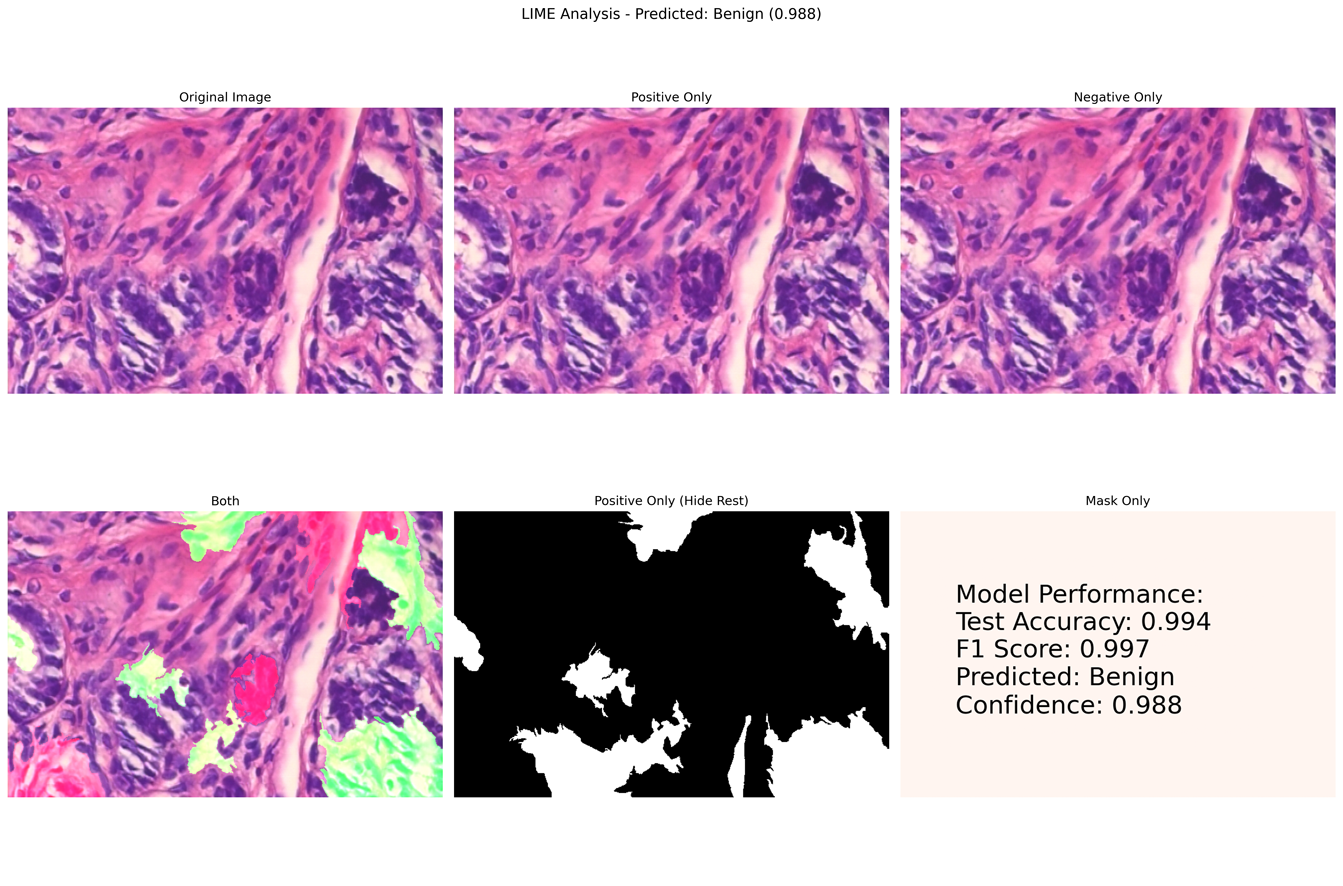}
}\hfill
\subfloat[LRP]{%
 \includegraphics[width=0.45\textwidth]{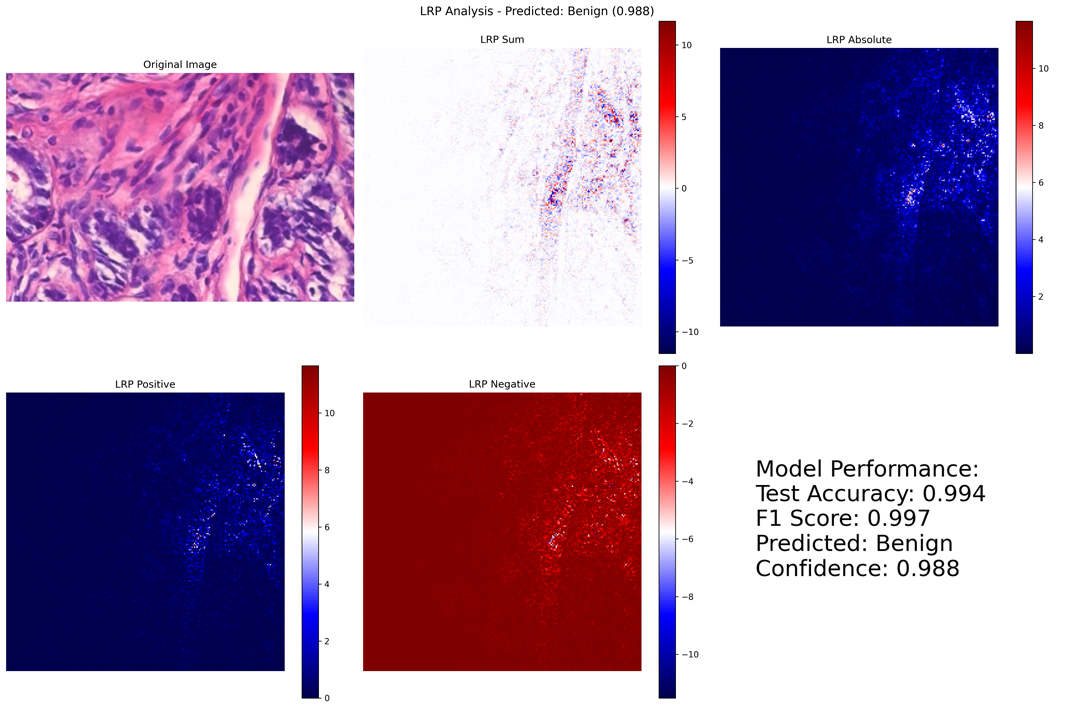}
}

\caption{Explainable AI (XAI) visualizations of DenseNet121 predictions on the BreakHis dataset using (a) Grad-CAM, (b) Integrated Gradients, (c) LIME, and (d) Layer-wise Relevance Propagation (LRP).}
\label{fig:densenet_xai}
\end{figure}

The EfficientNet model, trained on the BreakHis dataset, attained a test accuracy of 0.862, recall of 0.860, precision of 0.836, F1-score of 0.848, and AUC-ROC of 0.940, with benign predictions exhibiting high confidence levels (0.956–0.969) (Figure 40). Grad-CAM representations at layers 2, 4, 6, and 8 demonstrated an increasingly finer concentration on nuclei-rich and glandular structures, whereas Integrated Gradients consistently highlighted epithelial boundaries and clustered nuclei across various baselines. LIME validated localized decision boundaries via superpixel-level attributions, while Saliency Maps emphasized pixel-level gradients indicative of tissue shape. The XAI approaches collectively demonstrate that EfficientNet, while slightly less accurate than DenseNet121, relies on clinically significant histopathological traits, hence ensuring interpretability and robust performance.

\begin{figure}[H]
\centering

\subfloat[Grad-CAM]{%
 \includegraphics[width=0.45\textwidth]{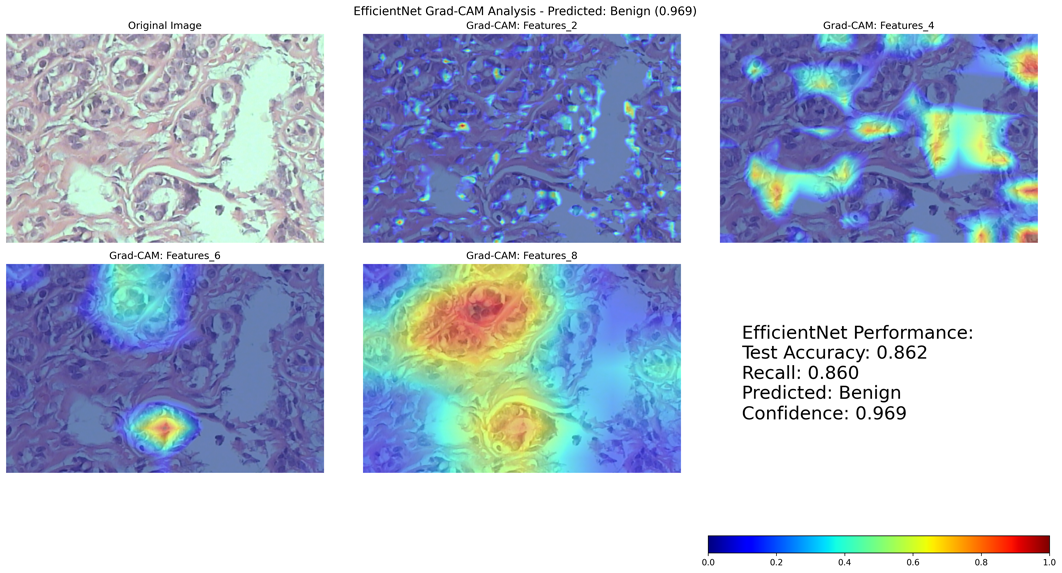}
}\hfill
\subfloat[Integrated Gradients]{%
 \includegraphics[width=0.45\textwidth]{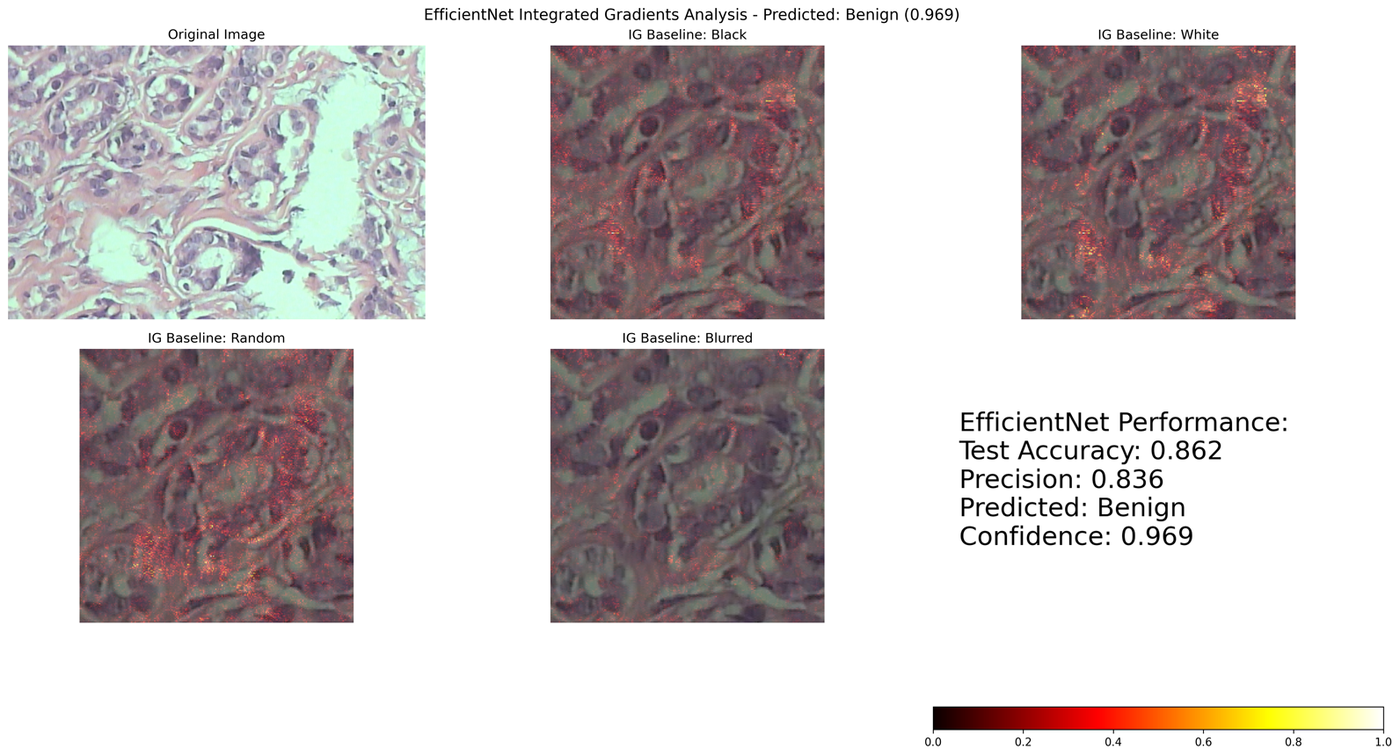}
}

\vspace{0.4cm}

\subfloat[LIME]{%
 \includegraphics[width=0.45\textwidth]{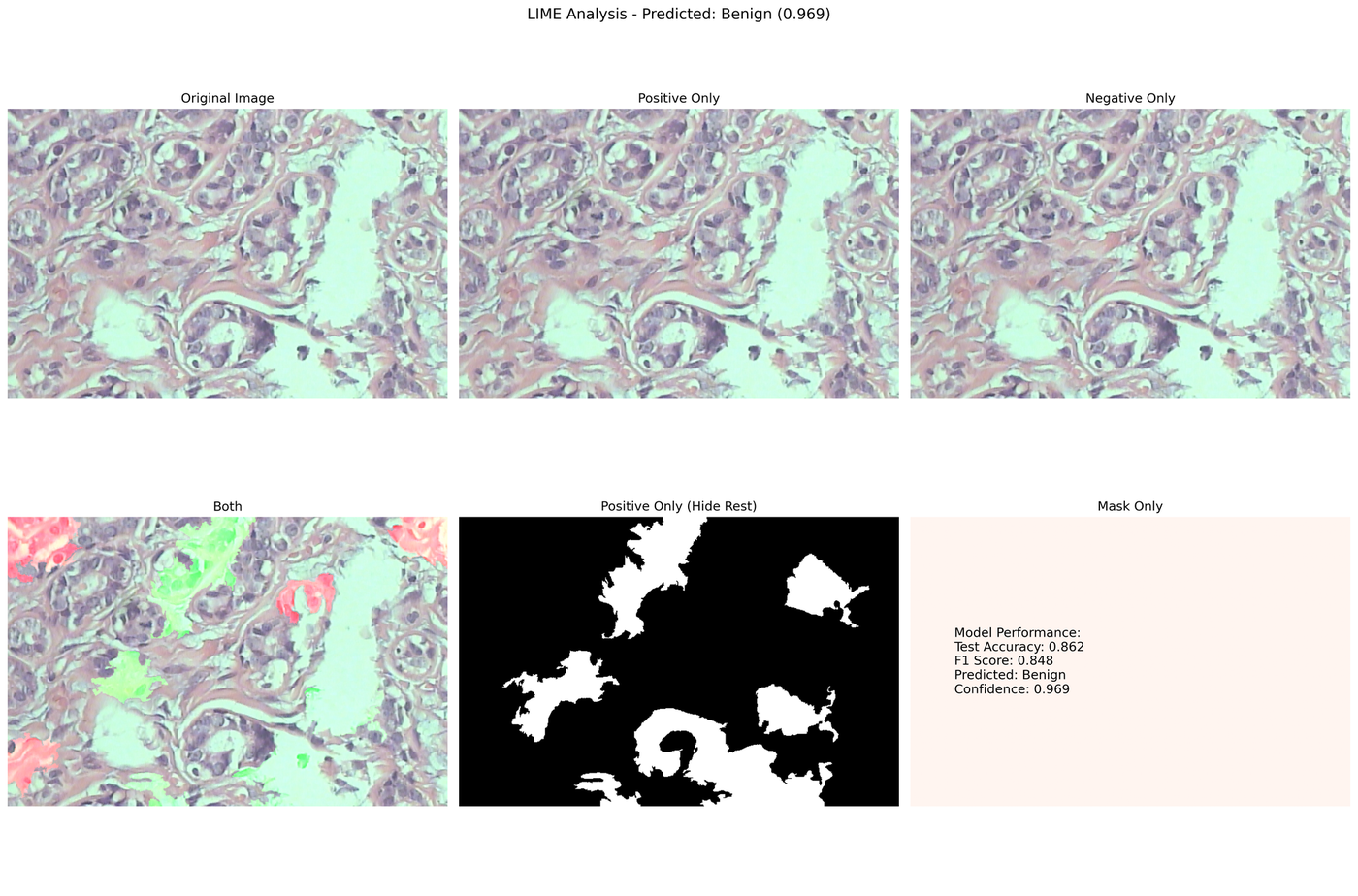}
}\hfill
\subfloat[Saliency Map]{%
 \includegraphics[width=0.45\textwidth]{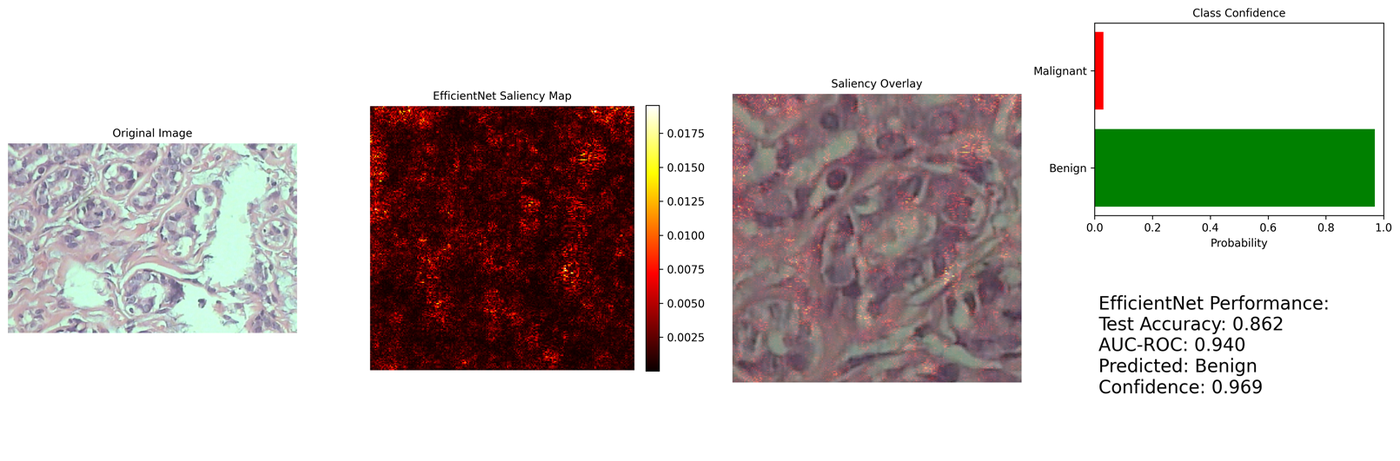}
}

\caption{XAI visualizations of EfficientNet predictions on the BreakHis dataset using (a) Grad-CAM, (b) Integrated Gradients, (c) LIME, and (d) Saliency Maps.}
\label{fig:efficientnet_xai}
\end{figure}

The InceptionV3 model, trained on the BreakHis dataset, attained a test accuracy of 0.846, recall of 0.884, precision of 0.795, and an F1-score of 0.837, with malignant predictions exhibiting confidence levels ranging from 0.775 to 0.851. Grad-CAM indicated a transition from textural patterns in first layers to nuclei-dense areas in deeper modules (Mixed5d–Mixed7c), whereas Integrated Gradients consistently underscored clustered nuclei and cell boundaries across various baselines (Figure 41). LIME detected superpixel regions associated with malignant forms compared to suppressed backgrounds, while LRP further analysed positive attributions to nuclear clusters and epithelial anomalies, removing irrelevant tissue. These XAI approaches collectively demonstrate that InceptionV3 bases its malignant predictions on clinically significant histological characteristics, hence ensuring interpretability while maintaining robust performance.

\begin{figure}[H]
\centering

\subfloat[Grad-CAM]{%
 \includegraphics[width=0.45\textwidth]{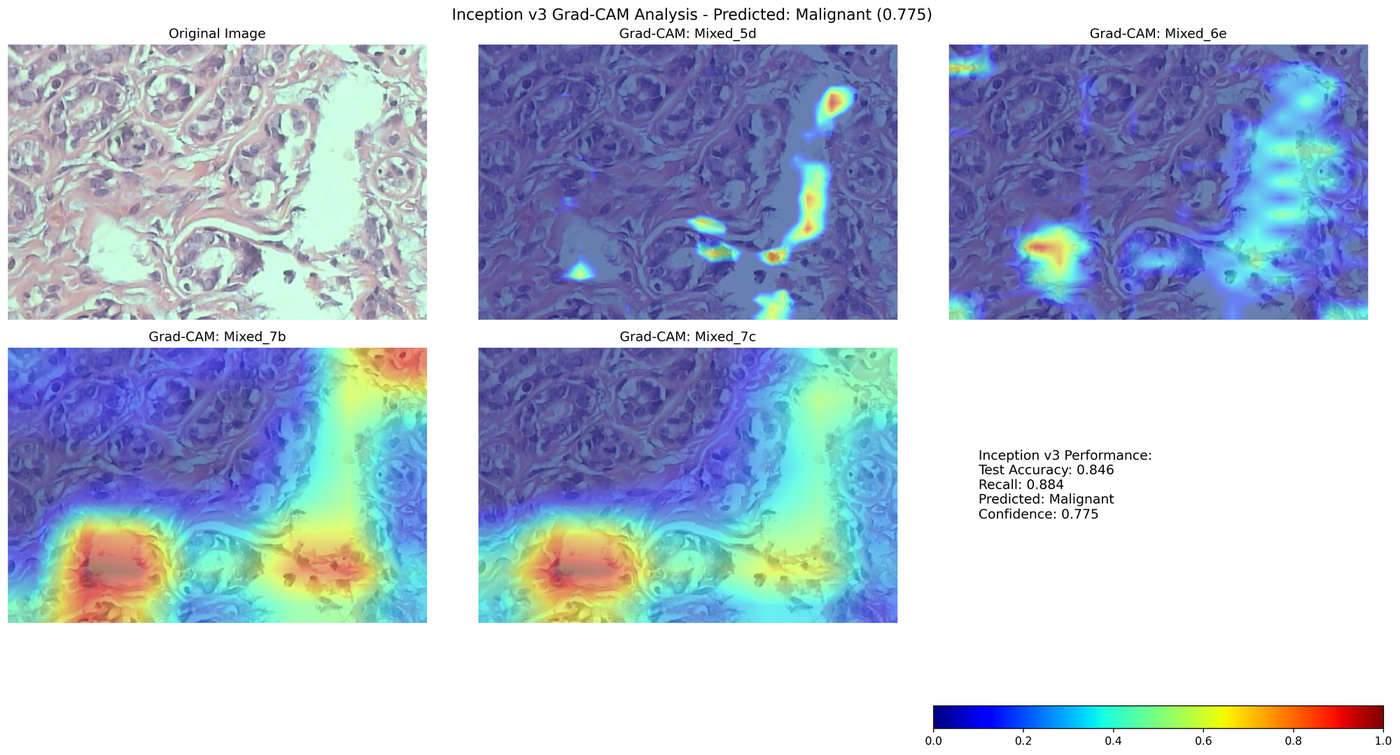}
}\hfill
\subfloat[Integrated Gradients]{%
 \includegraphics[width=0.45\textwidth]{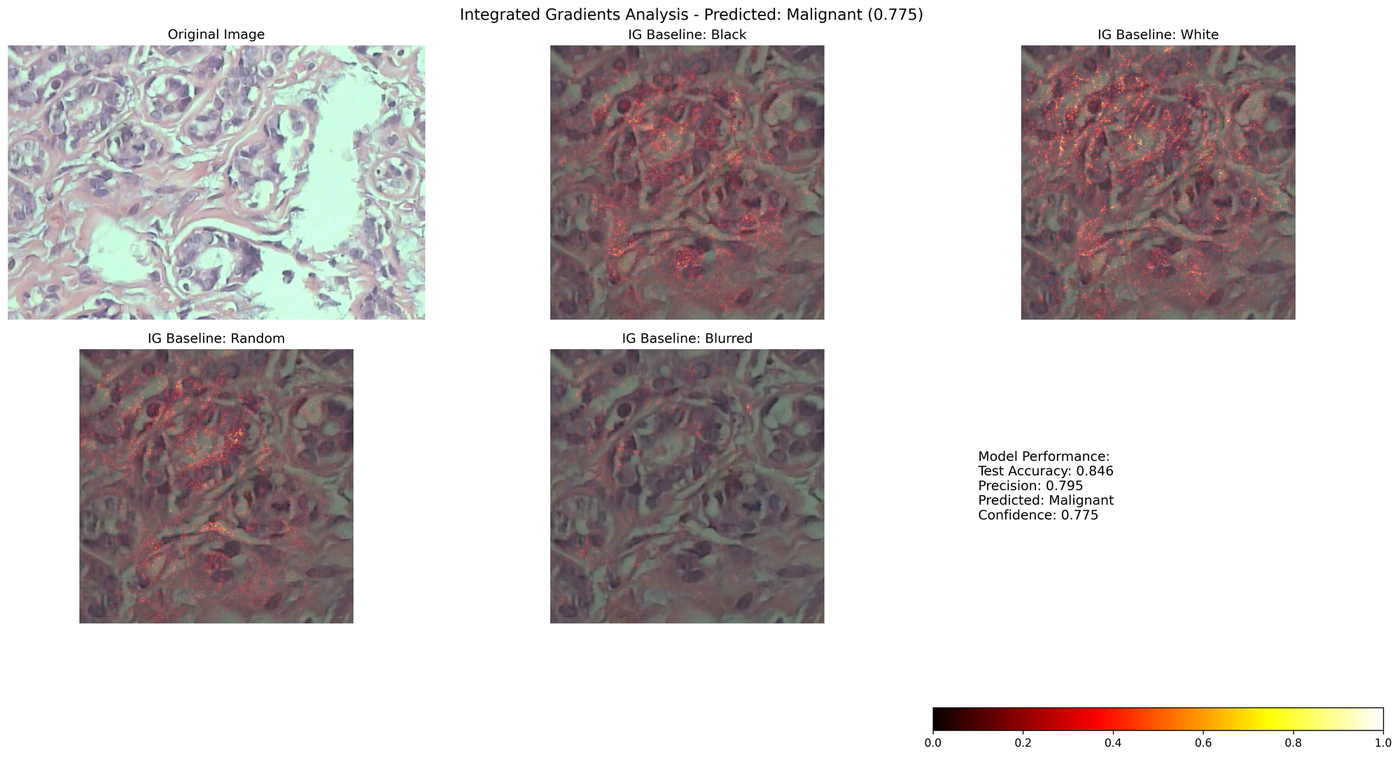}
}

\vspace{0.4cm}

\subfloat[LIME]{%
 \includegraphics[width=0.45\textwidth]{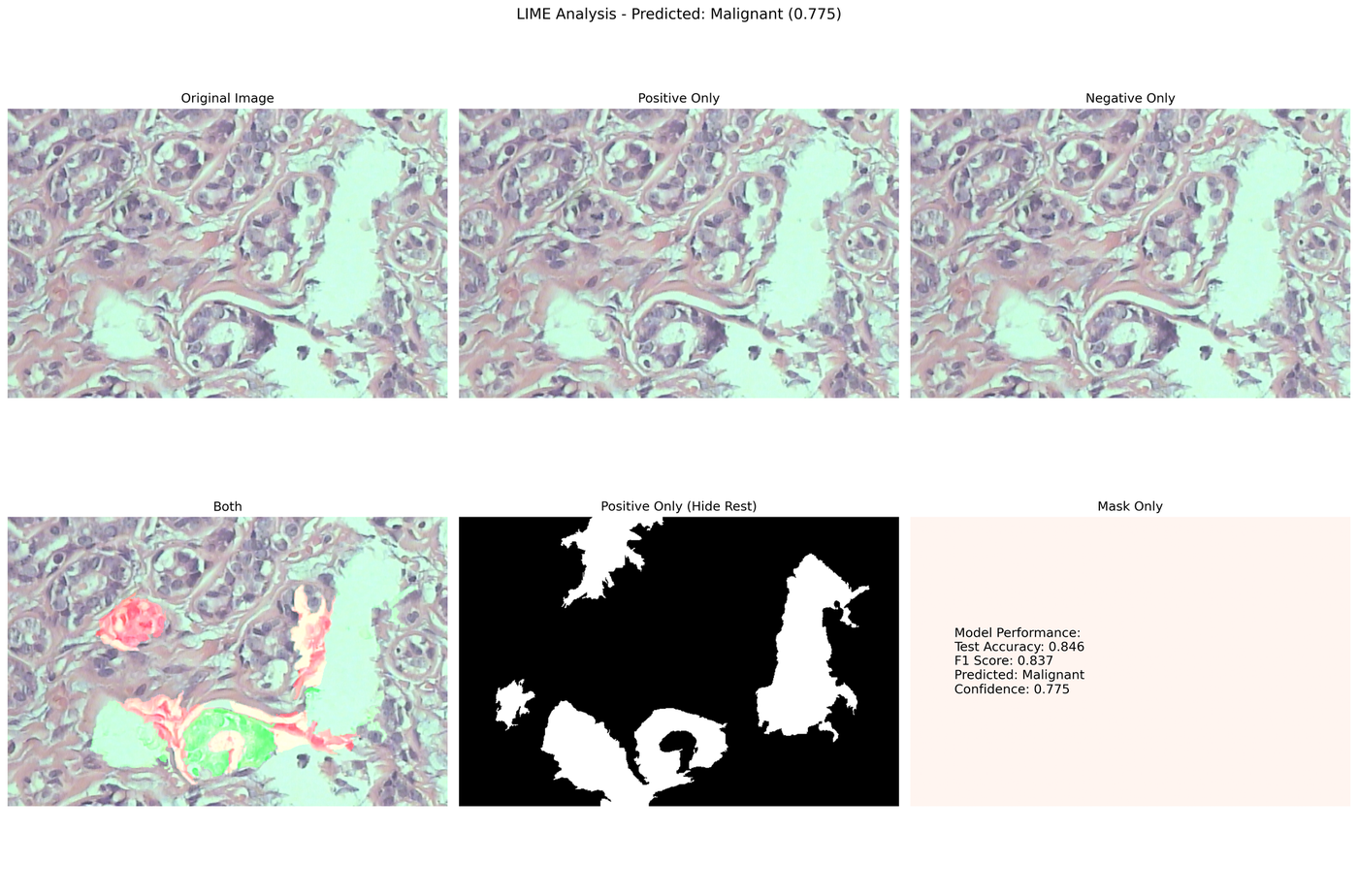}
}\hfill
\subfloat[LRP]{%
 \includegraphics[width=0.45\textwidth]{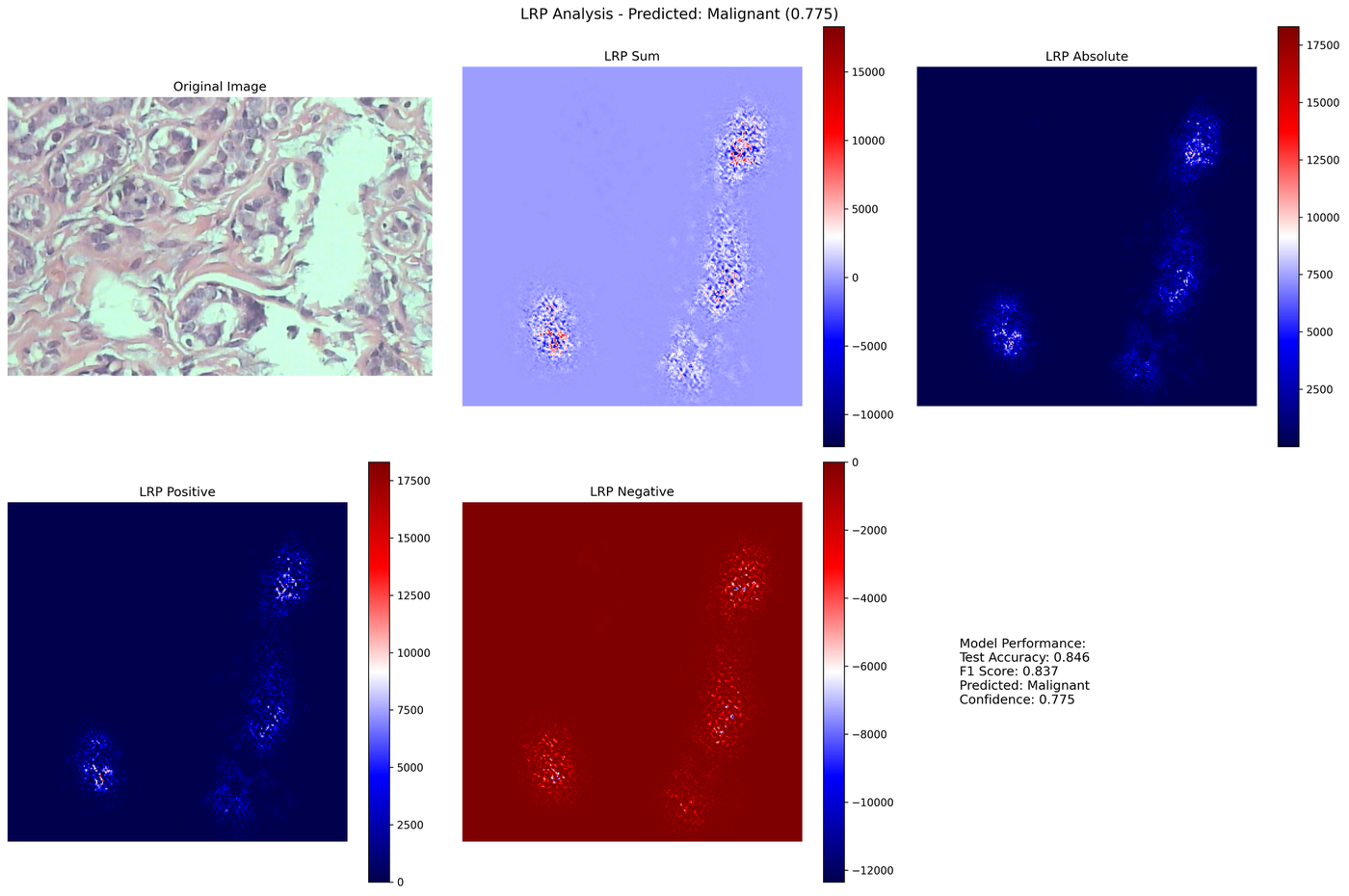}
}

\caption{XAI visualizations of InceptionV3 predictions on the BreakHis dataset using (a) Grad-CAM, (b) Integrated Gradients, (c) LIME, and (d) Layer-wise Relevance Propagation (LRP).}
\label{fig:inception_xai}
\end{figure}

The MobileNetV2 model, trained on the BreakHis dataset, attained a test accuracy of 0.847, precision of 0.835, recall of 0.819, and an F1-score of 0.827, with benign predictions exhibiting high confidence levels (0.952–0.953). Grad-CAM demonstrated a gradual localization of features from textures to regions packed with nuclei, whereas Integrated Gradients validated consistent attributions to nuclear boundaries across various baselines (Figure 42). LIME emphasized superpixel-level positive (green) and negative (red) areas, while LRP dissected predictions into clinically pertinent positive nuclear attributions and background exclusions. The XAI techniques collectively validated MobileNetV2's dependence on diagnostically relevant histopathological features, ensuring interpretability while maintaining competitive performance.

\begin{figure}[H]
\centering

\subfloat[Grad-CAM]{%
 \includegraphics[width=0.45\textwidth]{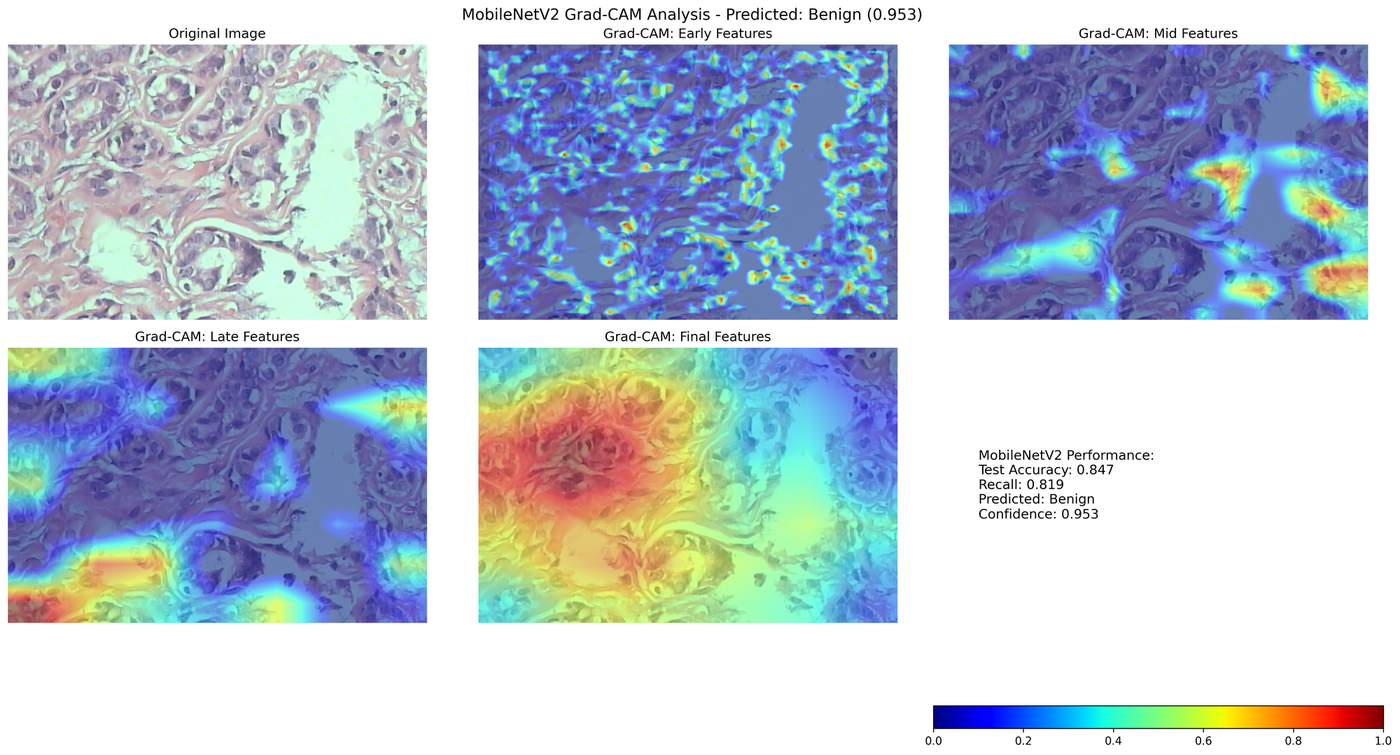}
}\hfill
\subfloat[Integrated Gradients]{%
 \includegraphics[width=0.45\textwidth]{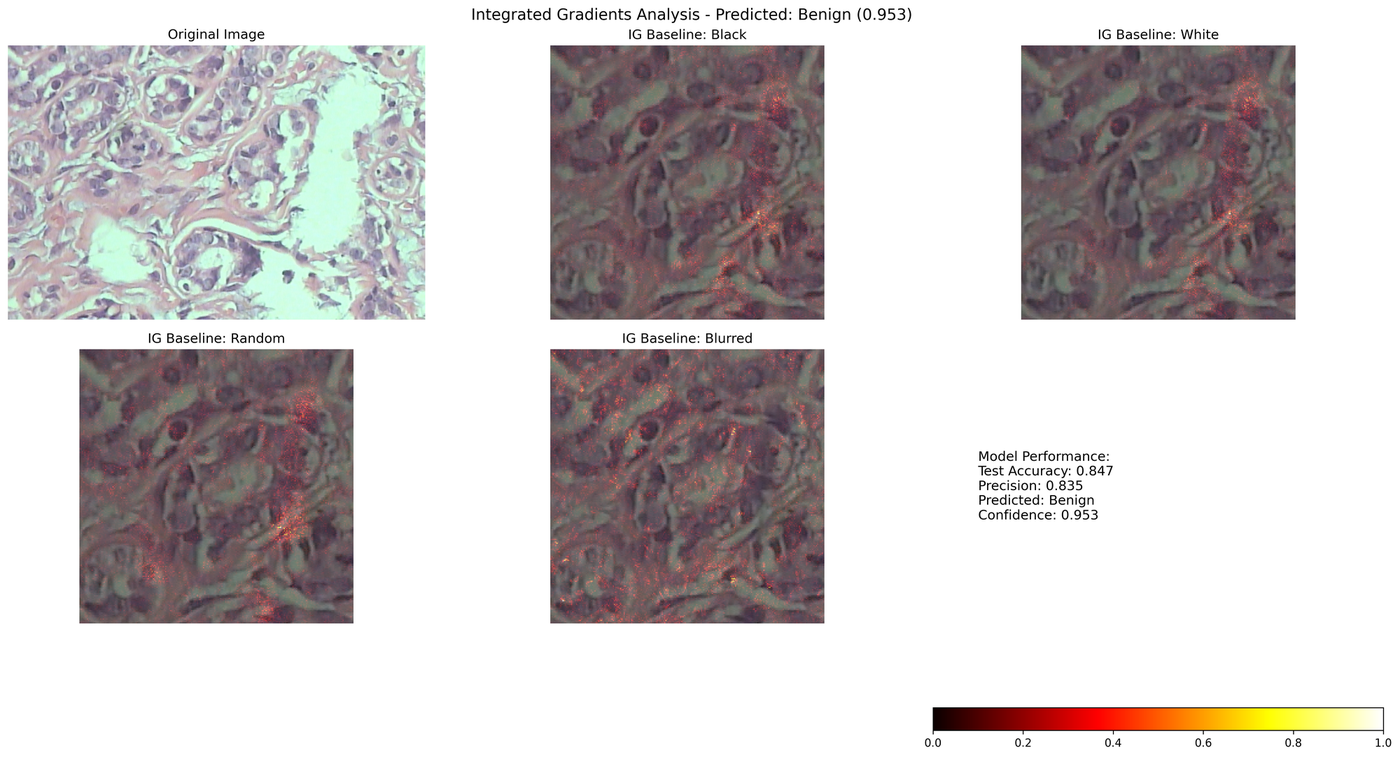}
}

\vspace{0.4cm}

\subfloat[LIME]{%
 \includegraphics[width=0.45\textwidth]{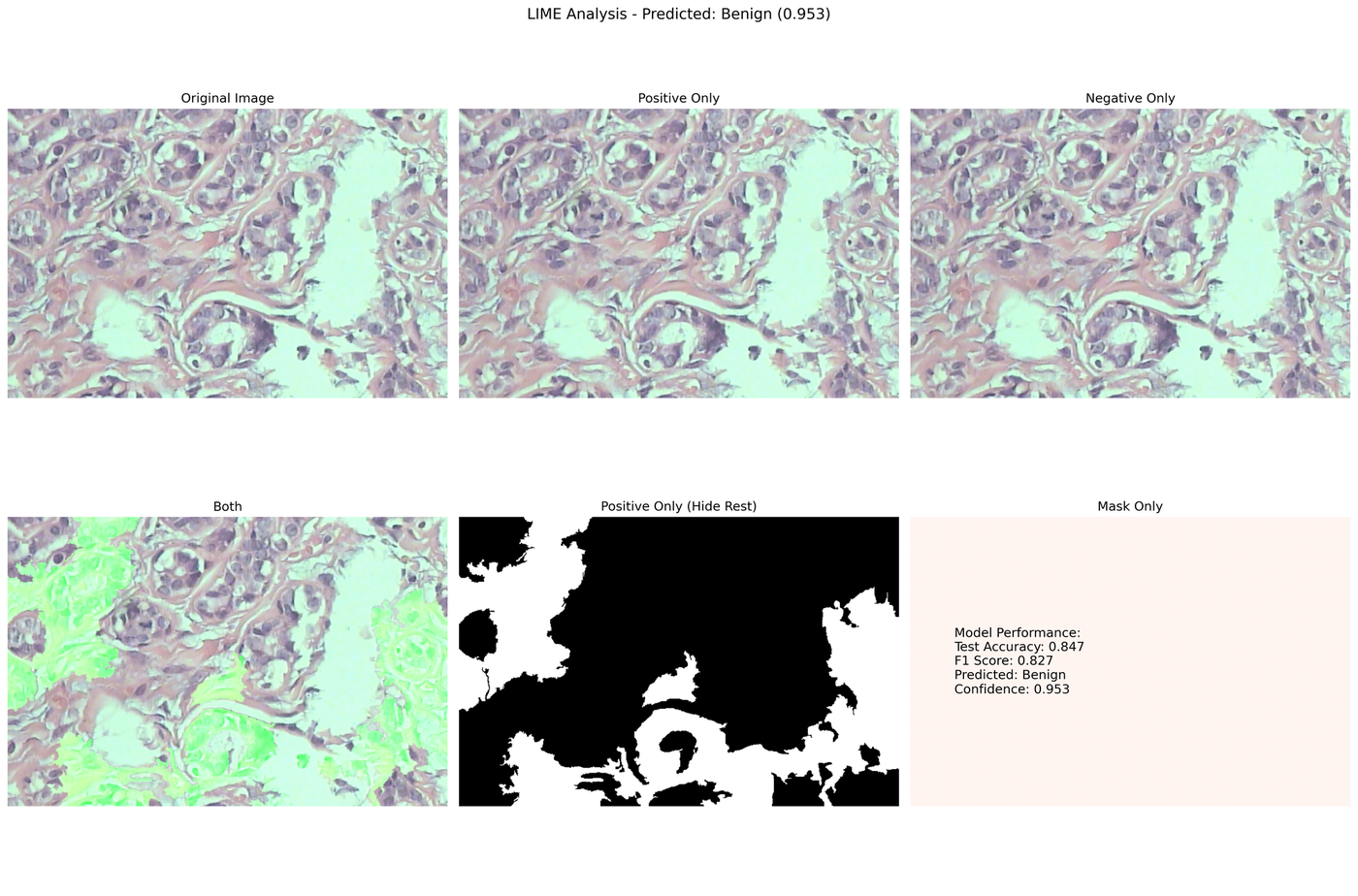}
}\hfill
\subfloat[LRP]{%
 \includegraphics[width=0.45\textwidth]{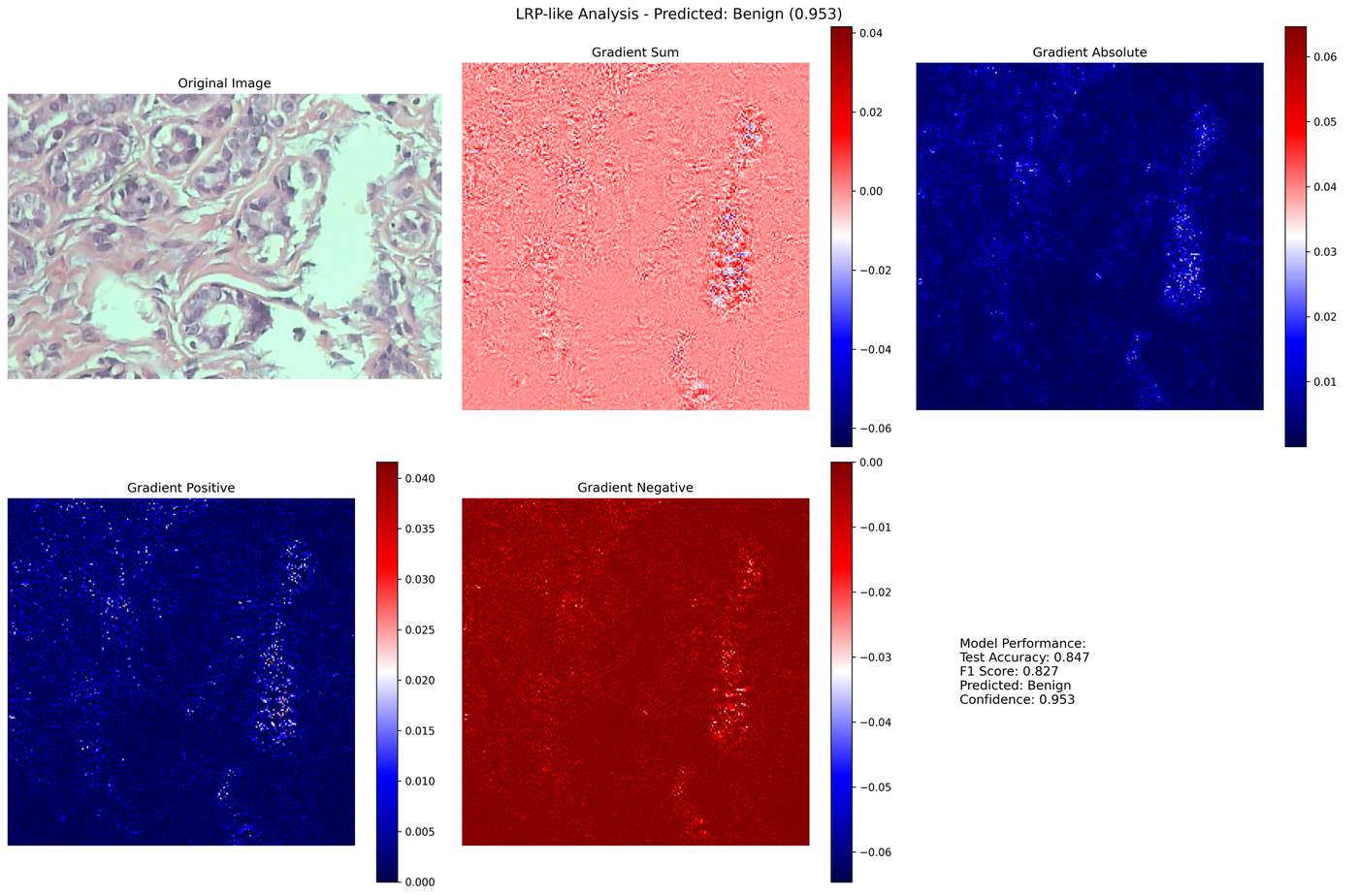}
}

\caption{XAI visualizations of MobileNetV2 predictions on the BreakHis dataset using (a) Grad-CAM, (b) Integrated Gradients, (c) LIME, and (d) LRP.}
\label{fig:mobilenet_xai}
\end{figure}

\subsection{Statistical Significance, Robustness, and Generalization Analysis}

The overall performance ranks derived from the classification experiments are well supported by the statistical validation results. Among all evaluated architectures, DenseNet121 consistently emerged as the best-performing model, achieving an average test accuracy of 99.4\% (95\% CI: 98.6-99.7\%), an AUC-ROC of 0.999 (95\% CI: 0.9963--1.0000), precision of 99.9\%, recall of 99.5\%, F1-score of 99.6\%, Cohen's $\kappa$ of 0.986, and an MCC of 0.984. The high agreement coefficients show the high level of consistency between predicted labels and true labels, and is the reason why DenseNet121 achieved the lowest Friedman rank (1.00) and statistically outperformed all other models ($p<0.001$). Furthermore, it is magnifications sensitive under 1.08\% across all the magnifications of 40$\times$, 100$\times$, 200$\times$ and 400$\times$ levels, showing excellent robustness and generalization capability in different tissue magnifications.

\begin{table*}[htbp]
\centering
\caption{Comprehensive statistical performance summary of all models on the BreakHis test set.}
\label{tab:statistical-performance-summary}
\renewcommand{\arraystretch}{1.15}
\setlength{\tabcolsep}{4pt}
\scriptsize
\begin{tabular}{llccccccccc}
\toprule
\textbf{Model} & \textbf{Arch.} & \textbf{Acc. (\%)} & \textbf{Prec. (\%)} & \textbf{Recall (\%)} & \textbf{Spec. (\%)} & \textbf{F1 (\%)} & \textbf{AUC} & $\boldsymbol{\kappa}$ & \textbf{MCC} & \textbf{F-Rank} \\
\midrule
DenseNet121 & CNN & 99.4 [98.6, 99.7] & 99.9 & 99.5 & 99.2 & 99.6 & 0.999 [0.9963, 1.0000] & 0.986 & 0.984 & 1.00 \\
EfficientNet & CNN & 86.2 [83.6, 88.4] & 83.6 & 86.0 & 86.3 & 84.8 & 0.940 [0.9192, 0.9608] & 0.685 & 0.700 & 3.62 \\
Xception & CNN & 85.8 [83.2, 88.1] & 80.0 & 91.2 & 81.5 & 85.2 & 0.939 [0.9180, 0.9600] & 0.676 & 0.689 & 3.50 \\
InceptionV3 & CNN & 84.6 [81.9, 86.9] & 79.5 & 88.4 & 81.5 & 83.7 & 0.933 [0.9111, 0.9549] & 0.648 & 0.665 & 4.00 \\
MobileNetV2 & CNN & 84.7 [82.0, 87.0] & 83.5 & 81.9 & 86.9 & 82.7 & 0.933 [0.9111, 0.9549] & 0.651 & 0.673 & 4.12 \\
ResNet50 & CNN & 83.6 [80.9, 86.0] & 79.1 & 86.2 & 81.5 & 82.5 & 0.919 [0.8950, 0.9430] & 0.625 & 0.645 & 5.25 \\
VGG19 & CNN & 81.3 [78.4, 83.9] & 76.5 & 84.1 & 79.0 & 80.1 & 0.905 [0.8792, 0.9308] & 0.573 & 0.599 & 7.50 \\
Swin Transformer & ViT/Swin & 79.5 [76.5, 82.2] & 73.6 & 84.6 & 75.3 & 78.7 & 0.860 [0.8293, 0.8907] & 0.532 & 0.564 & 8.00 \\
ViT & ViT/Swin & 77.6 [74.6, 80.4] & 71.1 & 84.3 & 72.2 & 77.2 & 0.858 [0.8271, 0.8889] & 0.488 & 0.530 & 8.75 \\
ConvNeXt & CNN & 77.1 [74.0, 79.9] & 85.1 & 89.1 & 2.4 & 87.0 & 0.653 [0.6110, 0.6950] & 0.477 & -0.180 & 9.25 \\
\bottomrule
\end{tabular}
\end{table*}

EfficientNet, Xception, InceptionV3 and MobileNetV2 were the second tier of models, with an accuracy of 84.6\% to 86.2\% and AUC ranging between 0.933 and 0.940. They were not far off their statistical measures from their predictive performance. The best agreement metrics with DenseNet ($\kappa = 0.685$ and MCC $= 0.700$) and the lowest magnification variability ($\sigma = \pm0.75\%$), EfficientNet earned the top ranking (Friedman rank $= 3.62$), with high performance at all magnifications. Likewise, Xception achieved a high sensitivity (91.2\%) and an MCC of 0.689, meaning that it had a high recall value and good predictive consistency. The moderate agreement between both models (InceptionV3 and MobileNetV2, $\kappa = 0.65$) and similar AUC values (0.933) helps to justify the models' statistically similar performance when compared pairwise.

The statistical results also account for the relatively poor performance of ResNet50 and VGG19. Both models had satisfactory accuracy (83.6\% and 81.3\%) and relatively low agreement coefficient (0.625 and 0.573), and moderate MCC values (0.645 and 0.599), which showed less reliable classification boundaries and less robustness in variable magnification levels. Because larger magnification sensitivities, $\sigma = \pm2.36\%$ and $\pm1.99\%$ on image magnification were suggested with their larger effects of dependence on image scale variations, they had lower Friedman rankings.

\begin{figure}[H]
 \centering
 \includegraphics[width=0.75\linewidth]{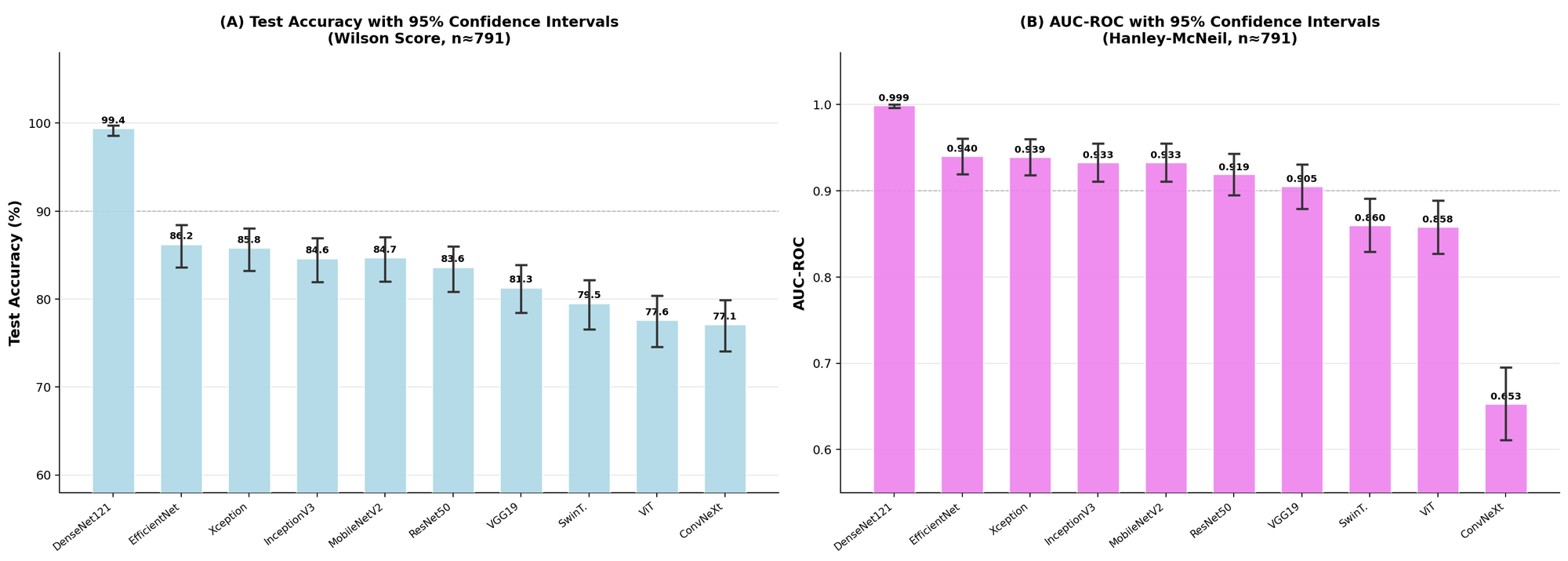}
 \caption{Statistical comparison of the 10 models tested in the patient-stratified BreakHis test set, with test accuracy and AUC-ROC and their 95\% confidence intervals quantifying predictive uncertainty and comparative performance.}
 \label{fig:single13}
\end{figure}

The transformer based architectures (Swin Transformer and Vision Transformer (ViT)) had relatively poor accuracy with 79.5\% and 77.6\%, respectively, and AUC values less than 0.86. There is moderate agreement between these results and their moderate agreement statistics ($\kappa = 0.532$ and $0.488$) and lower MCC scores (0.564 and 0.530). In addition, both architectures had the largest magnification sensitivities ( $\sigma = \pm2.99\%$ and $\pm3.18\%$), and very poor Friedman rankings (8.00 and 8.75) which signify that they were not stable and showed the least robust results for different image resolutions. The Mann--Whitney U test also found that CNN architectures were significantly more accurate ($p = 0.0032$) and more accurate for the AUC ($p = 0.0047$) than the transformers, indicating the advantage of convolutional inductive biases for relatively small histopathological datasets like BreakHis.

\begin{figure}[H]
 \centering
 \includegraphics[width=0.75\linewidth]{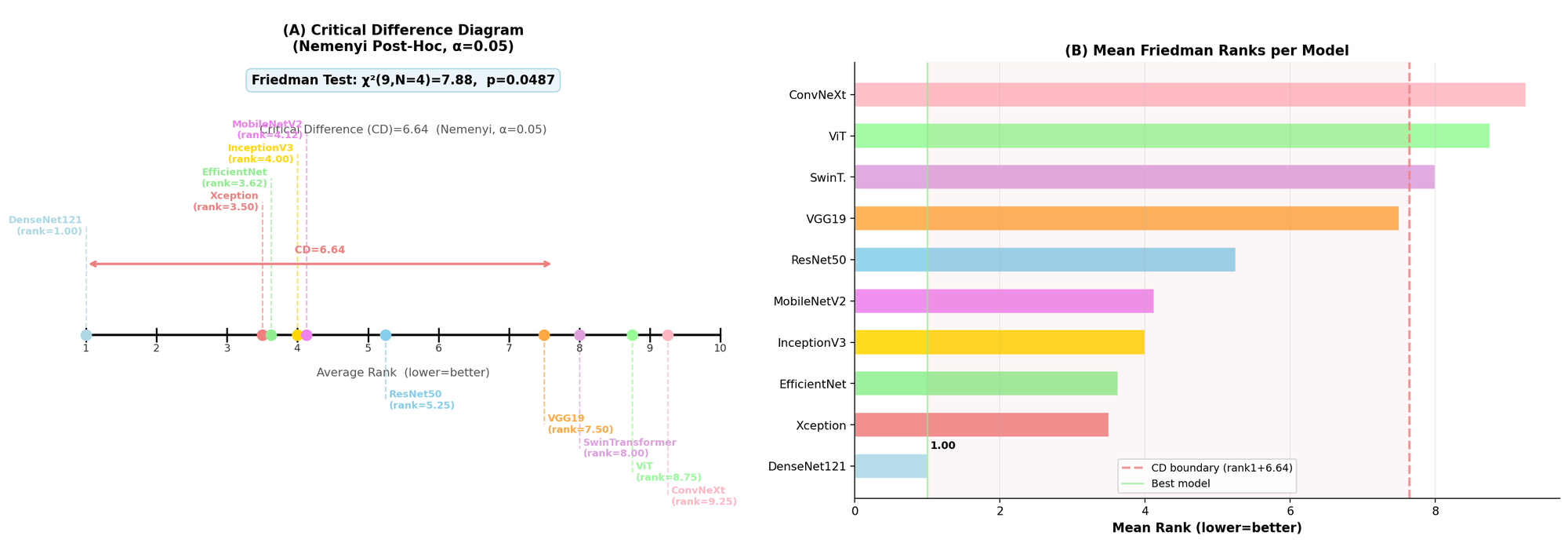}
 \caption{Friedman and Nemenyi non-parametric analyses comparing the ten architectures across four magnification levels that display statistically significant differences and average performance ranking of the evaluated models.}
 \label{fig:single14}
\end{figure}

Lastly, ConvNeXt was the worst-performing architecture in terms of accuracy (77.1\%), AUC (0.653) and MCC (-0.180) although it had a decent F1 score (87.0\%). The negative MCC value and low value of $\kappa$ (0.477) suggest poor discriminator capability and prediction bias to the majority class. This is also supported by its low Friedman ranking (9.25) and the high dispersion in the AUC values at different magnifications. The combined results of these statistical analyses provide further validation and corroboration of the experimental results: Conventional classification metrics validate the performance hierarchy as well as rigorous significance testing, agreement measures, confidence intervals and robustness analyses, which significantly adds to the credibility and reproducibility of the results reported.

\begin{figure}[H]
 \centering
 \includegraphics[width=0.75\linewidth]{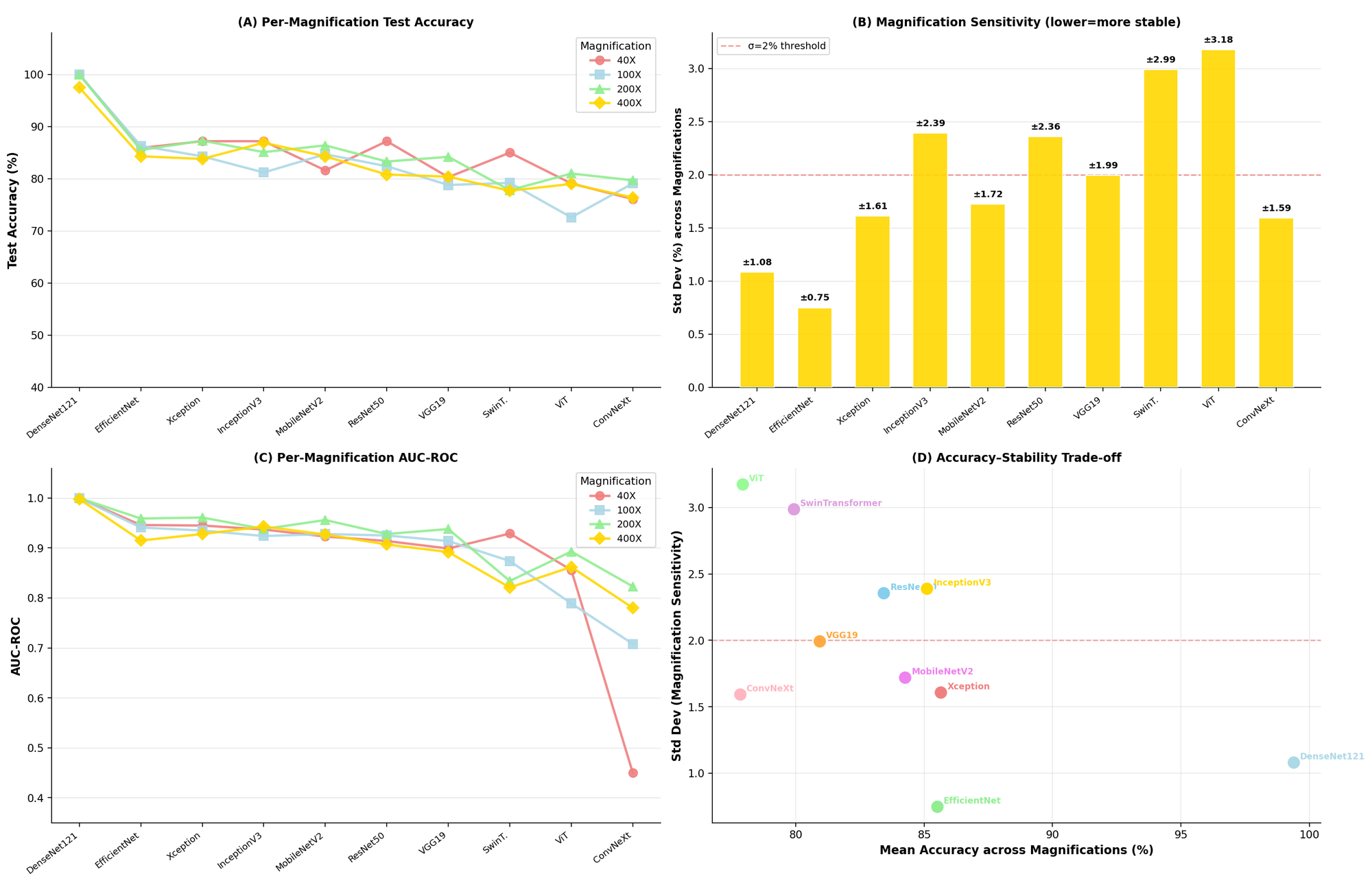}
 \caption{Magnification sensitivity analysis showing the robustness and stability of the model performance at different histopathological magnifications (40×, 100×, 200× and 400×) using the accuracy, AUC-ROC and deviation from accuracy metrics.}
 \label{fig:single15}
\end{figure}

\section{Limitations}
Although different transfer learning architectures were comprehensively benchmarked on BreakHis and various explainable AI techniques were used, there are still limitations. Firstly, the study was based on a single histopathological data set and a binary classification scenario to limit the generalizability of the result to other imaging modalities, multi-class classification of subtypes, tumour grading, and staging for clinical scenarios. All statistical measures were conducted in ways that ensure the superiority of DenseNet121, and these tests and the perfect scores at several magnifications in itself should be treated with a grain of salt due to near-perfect performance (99.4\% accuracy, AUC-ROC of 0.999, $\kappa=0.986$ and MCC = 0.984). Histopathological datasets, like BreakHis, may have nuances within and between magnifications that can create implicit correlations and result in inflated performance metrics. Moreover, the difference between the positive validation and test performance, as well as the very high agreement scores indicate that some degree of learning of the validation set may exist, in spite of the statistically significant improvement in performance. Furthermore, the XAI methods used are mostly qualitative and do not measure explanation faithfulness and clinical utility, quantitatively. Lastly, the complexity of the memory and computational hardware of large transformer-based architectures limited the exploration of a wide range of hyperparameters and external experiments. However, additional testing on independent multi-institutional data and quantitative assessment of interpretability would be required before confidently implementing for clinical use.

\section{Conclusion \& Future Work}
In this work, we benchmarked 10 state-of-the-art transfer learning architectures in order to classify breast cancer on the BreakHis dataset: CNN-based models (ResNet50, DenseNet121, EfficientNetV2, MobileNetV2, VGG19, Xception, ConvNeXt and InceptionV3) and transformer-based models (Vision Transformer and Swin Transformer). Both experimental and statistical analyses showed that DenseNet121 yielded the highest accuracy (99.41\%), AUC-ROC (0.9997), Cohen's $\kappa$ (0.986), and MCC (0.984) and was statistically significantly better than all models compared ($p < 0.001$). It was also shown to be robust in terms of magnification in tests at all the image resolutions and was found to be only slightly sensitive to variations in magnification ($\sigma = \pm1.08\%$). Moreover, Grad-CAM and LIME provided clinically relevant visual explanations for DenseNet121, EfficientNet and Swin Transformer, consistently identifying nuclei-dense and glandular regions that are crucial for diagnostic decisions. Together, these results highlight the potential of combining the strengths of transfer learning architectures with explainable AI approaches to achieve reliable clinical decision-making support systems for breast cancer diagnosis that are accurate, transparent, and clinically relevant.

However, the results are to be viewed with some caution as the study is based upon one histopathological dataset and binary classification. The proposed framework will be validated with independent multi-institutional and multi-modal data sets in future, extended to subtyping and grading tasks, explored on the hybrid CNN-Transformer architectures, and validated quantitatively and clinically to increase the generalization capabilities, reliability, and clinical utility.

\section*{Author contributions statement}

Author 1 analyzed the results and reviewed the manuscript draft. Authors 2 and 4 contributed to the overall methodology, original draft preparation, and data analysis. Author 3 conducted the experiments and generated the results. Authors 5 and 6 were responsible for project administration, results evaluation. All authors reviewed, edited, and approved the final version of the manuscript.

\section*{Data Availability}
The datasets analysed during the current study are publicly available. This study utilised the BreakHis (Breast Cancer Histopathological Image Classification) dataset, which is accessible through the public repository at:
\url{https://web.inf.ufpr.br/vri/databases/breast-cancer-histopathological-database-breakhis/}

No new datasets were generated during this study. All processed data supporting the findings of this research are available from the corresponding author upon reasonable request.

\section*{Ethics Statement}
This study used the publicly available BreakHis dataset consisting of de-identified histopathological images. No direct human participation, patient recruitment, or intervention was conducted by the authors. The dataset was originally collected following relevant clinical and ethical guidelines, and all images are anonymised. Therefore, additional ethical approval and informed consent were not required for this secondary analysis.

\section*{Acknowledgments}
This work was supported by the Deanship of Scientific Research, Vice Presidency for Graduate Studies and Scientific Research, King Faisal University, Saudi Arabia [Grant No.XXXXXX].

\bibliographystyle{unsrt}
\bibliography{references}

\end{document}